\documentclass[pdflatex,sn-mathphys-num]{sn-jnl}

\usepackage{hyperref}
\usepackage{graphicx}%
\usepackage{multirow}%
\usepackage{amsmath,amssymb,amsfonts}%
\usepackage{amsthm}%
\usepackage{mathrsfs}%
\usepackage[title]{appendix}%
\usepackage{xcolor}%
\usepackage{textcomp}%
\usepackage{manyfoot}%
\usepackage{booktabs}%
\usepackage{algorithm}%
\usepackage{algorithmicx}%
\usepackage{algpseudocode}%
\usepackage{listings}%

\usepackage[utf8]{inputenc} 
\usepackage[T1]{fontenc}    
\usepackage{hyperref}       
\usepackage{url}            
\usepackage{nicefrac}       
\usepackage{microtype}      
\usepackage{cleveref}
\usepackage{caption}
\usepackage[nolist]{acronym}
\usepackage[acronym]{glossaries}
\usepackage{siunitx}
\usepackage{longtable}
\usepackage{enumitem}       
\usepackage{wrapfig}        
\usepackage[font={small}]{caption}
\usepackage{comment}
\usepackage{framed}
\usepackage{xltabular}
\usepackage{chngcntr}
\usepackage{orcidlink}

\begin{document}

\title[One Law, Many Languages]{One~Law,~Many~Languages: Benchmarking~Multilingual~Legal~Reasoning for~Judicial~Support}

\author[1,4]{\fnm{Ronja} \sur{Stern}~\orcidlink{0009-0006-3980-5182}}
\equalcont{These authors contributed equally to this work.}

\author[1]{\fnm{Rasiah} \sur{Vishvaksenan}}
\equalcont{These authors contributed equally to this work.}

\author[2]{\fnm{Veton} \sur{Matoshi}~\orcidlink{0009-0002-6613-5701}} 

\author[3]{\fnm{Srinanda} \sur{Brügger Bose}~\orcidlink{0000-0002-7825-791X}}

\author[2]{\fnm{Matthias} \sur{Stürmer}~\orcidlink{0000-0001-9038-4041}}

\author[4]{\fnm{Ilias} \sur{Chalkidis}~\orcidlink{0000-0002-0706-7772}}

\author[5]{\fnm{Daniel E.} \sur{Ho}~\orcidlink{0000-0002-2195-5469}}

\author[1,2,5]{\fnm{Joël} \sur{Niklaus}~\orcidlink{0000-0002-2779-1653}}
\equalcont{These authors contributed equally to this work.}
\affil[1]{\orgname{University of Bern}, \country{Switzerland}}

\affil[2]{\orgname{Bern University of Applied Sciences}, \orgaddress{\country{Switzerland}}}

\affil[3]{\orgname{University of Fribourg}, \orgaddress{\country{Switzerland}}}

\affil[4]{\orgname{University of Copenhagen}, \orgaddress{\country{Denmark}}}

\affil[5]{\orgname{Stanford University}, \orgaddress{\country{USA}}}



\abstract{
Recent strides in Large Language Models (LLMs) have saturated many Natural Language Processing (NLP) benchmarks, emphasizing the need for more challenging ones to properly assess LLM capabilities. 
However, domain-specific and multilingual benchmarks are rare because they require in-depth expertise to develop.  
Still, most public models are trained predominantly on English corpora, while other languages remain understudied, particularly for practical domain-specific NLP tasks.
In this work, we introduce a novel NLP benchmark for the legal domain that challenges LLMs in five key dimensions: processing \emph{long documents} (up to 50K tokens), using \emph{domain-specific knowledge} (embodied in legal texts), \emph{multilingual} understanding (covering five languages), \emph{multitasking} (comprising legal document-to-document Information Retrieval, Court View Generation, Leading Decision Summarization, Citation Extraction, and eight challenging Text Classification tasks) and \emph{reasoning} (comprising especially Court View Generation, but also the Text Classification tasks). 
Our benchmark contains diverse datasets from the Swiss legal system, allowing for a comprehensive study of the underlying non-English, inherently multilingual legal system. 
Despite the large size of our datasets (some with hundreds of thousands of examples), existing publicly available multilingual models struggle with most tasks, even after extensive in-domain pre-training and fine-tuning. 
We publish all resources (benchmark suite, pre-trained models, code) under permissive open CC BY-SA licenses.
}




\maketitle

\begin{acronym}[UMLX]
    \acro{FSCS}{Federal Supreme Court of Switzerland}
    \acro{SCI}{Supreme Court of India}
    \acro{ECHR}{European Convention of Human Rights}
    \acro{ECtHR}{European Court of Human Rights}
    \acro{SCOTUS}{Supreme Court of the United States}
    \acro{SPC}{Supreme People's Court of China}
    \acro{SJP}{Swiss-Judgment-Prediction}
    \acro{ASO}{Almost Stochastic Order}
    \acro{ILDC}{Indian Legal Documents Corpus}
    
    \acro{US}{United States}
    \acro{EU}{European Union}

    \acro{NLP}{Natural Language Processing}
    \acro{ML}{Machine Learning}
    \acro{LJP}{Legal Judgment Prediction}
    \acro{SJP}{Swiss-Judgment-Prediction}
    \acro{PJP}{Plea Judgment Prediction}
    
    \acro{BERT}{Bidirectional Encoder Representations from Transformers}
    \acro{LSTM}{ Long Short-Term Memory }
    \acro{GRU}{Gated Recurrent Unit}
    \acro{BiLSTM}{Bidirectional Long Short-Term Memory}
    \acro{CNN}{Convolutional Neural Networks}

    \acro{PLM}{pre-trained Language Model}
    \acro{LM}{Language Model}

    \acro{RTD}{Replaced Token Detection}

    \acro{CLT}{Cross-Lingual Transfer}
    \acro{HRL}{high resource language}
    \acro{LRL}{low resource language}

    \acro{POS}{Part-of-Speech}
    
    \acro{SLTC}{Single Label Text Classification}
    \acro{MLTC}{Multi Label Text Classification}
    \acro{TC}{Text Classification}
    \acro{NLU}{Natural Language Understanding}
    \acro{IR}{Information Retrieval}
    \acro{NER}{Named Entity Recognition}
    \acro{NLU}{Natural Language Understanding}
    \acro{QA}{Question Answering}
    \acro{NLI}{Natural Language Inference}

    \acro{GNB}{Gaussian Naive Bayes}
    \acro{DT}{Decision Tree}
    \acro{SVM}{Support-Vector Machine}
    \acro{RF}{ Random Forest}
    \acro{XGBoost}{eXtreme Gradient Boosting}
    \acro{MLIR}{Multilingual Information Retrieval}
    \acro{IR}{Information Retrieval}
    \acro{NDCG}{Normalized Discounted Cumulative Gain}
    \acro{LD}{Leading Decision}
    \acro{FSCD}{Federal Supreme Court Decisons}
    \acro{SFCS}{Swiss Federal Supreme Court}
    \acro{LLM}{Large Language Model}
    \acro{LM}{Language Model}

    \acro{CVG}{Court View Generation}
    \acro{JP}{Judgment Prediction}
    \acro{LAP}{Law Area Prediction}
    \acro{SLAP}{Sub Law Area Prediction}
    \acro{CP}{Criticality Prediction}
    \acro{LDS}{Leading Decision Summarization}
    \acro{CE}{Citation Extraction}
    
    \acro{SBERT}{Sentence-Bert}
    \acro{Doc2Doc}{Document-to-Document}
    \acro{mUSE}{Multilingual Universal Sentence Encoder}

\end{acronym}


\section{Introduction}
\label{sec:introduction}

\begin{figure}[b]
    \centering
    \includegraphics[width=1\textwidth]{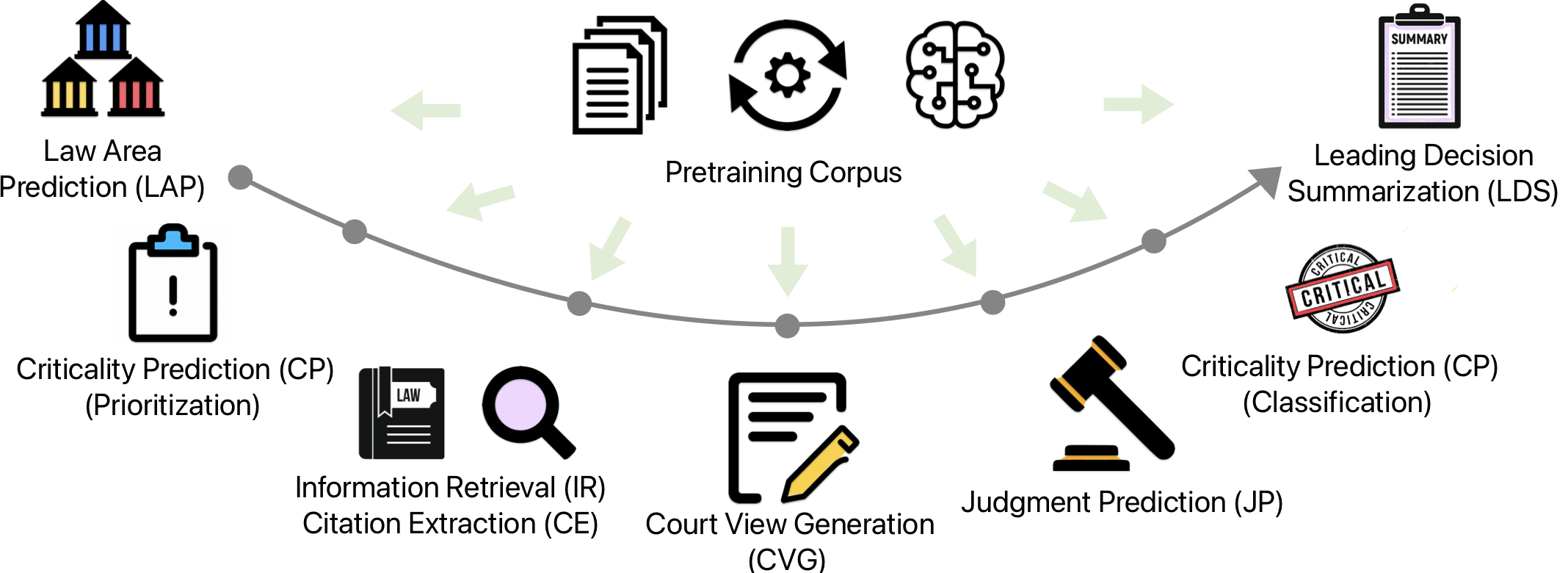}
    \caption{We showcase how we picture this benchmark supporting the judicial system end-to-end. 1) we route incoming complaints to the correct chamber (LAP), 2) we prioritize easier cases for preliminary automated processing (CP), 3) IR and CE help judges and clerks in finding and citing relevant legislation and case law, 4) we help the judiciary in drafting decisions (CVG), 5) we predict and verify judgments based on the history of the court (JP), 6) we predict which decisions are likely going to have a big impact on future jurisprudence (CP) and 7) we summarize the leading decisions (LDS). The LLMs performing all these tasks are trained on and retrieve from the pretraining corpus containing most publicly available Swiss legal data.}
    \label{fig:bigger-picture-1}
\end{figure}

The history of \ac{NLP} within the legal domain is extensive \citep{ashley_2017}, with remarkable progress recently \citep{katz2023natural}. Notably, the introduction of datasets containing legal data from various jurisdictions worldwide \citep{paul2021lesicin,chalkidis_extreme_2019}, as well as the development of more domain-specific tasks and benchmarks \citep{hendrycks2021cuad, Li2021CourtOG, semo_classactionprediction_2022, brugger_multilegalsbd_2023, hwang_multi-task_2022, niklaus2023lextreme, thakur_beir_2021, chen_mtg_2022, guha_legalbench_2022} have significantly contributed to the progress in the field.
General benchmarks such as SuperGLUE \citep{wang_superglue_2019} are saturated and ineffective at differentiating \acp{LLM}. Hence, larger, challenging benchmarks are urgently needed, especially in the domain-specific context. In Switzerland, the availability of only one dataset for evaluating legal \acp{LLM} \citep{niklaus_swiss-judgment-prediction_2021, niklaus_empirical_2022} hampers the assessment of their performance and effectiveness within the country's diverse linguistic and legal landscape.

In this  paper, we introduce seven new datasets covering a range of tasks and spanning across five languages (German, French, Italian, Romansh, English) within the same overarching jurisdiction. These datasets are derived from 26 cantons and the \ac{SFCS} in the uniquely multilingual and multi-jurisdictional context of Switzerland. The country's multiple official languages
and a wealth of data for its size, position Switzerland as an exemplary testbed for assessing \acp{LLM} in a multilingual environment with distinct legal frameworks. Our assessment concentrates on three classification tasks -- \ac{CP}, \ac{JP}, and \ac{LAP} -- an \ac{IR} task and two generative tasks -- \ac{CVG} and \ac{LDS} (see \Cref{fig:bigger-picture-1} for an overview).
To facilitate a comprehensive analysis and provide baselines for future research, we evaluate an array of models on our datasets similar to \citet{hwang_multi-task_2022} or \citet{niklaus2023lextreme}. Furthermore, we have pretrained our own Swiss legal models, Legal Swiss RoBERTa\textsubscript{Base/Large} and Legal Swiss Longformer\textsubscript{Base}. 
Our tasks challenge current models significantly, with the best performing model only achieving an aggregated Macro F1 score of 48.4. ChatGPT was not able to solve the \ac{TC} tasks well, considerably lagging behind fine-tuned models.
The results for \ac{CP}, \ac{IR} and \ac{CVG} are particularly underwhelming, seeming rather arbitrary. We invite the research community to develop new methods to tackle these hard tasks.
All data employed in this study is from publicly available sources (see \textit{\href{https://entscheidsuche.ch/dataUsage}{https://entscheidsuche.ch/dataUsage}} and \textit{\href{https://www.fedlex.admin.ch/en/legal-information}{https://www.fedlex.admin.ch/en/legal-information}}) and is published on the HuggingFace Hub under a CC BY-SA license (link released upon acceptance). 

This paper makes three contributions. 
First, we present seven public multilingual datasets containing Swiss legal documents. 
Second, we release two large, in-domain pretraining datasets, and pretrain three new models: Legal-Swiss-RoBERTa\textsubscript{Base/Large} and Legal-Swiss-LongFormer\textsubscript{base}.
Third, we evaluate multilingual baselines on our datasets and compare them to our models. Although in-domain pretraining improves performance, significant room for improvement remains in most tasks.

The article below is structured as follows: Next, we provide an overview of related work (Chapter 2) and explain important background information of the Swiss legal system (Chapter 3). We then introduce our new legal datasets by describing how they were created and what they can be used for (Chapter 4). In the experiments section, we present how we pretrained and tested our legal models (Chapter 5). And finally we illustrate our results (Chapter 6), provide an error analysis (Chapter 7), and conclude our work with recommendations for future work (Chapter 8).



\section{Related Work}
\label{sec:related_work}

In this section, we discuss prior work on benchmarks for long documents, domain specificity, multilinguality, multitasking and reasoning. Additional task-specific related work is presented in \Cref{sec:add_related_work}.

\paragraph{Long Documents}

\textsc{SCROLLS} (\textbf{S}tandardized \textbf{C}ompa\textbf{R}ison \textbf{O}ver \textbf{L}ong \textbf{L}anguage \textbf{S}equences) 
consists of  summarization, \ac{QA}, and \ac{NLI} tasks with example inputs typically in the thousands of English words \citep{shaham2022scrolls}. 
\textsc{MuLD} (\textbf{Mu}ltitask \textbf{L}ong \textbf{D}ocument)
is a set of six tasks (twice \ac{QA}, style change detection, classification, summarization, and translation) where each input is at least 10K tokens, with some up to almost 500K tokens \citep{hudson2022muld}. While valuable, both benchmarks are for the English language only and they do not cover court rulings nor legislation.


\paragraph{Domain Specificity}

\textsc{LexGLUE} covers six predictive tasks over five datasets made of English documents from the US, EU, and Council of Europe \citep{chalkidis2022lexglue}. 
LEXTREME is a multi-lingual and multi-task benchmark for the legal domain \citep{niklaus2023lextreme}. 
LegalBench \citep{guha_legalbench_2022} covers zero-shot and few-shot \ac{LM} evaluation for diverse realistic legal tasks in English.
LBOX OPEN \citep{hwang_multi-task_2022} consists of five legal tasks from South Korea. 
Although there exist legal benchmarks, they are mostly focused on understanding tasks, while we also provide challenging generation and information retrieval tasks.

\paragraph{Multilinguality}
\textsc{XTREME}~\citep{hu_xtreme_2020} evaluates cross-lingual generalization across six tasks and ten datasets, covering 40 languages. Some datasets were cross-lingual, others were extended via professional and automatic translations.
XTREME-UP (\textbf{U}nder-Represented and \textbf{U}ser-Centric with \textbf{P}aucal Data) 
expands XTREME, focusing on few-shot evaluation of multilingual models for user-centric tasks \citep{ruder2023xtremeup}. It includes 88 underrepresented languages like Swahili, Burmese, and Telugu. While these benchmarks contain many languages, they do not cover any legal data. 


\paragraph{Multitasking}

\textsc{GLUE}
(\textbf{G}eneral \textbf{L}anguage \textbf{U}nderstanding \textbf{E}valuation)
\citep{wang_glue_2018}, an early benchmark of sentence NLU tasks evaluating general-purpose neural \acp{LM}, quickly became obsolete due to advanced models like BERT \citep{devlin_bert_2019}. Its updated version, \textsc{SuperGLUE}~\citep{wang_superglue_2019}, introduced new tasks challenging for machines yet solvable by humans.
\textsc{MMLU} (\textbf{M}assive \textbf{M}ultitask \textbf{L}anguage \textbf{U}nderstanding) 
features only zero-shot and few-shot learning tasks~\citep{hendrycks_measuring_2021}, containing 16K multiple-choice questions across 57 subtasks, in subjects such as humanities, social and hard sciences.
BIG-Bench \citep{srivastava_beyond_2022}
(Beyond the Imitation Game)
consists of 204 language tasks created by 450 authors from 132 institutions. The tasks cover topics such as linguistics, childhood development, math, common-sense reasoning, biology, physics, social bias, software development.
Although there exist many multi-task benchmarks, they are often restricted to language understanding while we also cover information retrieval and generation tasks.

\paragraph{Reasoning}
Reasoning refers to the capability of AI systems to understand, interpret, and generate reasoning based on natural language input. Recent research has made significant strides in enhancing the reasoning capabilities of LLMs, mostly via different prompting techniques. \citet{Wei2022} introduced the concept of Chain-of-Thought prompting, which has been shown to improve the performance of \acp{LM} on reasoning tasks. Later, more sophisticated techniques, such as least-to-most prompting \cite{zhou_least--most_2023}, Self-Consistency \cite{wang_self-consistency_2023}, Tree of Thoughts \cite{yao_tree_2023, Long2023} or Graph of Thoughts \cite{besta_graph_2023}, proved to be even more capable, based on tasks related to symbolic manipulation, compositional generalization, and math reasoning. 
A range of studies have explored legal reasoning in the context of AI and law. \citet{guha_legalbench_2023} proposed LegalBench, a collaborative benchmark for legal reasoning, based on the IRAC (Issue, Rule, Application, Conclusion) framework. Chain-of-thought prompting has been further explored by \citet{Yu2022} in the context of legal reasoning tasks, where it was found to be effective, particularly when combined with specific legal reasoning techniques.
While reasoning has been studied in legal tasks, previous work mostly focused on the English language. We study diverse challenging legal reasoning tasks in the inherently multilingual Swiss jurisdiction.

\section{Background on the Swiss Legal System}
\label{background-legal-system}

In order to understand the uniqueness of the datasets and its context, this chapter briefly explains organizational aspects of the Swiss Federal Supreme Court (SFSC), the structure of Swiss court rulings, and relevant processes within Swiss legal system.

\subsection{Structure and Function of the Swiss Federal Supreme Court}
Switzerland comprises 26 cantons, each with unique jurisdiction and court organization. The \ac{SFCS} is Switzerland's highest legal authority and final arbiter for federal criminal, administrative, patent, and cantonal courts. Its decisions bridge gaps in legislation, and shape the development of the law and its adaptation to changing circumstances. The \ac{SFCS} has seven divisions, specializing in public, penal, and civil law \citep{SwissFederalSupremeCourt}.
The Supreme Court is responsible for addressing appeals stemming from the Cantonal Courts. Some \ac{SFCS} cases over time are designated as \acp{LD} and are published separately in the official portal of the Federal Tribunal (Bundesgerichte). The names of the parties are anonymized for concerns of privacy and the decisions published have the potential of influencing future jurisprudence. Cantonal court proceedings begin at the lowest instance and may be appealed higher, with appeal stages varying by canton and legal area. 
Courts, especially the Supreme Court, first check if existing judgments have already settled the legal principles of the case at hand. Some very relevant judgements hailing from the Cantonal Courts are also published on their official websites. Common Law Courts recognize precedents established over time. Many countries have official Supreme Court portals listing key decisions that could influence or determine future legal directions.

\subsection{Structure of Swiss Court Rulings}
Swiss court decisions typically consist of four major sections: 1) The \emph{rubrum} (introduction) contains the date and chamber, mentions the involved judge(s) and parties and finally the topic of the decision. 2) The \emph{facts} describe what happened in the case and form the basis for the considerations of the court. The higher the level of appeal, the more general and summarized the facts. 3) The \emph{considerations} reflect the formal legal reasoning, citing laws and other influential rulings, and forming the basis for the final ruling. 4) The \emph{rulings}, finally, are an enumeration of the binding decisions made by the court. This section is usually rather short and summarizes the considerations.

\subsection{Jurisdiction and Admissibility Before the Swiss Courts}
This section covers jurisdiction (the authority to decide a case as defined by law) and admissibility in the light of the Swiss Civil Procedure Code (SCPC). Jurisdiction is equally important in criminal and administrative matters. Determining the jurisdiction of the court forms an integral part of the Swiss legal system. Cantonal law is responsible for authorising the courts for specific disputes. The decision pertaining to the jurisdiction is taken by the court on its own accord, also known as "ex officio". 
The court needs to verify that the case is not currently being debated in another proceeding or has already been resolved in a previous binding decision. Additionally, the court must ensure that all procedural requirements outlined in the SCPC are met before accepting a case.

\subsection{Swiss Legal Proceedings: From Initiation to Judgment}
Swiss law does not mandate oral pleadings, written pleadings mark the initiation of the ordinary proceedings before the court. The parties must present a detailed description of the facts of the case along with supporting evidence. The statement of claim forms the first set of the pleadings in claiming a prayer of relief to a dispute. Following which the court serves the statement of claim to the defendants along with a deadline to the filing of the statement of defence along with the liberty of filing a counter claim. 
The judge exercises discretion to call in for additional evidence and a clarification if needed. Upon exchanging the pleadings among the parties, the court allows admission of evidence followed by the opportunity to the respective parties to comment on the evidence and the merits of the given case; this forms part of the closing submissions. An 'instruction hearing' or a settlement meeting is often proposed by the courts in the German speaking cantons of Switzerland after the closing submissions. After taking into consideration the pleadings made by the parties and the evidence produced, the judge delivers the final decision. At this juncture it is important to note the contribution of the court clerks. To formulate a decision, the judge has to carefully consider every submission. They have to take into account the evidence and weigh in the law for a just and fair decision. However, this process is quite exhaustive and time consuming. To expedite this process, the judicial clerks prepare a memorandum at the beginning of the hearing and draft a proposed judgement under the supervision of a given judge. They participate in the hearings and keep a record of the proceedings pursuant to which they derive at the judicial reasoning behind the decision. In Switzerland, the judgment is also co-signed by a law clerk working on the case with the judge.

In the following we show a simplified structure of the flow of a case and link to possible automated support through datasets proposed in \Cref{sec:datasets}.

\begin{enumerate}
    \item \textbf{Complaint Filing}\\
    \emph{The statements of claim (complaint) are filed before the court by the plaintiff}
    \item \textbf{Routing} (\acf{LAP}, see \Cref{sec:lap})\\
    \emph{The case is routed to the responsible chamber within the court} 
    \item \textbf{Informing Defendant}
    \emph{The statements of claim are sent to the defendant along with a date to file the statements of defense and counter-claims}
    \item \textbf{Legal Research} (\acf{IR}, see \Cref{sec:ir})\\
    \emph{The clerk researches relevant case law and legislation}
    \item \textbf{Write Facts}\\
    \emph{The clerk writes the facts description}
    \item \textbf{Write Legal Reasoning} (\acf{CVG}, see \Cref{sec:cvg})\\
    \emph{The clerk writes the considerations}
    \item \textbf{Clarifications}\\
    \emph{The judge asks for clarifications and additional evidence}
    \item \textbf{Settlement}\\
    \emph{The judge proposes a meeting for settlement}
    \item \textbf{Verdict} (\acf{JP}, see \Cref{sec:jp})\\
    \emph{If no settlement is reached, the judge delivers the verdict}
    \item \textbf{Leading Decisions} (\acf{CP}, see \Cref{sec:cp})\\
    \emph{The court's department president determines whether the case will be published as a leading decision (usually when the law is interpreted in a new way)}
    \item \textbf{Summarization} (\acf{LDS}, see \Cref{sec:lds})\\
    \emph{The leading decision is summarized}
\end{enumerate}

\section{SMILED: The Datasets}
\label{sec:datasets}

\begin{table*}[ht]
\centering
\vspace{-5ex}
    \caption{Overview of all datasets and their multilingualism: 
    Abbreviations: \textbf{Cant}onal, \textbf{Fed}eral, \textbf{Fac}ts, \textbf{Cons}iderations.
    Column Fac and Cons report the mean number of tokens. Sections Facts and Considerations are not available for Summarization, Legislation and Rulings due to different format, thus we report the mean number of tokens for the full text and mark it with *.}
\resizebox{1\textwidth}{!}{
\begin{tabular}{llrrrrrrrr}
         \toprule
         \textbf{Dataset} & \textbf{Level} & \textbf{Total} & \textbf{DE}   & \textbf{FR} & \textbf{IT} & \textbf{RM} & \textbf{EN} & \textbf{Fac} & \textbf{Cons}\\
         \midrule
         Rulings                & Cant + Fed & 638K & 320K & 247K & 71K & - & 180 & - & *7K\\ 
         Leading Decisions      & Fed & 21K    & 14K       & 6K  & 1K  & - & -  & 689 & 3K \\
         Legislation            & Cant + Fed & 36K    & 18K      & 11K  & 6K  & 534 & 207 & - & *7K \\
        \midrule
         Doc2Doc IR             & Fed & 141K   & 87K    & 46K  & 8K  & - & - & 847 & 3K\\
         \midrule
        Citation Extraction     & Fed & 131K & 85K & 38K & 8K & - & - & - & 204 \\ 
         \midrule
         Criticality Prediction            & Fed & 139K   & 85K     & 45K     & 8K  & - & - & 828 & 3K\\
         Law Area Prediction              & Cant + Fed & 329K & 127K  & 156K  & 46K & - & - & 2K & 4K\\
         Judgment Prediction & Cant + Fed & 329K & 160K  & 128K  &  41K & - & - & 2K & 4K\\
         \midrule
         Court View Gen         & Cant + Fed & 404K & 197K  & 163K  & 44K & - & - & 2K & 5K\\
         Court View Origin Gen        & Fed & 270 & 49  & 221  & - & - & - & 1K & 6K\\
         Leading Decisions Summ      & Fed & 18K & 12K & 5K & 835 & - & - & - & *3K\\
        \bottomrule
    \end{tabular}
    }
    \label{tab:lang-dist}
\vspace{-2ex}
\end{table*}

In this section, we introduce SMILED (\textbf{S}wiss \textbf{M}ultil\textbf{I}ngual \textbf{LE}gal \textbf{D}ata) our new datasets, how we created them and what they can be used for.
\Cref{tab:lang-dist} presents our eleven datasets, collected from 26 cantons (in addition to federal decisions), 184 courts, 456 chambers, four main law areas, and five languages as seen in \Cref{tab:metadata} in the Appendix. Document availability varies significantly across cantons and courts.\footnote{\tiny{The \ac{SFCS} is the only court where we have complete data, since all decisions since 2007 have been published.}} Most courts are monolingual, but there are cantons where multiple languages are used in documents. 
Besides decision-based datasets, we also provide a collection of approx. 35K laws from cantonal and federal jurisdictions in Switzerland.

While most \ac{SFCS} cases are written in German, French is more common for cantonal cases. We partitioned downstream datasets into training (until 2015), validation (2016-2017), and test (2018-2022) sets. We opted for a relatively large test split because \acp{LLM} seem to need relatively little training data \cite{brown-etal-gpt3}. A large test set allows longitudinal studies, including COVID-19 pandemic years. This large temporal gap between the newest training (2015) and test (2022) samples might contribute to model difficulties (see \Cref{sec:results}). Source data undergo rigorous curation by established Swiss institutions such as courts and administrative bodies, including manual anonymization at an approximate cost of 45 minutes per case \citep{niklaus_swiss-judgment-prediction_2021}. While we extract labels from the texts and from metadata semi-automatically, mostly using regular expressions (regex), the underlying data is of highest quality, since it is produced by highly trained legal professionals such as judges and court clerks. Collecting such volume of high-quality data would cost billions of dollars to annotate privately. 

In contrast to previous datasets and benchmarks, our dataset closely mimics the main processes of the Swiss judicial system, making the benchmark highly relevant for real-world applications. Given the unsatisfactory performance of various LLMs on many tasks, we believe that this benchmark addresses a critical gap by still containing ample discriminative room across models.


\subsection{Database Creation Pipeline}
\label{sec:dataset_creation}

\begin{wrapfigure}{r}{0.65\textwidth}
    \vspace{-4ex}
    \centering
    \includegraphics[width=0.65\textwidth]{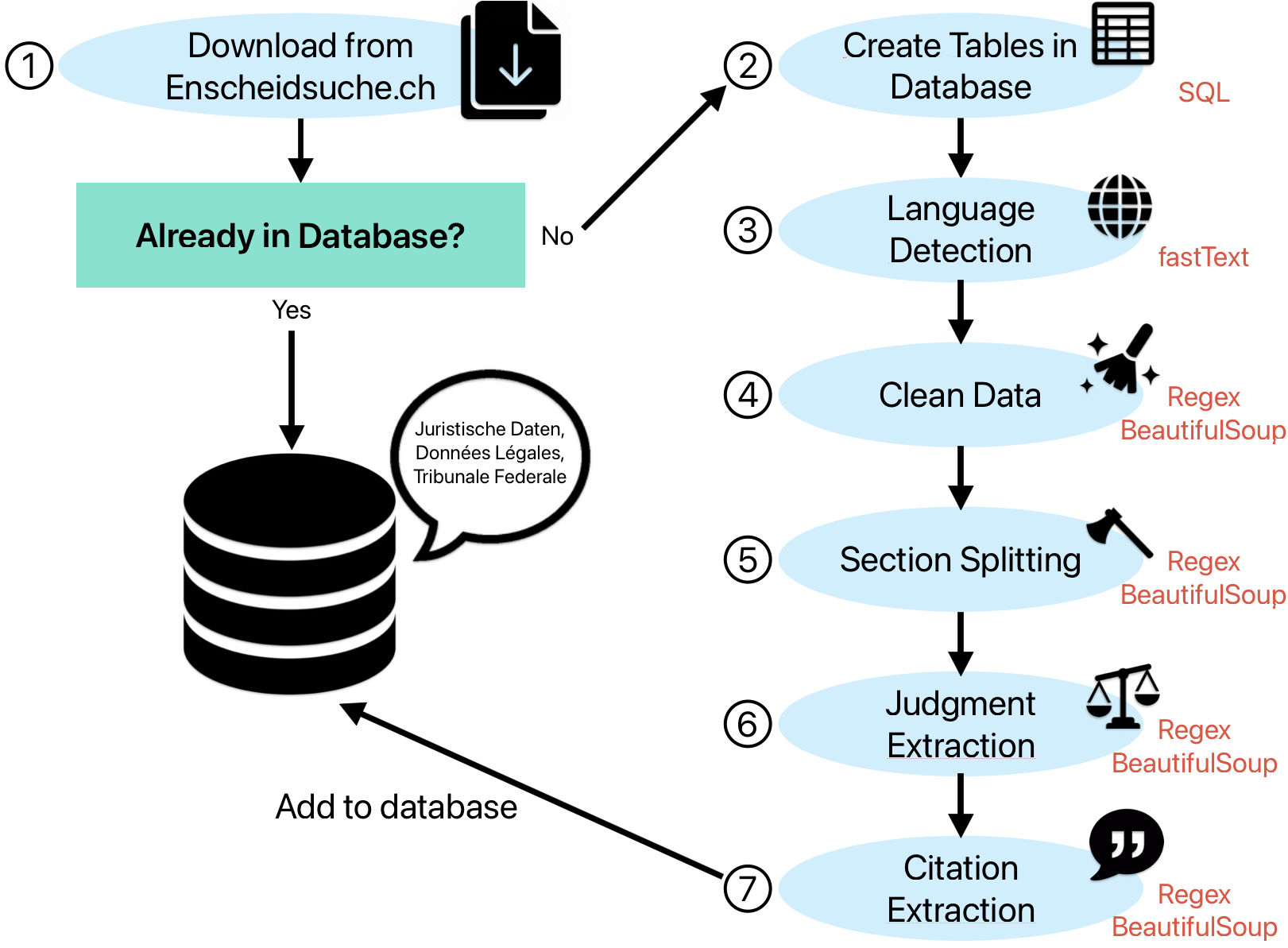}
    \caption{Database Creation Pipeline}
   \label{fig:flow}
   \vspace{-5ex}
\end{wrapfigure}

Every day, new cases are published on Entscheidsuche.ch, allowing for daily document retrieval (see \Cref{fig:flow}). \textbf{(1)} We scrape all files from Entscheidsuche.ch, including each court's folder metadata. Only new case documents are sent through the pipeline.
\textbf{(2)} We used BeautifulSoup / tika-python library to extract text from HTML / PDF files. 
\textbf{(3)} We detect the language using fastText \citep{grave2018learning} for subsequent tasks. \textbf{(4)} A cleaner removes irregular patterns or redundant text to avoid extraction errors. \textbf{(5)} Cases are segmented into header, facts, considerations, rulings, and footer via regex patterns.
\textbf{(6)} To extract the judgement outcome, a word set is defined for each outcome. As these indicators are not context-exclusive, considering only the ruling section is crucial to avoid false positives. Therefore, accurate judgment outcome extraction relies on precise section splitting.
\textbf{(7)} \ac{LD} and law citations are obtained through regex (cantonal) or BeautifulSoup (federal). The \ac{SFCS} labels citations with HTML tags, leading to a high quality of citations for federal cases in our datasets.

Providing an objective metric for quality is hard and expensive to obtain. Multiple people repeated quality checks over multiple months during this process to ensure the highest quality. The parsers and regexes were double-checked by senior authors before integration. We wrote a series of tests to make sure that the pipeline is robust to changes (test\_utils.py).
Finally, we wrote code to easily inspect samples at various stages of the pipeline to ensure quality (debug\_utils.py).

\subsection{Pretraining}
\label{sec:pretraining-datasets}
We release two large datasets for pretraining, used for our Swiss legal models: one containing legislation (116M words), the other containing court rulings (2.1B words). 

\paragraph{Rulings}
The Swiss Rulings dataset is a comprehensive collection of Swiss court rulings designed for pretraining purposes. It consists of 638K cases (3.3B tokens) distributed across three languages: German (319K), French (247K) and Italian (71K) (see \Cref{fig:rulings_lang}). Spanning several decades and covering multiple areas of law, this dataset provides an extensive representation of Swiss law practice. Figures~\ref{fig:rulings_lang} and \ref{fig:rulings_text_length_distribution} provide an overview of the distributions of cantons and of text lengths.

\paragraph{Legislation}
The Swiss Legislation dataset comprises 35.7K legislative texts (182M tokens). \Cref{fig:legislation_text_length_distribution} shows the length distribution of the legislation texts. The legislation texts cover five languages: German, French, Italian, Romansh, and English (see \Cref{fig:legislation_lang}). \Cref{fig:legislation_canton} details its coverage of federal, cantonal, and inter-cantonal legislation on a broad array of legal topics including public health, education, civil rights, societal matters, energy, environment, infrastructure, and visa regulations. It also includes instances of the same legislation texts across different languages, useful for enhancing the multilingual capabilities of legal \acp{LM}.\\


\begin{minipage}{\textwidth}
\begin{framed} 
\begin{figure}[H]
    \centering
    \label{fig:pretraining_corpora}
    \vspace{-4ex}
    \begin{minipage}[t]{\textwidth}
        \textbf{Pre-training Corpora} 
        \vspace{-1ex}\\
        \rule{\textwidth}{0.4pt}
        A large corpus of high-quality domain-specific text is crucial for training LLMs capable of performing tasks in a given domain. This dataset collects a large part of publicly available legal text relevant for Switzerland.
        \vspace{1ex}
    \end{minipage}
    \rule{\textwidth}{0.4pt}
\end{figure}

\begin{figure}[H]
    \centering
    \begin{minipage}[t]{0.60\textwidth}
        \vspace{-7ex} 
        \textbf{Legislation text:} \\
        \tiny{
            Der Grosse Rat des Kantons Aargau, gest\"{u}tzt auf die \textsection{}\textsection{} 72 Abs. 3 und 78 Abs. 1 der Kantonsverfassung, beschliesst: \par 1. Allgemeine Bestimmungen  \par \textsection{} Gegenstand und Zweck \par 1 Dieses Gesetz regelt a) die amtliche Information der \"{O}ffentlichkeit und den Zugang zu amtlichen Dokumenten [...] \par
            \textsection{} 15 Bekanntgabe an Private \par \"{O}ffentliche Organe geben Privaten Personendaten nur bekannt, wenn \par a) sie dazu gesetzlich verpflichtet sind, oder  \par b) die Bekanntgabe n\"{o}tig ist, um eine gesetzliche Aufgabe [...]
        }\\
       \vspace{-3ex}\\
    \end{minipage}
    \hfill
    \begin{minipage}[t]{0.35\textwidth}
        \vspace{-7ex} 
        \textbf{Metadata:} \\
        \tiny{
            UUID: \textit{58450ad4-108d-4e10-b559-a7efece689d7} \\
            Year: \textit{2015} \\
            Language: \textit{German} \\
            Canton: \textit{AG} \\
            Title: \textit{Gesetz über die Information der Öffentlichkeit, den Datenschutz und das Archivwesen} \\
            Abbreviation: \textit{IDAG} \\
            SR Number: \textit{150.700}
        }
    \end{minipage}
    \rule{\textwidth}{0.4pt}
\end{figure}
\begin{figure}[H]
    \centering
    \vspace{-6ex}
    \begin{minipage}[t]{0.49\textwidth}
       \includegraphics[width=\textwidth]{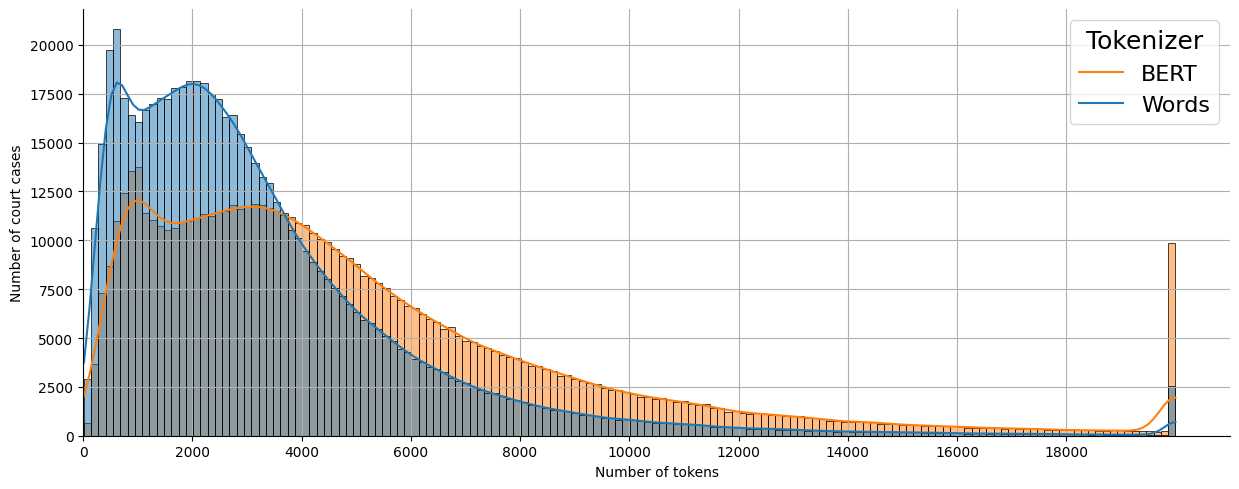}  
        \caption{{Rulings text length distribution}}
        \label{fig:rulings_text_length_distribution}
    \end{minipage}
    \hfill
    \begin{minipage}[t]{0.49\textwidth}
            \includegraphics[width=\textwidth]{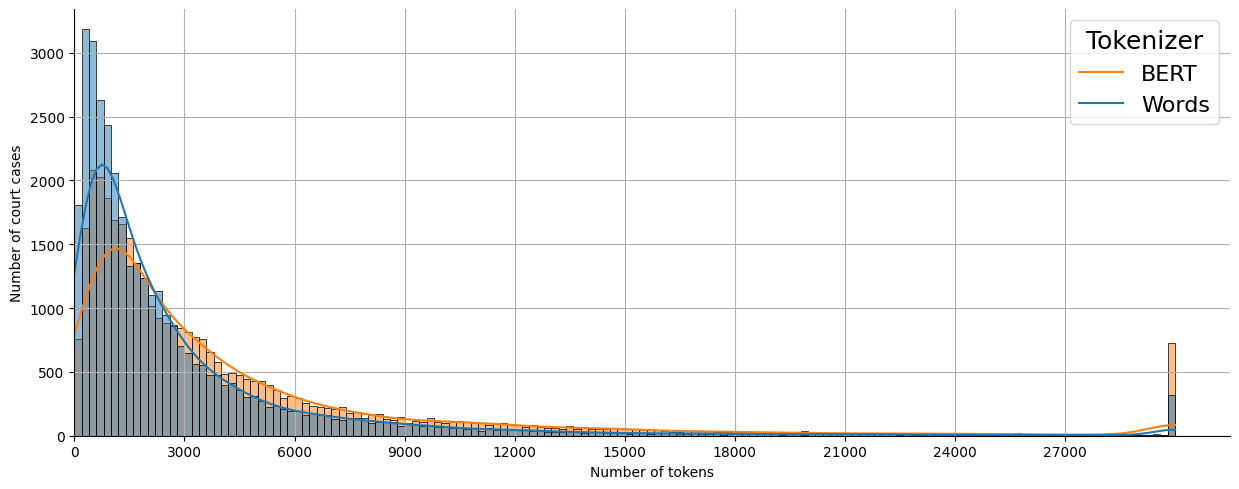}  
        \caption{{Legislation text length distribution}}
        \label{fig:legislation_text_length_distribution}
    \end{minipage}
    \rule{\textwidth}{0.4pt}
\end{figure}

\begin{figure}[H]
    \centering
    \vspace{-6ex} 
    \begin{minipage}[t]{0.49\textwidth}
        \includegraphics[width=\textwidth]{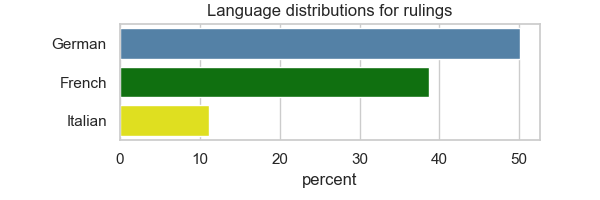}
        \caption{Language dist. rulings texts}
        \label{fig:rulings_lang}
    \end{minipage}
    \hfill
    \begin{minipage}[t]{0.49\textwidth}
        \includegraphics[width=\textwidth]{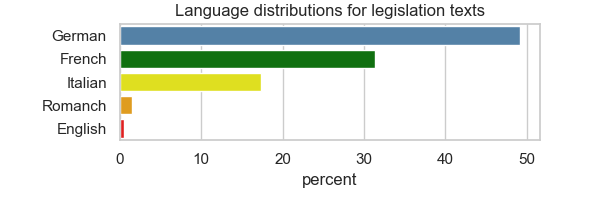}
        \caption{{Language dist. legislation texts}}
        \label{fig:legislation_lang}
    \end{minipage}
\end{figure}

\begin{figure}[H]
    \centering
    \vspace{-5ex} 
    \begin{minipage}[t]{0.49\textwidth}
        \includegraphics[width=\textwidth]{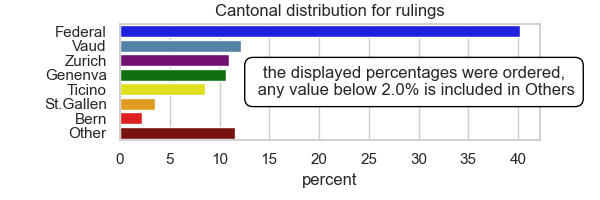}
        \vspace{-2ex}
        \caption{{Cantonal dist. rulings texts}}
        \label{fig:rulings_canton}
    \end{minipage}
    \hfill
    \begin{minipage}[t]{0.49\textwidth}
        \includegraphics[width=\textwidth]{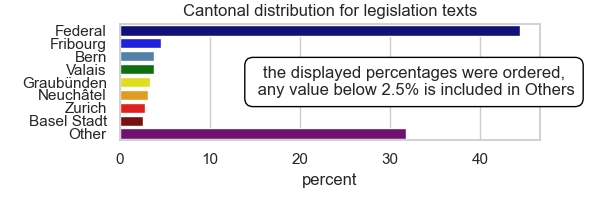}
        \vspace{-2ex}
        \caption{{Cantonal dist. legislation texts}}
        \label{fig:legislation_canton}
    \end{minipage}
    \vspace{-5ex}
\end{figure}
\end{framed}
\end{minipage}

\subsection{Text Generation}
\paragraph{Court View Generation}
\label{sec:cvg}

Clerks and judges spend about 50\% of their time on writing considerations in penal law and an estimated 85\% in other legal areas \citep{niklaus_swiss-judgment-prediction_2021}. Crafting considerations is central to a judge's role, requiring deep knowledge of legislation, caselaw, legal analysis, and advanced reasoning to synthesize this information. The task's complexity, particularly in the Supreme Court, manifests in the high average appointment age of judges of 50 years, highlighting the length and difficulty of their professional journey.\footnote{\tiny{\emph{Mit 28 Jahren Mitglied des Bundesgerichts}, SonntagsZeitung, December 19, 2019, p. 24}}
Given the time and expertise demands, the need for the \ac{CVG} task arises, targeting the generation of case considerations from facts.
Generating court views is challenging due to the length and complexity of both facts (input) and considerations (output). Current models, limited in handling long context, frequently fail to process this extensive input. This limitation, combined with the input's complexity, highlights the current models' deficiencies.
To benchmark models in long-context complex reasoning, we introduce a novel \ac{CVG} dataset with over 400K cases spanning various legal scenarios. Averaging 1522 tokens for facts and 4673 for considerations, the dataset serves as a challenging benchmark for generating coherent, accurate case considerations.
Additionally, we present a court view origin dataset with federal rulings, augmented by lower court data, including facts and considerations from both levels. This adds a multilevel judicial perspective, enhancing understanding of case progression and increasing the \ac{CVG} challenge. \Cref{fig:cvg_facts_length_distribution} and \Cref{fig:cvg_considerations_length_distribution} show the length distributions for facts and considerations of the CVG dataset respectively.\\

\begin{minipage}{\textwidth}
\begin{framed}
\begin{figure}[H]
\ContinuedFloat
    \centering
    \label{tab:CVG_task}
    \vspace{-4ex}
    \begin{minipage}[t]{\textwidth}
        \textbf{Illustration of the \acf{CVG} task} \\
        \vspace{-4ex}\\
        \rule{\textwidth}{0.4pt}
        Court view generation is arguably one of the most difficult NLP tasks. It requires extensive legal reasoning capabilities, significant experience, and good knowledge of the specific law area a judge is operating in. Machines may assist judges and clerks by suggesting word or sentence continuations while they are typing or even by setting up complete drafts. To solve this task well, the following ingredients are likely necessary: First, a strong retrieval system capable of providing the necessary legal context based on legislation and influential previous rulings. Second, a strong legal reasoning system capable of analyzing the facts, any lower court decisions, and the retrieved documents. This matter is complicated further in our dataset due to lengthy documents in multiple languages.
    \vspace{1ex}
    \end{minipage}
    \rule{\textwidth}{0.4pt}
\end{figure}

\begin{figure}[H]
    \ContinuedFloat
    \centering
    \begin{minipage}[t]{0.74\textwidth}
        \vspace{-7ex} 
        \textbf{Input}\\
        \tiny{
           [Facts]:
            Zum Sachverhalt:
            1. Am Dienstag, 23. August 2005, um 12.16 Uhr, lenkte X seinen Personenwagen von
            Bronschhofen her kommend auf der Hauptstrasse in Richtung Wil. Auf der Höhe
            Hauptstrasse 64 wurde er von der Kantonspolizei St. Gallen anlässlich einer
            Geschwindigkeitskontrolle innerorts mit einer Geschwindigkeit von 80 km/h gemessen. Nach Abzug der technisch bedingten Sicherheitsmarge von 5 km/h resultierte eine rechtlich relevante Geschwindigkeit von 75 km/h.
            2. Mit Strafbescheid des Untersuchungsamtes Gossau wurde der Angeklagte am 10. Mai 2006 wegen grober Verletzung der Verkehrsregeln zu einer Busse von Fr. 610.00 verurteilt. Dagegen erhob er Einsprache. Der Einzelrichter des Kreisgerichtes Alttoggenburg-Wil verurteilte ihn mit Urteil vom 14. September 2006 wegen grober Verkehrsregelverletzung und fällte eine Busse von Fr. 600.00 aus. Für die Löschung im Strafregister wurde eine Probezeit von zwei Jahren angesetzt, die Kosten des Verfahrens wurden dem Angeklagten auferlegt. [....]
            \\
        }
    \vspace{-4ex}\\
    \end{minipage}
    \hfill
    \begin{minipage}[t]{0.23\textwidth}
        \vspace{-7ex}
        \textbf{Metadata:} \\
        \tiny{
            Decision ID: \textit{0f86bb1e-ed24-52a1-bec7-e04451485a7f}\\
    		Year: \textit{2007}\\
            Language: \textit{German}\\
            Law Area: \textit{Penal}\\
            Court: \textit{SG\_KG}\\
            Chamber: \textit{SG\_KG\_001,}\\
            Canton: \textit{SG}\\
            Region: \textit{Eastern Switzerland}
        }
    \end{minipage}
    \rule{\textwidth}{0.4pt}
\end{figure}

\begin{figure}[H]
    \ContinuedFloat
    \centering
    \begin{minipage}[t]{\textwidth}
    \vspace{-7ex} 
     \textbf{Target}:
        \tiny{
        [Considerations]:
        Aus den Erwägungen:
        1. Nach Art. 90 Ziff. 2 SVG wird mit Freiheitsstrafe bis zu drei Jahren oder Geldstrafe bestraft, wer [...] Der subjektive Tatbestand der groben Verkehrsregelverletzung ist hier deshalb regelmässig zu bejahen. Eine Ausnahme kommt etwa da in Betracht, wo [...]
        2. Der Angeklagte bringt vor, die Vorinstanz habe den Grundsatz in dubio pro reo verletzt, wenn [...]
        Indem der Angeklagte innerorts mit mindestens 25 km/h zu schnell gefahren ist, hat er
        den objektiven Tatbestand der groben Verkehrsregelverletzung erfüllt.
        […] Aus dem gleichen Grund ist auch der Beweisantrag zur Vornahme eines Augenscheins abzuweisen.
        III. 
        1. Der Angeklagte hat eine grobe Verkehrsregelverletzung begangen. Sein Verschulden wiegt schon deshalb nicht mehr leicht, weil […], so
        erscheint eine Geldstrafe von 4 Tagessätzen angemessen (Art. 34 i.V.m. Art. 47 StGB).
        […] 
        }\\
        \vspace{1ex}
        \rule{\textwidth}{0.4pt}
    \end{minipage}
\end{figure}

\begin{figure}[H]
    \ContinuedFloat
    \centering
    \vspace{-6ex}
    \begin{minipage}[b]{0.49\textwidth}
       \includegraphics[width=\textwidth]{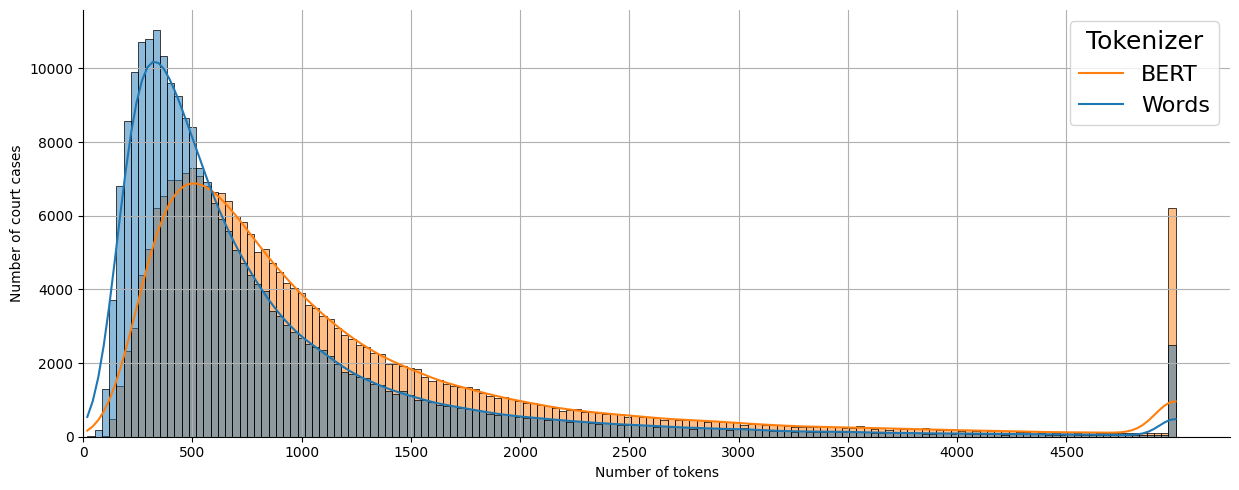}
        \captionof{figure}{CVG facts length distribution}
        \label{fig:cvg_facts_length_distribution}
    \end{minipage}
    \hfill
    \begin{minipage}[b]{0.49\textwidth}
        \includegraphics[width=\textwidth]{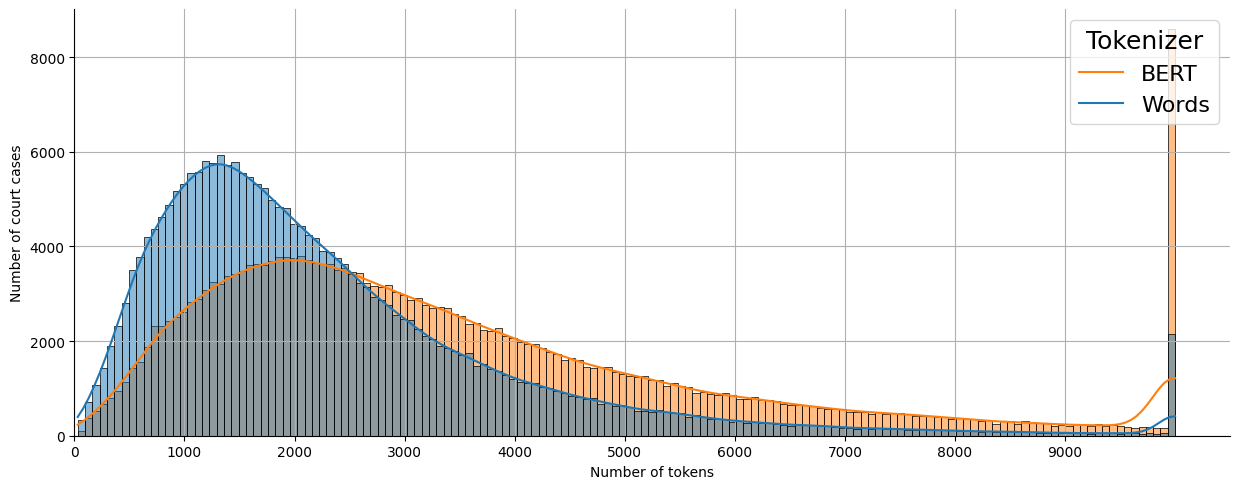}
        \captionof{figure}{CVG considerations length dist.}
        \label{fig:cvg_considerations_length_distribution}
    \end{minipage}
    \vspace{-5ex}
\end{figure}
\end{framed}
\end{minipage}

\paragraph{Leading Decision Summarization}
\label{sec:lds}

\ac{LD} are crucial in the Swiss legal system, often cited to clarify legislative gaps. Access to their summaries simplifies searching and understanding key concepts, the most important citations, and main themes. In the \ac{LDS} dataset, we include 18K \ac{LD}s with their summaries, written by \ac{SFCS} clerks and judges. The summaries contain three parts: 1) a short list of important citations, 2) keywords from a thesaurus, and 3) a free-form summary containing references to the most important numbered considerations. Figures \ref{fig:lds_input_length_distribution} and \ref{fig:lds_summary_length_distribution} show the length distributions for the input text and summary of the LDS dataset respectively.\\

\begin{minipage}{\textwidth}
\begin{framed}
\begin{figure}[H]
\ContinuedFloat
    \renewcommand{\arraystretch}{1.5} 
    \centering
    \label{tab:LDS_task}
    \vspace{-4ex}
    \begin{minipage}[t]{\textwidth}
        \textbf{Illustration of the \acf{LDS} task} \\
        \vspace{-4ex}\\
        \rule{\textwidth}{0.4pt}
        Summarizing cases is very important for lawyers to absorb the most relevant information in less time. Lawyers need to read many cases during their research. Reducing the time needed to comprehend the gist of a case brings direct economic value.
        \vspace{1ex}
    \end{minipage}
    \rule{\textwidth}{0.4pt}
\end{figure}

\begin{figure}[H]
    \ContinuedFloat
    \centering    
    \begin{minipage}[t]{0.74\textwidth}
        \vspace{-7ex} 
        \textbf{Input}\\
        \tiny{
           [Case Text]:
    BGE 141 IV 201 S. 201
    Dai considerandi:
    8.
    8.2.1 
    \`{E} stato accertato, senza arbitrio, che la ricorrente ha pi\`{u} volte chiesto a F. di trovare, nel senso di contattare e ingaggiare (avendo precisato che aveva i soldi per pagare), qualcuno che potesse uccidere il marito e che egli rifiut\`{o} di fare quello che gli si domandava.
    8.2.2 
    Contrariamente a quanto sostenuto nel gravame, la contestata richiesta risulta tutt'altro che generica: permetteva di ben comprendere sia il genere di infrazione finale prospettata (reato contro la vita) sia la vittima designata sia il comportamento da assumere, ossia reperire e ingaggiare qualcuno allo scopo, atteso che vi era a disposizione denaro. F. non si \`{e} risolto a commettere alcunch\'{e}, motivo per cui si \`{e} di fronte solo a un tentativo di istigazione e la questione del nesso causale tra l'atto di persuasione e la decisione dell'istigato di commettere il reato neppure si pone.
    [...]
    }\\
    \vspace{-3ex}\\
    \end{minipage}
    \hfill
    \begin{minipage}[t]{0.22\textwidth}
    \vspace{-7ex}
        \textbf{Metadata:} \\
        \tiny{
            Decision ID:\\ \textit{91ae0d9f-9aec-4b2b-a7ee-042abc42adaa}\\
            Year: \textit{2015}\\
            Language: \textit{Italian}\\
            Court: \textit{CH\_BGE}\\
            Chamber: \textit{CH\_BGE\_006,}\\
            Canton: \textit{CH}\\
            Region: \textit{Federation}
        }
    \end{minipage}
    \rule{\textwidth}{0.4pt}

    \begin{minipage}[t]{\textwidth}
    \textbf{Target}:
        \tiny{[Regeste]: Regeste Art. 24 Abs. 2 StGB; indirekte Anstiftung (Kettenanstiftung), Versuch. Auch die versuchte indirekte Anstiftung (Kettenanstiftung) zu einem Verbrechen ist strafbar (E. 8.2.2).}\\
        \rule{\textwidth}{0.4pt}
    \end{minipage}
\end{figure}

\begin{figure}[H]
    \ContinuedFloat
    \centering
    \vspace{-7ex}
    \begin{minipage}[b]{0.49\textwidth}
       \includegraphics[width=\textwidth]{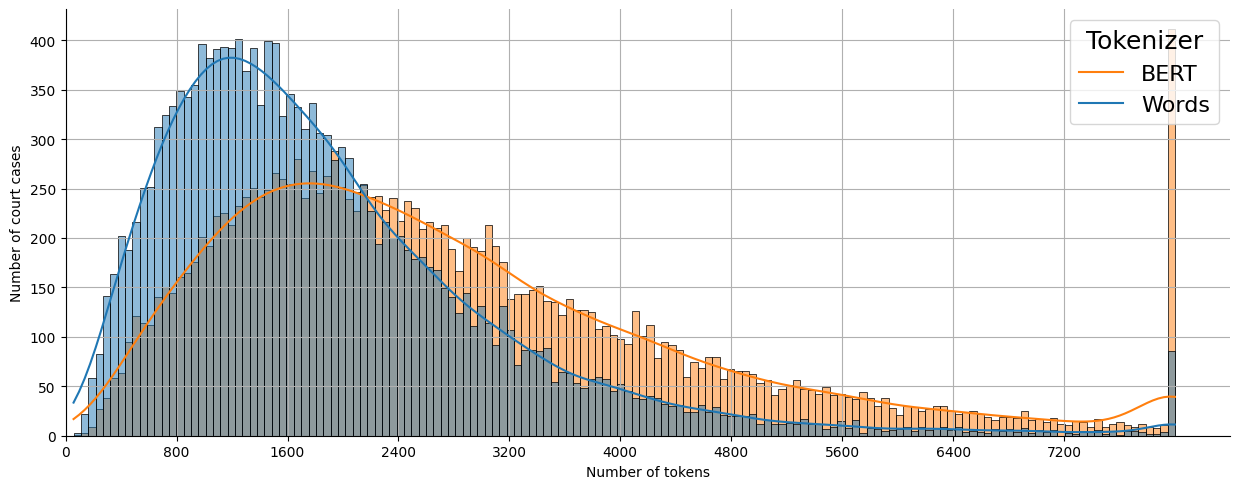}  
        \captionof{figure}{LDS input length distribution}
        \label{fig:lds_input_length_distribution}
    \end{minipage}
    \hfill
    \begin{minipage}[b]{0.49\textwidth}
        \includegraphics[width=\textwidth]{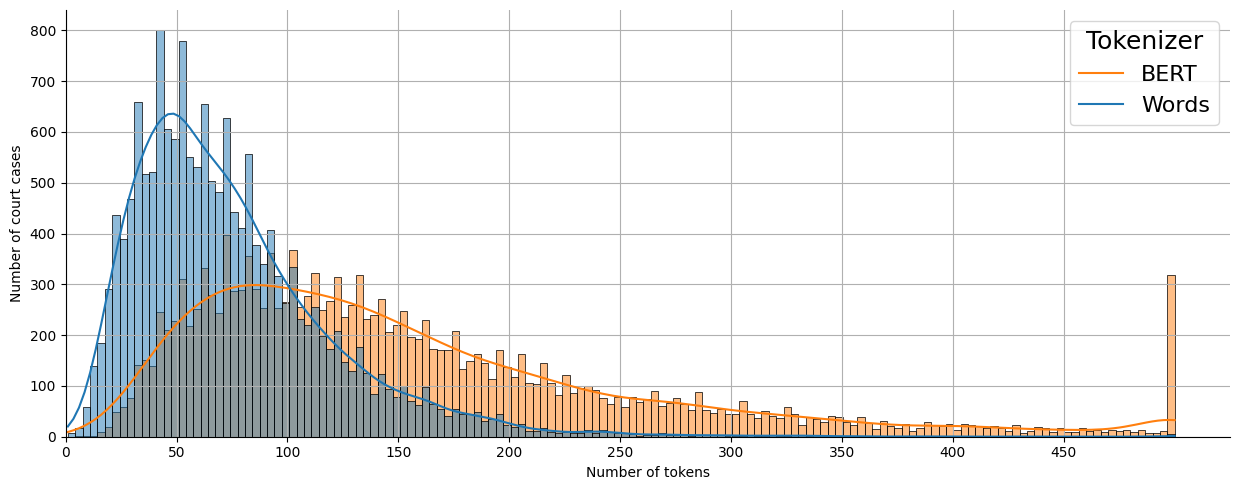}  
        \captionof{figure}{LDS summary length distribution}
        \label{fig:lds_summary_length_distribution}
    \end{minipage}
    \vspace{-5ex}
\end{figure}
\end{framed}
\end{minipage}

\subsection{Text Classification}

We use eight configurations from the \acf{LAP}, \acf{CP}, and \acf{JP} datasets. \ac{LAP} and \ac{JP} datasets include federal and cantonal cases, while the \ac{CP} dataset focuses on \ac{SFCS} cases. All tasks presented in this section involve \ac{SLTC}, which required either extracting or defining labels. For each of the tasks, we consider the facts and the considerations as input. The facts represent the most similar available proxy to the complaints, useful for predictive tasks. The considerations as input make the tasks considerably easier, since they include the legal reasoning. These tasks can be used as post-hoc analyses for verification (e.g., in \ac{JP} whether the judgment is congruent with the given reasoning).

\paragraph{Law Area Prediction}
\label{sec:lap}

The \textbf{Law Area} label was established by associating a law area to each chamber where a case was adjudicated. Using metadata from Entscheidsuche.ch, a lawyer helped define the law areas for each chamber, resulting in chambers being classified into one of four main law area categories (civil, public, criminal and social law) and 12 sub-areas. Due to many chambers operating in various law areas, it was not always feasible to assign a single law area label to each chamber. Particularly for the more detailed sub-areas, where several chambers could not be uniquely linked, this resulted in a small subset of cases with the subset label. Initial results on the full dataset including the four main law areas showed that current models achieve near perfect accuracy, which is why we only consider the smaller filtered dataset of sub-areas for this benchmark. \Cref{fig:lap_facts_length_distribution} shows the length distribution for the facts of the LAP dataset.\\

\begin{minipage}{\textwidth}
\begin{framed}
\begin{figure}[H]
    \centering
    \label{tab:SLAP_task}
    \vspace{-4ex}
    \begin{minipage}[t]{\textwidth}
        \textbf{Illustration of the \acf{SLAP} task} 
        \vspace{-1ex}\\
        \rule{\textwidth}{0.4pt}
        Before the judge even sees a complaint, it is first handled by the court's administrative staff, deciding to which chamber (suborganisation inside the court hearing matters in a specific subpart of the law) the complaints should be routed. For this task, models trained on a dataset like ours could assist by providing a suggestion.
        \vspace{1ex}
    \end{minipage}
    \rule{\textwidth}{0.4pt}
\end{figure}

\begin{figure}[H]
    \ContinuedFloat
    \centering
    \begin{minipage}[t]{0.72\textwidth}
    \vspace{-7ex} 
    \textbf{Input}\\
        \tiny{
           [Facts]:
    I. Faits\
    1. En date du 11 novembre 2013, l'intim\'{e} a d\'{e}pos\'{e} \`{a} la Commune de Corcelles une
    demande de permis de construire pour la pose d'un rev\^{e}tement bitumineux sur l'acc\`{e}s \`{a}
    son immeuble, le prolongement d'un chemin existant et l'installation d'une piscine sur les
    parcelles n\textdegree{} C.\_ et D.\_ du registre foncier de la commune de Corcelles.
    Les parcelles se situent en zone agricole. Le recourant a form\'{e} opposition contre ce projet
    de construction. Dans sa d\'{e}cision globale du 1er d\'{e}cembre 2014, la Pr\'{e}fecture du Jura
    bernois a accept\'{e} la demande d'octroi du permis de construire.
    2. Le 31 décembre 2014, le recourant a déposé un recours contre cette décision auprès
    de la Direction des travaux publics, des transports et de l'énergie du canton de Berne
    (TTE). Il fait valoir, en substance, que différentes conditions de la décision globale du 1er
    décembre 2014 n'auraient pas été respectées.[....]
    }
    \vspace{-1ex}\\
    \end{minipage}
    \hfill
    \begin{minipage}[t]{0.25\textwidth}
        \vspace{-7ex} 
        \textbf{Metadata:} \\
        \tiny{
            Decision ID: \textit{519d0350-6e0e-5551-9bc9-1df033382168}\\
            Year: \textit{2015}\\
            Language: \textit{French}\\
            Law Area: \textit{Public}\\
            Law Sub Area: \textit{Urban Planning and Environmental}\\
            Court: \textit{BE\_VB}\\
            Chamber: \textit{BE\_VB\_001,}\\
            Canton: \textit{BE}\\
            Region: \textit{Espace Mittelland}\\
        }
    \end{minipage}
    \rule{\textwidth}{0.4pt}
 \end{figure}

 \begin{figure}[H]
    \centering
    \begin{minipage}[t]{\textwidth}
    \vspace{-7ex} 
    \textbf{Target}:
    \tiny{Urban Planning and Environmental}\\
    \tiny{
        Possible Targets: Tax, Urban Planning and Environmental, Expropriation, Public Administration, Other Fiscal, Rental and Lease, Employment Contract, Bankruptcy, Family, Competition and Antitrust, Intellectual Property, Substantive Criminal, Criminal Procedure
        }
        \vspace{-0ex}\\
        \rule{\textwidth}{0.4pt}
    \end{minipage}
    \begin{minipage}[t]{0.6\textwidth}
        \vspace{0ex}
        \includegraphics[width=\textwidth]{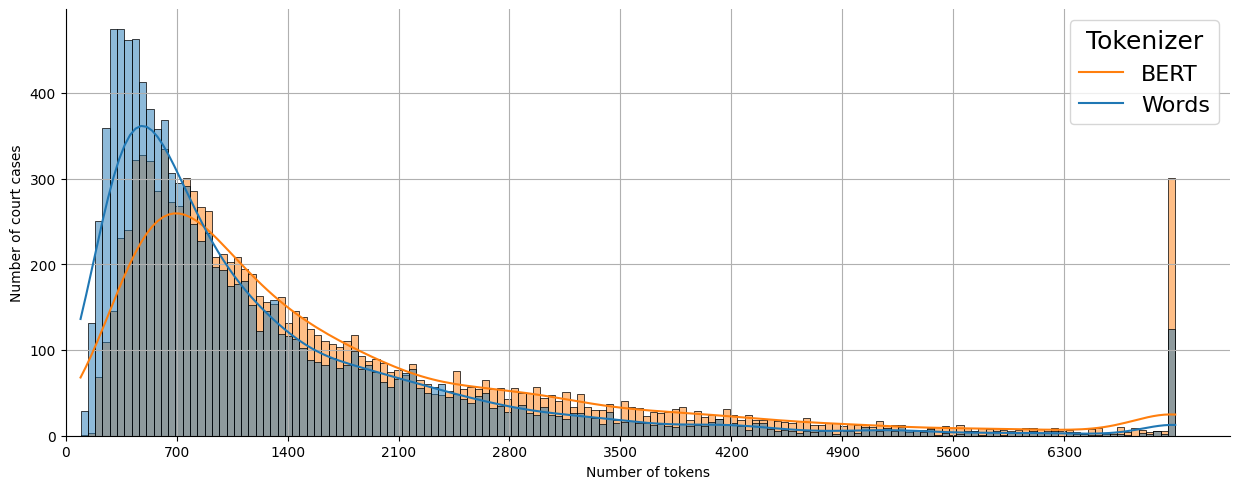}
        \caption{Law Area Prediction facts length distribution}
        \label{fig:lap_facts_length_distribution}
    \end{minipage}
    \vspace{-6ex}
\end{figure}
\end{framed}
\end{minipage}

\paragraph{Judgment Prediction}
\label{sec:jp}

We generated the \textbf{Judgment} label by using regex patterns to extract the judgment outcome, assigning it a binary label: approval or dismissal, akin to \citet{niklaus_swiss-judgment-prediction_2021}. Partially approved or dismissed judgments were labeled as either approval or dismissal. Figure \ref{fig:jp_facts_length_distribution} shows the length distribution for the facts of the JP dataset.\\

\begin{minipage}{\textwidth}
\begin{framed}
\begin{figure}[H]
\ContinuedFloat
    \renewcommand{\arraystretch}{1.5} 
    \centering
    \label{tab:JP_task}
    \vspace{-4ex}
    \begin{minipage}[t]{\textwidth}
        \textbf{Illustration of the \acf{JP} task} \\
        \vspace{-4ex}\\
        \rule{\textwidth}{0.4pt}
        Judgment Prediction might be used in the future in jurisdictions that are experiencing extremely high case loads such as US immigration. The legal maxim "justice delayed is justice denied" may provide motivation for judgment prediction being applied in such highly overloaded jurisdictions by giving affected people the opportunity to have their case heard much earlier (in some jurisdictions wait times are \href{https://trac.syr.edu/phptools/immigration/court_backlog/apprep_backlog_avgdays.php}{years}). For example, consider a scenario where a person is held in pretrial detention awaiting their case to be heard. With Judgment Prediction, it may be possible to identify cases where there is a high likelihood of the individual being not guilty. These cases can then be prioritized for judicial review, potentially reducing the time innocent individuals spend in detention due to system backlogs.
        \vspace{1ex}
    \end{minipage}
    \rule{\textwidth}{0.4pt}
\end{figure}

\begin{figure}[H]
    \ContinuedFloat
    \centering    
    \begin{minipage}[t]{0.72\textwidth}
    \vspace{-7ex} 
        \textbf{Input}\\
        \tiny{
           [Facts]: En fait :
            A.
            Le 5 f\'{e}vrier 2015 \`{a} 21 h 30, \`{a} [...],A.H.\_ a \'{e}t\'{e} appr\'{e}hend\'{e} par la police, qui l'a entendu le lendemain vers 0 h 45 comme pr\'{e}venu notamment d'infraction \`{a} la LStup (Loi f\'{e}d\'{e}rale sur les stup\'{e}fiants ; RS 812.121). L'int\'{e}ress\'{e} a d\'{e}clar\'{e} qu'alors qu'il se trouvait dans un bar \`{a} [...], un homme s'\'{e}tait assis \`{a} c\^{o}t\'{e} de lui et lui avait demand\'{e} de la coca\"{\i}ne. Le pr\'{e}venu s'\'{e}tait rendu dans l'appartement occup\'{e} notamment par B.H.\_, un compatriote qui l'h\'{e}bergeait \`{a} l'occasion, pour prendre une boulette de coca\"{\i}ne, qu'il avait ensuite vendue \`{a} l'inconnu pour 100 francs. Les policiers lui avaient finalement indiqu\'{e} que le client en question \'{e}tait en r\'{e}alit\'{e} un agent de police en civil.
            Sur la base d'indications fournies par A.H.\_ au client lors de cette transaction, l'appartement occup\'{e} par B.H.\_, rue de [...] \`{a} [...], avait fait l'objet, la veille vers 22 h 45, d'une perquisition qui avait amen\'{e} la d\'{e}couverte de 29.8 g de coca\"{\i}ne et de plusieurs t\'{e}l\'{e}phones portables. [....]
            }
    \vspace{-1ex}\\
    \end{minipage}
    \hfill
    \begin{minipage}[t]{0.25\textwidth}
        \vspace{-7ex} 
        \textbf{Metadata:} \\
        \tiny{
            Decision ID: \textit{0dd2f9f7-872e-4200-9f9c-f1c12520c267}\\
    		Year: \textit{2015}\\
            Language: \textit{French}\\
            Law Area: \textit{Penal}\\
            Judgment: \textit{Dismissal}\\
            Court: \textit{VD\_TC}\\
            Chamber: \textit{VD\_TC\_013,}\\
            Canton: \textit{VD}
            Region: \textit{Région lémanique}
            }
    \end{minipage}
    \rule{\textwidth}{0.4pt}
\end{figure}

\begin{figure}[H]
    \ContinuedFloat
    \centering
    \begin{minipage}[t]{\textwidth}
        \vspace{-10ex} 
        \textbf{Target}:
        \tiny{\textbf{Dismissal}}\\
        \tiny{Possible SP label: Approval, Dismissal}\\
        \rule{\textwidth}{0.4pt}\\
        \vspace{-12ex}\\
    \end{minipage}
\end{figure}

\begin{figure}[H]
    \centering
    \ContinuedFloat
    \begin{minipage}[t]{0.69\textwidth}
    \vspace{-12ex} 
        \includegraphics[width=\textwidth]{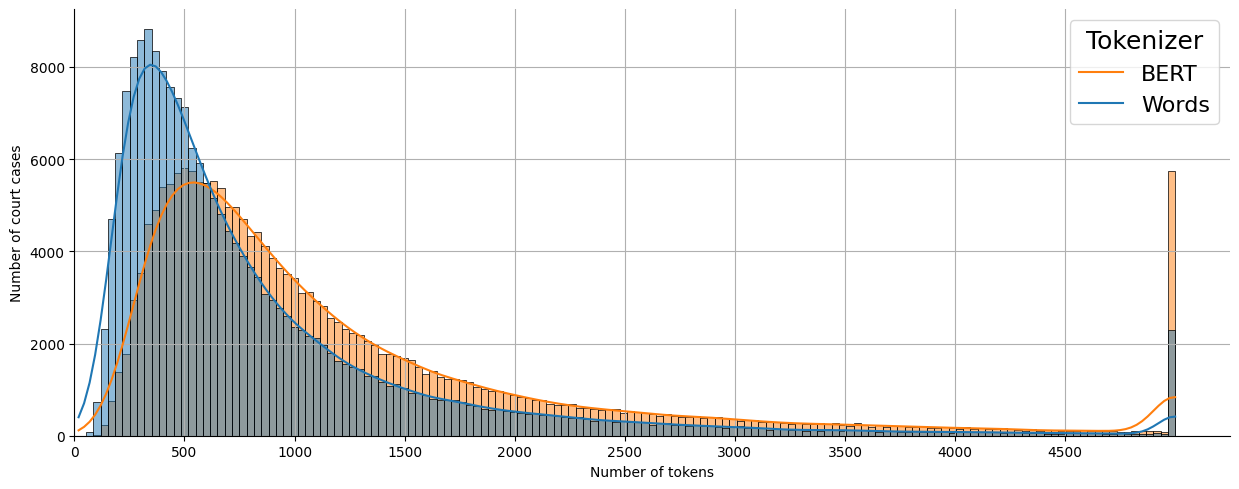}
        \caption{Judgment Prediction facts length distribution}
        \label{fig:jp_facts_length_distribution}
    \end{minipage}
    \vspace{-6ex}
\end{figure}
\end{framed}
\end{minipage}

\paragraph{Criticality Prediction}
\label{sec:cp}
First and foremost, it's essential to define the concept of criticality within the Swiss legal framework. Critical cases go beyond mere importance; they are cases with the potential to significantly influence future jurisprudence. It's therefore crucial to distinguish them from merely important cases.
We quantified \textbf{Criticality} in two ways:
First, the \textbf{LD-Label} is binary: \emph{critical} and \emph{non-critical}. \ac{SFCS} cases are labeled as \emph{critical} if additionally published as \acl{LD} (see \Cref{background-legal-system}). We extracted \ac{SFCS} file names from \ac{LD} headers using regex. Cases absent in \ac{LD} headers are labeled \emph{non-critical} (missing \emph{critical} cases exist as we have all \ac{LD} but not all \ac{SFCS} cases).
Second, to create a more precise adaptation of the LD-Label, we developed the \textbf{Citation-Label}, which involved counting all citations of \ac{LD} in all \ac{SFCS} cases. The \ac{LD} frequency was weighted based on recency, with older citations receiving a smaller weight: \begin{math}score=count*\frac{year-2002+1}{2023-2002+1}\end{math}. 
This resulted in a ranking of \acp{LD}, which were then divided into four categories of criticality \emph{critical-1} to \emph{critical-4}. We used the 25, 50 and 75\% quartiles as separation for our four classes. 
\Cref{fig:cp_facts_length_distribution} and \Cref{fig:cp_considerations_length_distribution} show the length distributions for the facts and the considerations of the \ac{CP} dataset respectively.\\

\begin{minipage}{\textwidth}
\begin{framed}
\begin{figure}[H]
\ContinuedFloat
    \centering
    \label{tab:CP_task}
    \vspace{-4ex}
    \begin{minipage}[t]{\textwidth}
        \textbf{Illustration of the \acf{CP} task} \\
        \vspace{-4ex}\\
        \rule{\textwidth}{0.4pt}
        We see two potential applications of the Criticality Prediction task: Prioritization and Classification. The prioritization task takes as input the facts as a proxy for a complaint and produces a prioritization score judging how critical/important this case is. This prioritization might help decide which cases should be heard earlier or by more experienced judges. In a futuristic scenario where automatic judgment prediction is accepted, cases with low priority could be sent to automated solutions, while high priority cases would be sent to human judges. The classification task takes as input the considerations or the entire ruling and performs a post-hoc analysis comparing it to prior caselaw and judging its potential impact on future jurisprudence. 
        \vspace{1ex}
    \end{minipage}
    \rule{\textwidth}{0.4pt}
\end{figure}

\begin{figure}[H]
    \ContinuedFloat
    \centering
    \begin{minipage}[t]{0.74\textwidth}
        \vspace{-7ex} 
        \textbf{Input}\\
        \tiny{
           [Consideraions]: Erw\"{a}gungen:
        1. Angefochten ist der in einem kantonal letztinstanzlichen Scheidungsurteil festgesetzte nacheheliche Unterhalt in einem Fr. 30'000.-- \"{u}bersteigenden Umfang; auf die Beschwerde ist somit einzutreten (Art. 72 Abs. 1, Art. 74 Abs. 1 lit. b, Art. 75 Abs. 1 und Art. 90 BGG).
        2. Die Parteien pflegten eine klassische Rollenteilung, bei der die Ehefrau die Kinder grosszog und sich um den Haushalt k\"{u}mmerte. Infolge der Trennung nahm sie im November 2005 wieder eine Arbeitst\"{a}tigkeit auf und erzielt mit einem 80\%-Pensum Fr. 2'955.-- netto pro Monat. Beide kantonalen Instanzen haben ihr jedoch auf der Basis einer Vollzeitstelle ein hypothetisches Einkommen von Fr. 3'690.-- angerechnet. Das Obergericht hat zwar festgehalten, der Ehefrau sei eine Ausdehnung der Arbeitst\"{a}tigkeit kaum m\"{o}glich, gleichzeitig aber erwogen, es sei nicht ersichtlich, weshalb sie nicht einer Vollzeitbesch\"{a}ftigung nachgehen k\"{o}nne. Ungeachtet dieses Widerspruches wird das Einkommen von Fr. 3'690.-- von der Ehefrau ausdr\"{u}cklich anerkannt, weshalb den nachfolgenden rechtlichen Ausf\"{u}hrungen dieser Betrag zugrunde zu legen ist. [...]
        }
        \vspace{-0ex}\\
    \end{minipage}
    \hfill
    \begin{minipage}[t]{0.22\textwidth}
        \vspace{-7ex} 
        \textbf{Metadata:} \\
        \tiny{
            Decision ID: \textit{65aad3f6-33c2-4de2-91c7-436e8143d6ea}\\ Year: \textit{2007}\\
            Language: \textit{German}\\
            Law Area: \textit{Civil}\\
            LD Label: \textit{Critical}\\
            Citation Label: \textit{Citation-1}\\
            Court: \textit{CH\_BGer}\\
            Chamber: \textit{CH\_BGer\_005,}\\
            Canton: \textit{CH}\\
            Region: \textit{Federation}
        }
    \end{minipage}
    \rule{\textwidth}{0.4pt}
\end{figure}

\begin{figure}[H]
    \centering
    \begin{minipage}[t]{\textwidth}
        \vspace{-7ex} 
        \textbf{Target}:
        \tiny{\textbf{critical-1}}\\
        \tiny{
            {Possible LD label: critical, non-critical, Possible citation label: critical-1, critical-2, critical-3, critical-4}
        }
        \vspace{01ex}\\
        \rule{\textwidth}{0.4pt}
    \end{minipage}
   
    \begin{minipage}[t]{0.49\textwidth}
        \includegraphics[width=\textwidth]{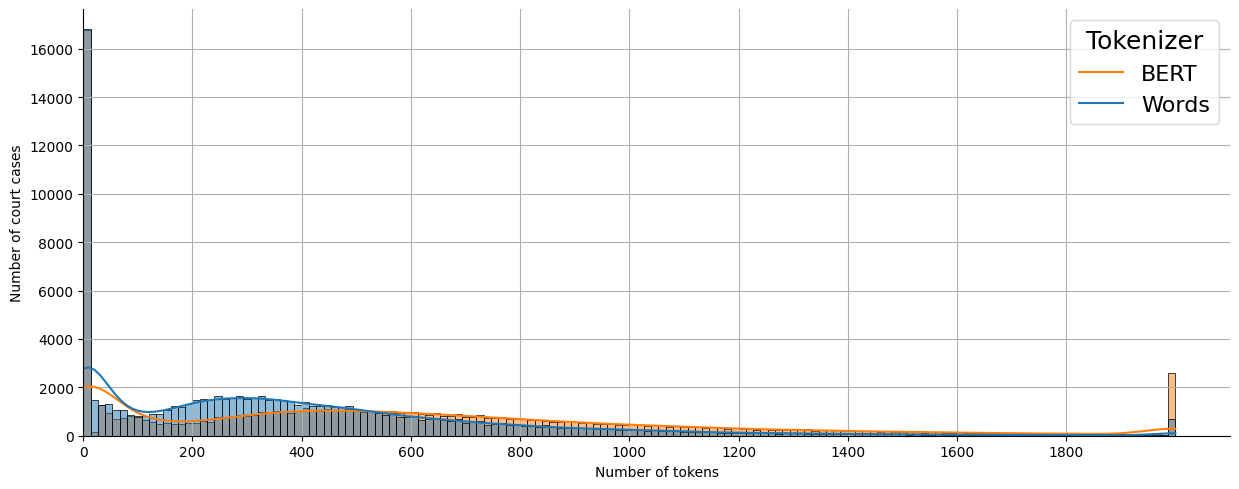}
        \caption{CP facts length distribution}
        \label{fig:cp_facts_length_distribution}
    \end{minipage}
    \hfill
    \begin{minipage}[t]{0.49\textwidth}
        \includegraphics[width=\textwidth]{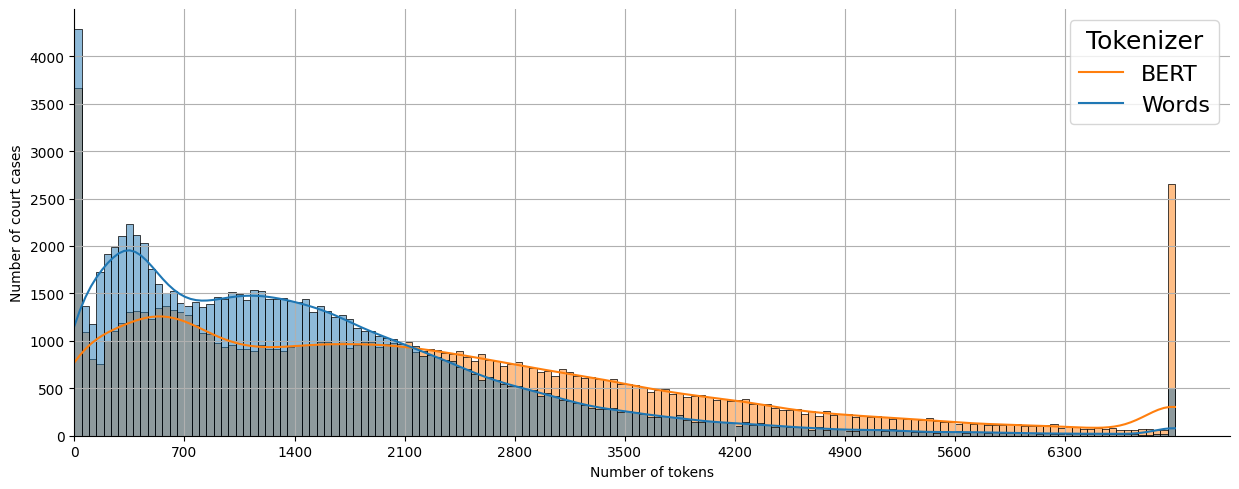}
        \caption{CP considerations length distr.}
        \label{fig:cp_considerations_length_distribution}
    \end{minipage}
    \vspace{-6ex}
\end{figure}
\end{framed}
\end{minipage}

\subsection{Information Retrieval}
\label{sec:ir}

Our \ac{IR} task is organized into queries, qrels, and corpus (see \Cref{fig:ir-data} in the Appendix). The corpus includes all Swiss legislation and leading decisions; queries come from \ac{SFCS} cases in German, French, and Italian. The mean token count for our queries significantly exceeds that in other \ac{IR} benchmarks, due to the use of entire documents as queries (see \Cref{tab:lang-dist}).
The goal is to find laws and decisions cited in a given case. We use the facts as a proxy for a lawyer-drafted appeal. Ground truth is based on citations from the considerations. We extract relevant cited laws and decisions from the Swiss legislation and leading decisions datasets, respectively. Document lengths resemble tasks like EU2UK \citep{chalkidis2021regulatory}. Laws in all three official languages yield cross-lingual query-corpus pairs, logged as qrels. Long documents and cross-lingual factors challenge retrieval models. We have 10K documents, 101K queries, and 2K qrels, averaging 19 relevant documents per query. Finding relevant legal references for \ac{SFCS} cases is challenging due to legal language complexity, multilinguality, and long documents.
\Cref{fig:ir_facts_length_distribution} shows the length distribution for the facts of the IR dataset.
Figures \ref{fig:ld_facts_length_distribution} and \ref{fig:ld_considerations_length_distribution} show the length distributions for the facts and considerations of Leading Decisions which are part of the corpus.\\

\begin{minipage}{\textwidth}
\begin{framed}
\begin{figure}[H]
\ContinuedFloat
    \renewcommand{\arraystretch}{1.5} 
    \centering
    \label{tab:MLIR_task}
    \vspace{-4ex}
    \begin{minipage}[t]{\textwidth}
        \textbf{Illustration of the \acf{MLIR} task} \\
        \vspace{-4ex}\\
        \rule{\textwidth}{0.4pt}
        Information retrieval is at the heart of the daily work of lawyers. Much like in scientific writing, lawyers base their arguments on prior caselaw, relevant legislation, and legal analyses. Thus, they spend a large part of their work searching for these documents, motivating the importance of legal IR. Annotating these data at scale is very costly, which is why this dataset is based on the citation graph of Swiss Supreme Court cases. In Switzerland lawyers operate in a trilingual jurisdiction with legislation and caselaw appearing in up to three official languages German, French and Italian. This means that for a complaint written in German, a case written in French might be relevant, leading to Multilingual IR, further complicating the task.\\
        \vspace{-1ex}
    \end{minipage}
    \rule{\textwidth}{0.4pt}
\end{figure}

\begin{figure}[H]
    \ContinuedFloat
    \centering   
    \vspace{-7ex} 
    \begin{minipage}[t]{0.74\textwidth}  
        \textbf{Input}\\
        \tiny{
            [Facts]:
    	    Fatti:
            A. Il 9 novembre 2006 G.\_, nata P.\_ (1959), ha contratto matrimonio con F.\_. Dall'unione non sono nati figli. Per contro il marito ha avuto figli (ormai adulti) dal primo matrimonio i quali non hanno per\`{o} vissuto in economia domestica con G.\_. Il 26 settembre 2011 \`{e} deceduto F.\_.
            Con domanda del 4 ottobre 2011 G.\_ ha chiesto alla cassa di compensazione Medisuisse l'erogazione di una rendita vedovile. Con decisione del 27 ottobre 2011, sostanzialmente confermata il 22 dicembre successivo in seguito all'opposizione dell'interessata, la cassa di compensazione ha respinto la richiesta di prestazione per il motivo che la richiedente non era stata sposata almeno cinque anni con il defunto marito, come invece prescritto dalla legge, bens\`{\i} "solo" 4 anni 10 mesi e 18 giorni.
            B. Osservando che il termine di cinque anni non era adempiuto per soli pochi giorni e invocando di conseguenza una applicazione della legge secondo [...]
            }
    \vspace{-7ex}\\
    \end{minipage}
    \hfill
    \begin{minipage}[t]{0.22\textwidth}
        \textbf{Metadata:} \\
        \tiny{
            Decision ID:\\\textit{6856ac58-5d12-48c4-acef-831d50c79886}\\
            Year: \textit{2012}\\
            Language: \textit{Italian}\\
            Law Area: \textit{Social}\\
            Court: \textit{BE\_VB}\\
            Chamber: \textit{CH\_BGer\_009,}\\
            Canton: \textit{BE}\\
            Region: \textit{Federation}
            }
    \end{minipage}
    \vspace{1ex}\\
    \rule{\textwidth}{0.4pt}
    \vspace{-3ex}\\ 
    \begin{minipage}[t]{\textwidth}
    \textbf{Target:}
        \tiny{Laws: [9 law ids]
    75488867-c001-4eb9-93b9-04264ea91f55, 
    e10ed709-8b11-47e3-8006-88b26d86e498, 
    [...]\par
    
    Cited Rulings: [2 ruling ids]
    54df6482-97d7-47eb-afb1-1ccb9369cb89, 
    921a799a-9077-4057-8e46-4919fd4f3263\\
    (There are 10K different laws and rulings see Section \ref{sec:ir} for more information)
    }
    \vspace{-1ex}\\
    \end{minipage}
    \rule{\textwidth}{0.4pt}
    \vspace{-3ex}\\
\end{figure}

\begin{figure}[H]
    \ContinuedFloat
    \centering
    \vspace{-6ex} 
    \begin{minipage}[b]{0.79\textwidth}
    \centering
       \includegraphics[width=0.7\textwidth]{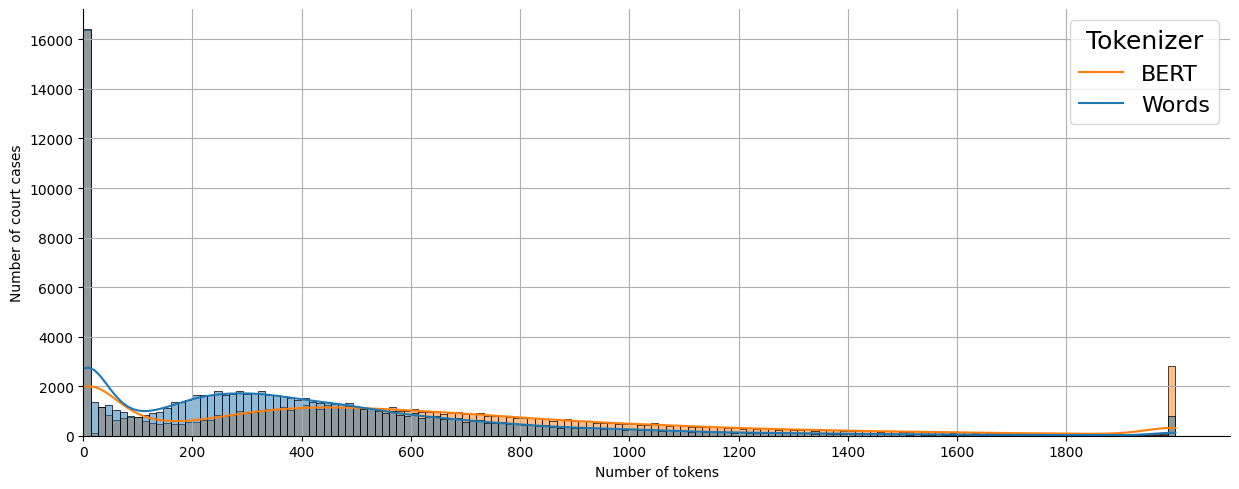}  
        \captionof{figure}{Information Retrieval facts length distribution}
        \label{fig:ir_facts_length_distribution}
    \end{minipage}
    \vspace{-6ex}
\end{figure}

\begin{figure}[H]
    \ContinuedFloat
    \centering
    \vspace{-0ex}
    \begin{minipage}[b]{0.49\textwidth}
       \includegraphics[width=\textwidth]{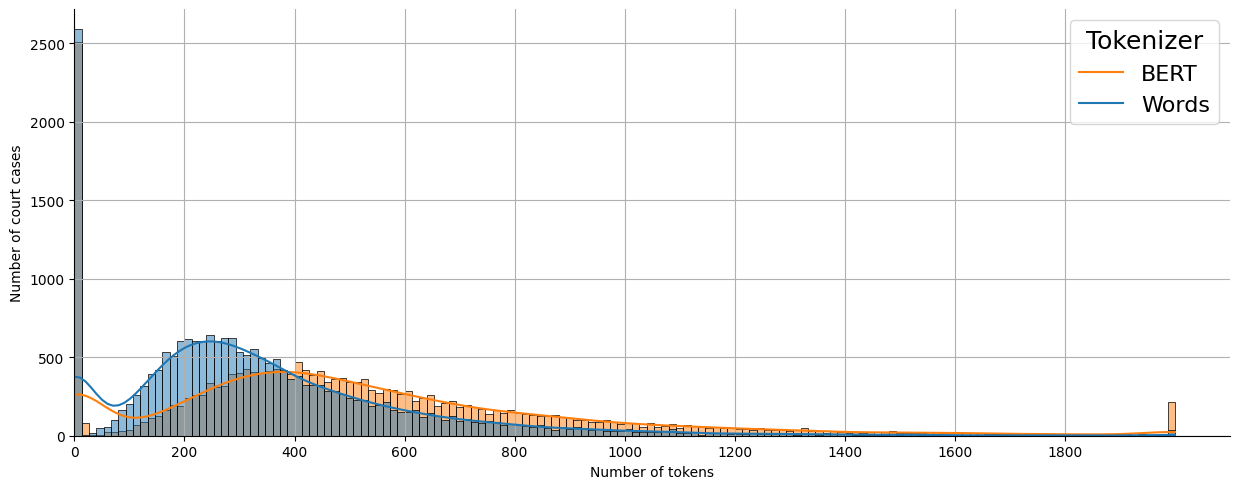}
        \captionof{figure}{LD facts length distribution}
        \label{fig:ld_facts_length_distribution}
    \end{minipage}
    \hfill
    \begin{minipage}[b]{0.49\textwidth}
        \includegraphics[width=\textwidth]{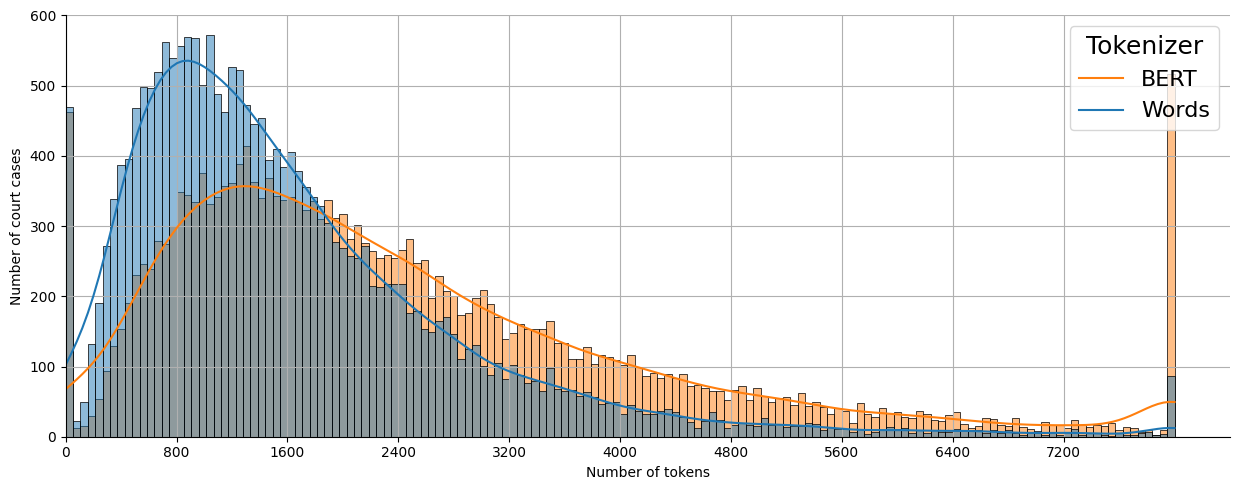}
        \captionof{figure}{LD considerations length distribution}
        \label{fig:ld_considerations_length_distribution}
    \end{minipage}
    \vspace{-6ex}
\end{figure}
\end{framed}
\end{minipage}

\subsection{Citation Extraction}
\label{sec:ce}

The \ac{SFCS} annotates citations with special HTML tags, which we used to create a token classification dataset for \ac{CE}. Solving the task with regexes is complicated due to extensive citation rules, but the transformer-based model MiniLM \citep{wang_minilm_2020} achieves over 95 macro F1. For brevity, we omit experiments on this dataset, but release the dataset and the trained model as a resource to the community under CC BY-SA licenses.
\Cref{fig:ce-cons-length-distribution} shows the token distribution for the CE dataset. \\

\begin{minipage}{\textwidth}
\begin{framed}
\begin{figure}[H]
\ContinuedFloat
    \centering
    \label{tab:ce-task}
    \vspace{-4ex}
    \begin{minipage}[t]{\textwidth}
        \textbf{Illustration of the \acf{CE} task} \\
        \vspace{-4ex}\\
        \rule{\textwidth}{0.4pt}
         Citation extraction is an important preprocessing step to collect information from legal documents. It enables easy semantic linking to relevant legislation, caselaw and analyses. Due to extensive rulebooks, simple regexes are often insufficient for accurate extraction of legal citations, motivating the need for more complex approaches. Addditionally, the dataset can be used to train models to suggest legal references while drafting text \cite{taylor_galactica_2022}.
         \vspace{1ex}
    \end{minipage}
    \rule{\textwidth}{0.4pt}
\end{figure}

\begin{figure}[H]
    \ContinuedFloat
    \centering
    \vspace{-7ex} 
    \begin{minipage}[t]{0.60\textwidth}
        \textbf{Input}\\
        \tiny{
           Considerations: \par
	['ergangen', 'ist', 'und', 'sich', 'das', 'Verfahren', 'daher', 'noch', 'nach', 'dem', 'Bundesgesetz', '\"{u}ber', 'die', 'Organisation', 'der', 'Bundesrechtspflege', '(', 'OG', ')', 'vom', '16', '.', 'Dezember', '1943', 'richtet', '(', 'vgl', '.', 'Art', '.', '132', 'Abs', '.', '1', 'BGG', ';', 'BGE', '132', 'V', '393', 'E', '.', '1', '.', '2', 'S', '.', '395', ')', 'dass', 'das', 'Verfahren', 'nicht', 'die', 'Bewilligung', 'oder', 'Verweigerung', 'von', 'Versicherungsleistungen', 'zum', 'Gegenstand', 'hat', 'und', 'deshalb', 'gem\"{a}ss', 'Art', '.', '134', 'Satz', '1', 'OG', '[', 'in', 'der', 'von', '1'] 
    }
    \vspace{-1ex}\\
    \end{minipage}
    \hfill
    \begin{minipage}[t]{0.37\textwidth}
        \textbf{Metadata:} \\
        \tiny{
        Decision ID:\\\textit{1572342e-a20d-4137-9593-47fc43b98af3}\\
        Year: \textit{2007}\\
	    Language: \textit{German}\\
	    Law Area: \textit{Social}\\
        Court: \textit{CH\_BGer}\\
	    Chamber: \textit{CH\_BGer\_009,}\\
	    Canton: \textit{CH}\\
	    Region: \textit{Federation}
     }
    \end{minipage}
    \rule{\textwidth}{0.4pt}
    \vspace{-3ex}\\ 
    \begin{minipage}[t]{\textwidth}
     \textbf{Target}:
        \tiny{[array of labels] [0, 0, 0, 0, 0, 0, 0, 0, 0, 0, 0, 0, 0, 0, 0, 0, 0, 0, 0, 0, 0, 0, 0, 0, 0, 0, 0, 0, 3, 4, 4, 0, 0, 0, 0, 0, 1, 2, 2, 2, 0, 0, 0, 0, 0, 0, 0, 0, 0, 0, 0, 0, 0, 0, 0, 0, 0, 0, 0, 0, 0, 0, 0,  0, 0, 3, 4, 4, 0, 0, 0, 0, 0, 0, 0, 0]\\
            possible labels: 0: O
            1: B-CITATION
            2: I-CITATION
            3: B-LAW 
            4: I-LAW\\
        }
    \vspace{1ex}
    \rule{\textwidth}{0.4pt}
    \end{minipage}
\end{figure}

\begin{figure}[H]
    \ContinuedFloat
    \centering
    \begin{minipage}[b]{0.79\textwidth}
    \centering
    \vspace{-6ex} 
       \includegraphics[width=0.70\textwidth]{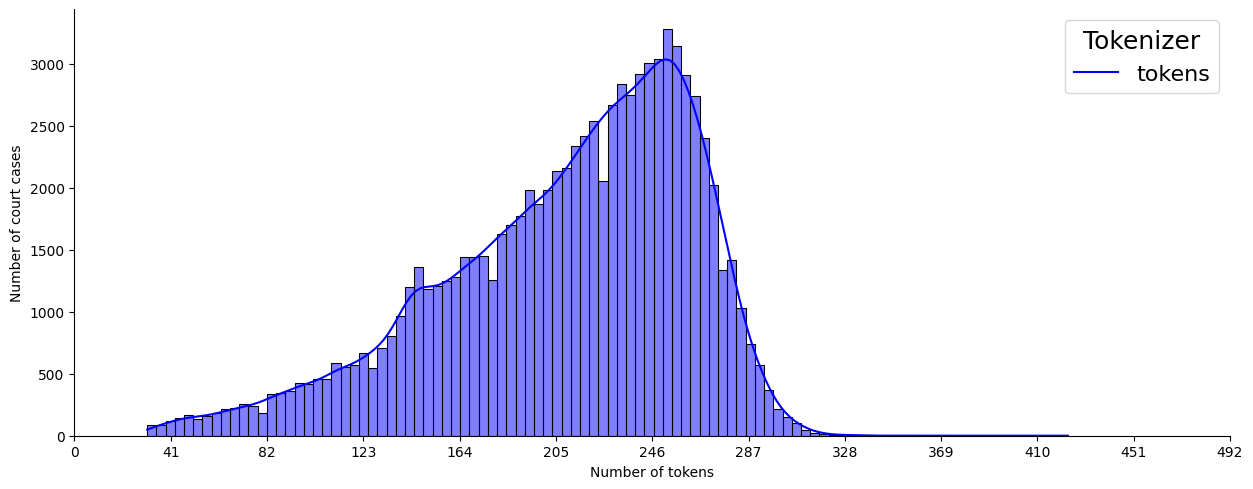} \\
    \captionof{figure}{Citation Extraction considerations length distribution} 
    \label{fig:ce-cons-length-distribution} 
    \end{minipage}
    \vspace{-6ex}
\end{figure}
\end{framed}
\end{minipage}

\subsection{The Big Picture}

The pretraining corpus and our seven datasets \ac{CVG}, \ac{LDS}, \ac{LAP}, \ac{JP}, \ac{CP}, \ac{IR}, and \ac{CE} form a unified framework to support officials at crucial stages in judicial system (see Figure~1). Pre-training serves as the foundation, equipping models with the ability to specialize in the respective tasks and thereby enhancing their performance. Models trained on the the other interconnected tasks, can provide crucial  support in a future semi-automated justice system.

\section{Experiments}
\label{sec:experiments}

In this section, we detail the pretraining of our legal models and the experimental setup for each task.
Besides BLOOM \citep{scao_bloom_2022} and mT5 \citep{xue_mt5_2021}, few open multilingual \acp{LLM} (> 500M parameters) exist, with most recent work pretraining on English only. LLaMA-2 \citep{touvron_llama_2023-1} uses only 4.5\% non-English data, while XLM-R \citep{conneau_unsupervised_2020} includes 87\% non-English text. \Cref{tab:models} provides an overview of models evaluated.

\begin{table*}[ht]
  \centering
  \vspace{-3ex}
    \caption{Models: 
    InLen is the maximum input length the model has seen during pretraining. \# Paramaters is the total parameter count (including embedding). Our models were built upon the pre-trained RoBERTa/Longformer. SwissBERT was further trained from X-MOD. Utilizing three language adapters with X-MOD and SwissBERT led to fewer parameters and languages. (\%de/\%fr/\%it) shows the percentages of the Swiss languages in the corpus. 
    }
  \footnotesize
  \resizebox{\textwidth}{!}{
  \begin{tabular}{llrrrrrrrr}
    \toprule
    \textbf{Model} & \textbf{Source} & \textbf{InLen} & \textbf{\# Parameters} & \textbf{Vocab} & \textbf{\# Steps} & \textbf{BS}  & 
  \textbf{Corpus (\%de/\%fr/\%it)} & \textbf{\# Langs} \\
    \midrule
    MiniLM                              & \citet{wang_minilm_2020}                  & 512 & 118M & 250K & 1M & 256       & 2.5TB CC100 (2.9/2.5/1.3) & 100\\
    DistilmBERT                         & \citet{sanh_distilbert_2020}              & 512 & 135M & 120K & n/a & < 4000           & Wikipedia (na/na/na) & 104\\
    mDeBERTa-v3                         & \citet{he_deberta_2021, he_debertav3_2021}& 512 & 278M & 128K & 500K & 8192    & 2.5TB CC100 (2.9/2.5/1.3) & 100\\
    XLM-R\textsubscript{Base/Large}     & \citet{conneau_unsupervised_2020}         & 512 & 278M/560M & 250K & 1.5M & 8192    & 2.5TB CC100 (2.9/2.5/1.3) & 100\\
    X-MOD\textsubscript{Base}           & \citet{pfeiffer_lifting_2022}             & 512 & 299M  &   250K    &   1M & 2048
            & 2.5TB CC100 (2.9/2.5/1.3) & 3 (81)\\
    SwissBERT \tiny{(XLM vocab)}        & \citet{vamvas2023swissbert}                      & 512 & 299M  &   250K    &   364K & 768
            & Swissdox (80/18/1) & 3 (4)\\
    \midrule
    mT5\textsubscript{Small/Base/Large} & \citet{xue_mt5_2021}                      & 1K & 300M/580M/1.2B  &   250K    &   1M & 1024 & mC4 (CC) (3.1/2.9/2.4) & 101\\
    \midrule
    BLOOM\textsubscript{560M}             & \citet{scao_bloom_2022}                 & 2K & 560M  &   250K    &   1.3M & 256 & ROOTS (0/5/0) & 59\\
    \midrule
    Legal-Swiss-R\textsubscript{Base}   & ours                                      & 512  & 184M & 128K & 1M & 512   & CH Legal (50/27/23) & 3\\
    Legal-Swiss-R\textsubscript{Large}  & ours                                      & 512  & 435M & 128K & 500K & 512 & CH Legal (50/27/23) & 3\\
    Legal-Swiss-LF\textsubscript{Base}  & ours                                      & 4096 & 208M & 128K & 50K & 512  & CH Legal (50/27/23) & 3\\
    \midrule
    GPT-3.5                             & \citet{brown-etal-gpt3}                 & 16K & 175B  &   na    &   na & na & na   & na\\
    GPT-4                               & \citet{openai_gpt-4_2023}                 & 32K & na & na    &   na & na & na   & na\\
    LLaMA-2                               & \citet{touvron_llama_2023-1}            & 4K & 7B/13B/70B  &   32K    &   na & na & LLaMA-2 (0.2/0.2/0.1)   & 27\\
    \bottomrule
  \end{tabular}
  }
  \label{tab:models}
  \vspace{-3ex}
\end{table*}

\subsection{Pretraining Legal Models}
\label{sec:pretraining}

We release two multi-lingual legal PLMs, Legal-Swiss-RoBERTa and Legal-Swiss-LF\textsubscript{Base} (see \Cref{tab:models} for details), trained on Swiss rulings, legislation, and EUR-LEX data \citep{niklaus_multilegalpile_2023}. We adhered to the following best-practices in \ac{LM} development:

\noindent (a) We warm-start (initialize)our models from the original XLM-R checkpoints (base or large) of \citet{conneau_cross-lingual_2019}. Model recycling is a standard process followed by many~\citep{wei-etal-2022, instructgpt} to benefit from starting from an available ``well-trained'' PLM, rather from scratch (random). XLM-R was trained on 2.5TB of cleaned CommonCrawl data in 100 languages. 

\noindent (b) We train a new tokenizer of 128K BPEs on the training subsets to better cover legal language across languages. However, we reuse the original XLM-R embeddings for all lexically overlapping tokens \citep{pfeiffer_unks_2021}, i.e., we warm-start word embeddings for tokens that already exist in the original XLM-R vocabulary, and use random ones for the rest.

\noindent (c) We continue pretraining our models on our pretraining corpus with batches of 512 samples for an additional 1M/500K steps for the base/large model. We do initial warm-up steps for the first 5\% of the total training steps with a linearly increasing learning rate up to $1e\!-\!4$, and then follow a cosine decay scheduling, following recent trends. For half of the warm-up phase (2.5\%), the Transformer encoder is frozen, and only the embeddings, shared between input and output (MLM), are updated. We also use an increased 20/30\% masking rate for base/large models respectively, where also 100\% of the predictions are based on masked tokens, compared to \citet{devlin_bert_2019}\footnote{\citet{devlin_bert_2019} -- and many other follow-up work -- used a 15\% masking ratio, and a recipe of 80/10/10\% of predictions made across masked/randomly-replaced/original tokens.}, based on the findings of~\citet{wettig2022should}.

\noindent (d) For both training the tokenizer and our legal models, we use a sentence sampler with exponential smoothing of the sub-corpora sampling rate following \citet{conneau_cross-lingual_2019} and \citet{raffel_exploring_2020}, since there is a disparate proportion of tokens across languages (Figure~\ref{fig:legislation_lang}) and we aim to preserve per-language capacity, i.e., avoid overfitting to the majority (almost 50\% of the total number of texts) German texts.

\noindent (e) We consider mixed cased models, i.e., both upper- and lowercase letters covered, similar to all recently developed large PLMs~\citep{conneau_cross-lingual_2019, raffel_exploring_2020, brown-etal-gpt3}.

\noindent (f) To better account for long contexts in legal documents, we continue training the base-size multilingual model on long contexts (4096 tokens) with windowed attention (128 tokens window size) \citep{beltagy_longformer_2020} for 50K steps, dubbing it Legal-Swiss-LF-base. We use the standard 15\% masking probability and increase the learning rate to $3e\!-\!5$ before decaying but otherwise use the same settings as for training the short-context models. 


We will release all models on the
HuggingFace Hub under a CC BY-SA license including intermediate checkpoints (every 50K/10K training steps for RoBERTa/Longformer models) upon acceptance.
Limited resources prevented us from pretraining a large generative model, so we leave this to future work.

\subsection{Text Generation}
We evaluated using \textbf{BERT}Score \citep{zhang_bertscore_2020}, \textbf{BLEU} \citep{BLEU}, \textbf{MET}EOR \citep{banerjee_meteor_2005}, and \textbf{R}OUGE \citep{lin_rouge_2004}. Each individual metric has inherent weaknesses \citep{zhang_bertscore_2020}, so it is necessary to employ multiple metrics for a more comprehensive assessment. We suggest future work to evaluate predictions with trained lawyers.

\paragraph{Court View Generation}
\label{par:exp-cvg}
Due to lengthy input (avg. 1522 tokens) and output (avg. 4673 tokens) for CVG, we truncated input facts to 2048 tokens and output considerations to 512 tokens. In 90\% of cases, complete facts were retained. This truncation was driven by resource constraints and the task's complexity, which remained challenging with only 512 output tokens. Owing to test data volume and compute limits, the evaluation was limited to a subset of 1K instances. For the origin dataset, input was evenly divided between origin facts and considerations.

\paragraph{Leading Decision Summarization}
In our \ac{LDS} experiments, we faced large input text (avg. 3081 tokens) but shorter output text (avg. 168 tokens). To manage this, we truncated input to 4096 tokens and output to 256 tokens, preserving full output in over 80\% of cases.

\subsection{Text Classification}
\label{sec:text_classifcation_setup}

For our \ac{TC} tasks, namely \ac{LAP}, \ac{JP}, and \ac{CP}, we adopted the LEXTREME benchmark setup \citep{niklaus2023lextreme}, namely hierarchical aggregation of macro-averaged F1 scores using harmonic mean for fairness (the harmonic mean is biased more towards lower scores than the geometric or arithmetic mean). We averaged in order over random seeds, languages (de, fr, it), configurations (e.g., JP-F and JP-C), and datasets (LAP, JP, and CP). This setup punishes models with outlier low scores in certain languages, or configurations, thus promoting fairer models. 
We fine-tuned all models below 2B parameters per task on our training datasets with early stopping on the validation dataset. 
We evaluated closed models 0-shot as per the setup of \citet{Chalkidis2023ChatGPTMP}, using one instruction and example as input. Samples were randomly selected from the validation set to prevent test set leakage for future evaluations. For each sample, we checked whether it exceeded the model’s maximum token limit of 4096 and truncated if necessary. To manage costs, we limited the validation set to 1000 samples. Our experiments focused solely on zero-shot classification due to the long input lengths. For experiments involving few-shot classification approaches, we refer to \citet{Trautmann2023}. We show an overview of the prompts used in \Cref{app:prompts}. Since effective prompts can be quite dependent on the LLM, we did not do prompt engineering and used the same prompt across LLMs. We used the \href{https://platform.openai.com/docs/guides/gpt/chat-completions-api}{ChatCompletion API} for GPT-3.5 (gpt3.5-turbo as of June 7, 2023), 
and ran LLaMA-2 locally with 4-bit quantization.

\subsection{Information Retrieval}

In our IR task we expect performance to decline as the document count increases. We conducted an ablation study to assess minor dataset adjustments on performance (see \Cref{app:add_results}).
We use BM25 for scalability to long documents but note its limitations in contextual processing and multilingual handling \citep{INR-019}. Neural methods like \ac{SBERT} show promise but degrade on long texts due to truncation-induced context loss.
We also investigate training with hard negatives, using the distiluse-base-multilingual-cased-v1 SBERT model \citep{10.1145/3404835.3462880}. This 135M parameter model employs DistilmBERT as the student and \ac{mUSE} as the teacher, trained on 15 languages. It has a 128 token maximum sequence length and outputs 512-dimensional embeddings via mean pooling, suitable for cosine similarity scoring.
The BEIR datasets \citep{thakur_beir_2021} feature query lengths of 3-192 words and document lengths of 11-635 words. Our dataset surpasses these by 4-282x for queries (847 words on average) and approx. 8-500x for documents (approx. 4K/7K words on average for rulings/legislation). We therefore excluded cross-encoder models \citep{10.5555/3495724.3496209}, due to their high computational cost especially with longer texts. We evaluate models with \ac{NDCG} \citep{Wang2013ATA} and Capped Recall@k \citep{thakur_beir_2021}.\footnote{\tiny{The Capped Recall@k is computed as the proportion of relevant documents for a specific query, retrieved from the top k scored list of documents generated by the model. This is a good representation of model success in our specific task, as each query has multiple relevant documents without the need for intra-document ranking.}}
\section{Results}
\label{sec:results}


\subsection{Text Generation}


\begin{table*}[ht]
\vspace{-4ex}
\centering
\caption{Results on the Court View Generation task. The input is truncated to 2048 tokens.
\textbf{Bold}: best within setup; \underline{underlined}: best overall. (*) These models were fine-tuned on only 1'000 samples for 3 epochs. All models, except the mT5 models, were evaluated on the validation set.
}
\resizebox{1\textwidth}{!}{
\begin{tabular}{lrrrrr}
\toprule
\textbf{Model} & \textbf{Setup} & \textbf{BERT} $\uparrow$ & \textbf{BLEU} $\uparrow$ & \textbf{MET} $\uparrow$ & \textbf{R1 / R2 / RL} $\uparrow$ \\
\midrule
mT5\textsubscript{Large} & Fine-tuned & \bf \underline{75.74} & \bf \underline{66.92} & \bf \underline{34.44} & \bf \underline{34.91} / \bf 15.58 / \bf \underline{33.53} \\
mT5\textsubscript{Base} & Fine-tuned & 75.01 & 65.48 & 32.89 & 33.23 / 13.57 / 31.89 \\
mT5\textsubscript{Small} & Fine-tuned & 74.13 & 63.97 & 30.96 & 31.29 / 11.01 / 29.90 \\
\midrule
GPT-3.5-Turbo & Fine-tuned\textsuperscript{\tiny{*}} & 72.31 & 62.23 & 28.08 & 26.06 / 7.19 / 24.54 \\
LLaMA-2-13B Chat & Fine-tuned\textsuperscript{\tiny{*}} & \bf 74.22 & \bf 63.51 & \bf 33.33 & \bf 34.36 / \bf \underline{16.68} / \bf 33.20 \\
\midrule
GPT-4 & 1-shot & \bf 70.39 & 59.69 & 24.63 & 23.87 / 4.64 / 22.32 \\
\midrule
GPT-4 & 0-shot & 69.41 & 58.16 & 23.25 & 22.61 / 3.95 / 21.10 \\
LLaMA-2-13B Chat & 0-shot & 67.23 & 55.01 & 20.18 & 19.76 / 3.26 / 18.57 \\
\bottomrule
\end{tabular}%
}
\label{tab:court-view-generation-main}
\vspace{-3ex}
\end{table*}

\paragraph{Court View Generation}

We present CVG results in \Cref{tab:court-view-generation-main}. Fine-tuning models generally leads to higher scores, with even small models like mT5\textsubscript{Small} outperforming 1-shot GPT-4. 1-shot prompting offers marginal gains over 0-shot for both GPT-4 and Claude-2 (LLaMA-2 1-shot was limited by context width).
The generated text generally showed stylistic authenticity, resembling typical legal language, but often lacked logical coherence, highlighting current models' limitations in generating coherent court views. In multiple court cases, target considerations contained similar paragraphs, which were generally well-predicted (see examples in \Cref{tab:examples_court_view} in the Appendix).
While fine-tuned models proficiently predict specific textual patterns, LLaMA-2-13B-Chat in the zero-shot setup struggles, often reverting from German to English and introducing linguistic errors, probably due to a highly English dominant training corpus. Despite their challenges, zero-shot models focus more on the main content, while fine-tuned models mirror target formalities. We provide a more detailed error analysis in Appendix \ref{app:cvg_error_analysis}.
Larger mT5 models consistently outperformed smaller ones, but performance increase with longer input was minimal, sometimes counterproductive (see \Cref{tab:court-view-generation-app} in the Appendix).
The results from the origin dataset were less conclusive (see \Cref{tab:court-view-generation-origin} in the Appendix), likely due to the smaller dataset size.


\begin{table*}[ht]
\centering
\vspace{-2ex}
\caption{Results on the Leading Decision Summarization task.  The input is truncated to 4096 tokens.
\textbf{Bold}: best within setup; \underline{underlined}: best overall.
}
\resizebox{1\textwidth}{!}{
\begin{tabular}{lrrrrrrr}
\toprule
\textbf{Model} & \textbf{Setup} & \textbf{BERT} $\uparrow$ & \textbf{BLEU} $\uparrow$ & \textbf{MET} $\uparrow$ & \textbf{R1 / R2 / RL} $\uparrow$ \\
\midrule
mT5\textsubscript{Base} & Fine-tuned  & \bf{73.33} & \bf{30.81} & \bf{23.50} & \bf \underline{32.43} / \bf \underline{12.78} / \bf \underline{30.87} \\
mT5\textsubscript{Small} & Fine-tuned & 72.04 & 28.68 & 21.29 & 29.61 / 10.31 / 28.12 \\
\midrule
GPT-4 & 1-shot                      & \bf \underline{73.55} & \bf 47.75 & \bf \underline{34.72} & \bf 30.82 / \bf 9.68 / \bf 28.89 \\
GPT-3.5-Turbo-16K & 1-shot          & 72.89 & 45.21 & 32.76 & 29.69 / 9.25 / 27.94 \\
\midrule
GPT-4 & 0-shot                      & \bf 71.56 & 48.35 & \bf 32.97 & 26.52 / \textbf{8.93} / 24.51 \\
GPT-3.5-Turbo-16K & 0-shot          & 70.28 & 46.08 & 30.60 & 25.18 / 7.58 / 23.59 \\
\bottomrule
\end{tabular}
}
\label{tab:lds-main}
\vspace{-2ex}
\end{table*}

\paragraph{Leading Decision Summarization}
In contrast to the very specific CVG task, requiring long-form output, the closed large models perform very well on LDS, at least in BLEU and METEOR (see \Cref{tab:lds-main}). According to ROUGE, fine-tuned mT5 models are still better, while BERT-Score does not discriminate clearly. We assume that summarization is a much larger portion of internal instruction tuning datasets used for optimizing these models. 
The quality of the generated text demonstrated a good stylistic imitation of legal language and more consistent logical coherence compared to the \ac{CVG} task (see examples in Table \ref{tab:lds-analysis} and \ref{tab:examples_lds} in the Appendix). GPT-3.5-Turbo (zero-shot) offered a narrative-style summary, while others adhered to the traditional 'Regeste' format. Notably, GPT-3.5-Turbo made a factual error by negating a crucial element, and Claude-2 referenced an outdated legal provision. We conduct a detailed error analysis in \Cref{sec:error_analysis}.
\Cref{tab:lds-app} in the Appendix shows two trends for fine-tuned mT5 models. First, longer input generally improved scores across models. Second, larger models outperformed smaller ones, although the differences between base and large models were subtle.

\subsection{Text Classification}

\begin{table*}[ht]
  \vspace{-4ex}
  \centering
    \caption{Results on the Text Classification datasets using Macro F1 with the highest values in \textbf{bold}.
        'F' or 'C' after the dash indicate input from 'Facts' or 'Considerations'. 'CPLD' and 'CPC' denote CP tasks using LD and Citation labels, while 'SLAP' refers to Sub Law Area Prediction. Models with an asterisk (*) are zero-shot LLMs predicting on validation dataset samples (see Section \ref{sec:text_classifcation_setup}). 
        }
  \footnotesize
  \resizebox{\textwidth}{!}{
  \begin{tabular}{lrrrrrrrrrrr}
    \toprule
        \bf{Model} & \bf{CPLD-F} & \bf{CPLD-C} & \bf{CPC-F} & \bf{CPC-C} & \bf{SLAP-F} & \bf{SLAP-C} &  \bf{JP-F} &  \bf{JP-C} &  \bf{Agg.} \\
    \midrule
        XLM-R\textsubscript{Base}            &       57.2 &       65.9 &       21.3 &       23.7 &        67.2 &        73.4 &       60.9 &       79.7 &       44.6 \\
        XLM-R\textsubscript{Large}           &       56.4 &       67.9 &       24.4 &  \bf{29.1} &        65.1 &        78.9 &       60.8 &       80.9 &  \bf{48.6} \\
        X-MOD\textsubscript{Base}            &       56.6 &       67.8 &       20.0 &       20.6 &        63.9 &        64.4 &       60.5 &       79.1 &       41.9 \\
        SwissBERT\textsubscript{(xlm-vocab)} &       56.9 &       67.3 &       25.7 &       23.0 &        61.5 &        73.2 &       61.4 &       79.4 &       46.1 \\
    \midrule
        mT5\textsubscript{Small}      &       52.2 &       62.1 &       13.2 &       17.9 &        53.1 &        60.9 &       58.9 &       74.2 &       34.4 \\
        mT5\textsubscript{Base}       &       52.1 &       61.5 &       14.0 &       19.7 &        58.4 &        61.8 &       54.5 &       72.0 &       35.9 \\
    \midrule
        BLOOM\textsubscript{560M}            &       53.0 &       61.7 &       10.7 &        8.0 &        52.6 &        53.2 &       60.5 &       73.4 &       24.9 \\
    \midrule
        Legal-Swiss-RoBERTa\textsubscript{Base}       &       57.7 &       70.5 &       16.2 &       20.1 &        77.0 &        79.7 &       64.0 &       86.4 &       40.9 \\
        Legal-Swiss-RoBERTa\textsubscript{Large}      &       55.9 &       68.9 &  \bf{25.8} &       16.3 &        76.9 &   \bf{84.9} &       62.8 &  \bf{87.1} &       43.3 \\
        Legal-Swiss-LF\textsubscript{Base}      &  \bf{58.1} &  \bf{70.8} &       21.4 &       17.4 &   \bf{80.1} &        77.1 &  \bf{65.4} &       86.4 &       42.5 \\
    \midrule
        GPT-3.5*      &       46.6 &       44.8 &  25.7 &       16.7 &        67.9 &   69.5 &       51.3 &  61.9 &       38.6 \\
        LLaMA-2*       &       45.2 &       26.6 &  7.0 &       8.5 &        58.7 &   55.6 &       40.3 &  37.8 &       19.7 \\
    \bottomrule
    \end{tabular}
    }
    \label{tab:text-clasification-results}
    \vspace{-3ex}
\end{table*}

We present results in \Cref{tab:text-clasification-results}, with detailed information including standard deviations in \Cref{tab:config_aggregate_scores_with_std} in the Appendix. Language-specific scores are in \Cref{tab:config_aggregate_scores_language_specific} in the Appendix. Scores on the validation dataset are in \Cref{tab:config_aggregate_scores_of_validation_set} in the Appendix.
As expected, larger models generally perform better, with XLM-R\textsubscript{Large} emerging on top. 
Our pre-trained model Legal-ch R\textsubscript{Base} outperformed XLM-R\textsubscript{Base}, indicating that domain-specific pre-training enhances performance. Overall, our pre-trained models showed better aggregated results compared to other models.
However, unexpectedly, Legal Swiss RoBERTa\textsubscript{Large} underperformed compared to its base model XLM\textsubscript{Large}.
Due to the high weight to outliers allotted by the harmonic mean, Legal-ch-R\textsubscript{Large} is severely penalized by its relatively low performance in CPC-C compared to XLM-R\textsubscript{Large}. 
Despite extra training on longer texts up to 4096 tokens, Legal-ch-LF did not surpass the hierarchical Legal-ch-R-\textsubscript{Base} model.
Large models such as GPT-3.5, Claude-2, and LLaMA-2 underperform fine-tuned models, underlining the need for specialized models for these tasks.
The difference is largest in the \ac{JP} and \ac{SLAP} tasks where the fine-tuned models are best.



\subsection{Information Retrieval}

\begin{table*}[ht]
\centering
\vspace{-4ex}
\caption{Results on Information Retrieval with best scores per section in \textbf{bold}.
Abbreviations: \textbf{distil}use\textsubscript{Base}-multilingual-cased-v1, \textbf{swiss}-legal-roberta\textsubscript{Base}}
\resizebox{1\textwidth}{!}{
\begin{tabular}{lrr}
        \toprule
        \textbf{Model} & \textbf{RCap\texttt{@} 1 / 10 / 100} $\uparrow$ & \textbf{NDCG\texttt{@} 1 / 10 / 100} $\uparrow$\\
         \midrule
            BM25 (fr lang analyzer) & \textbf{11.37} / \textbf{7.74} / \textbf{16.54} & \textbf{11.37} / \textbf{8.34} / \textbf{11.51} \\
            \midrule
            SBERT distil & 0.90 / 0.75 / \hspace{1.5mm}2.64 & 2.06 / 1.70 / \hspace{1.5mm}3.31 \\
            SBERT distil + pos & \textbf{4.40} / 3.92 / 12.64 & \textbf{10.11} / 8.76 / 16.16 \\
            SBERT distil + pos + h-neg & 3.97 / \textbf{4.46} / \textbf{13.36} & 9.12 / \textbf{9.21} / \textbf{16.87} \\
            SBERT swiss + pos & 3.97 / 3.47 / 12.28 & 9.12 / 7.76 / 15.16\\
            \midrule
            SBERT distil eval on de queries & \textbf{4.22} / \textbf{4.49} / \textbf{15.21} & \textbf{8.21} / \textbf{8.15} / \textbf{15.86}\\
            SBERT distil eval on fr queries & 1.88 / 2.20 / \hspace{1.5mm}9.19 & 5.77 / 6.22 / 13.94 \\
            SBERT distil eval on it queries & 0.22 / 0.24 / \hspace{1.5mm}0.79 & 5.43 / 5.74 / 11.44\\
        \bottomrule
\end{tabular}
}
\label{tab:mlir-results}
\vspace{-3ex}
\end{table*}

\Cref{tab:mlir-results} shows that most models failed to retrieve relevant documents, even with k=100. Lexical models outperformed others even without hyperparameter optimization for BM25 \citep{chalkidis_regulatory_2021}.  Surprisingly, despite German prevalence in our dataset, a French \href{https://www.elastic.co/guide/en/elasticsearch/reference/current/analysis-lang-analyzer.html}{language analyzer} (used for stemming and stopword removal) demonstrated superior performance. For \ac{SBERT}, truncation led to context loss, negatively affecting scores, a problem absent in lexical models. Training \ac{SBERT} models using Multiple Negative Ranking Loss \citep{Henderson2017EfficientNL} significantly improved performance, with hard negative examples beneficial. \ac{SBERT} evaluation on single languages, denoted as DE, FR, and IT, revealed its inability to perform consistently across all languages, which could be caused by the training set consisting of more German than French or Italian documents.

More experiments are in \Cref{tab:mlir-results-100-shorten-corpus} and \Cref{tab:mlir-results-details} in the Appendix.
Overall, our study exposes limitations of models in handling multilingualism, long documents, and legal texts, areas relatively underexplored in previous research. These findings offer a foundation for the \ac{IR} community to address these challenges.

\section{Error Analysis}
\label{sec:error_analysis}

Automatic metrics are known to be limited in assessing the performance in text generation tasks \cite{schluter-2017-limits, zhang_bertscore_2020}. To mitigate this and to get qualitative insights into the generations of different models, an author of this work with a background in law performed the following error analysis.

\subsection{\acf{CVG}}
\label{app:cvg_error_analysis}

Generally, specific textual constructs, such as those related to asylum applications, appear frequently in a similar manner so fine-tuned models are able to proficiently predict these patterns, whereas zero-shot models face challenges. \textit{LLaMA-2-13B-Chat (0-shot)} switches to English after a few sentences, and the German segments contain linguistic and grammatical errors. A possible reason could be that LLaMA-2 was predominantly trained in English rather than other languages. Remarkably, the LLaMA zero-shot model doesn't revert to English when handling the French text, unlike its behavior in the German scenario.
It also made an unsupported claim regarding the appellant's age, which wasn't mentioned in the input. Zero-shot models tend to center more on the primary content, while fine-tuned models are tailored to predict formalities, mirroring the target.
The fine-tuned \textit{LLaMA-2-13B-Chat} references 'BFM' instead of the 'SEM' (State Secretariat for Migration), deviating from both \textit{GPT-3.5}'s outputs and the original input. 
To note, SEM emerged in 2004 from a merger between the BFM (Federal Office for Refugees) and the IMES (Federal Office for Immigration, Integration and Emigration), nevertheless our sample was from 2016.

As outlined in \Cref{par:exp-cvg}, constraints in model availability and computational resources led to truncation of the target output during fine-tuning and experimentation, thus accentuating formal aspects like court jurisdiction and the legitimacy of appeals. This causes zero-shot iterations to receive lower scores, even when they remain contextually accurate. In terms of content, both zero-shot models were on the right track, tending more towards rejection. Furthermore, zero-shot predictions provide clearer insights by minimizing emphasis on formal nuances.
\Cref{tab:cvg-analysis-de,tab:cvg-analysis-fr} in the Appendix show a comparison of generated text in German and French respectively from GPT-3.5-Turbo-16K, GPT-4, Claude Instant and Claude 2. In the following, we analyse the four models from a legal perspective.

\paragraph{German}
GPT-3.5 (Fine-tuned) identifies the correct provisions of law and resembles part of the target output. It does connect the legal reasoning to the facts from the input. The legal reasoning is very basic and not detailed enough; i.e., one can understand that it is an asylum petition but nothing more.
LLaMA-2-13B Chat (Fine Tuned) identifies some provisions of the law correctly but also makes some mistakes. For example the SEM (State Secretariat for Migration) is the rightful authority in dealing with this case but this output calls it the FOM instead; it fails to expand the abbreviation to this and as such it can be construed that it made a mistake. It outlines the procedural compliances but provides no background details.
CLAUDE 2 (0Shot) provides more reasonable output compared to the first two. It identifies the elements that are most relevant to judge the instant appeal and also lists down the case briefly so as to reach a legitimate consideration.
LLaMA-2-13B Chat (0Shot) lists down the factual background of the case in a concise manner. It does not take reference to the provisions of law that governs the instant petition but it has categorically stated the grounds on which the appeal was dismissed. For instance upon reading this output one can have a fair idea of what transpired in this particular asylum petition.

The target output identifies the legal provisions (laws and rules) and the rightful authority that shall in accordance to law preside over the instant matter. It states every legal provision along with any exceptions that govern the matter of asylum. The grounds on which an appeal can be lodged is also stated herein. Keeping the target output in mind, GPT 3.5 Fine Tuned, Claude 2(0 Shot) and LLaMA-2-13B Chat(0 Shot) have generated acceptable outputs. For instance GPT 3.5 Fine Tuned aligns closely with the target, it features most of the legal provisions and briefly states the matter of Asylum. On the other hand,  Claude 2(0 Shot) provides judicial reasoning by way of defining elements like "Refugee" in the output, which is integral to the case as an asylum is being sought for on that very ground. It has also briefly discussed the background of the case which is essential in order to reach a legitimate outcome. LLaMA-2-13B Chat(0 Shot) listed the factual backgrounds and also provided judicial grounds that would lead to the dismissal of the appeal. Bearing these aspects of the generated output these three have shown higher potential especially by identifying legal provisions and premises to reach a legit consideration.

\paragraph{French}
GPT-3.5 (Fine-tuned) does not provide extensive legal reasoning. It draws reference to Fribourg Public Law which does not form part of the factual background. It refers to quite a few articles from the SCPC but failed to connect them to the relevant facts of the case.
The LLaMA-2-13B Chat (Fine-tuned) records the verdict of the previous court, which is very relevant. The legal reasoning is connected to the chain of events which has been written along with the dates; this aspect is noteworthy and very professional. The legal principles and the application of specific statutes are not mentioned.
Claude 2 (0Shot) refers to the wrong articles, for example, Article 343 (1)(a) of the SCPC is not the correct article in this matter; articles 236 and 308 (1) would be the ones applicable in this case. This output describes a different question not quite connected to the main issue. Judicial reasoning has two important components, firstly the applicable law and second creating a liaison between the events and how the law has been applied. This output has not been able to tie the law and what happened in the case.
LLaMA-2-13B Chat (0 Shot) identifies the parties concerned and the question that has to be addressed at the beginning. It creates a good sequence of events without the dates though. Although it does not refer to any specific articles or citations it does base its legal reasoning on the SCPC. It does not analyse any evidence. This output has amateur judicial reasoning which is good compared to the rest of the outputs, but it is not a good fit for an appeals court.

The target output highlights the four essential elements of the instant case, it has introduced the dispute and the remedy at the very beginning. It has stated the reason as to why the instant appeal is a subject matter of a constitutional appeal. It has noted the procedural compliances and mentioned as to why the appeal would be admissable before the court. Keeping this sequence in mind the output generated by LLaMA-2-13B Chat (Fine Tuned) and LLaMA-2-13B Chat (0 Shot) are of good quality. Both these outputs have touched upon the essential elements necessary for case flow. For instance LLaMA-2-13B Chat (Fine Tuned) recorded the earlier verdict, which is very impressive, thereby establishing the reason behind the appeal. Despite lacking the legal provisions it has stated relevant legal reasoning. LLaMA-2-13B Chat (0 Shot) provides quite strong judicial reasoning, especially when it highlights the principle of trust in the context of the matter and also the justification of the grounds of dismissal in the light of the Swiss Code of Obligation. Although this is not part of the target output, it could be a nice explanation contributing to the matter.

\subsection{\acf{LDS}}
\label{app:lds_error_analysis}
\Cref{tab:lds-analysis} in the Appendix shows a comparison of generated text from GPT-3.5-Turbo-16K, GPT-4, Claude Instant and Claude 2.
In the side-by-side analysis, all models demonstrated an authentic stylistic representation. GPT-3.5-Turbo 16K provides a more narrative-style summary, while the other models stick more closely to a classic 'Regeste' format. However, when it came to the factual accuracy, GPT-3.5-Turbo-16K faltered by negating a crucial element (\textit{"wurde entschieden, dass 'kein' hypothetisches Einkommen.."}). Claude-2 cited an outdated legal provision, Article 137 of the ZGB, which has been inactive since 2011, even though the case decision was from 2017 and no such reference was present in the input. Despite this, the article cited remains relevant to the context. On the other hand, GPT-4's reference to BGE 137 III 118 and GPT-3.5's mention of BGE 143 III 233 S. 235 were accurate and existed in the provided input.

From the four generated outputs GPT4 (1 shot) stated the consideration in a very precise manner at the very inception. Despite the attempt at identifying the party, the maintenance debtor, which is a first given the other outputs, it has failed to identify it correctly. However attempting to identify the party is a good practice in terms of summarization. Overall it has managed to comprehend and explain the principle of law. 
Claude 2 (1 Shot) refers to the wrong article of the SCPC. Instead of article 127 which is relevant for the instant dispute it refers to article 137, it is a possibility that it may have confused the article number to the judgement cited (BGE 137 III 118 E.2.3). The output is not very precise, but just about touches the principle. It is more vague of a summarization than a concise deduction. 
GPT 3.5 Turbo 16K (0 SHOT) generates a summary on a wrong notion and fails to interpret the considerations correctly. For instance at no point of time was it recorded that the husband had an abusive behaviour. The Court may have explained through judgements, what could be considered abusive. The model may have misinterpreted the conditions contributing to abusive behaviour to the husband being abusive. In determining a modification application it has not been able to secure a result but has elaborated on the conditions when modification cannot be reversed.
CLAUDE INSTANT (0 SHOT) refers to article 107 of the SCPC, which is not mentioned with the factual framework of the instant case. However it has drawn reference to an important judgement recorded in the Bundesgerichte  very relevant to this context and fits perfectly. But it fails to spell out the outcome of the case, instead explains very vaguely the law.
In analysing the summarization generated by the four different models, GPT 4 (1 shot) generated the most precise version of the actual outcome and also showed the ability to understand the law behind the consideration.

\section{Conclusions and Future Work}

\subsection*{Conclusions}
\label{sec:conclusion}

We present SCALE, an end-to-end benchmark of seven datasets for the Swiss legal system, a worldwide unique possibility to study crosslinguality within the same jurisdiction. Our tasks require legal reasoning abilities and challenge models on four key aspects: long documents, domain-specificity, multilinguality, and multitasking. 
We evaluate 14 open and five closed multilingual models, including three in-domain pretrained, as a reference point. These models, including ChatGPT, 
show low performance, particularly in challenging tasks like \ac{CVG} and \ac{IR}. Our results highlight opportunities for improving models and set the stage for next-generation LLM evaluations in domain-specific, multilingual contexts.


\subsection*{Future Work}
\label{sec:future_work}

First, quantifying monetary amounts for example for alimony or eviction specifically or claims in general is a complicated task. An automated system to aid in suggesting amounts could greatly speed up the process.
Second, the summary (regeste) of the leading decisions are composed of important citations in the first part, keywords from a \href{https://www.bger.ch/index/juridiction/jurisdiction-inherit-template/jurisdiction-jurivoc-home.htm}{Thesaurus} in the second part and a text-based summary in the third part. For simplicity, we treat it as one string. The first two tasks could be framed as classification or retrieval tasks, possibly improving model performance.
Third, due to limited context width, we only considered the facts as input to the court view generation task. However, judges and clerks do not only look at the facts when drafting a decision. They consider a myriad of information including possible lower court decisions and relevant case law, legislation and legal analyses. This information is available in our dataset. In the future, we would like to develop systems that are capable of integrating all this information to write the legal reasoning.
Fourth, so far, to our knowledge, the largest model pretrained on legal data specifically is Legal-XLM-R\textsubscript{Large} (435M parameters) \cite{niklaus_multilegalpile_2023}. Future work should look at pretraining larger generative models in the billions and tens of billions of parameters.
Fifth, future work may investigate the more difficult Citation Prediction task in addition to the Citation Extraction task. In Citation Prediction, the model only gets the context up to the citation as input and is tasked to predict the citation. This may help lawyers in drafting their texts.
Sixth, we strongly suggest future work include relevant external information, like caselaw or legislation for solving these challenging tasks. Augmenting models with retrieval \citep{lewis_retrieval-augmented_2021} and models using tools \citep{schick_toolformer_2023} seem to be promising avenues.
Seventh, explainability and interpretability is important for widespread adoption of automated systems, especially in the legal domain \cite{tyss_towards_2024,chalkidis_fairlex_2022}. Future work could experiment more with explainability techniques like generating rationales as auxiliary outputs on our benchmark.
Finally, to provide a better perspective on the results, we suggest future work to collect human performance as an additional reference point. This may be done at several levels, i.e., laypeople, law students, early career lawyers, expert lawyers in the respective field.


\appendix
\clearpage
\onecolumn
\section{Appendix Table of Contents}
We include the following supplementary sections in addition to the main paper:\\

\begin{itemize}
\item[] \ref{app:resource_access}: Access to the Provided Resources
\item[] \ref{app:reproducibility}: Reproducibility Statement
\item[] \ref{app:ethical_considerations}: Ethical Considerations
\item[] \ref{app:limitations}: Limitations
\item[] \ref{sec:add_related_work}: Additional Related Work
\item[] \ref{app:add_experimental_setup}: More Detailed Experimental Setup
\item[] \ref{app:detailed_data_description}: Additional Detailed Data Descriptions
\item[] \ref{app:add_results}: Additional Results
\item[] \ref{app:err-ananlysis-examples}: Error Analysis Examples
\item[] \ref{app:prompts}: Prompts
\item[] \ref{app:example_generations}: Example Generations

\end{itemize}
\clearpage\section{Access to the Provided Resources}
\label{app:resource_access}

In this section, we provide the URLs to the data, models, and code upon acceptance.

\clearpage\section{Reproducibility Statement}
\label{app:reproducibility}
Datasets, models, and source code for both dataset curation and experiments will be publicly released upon acceptance. The exact links will be provided in \autoref{app:resource_access}. We detail experimental setup in \autoref{sec:experiments} and in more detail in \autoref{app:add_experimental_setup}. All prompts used for evaluation of models larger than 2B parameters are listed in \autoref{app:prompts}.
\clearpage\section{Ethical Considerations}
\label{app:ethical_considerations}

While our research has several positive applications, it is important to acknowledge potential negative societal impacts. \acp{LLM} and their applications in the legal domain could potentially automate certain tasks traditionally performed by legal professionals, such as legal \ac{IR} and \ac{LDS}. While our goal is to support lawyers, it could impact the job market for legal professionals.

Recent literature has identified potential ethical problems within legal NLP research. The study on legal judgement prediction in China regarding prison term duration demonstrated the criticalness of legal NLP datasets and analysis \citep{chen_charge-based_2019}. As a response to this publication at EMNLP 2019, concerned researchers pointed out important questions to respond when working with ethically delicate data and NLP tasks \citep{leins_give_2020}. They suggest asking a series of ethical questions to assess the potential societal risks associated with a publication.

For example, they asked: "Does the dataset contain information that might be considered sensitive or confidential?" \citep{leins_give_2020}. In our case, we only used publicly available court decisions that are anonymized. Therefore, the person should not be identifiable. Another aspect is concerned with the possibility of future updates of the court decision due to new facts or an appeal (going to a higher court): "Will the dataset be updated?" We could update our data anytime. However, this maintenance would not happen automatically. We would need to be informed that a new decision was made regarding a certain case.

Another ethical concern concerns precision: Thus, LLMs can occasionally deliver results that are not entirely precise. This can have severe implications when it comes to the legal domain, where precision and factual accuracy are paramount. This could potentially lead to misinformation or misinterpretation of legal texts, impacting legal proceedings and decisions.

While the benchmark focuses on the Swiss legal system, it is important to recognize that law systems are highly culturally and contextually dependent. The understanding and interpretations of legal texts by these models, especially in a multilingual context, might not accurately reflect the nuanced cultural aspects of different regions. This could potentially lead to misrepresentations or misinterpretations, particularly when applied to other legal systems.

Finally, like any AI model, LLMs could be misused to create misinformation or misleading content at scale, especially in languages and domains where automated content generation is still a novel concept. It is crucial to develop and implement robust ethical guidelines and policies to mitigate these risks.

Therefore, while the new benchmark presents exciting opportunities to improve LLMs, it is essential to carefully consider the implications of its use and manage the associated risks effectively. The developers and users of such technology should adhere to ethical guidelines to ensure its responsible use.

\clearpage\section{Limitations}
\label{app:limitations}

While we covered a lot of ground in this work, there are always limitations, which we list in this section.

\subsection{General} 
The research area for \acp{LM} and benchmarks continues to evolve, and while there is palpable enthusiasm in the field, it is critical to maintain a balanced perspective. Studies, including \citet{bender_climbing_2020}, have shed light on the limitations of \acp{LM} and benchmarks, stressing that \acp{LM} do not truly "learn" meaning and that communities often focus on limited datasets, some of which are borrowed from other fields. 

\subsection{Models} 
Even though English models are plentiful, LLMs pre-trained multilingually are very rare. To the best of our knowledge, mT5 is the only multilingual model with variants over 1B parameters, covering German, French, and Italian (BLOOM does not contain German and Italian). Additionally, we were limited by very large sequence lengths. We did not have the resources available to run mT5\textsubscript{XL} or mT5\textsubscript{XXL} with sequence lengths greater than 1K. 

\subsection{Data} 
The process of selecting tasks for benchmarks is typically influenced by the interests of the community or the convenience of available resources, rather than being informed by all-encompassing theories. These constraints present difficulties when trying to explore a model's broader applicability or its capacity for understanding. The data employed in benchmarks is often tied to a specific context and is naturally susceptible to inherent biases. Furthermore, the content of such data may vary significantly from real-world data, is de-contextualized and the uniformity of the task formats may not adequately reflect the diversity of human activities. Regarding our specific context, it is crucial to acknowledge that we cannot generalize Swiss legal data to other countries or different legal systems. 

Figures \ref{fig:rulings_lang} and \ref{fig:rulings_canton} show the language and cantonal distributions over rulings and Figures \ref{fig:legislation_lang} and \ref{fig:legislation_canton} over legislation. Note that the distribution is imbalanced for both rulings (50\% German, 39\% French, and 11\% Italian) and legislation (49\% German, 31\% French, and 17\% Italian). However, compared to \href{https://en.wikipedia.org/wiki/Languages_of_Switzerland}{Swiss Speaker Distribution} (63\% German, 23\% French, and 8\% Italian), the legal text is actually more balanced.
It looks similar in the cantonal distribution, with many sparsely inhabited cantons being represented above their weight, especially the ones in the non-German speaking regions such as Vaud, Geneva, and Ticino. 

The training sets of the LLMs we benchmarked have different cutoff points from before 2020 for XLM-R to 2023 for GPT-4. Therefore the models with the later cutoff dates may have seen some of our test data (since our test set is from the years 2018 to 2022). Ideally we would start the test set from after the last model's cutoff point. In absence of this possibility another option would have been to use different test sets for the different models, but that brings other confounders like differing time distance from train to test sets and potentially different difficulty of examples in the test sets. Therefore, we opted for a large test set covering a long time range, potentially mitigating these effects. However, we are aware that the more modern models might simply get better performance on our benchmarks because they have trained on the inputs already.

\subsection{Labels}
Annotating high-quality datasets is very expensive, especially when experts are needed, such as in the legal domain. Because of our limited budget and in order to arrive at a large amount of labels, we algorithmically generated labels based on metadata information present in the corpus. These metadata are of high quality, being provided by the courts themselves. 

\subsection{Text Classification}
\paragraph{Judgment Prediction}
While the judgment prediction task is arguably very interesting and also very challenging, it is unlikely to be deployed in practice anytime soon. Ideally, we would want the complaints as input (similar to \citet{semo_classactionprediction_2022}) instead of the facts description, since this is written by the court itself in part to justify its reasoning. Unfortunately, the complaints are not public in Switzerland, making us rely on the widely available facts description as a close proxy.

\paragraph{Law Area Prediction}
We used information about the chambers at the courts to determine the law areas. Predicting the main law area is not challenging for current models, leading to very high results and thus rendering this task unsuitable for a benchmark. Unfortunately, most chambers cover multiple sub areas, thus ruling them out for the sub area prediction task and considerably reducing dataset size. In conclusion, while this task is very useful in practice for routing requests to the different chambers inside a court, it relatively unsuitable for a challenging benchmark.

\paragraph{Criticality Prediction}
It is very difficult to estimate the importance of a case. By relying on proxies such as whether the case was converted to a leading decision (LD-label) and how often this leading decision was cited (Citation-label), we were able to create labels semi-automatically. While we discussed this with lawyers at length and implemented the solution we agreed on finally, this task remains somewhat artificial.

\subsection{Text Generation}
\paragraph{Court View Generation}
Court View Generation is an extremely challenging task and thus very well suitable for a benchmark. Current multilingual transformer-based models do not allow processing text in the tens of thousands of tokens. As a consequence, we were forced to look at a simplified version of this task, only considering the facts as input and ignoring relevant case law and legislation. Additionally, we were only able to generate the first 512 tokens of the considerations. We thus invite the community to develop new methods potentially capable of tackling harder versions of this task.

\paragraph{Leading Decision Summarization}
Due to limited resources, we limited our evaluation to mT5\textsubscript{Small/Base/Large}. Future work may investigate large multilingually pretrained generative models on this task. Additionally, one may want to conduct human evaluation on the generated summaries. However, due to extremely high lawyer salaries in Switzerland, this is not feasible at scale in an academic research project. Finally, we only considered the simple version of this task where we only generate a text-based output. Future work may treat the first and second parts of the summary as extreme multilabel classification problems of relevant citations and relevant keywords from the thesaurus \href{https://www.bger.ch/index/juridiction/jurisdiction-inherit-template/jurisdiction-jurivoc-home.htm}{Jurivoc} respectively, possibly increasing performance.

\subsection{Citation Extraction}
Even though the citations are annotated by the \ac{SFCS}, we encountered citations that were not marked. However, models achieved very high scores anyway, leading us to exclude it from the benchmark. Future work may investigate this in more detail.

\subsection{Information Retrieval}
The labels are constructed with the citations from the considerations. Due to most legal analyses being private, our corpus is restricted to case law and legislation. Constructing a ranking of relevant documents is challenging due to missing information, and thus probably requiring extensive human annotation. Additionally, S-BERT models are usually limited to 512 tokens, being a constraint for this task due to our long documents.



\clearpage\section{Additional Related Work}
\label{sec:add_related_work} 
    \subsection{Domain-Specific Pretraining}
    General-purpose \acp{LM} are trained on generic text corpora such as Wikipedia and evaluated on widely used benchmarks such as GLUE \citep{wang_glue_2018}. However, domain-specific models need focused datasets for training and specialized benchmarks for assessing the quality of the model. The following examples illustrate the increase in performance when using domain-specific datasets and benchmarks.

    In the biomedical area of natural language processing (BioNLP), \citet{leeBioBERTPretrainedBiomedical2020} created for the first time a domain-specific LM based on BERT \citep{devlin_bert_2019} by pre-training it on biomedical text corpora. They used PubMed abstracts (4.5B words) and PubMed Central (PMC) full-text articles (13.5B words). The resulting domain-specific LM BioBERT achieved higher F1 scores than BERT in the biomedical NLP tasks named entity recognition (0.62) and relation extraction (2.80), and a higher mean reciprocal rank (MRR) score (12.24) in the biomedical question-answering task. In 2022, those scores were outperformed. \citet{naseem2022benchmarking} conducted a domain-specific pre-training of ALBERT \citep{lan_albert_2020} using only text from the biomedical field (PudMed) and from the "Medical Information Mart for Intensive Care" (MIMIC-III), a large, de-identified and publicly-available collection of medical records \citep{johnson_mimic-iii_2016}. One domain-specific benchmark applied to test BioALBERT originates from \citet{Gu2021BLURBbenchmark} who created BLURB, the Biomedical Language Understanding and Reasoning Benchmark. \citet{naseem2022benchmarking} found that BioALBERT exceeded the state-of-the-art models by 11.09\% in terms of micro averaged F1-score (BLURB score). Another biomedical NLP benchmark is BLUE, the "Biomedical Language Understanding Evaluation" \citep{pengBLUE2019}. It covers five tasks (sentence similarity, named entity recognition, relation extraction, document classification, inference) with ten datasets from the biomedical and clinical area. BioALBERT also includes all datasets and tests from BLUE thus presenting the most comprehensive domain-specific model and benchmark in the biomedical area at the moment.
    
    In the financial domain, FinBERT was pretrained 2020 by \citet{yang-FinBERT-2020} using financial data. The text corpora consisted of 203’112 corporate reports (annual and quarterly reports from the Securities Exchange Commission SEC), 136'578 earnings call transcripts (conference call transcripts from CEOs and CFOs), and 488'494 analyst reports (textual analysis of the company) resulting in 3.3B tokens. For testing FinBERT, \citet{yang-FinBERT-2020} used the Financial Phrase Bank dataset with 4'840 sentiment classifications \citep{maloGoodDebtBad2014}, the AnalystTone Dataset with 10'000 sentences \citep{huangEvidenceInformationContent2014}, and FiQA Dataset with 1'111 sentences from an open challenge dataset for financial sentiment analysis (\hyperlink{https://sites.google.com/view/fiqa/home}{Financial Opinion Mining and Question Answering}). The results show that the domain-specific FinBERT outperforms the generic BERT models in all of these financial datasets. An improved financial domain LM was released 2022 by \cite{shah-FLANG-FLUE-2022} by introducing FLANG-BERT, the Financial LANGuage Model. They also created a domain-specific benchmark, Financial Language Understanding Evaluation (FLUE). Recently in May 2023, Bloomberg announced the BloombergGPT model, a \ac{LLM} for the financial domain \citep{wu-BloombergGPT-2023}. However, next to some experience on the training process no datasets, benchmarks, or weights have been released publicly.
    
    Numerous other domain-specific LMs have been created since the rise of BERT. They all outperform general-purpose LMs. For instance, SciBERT is a pretrained LM based on scientific publications and evaluated on a suite of tasks in difference scientific domains \citep{beltagy_scibert_2019}. ConfliBERT is built to improve monitoring of political violence and conflicts \citep{huConfliBERTPretrainedLanguage2022} and PoliBERTweet is used to analyze political content on Twitter \citep{kawintiranonPoliBERTweetPretrainedLanguage2022}. Cybersecurity is another important area thus \cite{aghaeiSecureBERTDomainSpecificLanguage2023} pretrained a M on a large corpus of cybersecurity text. To improve \ac{IR} tasks in the architecture, engineering, and construction (AEC) industry, \citet{zhengPretrainedDomainspecificLanguage2022} pretrained BERT on a corpus of regulatory text. Also, the domain-specific model BlueBERT \citep{pengBLUE2019} from the biomedical domain has been further pretrained and evaluated on more narrow, cancer-related vocabularies, resulting in CancerBERT \citep{zhouCancerBERTCancerDomainspecific2022}.

    In the legal domain \citet{chalkidis_legal-bert_2020} pretrained LegalBERT on EU and UK legislation, ECHR and US cases and US contracts. \citet{zheng_when_2021} pretrained CaseHoldBERT on US caselaw. \citet{henderson_pile_2022} trained PoL-BERT on their 256 GB diverse Pile of Law corpus. \citet{niklaus_budgetlongformer_2022} pretrained longformer models using the \ac{RTD} task \cite{clark_electra_2020} on the Pile of Law. \citet{hua_legalrelectra_2022} pretrained reformer models with RTD on 6 GB of US caselaw. Finally, \citet{niklaus_multilegalpile_2023} released a large multilingual legal corpus and trained various legal models on it.


    \subsection{Judgment Prediction}

    The domain of Legal Judgment Prediction (LJP) centers around the crucial task of predicting legal case outcomes given the provided facts. In the landscape of LJP research, there have been significant advances focusing on diverse languages, jurisdictions, and input types. Researchers have utilized a variety of datasets, each with their unique characteristics and annotations, to analyze and predict case outcomes \citep{ijcai2022p765, aletras_predicting_2016, sulea_predicting_2017, medvedeva_judicial_2018, chalkidis_neural_2019}.
    
    In the context of Chinese criminal cases, notable efforts have been made by \citet{xiao_cail2018_2018, xiao_lawformer_2021}, where they utilized the CAIL2018 dataset, which consists of over 2.6M cases and provides annotations for Law Article, Charge, and Prison Term, among others. 
    
    Focusing on the Indian and Swiss jurisdictions, \citet{malik_ildc_2021} and \citet{niklaus_swiss-judgment-prediction_2021, niklaus_empirical_2022} employed the ILDC and SJP datasets respectively, both using binary labels. The ILDC dataset, with over 34K Indian Supreme Court cases, offers sentence-level explanations along with Court Decision annotations, while the SJP corpus is trilingual, containing judgments from Switzerland in German, French, and Italian, and provides annotations like the publication year, legal area, and the canton of origin.
    
    European jurisdictions have been explored using the ECHR2019 and ECHR2021 datasets \citep{chalkidis_neural_2019, chalkidis2021paragraph}. These corpora feature cases from the European Court of Human Rights, annotated for Violation, Law Article, and Alleged Law Article, among others, with the latter also providing paragraph-level rationales.
    
    The FCCR dataset, containing over 126K cases from France, has been used to predict Court Decisions with different setups, offering additional annotations such as the date of the court ruling and the law area \citep{sulea_predicting_2017}.
    
    Recently, \citet{semo_classactionprediction_2022} introduced a new perspective on LJP, applying it to US class action cases. The proposed task involves predicting the judgment outcome based on the plaintiff's pleas, further expanding the scope of LJP research and making the task more realistic.

    \citet{Prasad2023} utilized the ILDC dataset and a subset of the LexGLUE dataset to evaluate a method termed 'Multi-stage Encoder-based Supervised with-clustering' approach. This method involved chunking long documents, embedding \ac{LLM} representations, and applying clustering techniques for automatic legal judgment prediction. It achieved a minimum total performance gain of approximately 2 points over the previous state-of-the-art methods. Furthermore, they proposed an 'Occlusion sensitivity-based Relevant Sentence Extractor' designed to extract sentences that explain the predictions, focusing on the most relevant sentences from the document.
    
    These efforts underscore the breadth of LJP research, demonstrating its applicability across multiple jurisdictions, languages, and legal systems, and its potential in assisting legal professionals and enhancing access to justice.

    \subsection{Criticality Prediction}
    \citet{chalkidis_neural_2019} introduced the Importance Prediction task, which predicts the importance of a ECtHR case on a scale from 1 (key case) to 4 (unimportant). Legal experts defined and assigned these labels for each case, representing a significant contrast to our approach where labels were algorithmically determined. This is to our knowledge the only comparable task to Criticality Prediction.

    \subsection{Law Area Prediction}
    Although not widely studied, several notable works have focused on \ac{LAP}. \citet{sulea_predicting_2017} worked on the Law Judgment Prediction (LJP) task, using a dataset of over 126K cases from the French Supreme Court. The study used Linear Support Vector Machines (SVMs) to classify cases into one of eight law areas, using the entire case description as input. This approach yielded an F1 score of 90\%. \citet{soh_legal_2019} conducted a similar study using a dataset of 6K judgments in English from the Singapore Supreme Court. These judgments were mapped into 30 law areas. Several text classifiers were used in the study, achieving a macro F1 score of up to 63.2\%.

    \subsection{Court View Generation}    
    Over the past decade, text generation (excluding summarization) in the field of Legal NLP has been underexplored \citep{katz_natural_2023}, especially in comparison to tasks such as classification and information extraction.
    One of the early contributions in this domain is by \citet{ye_interpretable_2018}, who introduced the task of generating court views from criminal case facts using a Seq2seq model. This work utilizes a Chinese dataset with factual descriptions averaging around 220 tokens and court views of approximately 30 tokens in length.
    Further contributions include \citet{wu_-biased_2020}'s approach in Chinese civil law, focusing on creating legal documents for private lending cases. To achieve the reduction of systemic bias, they utilized an encoder-decoder architecture, designed to generate two distinct types of court views: 'supportive', which aligns with the plaintiff's claims, and 'non-supportive', which counters them. 
    A key element of this approach is the judgment classifier, which determines the most suitable court view out of the two options.
    Additionally, \citet{Li_Zhang_2021} utilize Chinese case facts, as well as charge (formal accusations of crimes) and law article information, to generate court opinions, averaging between 31 to 34 tokens in length.
    Another notable work is by \citet{10.1145/3404835.3462984}, who introduced the Circumstances enhanced Criminal Court View Generation (C3VG) method, emphasizing the importance of Adjudging Circumstance (ADC) and Sentencing Circumstance (SEC) in the legal judgment process. ADC relates to establishing whether an act meets the legal criteria for a specific crime, while SEC involves factors considered during sentencing. The C3VG employs a two-stage architecture, incorporating a BERT-based Circumstances Selector and seq2seq models for ADC and SEC. The dataset used for this study is composed of Chinese legal documents focused on criminal law.
    In contrast to these Chinese-focused approaches, our research utilizes a trilingual dataset in German, French, and Italian. Also notably, our court views are substantially longer, averaging approximately 4,000 tokens, highlighting the increased complexity of our benchmark.
    With the emergence of powerful generative models, we expect a surge in research activity in this area, necessitating challenging tasks to assess progress effectively.
    
    \subsection{Leading Decisions Summarization}
    In the field of legal text summarization, several noteworthy contributions have been made \citep{grover_holj_2004, hachey-grover-2006-extractive, 10.1007/978-3-642-39931-2_14, JAIN2021100388, shukla2022legal}, with some being particularly significant:
    The creators of the BillSum dataset \citep{kornilova2019billsum} focused on summarizing 22K bills from the US Congress and the state of California. They also applied transfer learning in summarization from federal to state laws. Models based on BERT and TF-IDF, as well as a combination of both, have been evaluated on this dataset. The BillSum dataset focuses on English language documents related to the US legislative environment. 
    The Multi-LexSum dataset \citep{shen_multi-lexsum_2022} targets long civil rights lawsuits, with an average length of over 75K words. This 9K-document dataset allows for in-depth study at different summary lengths: short (25 words), medium (130 words), and long (650 words), a unique feature of the Multi-LexSum dataset. Models based on BART \cite{lewis2020bart} and PEGASUS \cite{zhang_pegasus_2020} were evaluated on this dataset.
    Like BillSum, the Multi-LexSum dataset is primarily for the English language and is relevant to the US legal setting. In addition to these datasets, the EUR-Lex-Sum \cite{aumiller_eur-lex-sum_2022} is based on documents and summaries from the European Union law platform (EUR-Lex), encompassing a diverse range of the 24 official European languages. The dataset comprises over 1,500 document/summary pairs per language and features 375 cross-lingually aligned legal acts available in all 24 languages. A key differentiator of EUR-Lex-Sum from our work is its focus on legislation from a supranational organization, as opposed to our emphasis on court cases within a single jurisdiction. Addressing language barriers in the Indian legal system, the MILDSum dataset \cite{datta-etal-2023-mildsum} introduces a novel approach by focusing on the cross-lingual translation of English legal documents into Hindi. It comprises 3,122 Indian legal case judgments. The study's significant finding is the effectiveness of the 'Summarize-then-Translate' method over direct cross-lingual summarization. The dataset includes documents with an average length of 4,696 tokens and summaries in both English and Hindi, averaging approximately 724 and 695 tokens, respectively.  Furthermore, \citet{bauer2023legal} made a significant contribution by focusing on extracting key passages from 430K U.S. court opinions to create concise summaries.

    \subsection{Citation Extraction}
    Early work from \citet{martinez-gonzalez_reference_2005} extract citations from legal text with patterns. \citet{nambanoor_kunnath_dynamic_2022} studied the effect of differing context size for citation classification in scientific text. \citet{taylor_galactica_2022} considered the more difficult Citation Prediction task on scientific text and found that larger models are more true to the real citation distribution, whereas smaller models tend to output the most frequent citations most of the times.

    \subsection{Information Retrieval}
    \citet{lawrie2023neural} revisited the challenges of multilingual IR and proposed neural approaches to address this issue. They demonstrated that combining neural document translation with neural ranking resulted in the best performance in their experiments conducted on the MS MARCO dataset \cite{bajaj2018ms}. However, this approach is computationally expensive. To mitigate this issue, they showed that using a pre-trained XLM-R multilingual model to index documents in their native language resulted in only a two percent difference in effectiveness. XLM-R is a transformer-based masked \ac{LM} that employs self-supervised training techniques for cross-lingual understanding \cite{conneau2020unsupervised}. \citet{lawrie2023neural} crucially utilized mixed-language batches from the neural translation of MS MARCO passages.

    A widely used technique is BM-25, which is an improved retrieval method that considers the term frequencies and takes into account the saturation effect and document length \cite{INR-019}. The saturation effect refers to the point where the relevance of a term stops increasing, even if it appears many times in a document. This issue is mitigated through the use of an additional parameter, k. Additionally, longer documents are more likely to contain a higher number of occurrences of a term simply because they contain more words, not necessarily because the term is more relevant to the document, which is why parameter b is used. The BM-25 score is calculated using the Inverse Document Frequency (IDF), Term Frequency (TF), queries Q, documents d, and term t.
    \begin{quote}
        \begin{math}
            BM25(d,Q,b,k) = \sum_{t\in Q}IDF(t)\frac{(k+1)TF(t,d)}{(1-b)(b*A)+TF(t,d)}\\
        \end{math}
    \end{quote}

    \citet{chalkidis_regulatory_2021} proposed a new \ac{IR} task called REG-IR, which deals with longer documents in the corpus and entire documents as queries. This task is an adaptation of  \ac{Doc2Doc} \ac{IR}, which aims to identify a relevant document for a given document. The authors observed that neural re-rankers underperformed due to contradicting supervision, where similar query-document pairs were labeled with opposite relevance. Additionally, they demonstrated for long documents that using BM25 as a document retriever in a two-stage approach often results in underperformance since the parameters k and b are often  not optimal when using standard values. The problem of noise filtering of long documents was also addressed by using techniques like stopwords removal. However, as seen in \cite{Leveling2012OnTE}, this approach can have a negative effect on performance. The best pre-fetcher for long documents was found to be C-BERTs \citep{chalkidis2021regulatory}, which are trained on classifying documents using predefined labels.

    In general, regarding information retrieval approaches, there is a trade-off between performance and efficiency. While bi-encoders are faster and more efficient, they lag behind cross-encoders in terms of performance. Late-interaction models, such as ColBERT \citep{ColBERT2020} and, to an even greater extent, COIL \citep{Gao2021}, attempt to mitigate this trade-off by delivering good performance while maintaining low computational costs. \citet{Lin2023} offer a hybrid approach in the form of low-dimensional dense lexical representations, combining lexical representations with dense semantic representations to preserve effectiveness while improving query latency.

    \citet{thakur_beir_2021} proposed a novel evaluation benchmark for IR that encompasses a wide range of approaches, including lexical, such as BM25 \citep{INR-019}, sparse,  dense, late-interaction, including ColBERT \citep{ColBERT2020}, and re-ranking models, including BM25+CE, introduced by \citet{thakur_beir_2021}. BM25+CE reranks the top-100 retrieved hits from a first-stage BM25 model using  a 6-layer, 384-h MiniLM \citep{MiniLM2020} as the ross-attentional re-ranking model. They found that while BM25+CE is computationally expensive, it provides a robust baseline, whereas other models failed to achieve comparable performance, except of ColBERT which performed only slightly weaker. Their findings suggest that there is still much room for improvement in this area of NLP. Efficient retrieval of relevant information is crucial for many NLP tasks, and these results highlight the need for continued research in this area.

\clearpage\section{More Detailed Experimental Setup}
\label{app:add_experimental_setup}

\subsection{Resources Used}
\label{appendix-resources}
The experiments were performed on internal university clusters on NVIDIA GPUs with the following specifications: 24GB RTX3090, 32GB V100, 48GB A6000, and 80GB A100. We used an approximate total of 160, 20, and 2 GPU days for the text classification, text generation and information retrieval experiments.

\paragraph{Text Generation}
For inference and fine-tuning LLaMA-2 on the text generation tasks, we used the \href{https://docs.together.ai/docs}{Together API}.

\subsection{Hyperparameters}
\label{appendix-hyperparameters}
\paragraph{Text Classification}
For all models and datasets, a learning rate of 1e-5 was used without any tuning. Each experiment was executed with three random seeds (1-3), and the batch size was tailored for each task and corresponding computational resource. Seeds that yielded very high evaluation losses were considered failed and, therefore, excluded from the analysis. If the GPU memory was inadequate, we used gradient accumulation as a workaround to arrive at a final batch size of 64. The training was conducted with early stopping based on validation loss, maintaining a patience level of 5 epochs. Due to the considerable size of the judgment prediction dataset and the extended duration of the experiment, training was limited to a single epoch with evaluations after every 1000th step. To reduce costs, we utilized AMP mixed precision during the training and evaluation phases whenever it did not lead to overflows (e.g., mDeBERTa-v3). We established the max-sequence-length (determined by the product of max-segment-length and max-segments in the hierarchical setup \citep{aletras_predicting_2016, niklaus_swiss-judgment-prediction_2021, niklaus_empirical_2022}) based on whether we used Facts: 2048 (128 X 16), or Considerations: 4096 (128 X 32). 

\paragraph{Text Generation}
For the main \ac{CVG} dataset, we trained our mT5 models for only one epoch (because of the large training set) with a final batch size of 16, using gradient accumulation as needed. We performed evaluations every 1000 steps. For the smaller origin dataset, we increased the number of epochs to 100 and evaluated every 100 steps. For the \ac{LDS} task, we adjusted the training to 10 epochs. 

\paragraph{Information Retrieval}

\begin{figure}
\vspace{-4mm}
  \centering
  \includegraphics[width=0.7\textwidth]{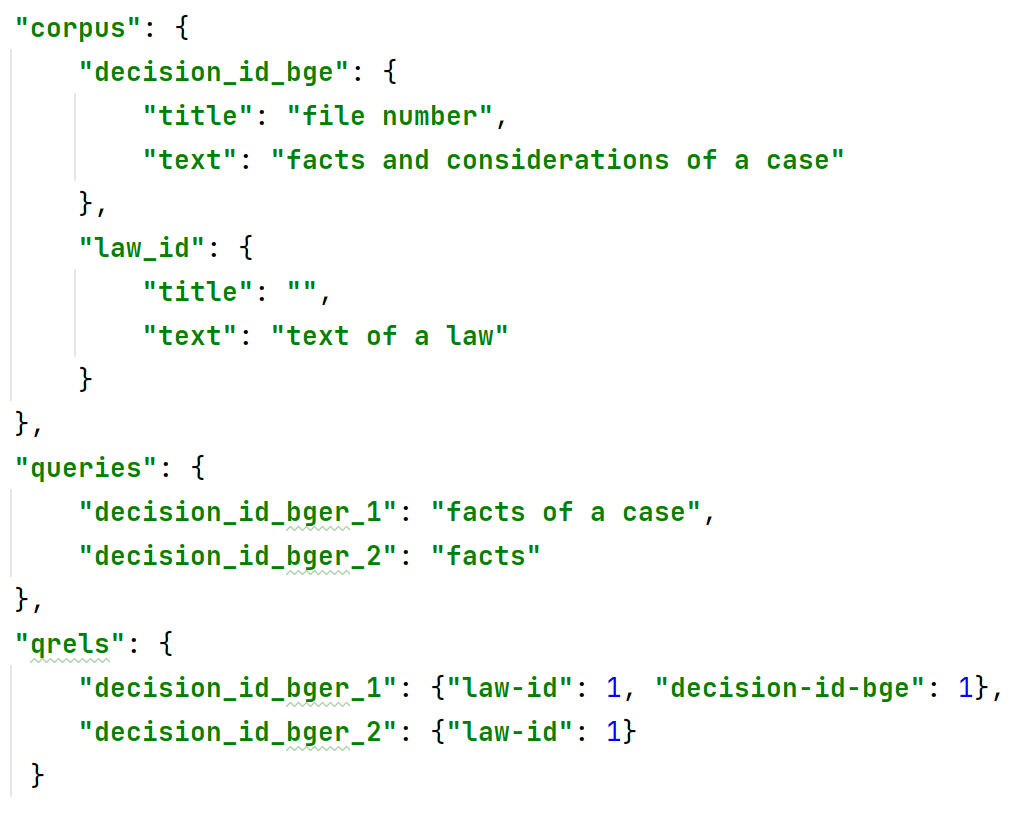}  
    \caption{Structure of corpus, queries and qrels for IR task}
    \label{fig:ir-data}
  \vspace{-3mm}
\end{figure}

For the BM25 model, we used the same parameters as used in the BEIR paper \citep{thakur_beir_2021}, chosen were k = 0.9 and b = 0.4.
For the SBERT model training, we employed the \href{https://github.com/beir-cellar/beir/blob/main/examples/retrieval/training/train_sbert.py}{BEIR} toolkit \citep{thakur_beir_2021}. Our training process was constrained by a maximum sequence length of 512 tokens. During the training phase, we completed a single epoch, comprising 5000 evaluation steps.

In the context of training with hard negative examples, we incorporated 5 negative examples for every query. The selection of these examples was based on the 5 highest-ranked erroneous predictions generated by the BM25 model. To facilitate training with these challenging negatives, we followed the guidelines provided by \citet{thakur_beir_2021}, utilizing the \href{https://github.com/beir-cellar/beir/blob/main/examples/retrieval/training/train_sbert_BM25_hardnegs.py}{Hardnegs} template.\\

\clearpage
\section{Detailed Data Description}
\label{app:detailed_data_description}

In this section, we provide additional information about the datasets. Table~\ref{tab:lextreme-task-dist} shows different classification task configurations, and Table~\ref{tab:metadata} provides additional information about general dataset metadata.

\begin{table*}[ht]
    \centering
    \vspace{-0ex}
        \caption{Task Configurations. Label names are \emph{Critical} (C), \emph{Non-critical} (NC), \emph{Critical-1} (C1) to \emph{Critical-4} (C4), \emph{Approval} (A), \emph{Dismissal} (D). For Law-Sub-Area we reported only the two most common labels \emph{Substantive Criminal} (SC), \emph{Criminal Procedure} (CP), and the two least common \emph{Intellectual Property} (IP), \emph{Other Fiscal} (OF). 
    Abbreviations: \textbf{Val}idation, \textbf{Con}siderations, \textbf{Fac}ts}
    \footnotesize
    \resizebox{\textwidth}{!}{%
    \begin{tabular}{lrrrrrrrrrrrrrrr}
         \toprule
         \textbf{Task Name} & \multicolumn{1}{l}{\textbf{Train}} & \multicolumn{3}{l}{\textbf{Labels Train}} && \multicolumn{1}{l}{\textbf{Val}} & \multicolumn{3}{l}{\textbf{Labels Val}}&& \multicolumn{1}{l}{\textbf{Test}} &  \multicolumn{4}{l}{\textbf{Labels Test}}\\
         \midrule
            && \textbf{C} & \textbf{NC} &&&& \textbf{C} & \textbf{NC} &&&& \textbf{C} & \textbf{NC} \\
            LD-Fac  & \textbf{75K} & 3K & 72K &-&-& \textbf{12K} & 580 & 13K &-&-& \textbf{26K} & 950 & 25K&-&-\\
            LD-Con & \textbf{91K} & 3K & 85K &-&-& \textbf{15K} & 580 & 13K &-&-& \textbf{32K} & 948 & 29K&-&-\\
        \midrule
           && \textbf{C-1} & \textbf{C-2} & \textbf{C-3} & \textbf{C-4} &&\textbf{C-1} & \textbf{C-2} & \textbf{C-3} & \textbf{C-4} && \textbf{C-1} & \textbf{C-2} & \textbf{C-3} & \textbf{C-4} \\
            Citation-Fac  & \textbf{2.5K} &782 & 626 & 585 & 513 & \textbf{563} &186 & 152 & 131 & 94 & \textbf{725} & 137 & 177 & 224 & 187\\
            Citation-Con  & \textbf{2.5K}&779 & 624 & 586 & 520 & \textbf{563} &186 & 154 & 131 & 92 & \textbf{723} &137 & 177 & 224 & 185\\
        \midrule
            && \textbf{D} & \textbf{A} &&&& \textbf{D} & \textbf{A} &&&& \textbf{D} & \textbf{A} \\
            Judgment-Fac  & \textbf{197K} & 135K & 62K &-&-& \textbf{37K} & 27K & 11K &-&-& \textbf{94K} & 67K & 27K &-&-\\
            Judgment-Con  & \textbf{188K} & 130K & 59K &-&-& \textbf{37K} & 26K & 11K &-&-& \textbf{92K} & 66K & 26K &-&-\\
        \midrule
            && \textbf{SC} & \textbf{CP} & \textbf{IP} & \textbf{OF} && \textbf{SC} & \textbf{CP} & \textbf{IP} & \textbf{OF} && \textbf{SC} & \textbf{CP} & \textbf{IP} & \textbf{OF} \\
            Law-Sub-Area-Fac  & \textbf{10K} & 3K & 3K & 6 & 2 & \textbf{9K} & 2K & 1K & 11 & 1 & \textbf{3K} & 1K & 509 & 5 & 1 \\
            Law-Sub-Area-Con  & \textbf{8K} & 2K & 1K & 6 & 2 & \textbf{7K} & 2K & 750 & 11 & 1& \textbf{3K} & 885 & 401 & 3 & 1 \\
        \bottomrule
    \end{tabular}
    }
    \label{tab:lextreme-task-dist}
\end{table*}

\begin{table*}[ht]
    \centering
    \vspace{-3ex}
    \caption{Listing of cantons, courts, chambers, law-areas}
    \footnotesize
    \resizebox{\textwidth}{!}{%
    \begin{tabular}{lll}
         \toprule
         \textbf{Metadata} &  \textbf{Number}&  \textbf{Examples}\\
         \midrule
            Cantons  & 26 (+1)& Aargau (AG), Bern (BE), Basel-Stadt (BS), Solothurn (SO), Ticino (Ti), Vaud (VD),... (+ Federation (CH))\\
            Courts  & 184 & Cantonal Bar Supervisory Authority, Supreme Court, administrative authorities, Tax Appeals Commission, \\
            & &Cantonal Court, Federal Administrative Court, ...\\
            Chambers  & 456 &  GR-UPL0-01, AG-VB-002, CH-BGer-011, ZH-OG-001, ZG-VG-004, VS-BZG-009, VD-TC-002, TI-TE-001, ...\\
            Law-Areas  & 4 & Civil, Criminal, Public, Social\\
            Languages  & 5 & German, French, Italian, Romansh, English \\
        \bottomrule
    \end{tabular}
    }
    \label{tab:metadata}
    \vspace{-2ex}
\end{table*}

\clearpage
\section{Additional Results}
\label{app:add_results}
\subsection{Court View Generation}
Table \ref{tab:court-view-generation-origin} shows the results of the \ac{CVG} task from both, the main and the origin dataset.
Table \ref{tab:cvg-results-lang} presents the \ac{CVG} task results split by language, detailing scores for German, French, and Italian.

\begin{table*}[!ht]
\centering
\vspace{-4mm}
\caption{Results of Court View Generation task. 'In Len' denotes input length in tokens. 
\textbf{Bold}: best within model; \underline{underlined}: best overall.
}
\resizebox{1\textwidth}{!}{
\begin{tabular}{lrrrrrrrrr}
\toprule
\multirow{2}{*}{\textbf{Model}} & \multirow{2}{*}{\textbf{In Len} $\uparrow$} & \multicolumn{4}{c}{\textbf{Main Scores} $\uparrow$} & \multicolumn{4}{c}{\textbf{Origin Scores} $\uparrow$} \\
\cmidrule(lr){3-6}
\cmidrule(lr){7-10}
& & \textbf{BERT} & \textbf{BLEU} & \textbf{MET} & \textbf{R1 / R2 / RL} & \textbf{BERT} & \textbf{BLEU} & \textbf{MET} & \textbf{R1 / R2 / RL} \\
\midrule
mT5\textsubscript{Large} & {2048} & \textbf{\underline{75.74}} & \textbf{\underline{66.92}} & \textbf{\underline{34.44}} & \textbf{\underline{34.91}} / \textbf{\underline{15.58}} / \textbf{\underline{33.53}} & 76.24 & \textbf{62.59} & 32.25 & 34.80 / 16.11 / 33.58 \\
mT5\textsubscript{Large} & 1024 & 75.56 & 66.68 & 34.02 & 34.26 / 14.72 / 32.87 & 74.99 & 58.35 & 31.06 & 33.35 / 14.80 / 32.16 \\
mT5\textsubscript{Large} & 512 & 75.27 & 66.12 & 33.48 & 33.61 / 14.26 / 32.21 & \textbf{\underline{76.33}} & 62.08 & \textbf{32.92} & \textbf{36.61} / \textbf{18.17} / \textbf{34.84} \\
\midrule
mT5\textsubscript{Base} & 2048 & 75.01 & 65.48 & 32.89 & 33.23 / 13.57 / 31.89 & 75.99 & \textbf{\underline{63.39}} & \textbf{\underline{34.15}} & 36.48 / \textbf{\underline{18.81}} / 35.58 \\
mT5\textsubscript{Base} & {1024} & \textbf{75.15} & \textbf{65.73} & \textbf{33.15} & \textbf{33.49} / \textbf{13.96} / \textbf{32.18} & 76.07 & 60.99 & 33.50 & \textbf{\underline{37.68}} / 18.79 / \textbf{\underline{36.58}} \\
mT5\textsubscript{Base} & 512 & 74.89 & 65.55 & 32.66 & 32.66 / 13.16 / 31.35 & \textbf{76.08} & 62.21 & 32.80 & 36.40 / 17.58 / 34.98 \\
\midrule
mT5\textsubscript{Small} & {2048} & \textbf{74.13} & \textbf{63.97} & \textbf{30.96} & \textbf{31.29} / \textbf{11.01} / \textbf{29.90} & 75.23 & 56.59 & 30.71 & 34.68 / 13.64 / 33.24 \\
mT5\textsubscript{Small} & 1024 & 74.00 & 63.70 & 30.68 & 31.05 / 10.77 / 29.64 & \textbf{75.75} & 58.99 & 31.17 & 34.62 / 14.25 / \textbf{33.91} \\
mT5\textsubscript{Small} & 512 & 73.92 & 63.83 & 30.57 & 30.58 / 10.35 / 29.20 & 75.63 & \textbf{61.12} & \textbf{32.33} & \textbf{35.16} / \textbf{14.45} / 33.72 \\
\bottomrule
\end{tabular}
}
\label{tab:court-view-generation-origin}
\end{table*}

\begin{table*}[ht]
\centering
\caption{Results of Court View Generation task. 'In Len' denotes input length in tokens. \textbf{Bold}: best within setup; \underline{underlined}: best overall. (*) These models were fine-tuned on only 1'000 samples for 3 epochs. All models, except for the mT5 models, were evaluated on the validation set. }
\label{tab:court-view-generation-app}
\resizebox{\textwidth}{!}{%
\begin{tabular}{lrrrrrr}
\toprule
\textbf{Model} & \textbf{Setup} & \textbf{In Len} $\uparrow$ & \textbf{BERT} $\uparrow$ & \textbf{BLEU} $\uparrow$ & \textbf{MET} $\uparrow$ & \textbf{R1 / R2 / RL} $\uparrow$ \\
\midrule
mT5\textsubscript{Large} & Fine-tuned & 2048 & \underline{\textbf{75.74}} & \underline{\textbf{66.92}} & \underline{\textbf{34.44}} & \underline{\textbf{34.91}} / \textbf{15.58} / \underline{\textbf{33.53}} \\
mT5\textsubscript{Large} & Fine-tuned & 1024 & 75.56 & 66.68 & 34.02 & 34.26 / 14.72 / 32.87 \\
mT5\textsubscript{Large} & Fine-tuned & 512 & 75.27 & 66.12 & 33.48 & 33.61 / 14.26 / 32.21 \\
\midrule
mT5\textsubscript{Base} & Fine-tuned & 2048 & 75.01 & 65.48 & 32.89 & 33.23 / 13.57 / 31.89 \\
mT5\textsubscript{Base} & Fine-tuned & 1024 & 75.15 & 65.73 & 33.15 & 33.49 / 13.96 / 32.18 \\
mT5\textsubscript{Base} & Fine-tuned & 512 & 74.89 & 65.55 & 32.66 & 32.66 / 13.16 / 31.35 \\
\midrule
mT5\textsubscript{Small} & Fine-tuned & 2048 & 74.13 & 63.97 & 30.96 & 31.29 / 11.01 / 29.90 \\
mT5\textsubscript{Small} & Fine-tuned & 1024 & 74.00 & 63.70 & 30.68 & 31.05 / 10.77 / 29.64 \\
mT5\textsubscript{Small} & Fine-tuned & 512 & 73.92 & 63.83 & 30.57 & 30.58 / 10.35 / 29.20 \\
\midrule
\midrule
GPT-3.5-Turbo & Fine-tuned\textsuperscript{\tiny{*}} & 2048 & 72.31 & 62.23 & 28.08 & 26.06 / 7.19 / 24.54 \\
LLaMA-2-13B Chat & Fine-tuned\textsuperscript{\tiny{*}} & 2048 & \textbf{74.22} & \textbf{63.51} & \textbf{33.33} & \textbf{34.36} / \underline{\textbf{16.68}} / \textbf{33.20} \\
LLaMA-2-7B Chat & Fine-tuned\textsuperscript{\tiny{*}} & 2048 & 73.27 & 62.34 & 31.6 & 32.31 / 14.47 / 31.38 \\
\midrule
\midrule
GPT-4 & 1-shot & 2048 & 70.39 & 59.69 & 24.63 & 23.87 / 4.64 / 22.32 \\
GPT-3.5-Turbo-16K & 1-shot & 8192 & \textbf{70.86} & 59.89 & \textbf{25.63} & 24.97 / 5.44 / 23.50 \\
GPT-3.5-Turbo-16K & 1-shot & 2048 & 70.73 & 59.92 & 25.55 & 24.95 / 5.43 / 23.49 \\
\midrule
\midrule
GPT-4 & 0-shot & 2048 & 69.41 & 58.16 & 23.25 & 22.61 / 3.95 / 21.10 \\
GPT-3.5-Turbo-16K & 0-shot & 8192 & 67.93 & 56.87 & 22.62 & 21.21 / 3.56 / 19.88 \\
GPT-3.5-Turbo-16K & 0-shot & 2048 & 67.80 & 56.52 & 22.32 & 20.99 / 3.46 / 19.74 \\
LLaMA-2-70B Chat & 0-shot & 2048 & 66.78 & 53.04 & 19.13 & 18.48 / 3.21 / 17.35 \\
LLaMA-2-13B Chat & 0-shot & 2048 & 67.23 & 55.01 & 20.18 & 19.76 / 3.26 / 18.57 \\
LLaMA-2-7B Chat & 0-shot & 2048 & 63.74 & 40.62 & 11.29 & 10.75 / 1.62 / 10.13 \\
\bottomrule
\end{tabular}%
}
\end{table*}

\begin{table*}[ht]
\centering
\caption{Results of the \ac{CVG} task split by language. Scores are presented in the order: German, French, and Italian.}
\label{tab:cvg-results-lang}
\renewcommand{\arraystretch}{1.5}  
\resizebox{\textwidth}{!}{%
\begin{tabular}{lrrrrrrr}
\toprule
\textbf{Model} & \textbf{Setup} & \textbf{BERT} $\uparrow$ & \textbf{BLEU} $\uparrow$ & \textbf{MET} $\uparrow$ & \textbf{R1} $\uparrow$ & \textbf{R2} $\uparrow$ & \textbf{RL} $\uparrow$ \\
\midrule
GPT-3.5-Turbo & Fine-tuned & 71.89 / 73.01 / 71.17 & 62.29 / 62.15 / 62.29 & 29.02 / 27.43 / 25.38 & 25.68 / 26.97 / 23.48 & 7.31 / 7.49 / 4.64 & 24.60 / 24.94 / 21.77 \\
LLaMA-2-13B-Chat & Fine-tuned & 75.06 / 73.76 / 71.03 & 66.24 / 61.05 / 58.81 & 36.44 / 30.69 / 26.95 & 36.22 / 33.37 / 27.12 & 19.51 / 14.55 / 9.44 & 35.35 / 31.92 / 25.63 \\
\midrule
GPT-4 & 1-shot & 69.73 / 71.31 / 69.22 & 58.83 / 60.89 / 58.23 & 24.20 / 25.48 / 21.76 & 22.40 / 25.86 / 21.85 & 3.56 / 6.15 / 2.88 & 21.26 / 23.82 / 20.35 \\
\midrule
GPT-4 & 0-shot & 69.12 / 69.87 / 68.54 & 57.90 / 58.58 / 57.18 & 23.25 / 23.42 / 21.83 & 21.48 / 24.14 / 21.08 & 2.97 / 5.26 / 2.78 & 20.25 / 22.32 / 19.44 \\
LLaMA-2-13B-Chat & 0-shot & 66.72 / 67.94 / 66.65 & 53.20 / 56.73 / 57.61 & 19.86 / 20.60 / 19.98 & 18.22 / 21.50 / 20.44 & 2.18 / 4.59 / 3.03 & 17.39 / 19.92 / 18.95 \\
\bottomrule
\end{tabular}%
}
\end{table*}

\subsection{\acf{LDS}}
Table \ref{tab:lds-app} shows all results of the \ac{LDS} task. Table \ref{tab:lds-results-lang} presents the \ac{LDS} task results split by language, detailing scores for German, French, and Italian.

\begin{table*}[ht]
\centering
\caption{Results of \acf{LDS} task. 'In Len' denotes input length in tokens. \textbf{Bold}: best within setup; \underline{underlined}: best overall. All models, except for the mT5 models, were evaluated on the validation set. }
\label{tab:lds-app}
\resizebox{\textwidth}{!}{%
\begin{tabular}{lrrrrrrr}
\toprule
\textbf{Model} & \textbf{Setup} & \textbf{In Len} $\uparrow$ & \textbf{BERT} $\uparrow$ & \textbf{BLEU} $\uparrow$ & \textbf{MET} $\uparrow$ & \textbf{R1 / R2 / RL} $\uparrow$ \\
\midrule
mT5\textsubscript{Large} & Fine-tuned & 4096 & \textit{OOM} & \textit{OOM} & \textit{OOM} & \textit{OOM} \\
mT5\textsubscript{Large} & Fine-tuned & 2048 & 73.10 & 27.21 & 21.88 & 31.47 / 12.22 / 29.94 \\
mT5\textsubscript{Large} & Fine-tuned & 512 & 70.67 & 26.89 & 18.31 & 24.76 / \hspace{1.5mm}6.15 / 23.48 \\
\midrule
mT5\textsubscript{Base} & Fine-tuned & 4096 & \textbf{73.33} & \textbf{30.81} & \textbf{23.50} & \underline{\textbf{32.43}} / \underline{\textbf{12.78}} / \underline{\textbf{30.87}} \\
mT5\textsubscript{Base} & Fine-tuned & 2048 & 72.45 & 30.13 & 21.94 & 30.09 / 10.79 / 28.71 \\
mT5\textsubscript{Base} & Fine-tuned & 512 & 70.60 & 27.10 & 18.31 & 24.72 / \hspace{1.5mm}6.15 / 23.55 \\
\midrule
mT5\textsubscript{Small} & Fine-tuned & 4096 & 72.04 & 28.68 & 21.29 & 29.61 / 10.31 / 28.12 \\
mT5\textsubscript{Small} & Fine-tuned & 2048 & 71.38 & 24.64 & 19.28 & 27.88 / \hspace{1.5mm}9.19 / 26.54 \\
mT5\textsubscript{Small} & Fine-tuned & 512 & 69.66 & 20.73 & 15.95 & 22.91 / \hspace{1.5mm}5.36 / 21.85 \\
\midrule\midrule
GPT-4 & 1-shot & 4096 & \underline{\textbf{73.55}} & 47.75 & 34.72 & 30.82 / 9.68 / 28.89 \\
GPT-3.5-Turbo-16K & 1-shot & 8192 & 72.92 & 46.15 & 33.68 & 29.69 / 9.47 / 27.97 \\
GPT-3.5-Turbo-16K & 1-shot & 4096 & 72.89 & 45.21 & 32.76 & 29.69 / 9.25 / 27.94 \\
\midrule\midrule
GPT-4 & 0-shot & 4096 & \textbf{71.56} & 48.35 & 32.97 & 26.52 / \textbf{8.93} / 24.51 \\
GPT-3.5-Turbo-16K & 0-shot & 4096 & 70.28 & 46.08 & 30.60 & 25.18 / 7.58 / 23.59 \\
\bottomrule
\end{tabular}%
}
\end{table*}

\begin{table*}[ht]
\centering
\caption{Results of the \ac{LDS} task split by language. Scores are presented in the order: German, French, and Italian.}
\label{tab:lds-results-lang}
\renewcommand{\arraystretch}{1.5}
\resizebox{\textwidth}{!}{%
\begin{tabular}{lrrrrrrrr}
\toprule
\textbf{Model} & \textbf{Setup} & \textbf{BERT} $\uparrow$ & \textbf{BLEU} $\uparrow$ & \textbf{MET} $\uparrow$ & \textbf{R1} $\uparrow$ & \textbf{R2} $\uparrow$ & \textbf{RL} $\uparrow$ \\
\midrule
GPT-4 & 1-shot & 73.89 / 72.84 / 74.08 & 49.32 / 45.85 / 38.08 & 36.50 / 32.01 / 28.75 & 31.14 / 30.03 / 32.43 & 9.91 / 9.13 / 10.82 & 29.25 / 27.99 / 30.84 \\
GPT-3.5 & 1-shot & 73.26 / 72.21 / 72.50 & 47.44 / 41.50 / 39.08 & 35.23 / 28.48 / 27.37 & 30.64 / 28.22 / 26.11 & 9.46 / 8.87 / 8.96 & 28.90 / 26.46 / 24.34 \\
\midrule
GPT-4 & 0-shot & 72.65 / 69.56 / 70.95 & 49.17 / 46.96 / 46.89 & 36.26 / 27.39 / 27.10 & 28.27 / 23.45 / 24.25 & 10.88 / 5.36 / 7.74 & 26.26 / 21.42 / 22.51 \\
GPT-3.5 & 0-shot & 71.25 / 68.39 / 69.84 & 47.39 / 43.68 / 44.17 & 33.31 / 25.51 / 27.91 & 27.11 / 21.73 / 21.86 & 9.22 / 4.64 / 4.75 & 25.47 / 20.22 / 20.48 \\
\bottomrule
\end{tabular}%
}
\end{table*}

\subsection{Text Classification}

Table~\ref{tab:config_aggregate_scores_with_std} shows more detailed results on the text classification datasets including standard deviations across seeds.

\begin{table*}[ht]
    \centering
    \caption{Configuration aggregate scores with standard deviations on the test set. The macro-F1 scores are provided.}
    \footnotesize
    \resizebox{\textwidth}{!}{%
    \begin{tabular}{lllllllllr}
\toprule
                \bf Model &                      \bf CPB-F &                      \bf CPB-C &                      \bf CPC-F &                      \bf CPC-C &                      \bf SLAP-F &                      \bf \bf SLAP-C &                       \bf JP-F &                       \bf JP-C &  \bf Agg. \\
\midrule
               MiniLM & 54.7\textsubscript{+/-1.9} & 65.8\textsubscript{+/-1.6} &  9.8\textsubscript{+/-2.8} & 20.8\textsubscript{+/-3.0} &  59.7\textsubscript{+/-3.8} &  61.1\textsubscript{+/-3.7} & 58.1\textsubscript{+/-0.4} & 78.5\textsubscript{+/-2.3} &  32.4 \\
           DistilmBERT & 56.2\textsubscript{+/-0.5} & 65.4\textsubscript{+/-1.7} & 19.6\textsubscript{+/-1.1} & 22.1\textsubscript{+/-0.4} & 63.7\textsubscript{+/-11.7} &  65.9\textsubscript{+/-6.4} & 59.9\textsubscript{+/-0.9} & 75.5\textsubscript{+/-3.3} &  42.1 \\
          mDeBERTa-v3 & 55.1\textsubscript{+/-2.0} & 69.8\textsubscript{+/-2.8} & 21.0\textsubscript{+/-3.6} & 17.5\textsubscript{+/-4.4} &  63.8\textsubscript{+/-6.3} &  59.3\textsubscript{+/-7.6} & 60.6\textsubscript{+/-0.9} & 77.9\textsubscript{+/-2.6} &  40.2 \\
           XLM-R\textsubscript{Base} & 57.2\textsubscript{+/-1.5} & 65.9\textsubscript{+/-3.2} & 21.3\textsubscript{+/-1.5} & 23.7\textsubscript{+/-1.9} & 67.2\textsubscript{+/-15.9} &  73.4\textsubscript{+/-2.5} & 60.9\textsubscript{+/-0.6} & 79.7\textsubscript{+/-2.5} &  44.6 \\
          XLM-R\textsubscript{Large} & 56.4\textsubscript{+/-1.8} & 67.9\textsubscript{+/-1.9} & 24.4\textsubscript{+/-7.2} & 29.1\textsubscript{+/-2.7} &  65.1\textsubscript{+/-8.5} &  78.9\textsubscript{+/-4.6} & 60.8\textsubscript{+/-0.6} & 80.9\textsubscript{+/-2.4} &  48.6 \\
           X-MOD\textsubscript{Base} & 56.6\textsubscript{+/-1.8} & 67.8\textsubscript{+/-2.9} & 20.0\textsubscript{+/-3.0} & 20.6\textsubscript{+/-3.5} & 63.9\textsubscript{+/-10.1} &  64.4\textsubscript{+/-7.0} & 60.5\textsubscript{+/-0.6} & 79.1\textsubscript{+/-2.6} &  41.9 \\
    SwissBERT\textsubscript{(xlm-vocab)} & 56.9\textsubscript{+/-0.7} & 67.3\textsubscript{+/-4.7} & 25.7\textsubscript{+/-8.3} & 23.0\textsubscript{+/-4.0} &  61.5\textsubscript{+/-9.5} &  73.2\textsubscript{+/-2.1} & 61.4\textsubscript{+/-0.6} & 79.4\textsubscript{+/-2.5} &  46.1 \\
    \midrule
     mT5\textsubscript{Small} & 52.2\textsubscript{+/-1.9} & 62.1\textsubscript{+/-5.2} & 13.2\textsubscript{+/-2.4} & 17.9\textsubscript{+/-1.7} & 53.1\textsubscript{+/-13.8} & 60.9\textsubscript{+/-15.9} & 58.9\textsubscript{+/-1.0} & 74.2\textsubscript{+/-3.6} &  34.4 \\
      mT5\textsubscript{Base} & 52.1\textsubscript{+/-1.6} & 61.5\textsubscript{+/-3.9} & 14.0\textsubscript{+/-2.8} & 19.7\textsubscript{+/-1.6} & 58.4\textsubscript{+/-17.2} & 61.8\textsubscript{+/-16.8} & 54.5\textsubscript{+/-1.5} & 72.0\textsubscript{+/-3.1} &  35.9 \\
    \midrule
           BLOOM-560m & 53.0\textsubscript{+/-1.7} & 61.7\textsubscript{+/-4.1} & 10.7\textsubscript{+/-3.7} &  8.0\textsubscript{+/-3.5} & 52.6\textsubscript{+/-10.7} &  53.2\textsubscript{+/-8.5} & 60.5\textsubscript{+/-0.7} & 73.4\textsubscript{+/-7.2} &  24.9 \\
    \midrule
      Legal-ch-R\textsubscript{Base} & 57.7\textsubscript{+/-1.6} & 70.5\textsubscript{+/-2.3} & 16.2\textsubscript{+/-5.8} & 20.1\textsubscript{+/-5.6} &  77.0\textsubscript{+/-3.6} &  79.7\textsubscript{+/-0.9} & 64.0\textsubscript{+/-1.3} & 86.4\textsubscript{+/-1.9} &  40.9 \\
     Legal-ch-R\textsubscript{Large} & 55.9\textsubscript{+/-2.2} & 68.9\textsubscript{+/-2.1} & 25.8\textsubscript{+/-7.8} & 16.3\textsubscript{+/-8.7} &  76.9\textsubscript{+/-2.3} &  84.9\textsubscript{+/-9.7} & 62.8\textsubscript{+/-0.9} & 87.1\textsubscript{+/-2.2} &  43.3 \\
     Legal-ch-LF\textsubscript{Base} & 58.1\textsubscript{+/-2.1} & 70.8\textsubscript{+/-2.9} & 21.4\textsubscript{+/-2.9} & 17.4\textsubscript{+/-8.6} & 80.1\textsubscript{+/-12.7} &  77.1\textsubscript{+/-4.8} & 65.4\textsubscript{+/-1.7} & 86.4\textsubscript{+/-1.8} &  42.5 \\
\bottomrule
\end{tabular}

    }
    \label{tab:config_aggregate_scores_with_std}
    \end{table*}

\subsubsection{Language specific results}

Table~\ref{tab:config_aggregate_scores_language_specific} shows more detailed results on the text classification datasets language specific scores.

SwissBERT, where pretraining tokens were most focused towards the dominant language German seems to have quite even results, with scores in Italian even being the highest. 
Models trained on CC100 (MiniLM, mDeBERTa, XLM-R and X-MOD) showed mixed results. For all models, German performance was very close to French performance. MiniLM, mDeBERTa, and X-MOD showed Italian underperformance whereas XLM-R showed very strong performance in Italian, especially the large variant. 
Even though the Legal-ch-R models are based on XLM-R, they show underperformance in Italian, but similar performance between French and German. 
mT5 models performed well in French, the base variant additionally also performed well on Italian. 
BLOOM was much better in French than in other languages, not surprising given it did not have German and Italian in the pretraining data. 

Overall, there only seems to be a weak trend connecting higher percentage of a given language in the pretraining corpus leading to better downstream results in that language.

\begin{table*}[ht]
    \centering
    \caption{Configuration aggregate scores. The macro-F1 scores from the language-specific subsets of the test set are provided. 
    }
    \footnotesize
    \resizebox{\textwidth}{!}{%
    \begin{tabular}{lccccccccc}
\toprule
\bf{Model} &          \bf{CPB-F} &          \bf{CPB-C} &          \bf{CPC-F} &          \bf{CPC-C} &         \bf{SLAP-F} &         \bf{SLAP-C} &           \bf{JP-F} &           \bf{JP-C} &           \bf{Agg.} \\
Languages & \hspace{1.2mm}de\hspace{1.3mm} / \hspace{1.2mm}fr\hspace{1.3mm} / \hspace{1.2mm}it\hspace{1.3mm}& \hspace{1.2mm}de\hspace{1.3mm} / \hspace{1.2mm}fr\hspace{1.3mm} / \hspace{1.2mm}it\hspace{1.3mm}& \hspace{1.2mm}de\hspace{1.3mm} / \hspace{1.2mm}fr\hspace{1.3mm} / \hspace{1.2mm}it\hspace{1.3mm}& \hspace{1.2mm}de\hspace{1.3mm} / \hspace{1.2mm}fr\hspace{1.3mm} / \hspace{1.2mm}it\hspace{1.3mm}& \hspace{1.2mm}de\hspace{1.3mm} / \hspace{1.2mm}fr\hspace{1.3mm} / \hspace{1.2mm}it\hspace{1.3mm}& \hspace{1.2mm}de\hspace{1.3mm} / \hspace{1.2mm}fr\hspace{1.3mm} / \hspace{1.2mm}it\hspace{1.3mm}& \hspace{1.2mm}de\hspace{1.3mm} / \hspace{1.2mm}fr\hspace{1.3mm} / \hspace{1.2mm}it\hspace{1.3mm}& \hspace{1.2mm}de\hspace{1.3mm} / \hspace{1.2mm}fr\hspace{1.3mm} / \hspace{1.2mm}it\hspace{1.3mm}& \hspace{1.2mm}de\hspace{1.3mm} / \hspace{1.2mm}fr\hspace{1.3mm} / \hspace{1.2mm}it\hspace{1.3mm}\\
\midrule
MiniLM                &  57.5 / 53.9 / 52.9 &  68.1 / 65.4 / 64.2 &   12.1 / 13.1 / 6.8 &  24.6 / 21.9 / 17.3 &  55.3 / 60.0 / 64.5 &  57.6 / 60.0 / 66.5 &  57.7 / 58.1 / 58.7 &  77.8 / 81.7 / 76.1 &  36.1 / 36.6 / 26.6 \\
DistilmBERT            &  56.3 / 55.6 / 56.8 &  67.8 / 63.9 / 64.7 &  20.2 / 18.2 / 20.7 &  22.6 / 21.6 / 22.2 &  50.9 / 67.2 / 79.6 &  57.8 / 68.8 / 72.9 &  60.5 / 60.7 / 58.6 &  75.6 / 79.8 / 71.7 &  41.4 / 41.3 / 43.6 \\
mDeBERTa-v3           &  57.6 / 55.1 / 52.7 &  73.9 / 68.1 / 67.7 &  25.4 / 22.8 / 16.8 &  22.1 / 21.6 / 12.6 &  59.7 / 60.1 / 73.3 &  59.4 / 69.9 / 51.3 &  59.5 / 61.8 / 60.4 &  78.8 / 80.7 / 74.5 &  44.8 / 43.8 / 33.9 \\
XLM-R\textsubscript{Base}            &  59.4 / 56.3 / 56.0 &  70.2 / 65.4 / 62.5 &  20.0 / 20.6 / 23.5 &  26.5 / 22.1 / 23.1 &  54.5 / 64.6 / 92.2 &  71.5 / 71.9 / 77.1 &  60.9 / 61.6 / 60.2 &  79.9 / 82.8 / 76.7 &  44.5 / 43.3 / 46.2 \\
XLM-R\textsubscript{Large}           &  58.4 / 56.8 / 54.1 &  70.5 / 67.3 / 66.0 &  22.5 / 19.7 / 36.2 &  26.7 / 28.2 / 33.0 &  65.5 / 56.1 / 77.0 &  73.7 / 78.8 / 84.9 &  60.8 / 61.6 / 60.1 &  81.3 / 83.7 / 77.9 &  46.9 / 45.1 / 54.9 \\
X-MOD\textsubscript{Base}            &  59.0 / 56.2 / 54.8 &  71.1 / 68.7 / 64.1 &  19.8 / 17.2 / 24.4 &  23.2 / 24.2 / 16.4 &  55.7 / 61.1 / 79.3 &  63.1 / 74.5 / 57.8 &  60.2 / 61.3 / 60.0 &  79.4 / 82.4 / 76.0 &  42.6 / 42.1 / 40.9 \\
SwissBERT\textsubscript{(xlm-vocab)} &  57.6 / 55.9 / 57.3 &  72.4 / 69.3 / 61.2 &  23.8 / 20.3 / 39.4 &  28.5 / 24.0 / 18.7 &  50.0 / 66.8 / 72.4 &  71.2 / 72.4 / 76.1 &  61.1 / 62.2 / 60.9 &  79.8 / 82.5 / 76.3 &  46.7 / 44.4 / 47.3 \\
\midrule
mT5\textsubscript{Small} & 54.8 / 51.7 / 50.3 & 69.2 / 61.9 / 56.4 & 14.2 / 16.2 / 10.5 & 15.9 / 18.1 / 20.2 & 37.6 / 67.6 / 66.2 & 51.7 / 86.6 / 54.4 & 59.8 / 59.5 / 57.5 & 75.9 / 77.7 / 69.4 & 33.1 / 38.3 / 32.3 \\
mT5\textsubscript{Base} & 54.1 / 52.1 / 50.3 & 66.4 / 61.9 / 56.8 & 10.6 / 16.3 / 16.9 & 18.7 / 18.7 / 22.1 & 40.4 / 80.8 / 70.6 & 47.2 / 87.9 / 62.7 & 56.2 / 55.0 / 52.6 & 73.4 / 75.3 / 67.9 & 31.0 / 39.0 / 38.9 \\
\midrule
BLOOM-560m & 55.1 / 53.2 / 50.9 & 64.6 / 65.3 / 56.2 &  12.6 / 16.1 / 7.1 &   9.5 / 13.6 / 5.1 & 39.9 / 61.8 / 63.2 & 42.7 / 61.8 / 59.6 & 59.8 / 61.5 / 60.3 & 68.8 / 84.2 / 69.1 & 26.8 / 34.8 / 18.4 \\
\midrule
Legal-ch-R\textsubscript{Base}       &  59.3 / 58.4 / 55.5 &  73.8 / 69.4 / 68.6 &  24.3 / 20.5 / 10.5 &  26.2 / 25.3 / 14.0 &  79.8 / 72.1 / 79.6 &  80.8 / 79.6 / 78.6 &  62.5 / 65.8 / 63.9 &  87.6 / 87.9 / 83.7 &  49.4 / 46.3 / 31.7 \\
Legal-ch-R\textsubscript{Large}      &  58.3 / 55.7 / 53.6 &  71.9 / 68.5 / 66.7 &  23.0 / 21.3 / 38.7 &   28.5 / 26.0 / 9.0 &  74.0 / 77.4 / 79.6 &  77.6 / 80.5 / 99.5 &  61.6 / 63.9 / 63.0 &  88.6 / 88.7 / 84.1 &  49.0 / 47.0 / 36.2 \\
Legal-ch-LF\textsubscript{Base}      &  60.7 / 58.3 / 55.5 &  74.8 / 70.0 / 67.8 &  25.3 / 21.5 / 18.4 &   29.2 / 26.7 / 9.9 &  75.9 / 70.3 / 99.7 &   82.2 / 72.6 / 100 &  63.1 / 67.1 / 66.0 &  87.5 / 87.9 / 84.0 &  51.2 / 47.2 / 31.0 \\
\bottomrule
\end{tabular}
}
\label{tab:config_aggregate_scores_language_specific}
\end{table*}

\begin{table*}[ht]
    \centering
    \caption{Configuration aggregate scores on the validation set. The macro-F1 scores are provided. The highest values are in bold. It is important to note that the scores presented here are calculated as the harmonic mean over multiple seeds.}
    \footnotesize
    \resizebox{\textwidth}{!}{%
    \begin{tabular}{lrrrrrrrrrrrrrrrrrrrrrrrrr}
\toprule
           \bf{Model} &                 \bf{CPB-F} &                 \bf{CPB-C} &                 \bf{CPC-F} &                 \bf{CPC-C} &                 \bf{SLAP-F} &                 \bf{SLAP-C} &                \bf{JP-F} &                \bf{JP-C} &  \bf{Agg.} \\
\midrule
               MiniLM                &       59.1 &       71.0 &       14.9 &       36.9 &        73.8 &        78.9 &       60.9 &       81.4 &       44.4 \\
DistilmBERT            &       59.6 &       70.1 &       26.3 &       35.8 &        74.1 &        90.3 &       60.8 &       78.8 &       53.1 \\
mDeBERTa-v3           &       60.1 &       73.0 &       30.4 &       36.0 &        77.4 &        82.0 &       63.3 &       81.1 &       55.5 \\
XLM-R\textsubscript{Base}            &       60.1 &       70.5 &       26.9 &       38.5 &        78.7 &        92.2 &       62.9 &       82.5 &       55.0 \\
XLM-R\textsubscript{Large}           &       60.5 &       71.7 &       27.2 &       39.7 &        74.0 &        96.2 &       63.2 &       83.1 &       55.5 \\
X-MOD\textsubscript{Base}            &       57.1 &       71.0 &       27.0 &       33.4 &        81.3 &        94.4 &       62.5 &       82.1 &       53.5 \\
SwissBERT\textsubscript{(xlm-vocab)} &       59.0 &       72.1 &       29.4 &       38.8 &        85.6 &        95.5 &       62.6 &       82.3 &       56.8 \\
\midrule
mT5\textsubscript{Small}      &       54.8 &       66.1 &       26.3 &       32.5 &        84.7 &        88.0 &       60.1 &       77.3 &       51.6 \\
mT5\textsubscript{Base}       &       55.7 &       64.4 &       24.3 &       29.3 &        83.0 &        82.6 &       47.7 &       66.1 &       47.3 \\
\midrule
BLOOM-560m            &       52.2 &       64.3 &       20.1 &       21.8 &        78.5 &        82.4 &       60.5 &       76.7 &       43.3 \\
\midrule
Legal-ch-R\textsubscript{Base}       &       61.2 &  \bf{73.6} &       27.7 &       41.0 &        99.0 &        96.1 &       65.3 &       88.8 &       58.2 \\
Legal-ch-R\textsubscript{Large}      &  \bf{61.8} &       73.5 &       29.8 &       32.0 &   \bf{99.3} &   \bf{98.8} &       65.1 &  \bf{89.7} &       56.6 \\
Legal-ch-LF\textsubscript{Base}      &       59.4 &       72.7 &  \bf{32.2} &  \bf{42.5} &        99.1 &        98.1 &  \bf{67.0} &       89.1 &  \bf{60.8} \\
\bottomrule
\end{tabular}

    }
    \label{tab:config_aggregate_scores_of_validation_set}
\end{table*}


\subsection{Information Retrieval}

\begin{table*}[ht]
    \centering
    \footnotesize
    \caption{Results IR: using a subset of 100 queries and only relevant documents in the corpus resulting in an easier task}
    \resizebox{\textwidth}{!}{
    \begin{tabular}{llllSSSSSS}
         \midrule
         \textbf{Model} & \textbf{Additional} &  \textbf{$Rcap@1$} $\uparrow$ & \textbf{$Rcap@10$} $\uparrow$ & \textbf{$Rcap@100$} $\uparrow$  & \textbf{$NDCG@1$} $\uparrow$  & \textbf{$NDCG@10$} $\uparrow$ & \textbf{$NDCG@100$} $\uparrow$\\
         \midrule
            Train + Evaluate S-BERT & sbert-legal-xlm-roberta-base &  32.32 & 32.34 & 81.77 & 32.32 & 30.89 & 49.11 \\
            Train + Evaluate S-BERT & sbert-legal-swiss-roberta-base & \textbf{36.36} & \textbf{35.68} & 76.03 & \textbf{36.36} & \textbf{34.54} & 49.90 \\
            Train + Evaluate S-BERT & distiluse-base-multilingual-cased-v1 & 22.22 & 30.35 & 84.38 & 22.22 & 25.72 & 48.66 \\
            Evaluate S-BERT & distiluse-base-multilingual-cased-v1 & 8.08 & 11.83 & 43.35 & 8.08 & 10.55 & 21.56 \\
            Train(HN) + Evaluate S-BERT & distiluse-base-multilingual-cased-v1 & 27.27 & 33.94 & \textbf{86.81} & 27.27 & 30.09 & \textbf{52.03} \\
            \midrule
            Dim Reduction & distiluse-base-multilingual-cased-v1 & 0.00 & 1.59 & 5.43 & 0.00 & 1.17 & 2.41 \\
            Cross Encoder & distiluse-base-multilingual-cased-v1 & 5.94 & 8.04 & 14.20 & 2.97 & 1.84 & 7.35 \\
            \midrule
            Lexical   &  & 5.94 & 8.04 & 14.20 & 5.94 & 8.52 & 10.64\\
            ML Lexical  & 'German' & 9.90 & 8.41 & 15.19 & 9.90 & 9.14 & 11.58\\
        \midrule
        \bottomrule
    \end{tabular}
    }
    \label{tab:mlir-results-100-shorten-corpus}
\end{table*}

\begin{table*}[ht]
\centering
\footnotesize
\caption{Results IR Additional: Results IR
        Abbreviations: Capped \textbf{R}ecall, \textbf{N}DCG, \textbf{dist}iluse-base-multilingual-cased-v1, swiss-legal-\textbf{rob}erta-base, legal-\textbf{xlm}-roberta-base, \textbf{T}rain, \textbf{H}ard \textbf{N}egative, \textbf{E}valuate, \textbf{S}-\textbf{B}ert, \textbf{Dim} Reduction}
\resizebox{\textwidth}{!}{
\begin{tabular}{lrrrrrrrr}
         \toprule
         \textbf{Model} & \textbf{} & \textbf{Adaption} & \textbf{$R@1$} $\uparrow$ & \textbf{$R@10$} $\uparrow$ & \textbf{$R@100$} $\uparrow$  & \textbf{$N@1$} $\uparrow$  & \textbf{$N@10$} $\uparrow$ & \textbf{$N@100$} $\uparrow$\\
         \midrule
            LR & &   & 8.38 & 6.43 & 15.76 & 8.38 & 6.66 & 10.23 \\
            LR & & S & \textbf{10.64}& 7.57 & 16.47 & \textbf{10.64} & 8.04 & 11.33\\
            LR & & SL & 7.91 & \textbf{9.99} & \textbf{32.46} & 9.13 & \textbf{9.65} & \textbf{18.03} \\
            \midrule
            MLR & 'German' & & 8.69 & 6.54 & 15.99 & 8.69 & 6.82 & 10.43\\
            MLR & 'German' & S & \textbf{10.88} & 7.65 & 16.80 & \textbf{10.88} & 8.14 & 11.53\\
            MLR & 'German' & SL & 8.05 & \textbf{9.94} & \textbf{32.63} & 9.30 & \textbf{9.70} &  \textbf{18.17}\\
            MLR & 'French' & & \textbf{11.37} & \textbf{7.74} & \textbf{16.54} & \textbf{11.37} & \textbf{8.34} & \textbf{11.51} \\
            MLR & 'French' & S & 10.97 & 7.60 & 16.52 & 10.97 & 8.14 & 11.41\\
            MLR & 'Italian' & & 10.08 & 7.118 & 16.294 & 10.08 & 7.582 & 11.021 \\
            MLR & 'English' & & 8.38 & 6.43 & 15.76 & 8.38 & 6.66 & 10.23 \\
            \midrule
            T+E SB & xlm & & 2.77 & 2.58 & 10.17 & 6.36 & 5.66 & 12.03 \\
            T+E SB & rob & & 3.97 &  3.47 & 12.28 & 9.12 & 7.76 & 15.16\\
            E SB & dist & & 0.90 & 0.75 & 2.64 & 2.06 & 1.70 & 3.31 \\
            T+E SB & dist & & 4.4 & 3.92 & 12.64 & 10.11 & 8.76 & 16.16 \\
            T+E SB & dist & S & 4.69 & 4.14 & 13.39 & 10.77 & 9.27 & 17.05 \\
            T+E SB & dist & SL & 1.79 & 3.92 & 14.17 & 4.03 & 6.17 & 12.91 \\
            \midrule
            SB T(HN)+E & dist & & 3.97 & 4.46 & 13.36 & 9.12 & 9.21 & 16.87 \\
            SB T(HN)+E & dist & S & 3.76 & 4.75 & 12.80 & 8.64 & 9.66 & 16.57 \\
            SB T(HN)+E & dist & SL & 2.34 & 4.37 & 14.43 & 5.27 & 6.99 & 13.75 \\
            \midrule
            T+E SB & dist & DE & 4.22 & 4.49 & 15.21 & 8.21 & 8.15 & 15.86\\
            T+E SB & dist & DE SL & 4.06 & 8.47 & 29.43 & 4.51 & 6.73 & 13.78\\
            T+E SB & dist & FR & 1.88 & 2.2 & 9.19 & 5.77 & 6.22 & 13.94 \\
            T+E SB & dist & FR SL & 2.69 & 5.68 & 27.28 & 3.0 & 4.59 & 11.11\\
            T+E SB & dist & IT & 0.22 & 0.24 & 0.79 & 5.43 & 5.74 & 11.44\\
            T+E SB & dist & IT SL & 1.71 & 4.54 & 16.24 & 1.91 & 3.38 & 6.83\\
            \midrule
            Dim & dist  & & 0.71 & 0.62 & 2.42 & 1.64 & 1.4 & 2.95 \\
        \bottomrule
\end{tabular}
}
\label{tab:mlir-results-details}
\end{table*}

For the ML Lexical Retrieval model a main language must be chosen, indicated with German, French and Italian. Dataset adaptions are indicated with: (S) stopword removal, (SL) using only single language links, (DE/FR/IT) using only queries in one language.
Table \ref{tab:mlir-results-100-shorten-corpus} shows the results of the IR task on a subset of 100 queries and with only relevant documents while Table \ref{tab:mlir-results-details} shows more detailed results using all queries.


\clearpage\section{Error Analysis Examples}
\label{app:err-ananlysis-examples}

In the following we present specific examples for the generation tasks. In \autoref{tab:cvg-analysis-de} and \autoref{tab:cvg-analysis-fr} the CVG task for German respectively French is depicted. In \autoref{tab:lds-analysis} an example for the LDS task is shown. All texts were also translated into English using Deepl.

\begin{table*}[h!]
\centering
\vspace{-3ex}
\caption{Comparison of the generated considerations (CVG task) from four LLMs in different settings. The scores of each example are stated below in the following order: BERTScore, BLEU-Score, METEOR, ROUGE-1 / ROUGE-2 / ROUGE-L. The first row contains the facts of the case which were given as input, and the second row contains the considerations, the target of this task. We provide an English translated version on the right.}
\label{tab:cvg-analysis-de}
\begin{tabular}{|p{0.48\textwidth}|p{0.48\textwidth}|}
\multicolumn{2}{p{\textwidth}}{\textbf{Generated Considerations - German}} \\
\hline
\tiny{Decision ID: \textit{26c65bde-9976-569d-920e-a3781b01536f}, Year: \textit{2016}, Language: German, Court: \textit{CH\_BVGE}, Source: \href{https://entscheidsuche.ch/docs/CH_BVGer/CH_BVGE_001_E-1176-2019_2019-03-27.pdf}{BVGer D-6530/2015}}
&
\tiny{Translated into English}\\
\hline
\tiny{
    \textbf{Input: }
    Sachverhalt:
        A.
        Der Beschwerdef\"{u}hrer ersuchte am 14. August 2015 in der Schweiz um
        Asyl. Am 27. August 2015 wurde er zur Person befragt (BzP) und am
        10. September 2015 fand \textendash{} im Beisein einer Vertrauensperson \textendash{} die Anh\"{o}-
        rung zu seinen Asylgr\"{u}nden statt.
        B.
        Der aus Albanien stammende Beschwerdef\"{u}hrer gab an, er sei am (...)
        geboren und somit noch minderj\"{a}hrig. Zur Begr\"{u}ndung seines Asylge-
        suchs f\"{u}hrte er aus, er habe eine Beziehung zu einer jungen Frau gehabt,
        worauf deren Vater, nachdem er von der Beziehung mit seiner Tochter er-
        fahren gehabt habe, von ihm verlangt habe, diese zu heiraten. Da er aber
        das M\"{a}dchen nicht liebe, habe er weder eine Ehe mit ihr eingehen noch
        eine Beziehung mit ihr weiterf\"{u}hren wollen. 
        Der Vater der jungen Frau habe
        sich mit seinen Eltern in Verbindung gesetzt und ihnen erklärt, dass ihr
        Sohn seine Tochter zu heiraten habe, da er mit ihr eine Beziehung geführt
        habe. Seine Eltern seien damit einverstanden gewesen. Seither habe ihn
        sein Vater täglich ermahnt, dass er dieses Mädchen nun heiraten müsse.
        Weil er sich jeden Tag dasselbe von seinem Vater habe anhören müssen
        und er dies nicht mehr ertragen habe, habe er sich zum Verlassen seines
        Heimatlandes entschlossen. Bei einer Rückkehr nach Albanien hätte er mit
        Schwierigkeiten zu rechnen, da man ihn zur Heirat zwingen würde.
        Zum Beleg seiner Personalien reichte der Beschwerdeführer eine Identi-
        tätskarte ein.
        C.
        Mit Verf\"{u}gung vom 5. Oktober 2015 \textendash{} dem Beschwerdef\"{u}hrer und seiner
        Vertrauensperson er\"{o}ffnet am darauffolgenden Tag \textendash{} stellte die Vorinstanz
        fest, der Beschwerdef\"{u}hrer erf\"{u}lle die Fl\"{u}chtlingseigenschaft nicht, lehnte
        das Asylgesuch ab und verf\"{u}gte die Wegweisung aus der Schweiz. Den
        zust\"{a}ndigen Kanton beauftragte sie mit dem Vollzug der Wegweisung.
        D.
        Mit Eingabe vom 13. Oktober 2015 (Poststempel) erhob der Beschwerde-
        f\"{u}hrer gegen diesen Entscheid beim Bundesverwaltungsgericht Be-
        schwerde und beantragte die Aufhebung der angefochtenen Verf\"{u}gung,
        sodann sei ihm Asyl zu gew\"{a}hren, eventualiter sei die Unzumutbarkeit des
        Wegweisungsvollzugs festzustellen und ihm die vorl\"{a}ufige Aufnahme zu
        D-6530/2015
        Seite 3
        gew\"{a}hren, subeventualiter sei die Sache an die Vorinstanz zur\"{u}ckzuwei-
        sen und diese anzuweisen, eine allf\"{a}llige Wegweisung des Beschwerde-
        f\"{u}hrers pflichtgem\"{a}ss abzukl\"{a}ren, dies unter W\"{u}rdigung seiner Minderj\"{a}h-
        rigkeit. In prozessualer Hinsicht beantragte er die unentgeltliche Prozess-
        f\"{u}hrung im Sinne von Art. 65 Abs. 1 VwVG sowie den Verzicht auf die Er-
        hebung eines Kostenvorschusses.
}
& 
\tiny{
Facts of the case:
        A.
        The complainant applied for asylum in Switzerland on 14 August 2015.
        asylum. On 27 August 2015, he was interviewed (BzP) and on
        10 September 2015, in the presence of a person of trust, the hearing on his reasons for asylum took place.
        hearing on his grounds for asylum took place.
        B.
        The complainant, who comes from Albania, stated that he was born on (...)
        and therefore still a minor. In support of his application for asylum
        asylum, he stated that he had had a relationship with a young woman,
        woman, whereupon her father, after learning of the relationship with his daughter
        daughter, demanded that he marry her. But since he
        girl, he had neither entered into a marriage with her nor continued a relationship with her.
        want to continue a relationship with her. 
        The young woman's father had
        contacted his parents and explained to them that their son should marry his
        son had to marry his daughter because he had been in a relationship with her.
        relationship with her. His parents had agreed to this. Since then
        Since then, his father had reminded him every day that he had to marry this girl.
        Because he had had to listen to the same thing from his father every day
        and he could no longer bear it, he decided to leave his home country.
        home country.
        If he returned to Albania, he would face
        difficulties, as he would be forced to marry.
        To prove his identity, the complainant submitted an identity card.
        identity card.
        C.
        By order of 5 October 2015, the complainant and his confidant were informed that he would be forced to return to Albania.
        the following day, the lower instance found that the appellant
        determined that the appellant did not fulfil the refugee status, rejected the application for asylum and
        rejected the asylum application and ordered his removal from Switzerland. The
        responsible canton with the execution of the removal order.
        D.
        By petition dated 13 October 2015 (postmark), the appellant lodged a complaint against the
        f\"{u}hrer appealed against this decision to the Federal Administrative Court and
        and requested that the contested decision be cancelled,
        He was then to be granted asylum or, alternatively, the unreasonableness of the deportation order was to be
        the unreasonableness of the deportation order and that he be granted provisional admission.
        D-6530/2015
        Page 3
        order that the case be remanded to the lower instance and that the lower instance be instructed to
        and to instruct it to order the dismissal of the appellant in all cases.
        the appellant in accordance with his duty, taking into account his minority.
        rity. From a procedural point of view, he requested that the proceedings be conducted free of charge.
        free of charge within the meaning of Art. 65 para. 1 VwVG as well as the waiver of an advance on costs.
        of an advance on costs.
}\\
\hline
\end{tabular}
\end{table*}

\begin{table*}[h!]
\ContinuedFloat
\caption{continuation}
\begin{tabular}{|p{0.48\textwidth}|p{0.48\textwidth}|}
\hline
\tiny{
    \textbf{Target: }
    Das Bundesverwaltungsgericht zieht in Erw\"{a}gung:
        1.
        1.1. Gem\"{a}ss Art. 31 VGG beurteilt das Bundesverwaltungsgericht Be-
        schwerden gegen Verf\"{u}gungen nach Art. 5 VwVG. Das SEM geh\"{o}rt zu den
        Beh\"{o}rden nach Art. 33 VGG und ist daher eine Vorinstanz des Bundesver-
        waltungsgerichts. Eine das Sachgebiet betreffende Ausnahme im Sinne
        von Art. 32 VGG liegt nicht vor. Das Bundesverwaltungsgericht ist daher
        zust\"{a}ndig f\"{u}r die Beurteilung der vorliegenden Beschwerde und entschei-
        det auf dem Gebiet des Asyls in der Regel so auch vorliegend endg\"{u}ltig
        (Art. 105 AsylG [SR 142.31]; Art. 83 Bst. d Ziff. 1 BGG).
        1.2. Das Verfahren richtet sich nach dem VwVG, dem VGG und dem BGG,
        soweit das AsylG nichts anderes bestimmt (Art. 37 VGG und Art. 6 AsylG).
        2. Die Kognition des Bundesverwaltungsgerichts und die zul\"{a}ssigen R\"{u}-
        gen richten sich im Asylbereich nach Art. 106 Abs. 1 AsylG und im Bereich
        des Ausl\"{a}nderrechts nach Art. 49 VwVG (vgl. BVGE 2014/26 E. 5),
        2.1. Mit Beschwerde kann die Verletzung von Bundesrecht (einschliesslich
        Missbrauch und \"{U}berschreiten des Ermessens) sowie die unrichtige und
        unvollst\"{a}ndige Feststellung des rechtserheblichen Sachverhalts ger\"{u}gt
        werden (Art. 106 Abs. 1 AsylG).
        3.
        3.1. Vorab ist die Frage der Prozessf\"{a}higkeit als Sachurteilsvoraussetzung
        von Amtes wegen zu pr\"{u}fen (FRITZ GYGI, Bundesverwaltungsrechtspflege,
        2. Aufl. 1983, S. 73). Als verfahrensrechtliches Korrelat der Handlungsf\"{a}-
        higkeit ist sie nach den einsch
}
&
\tiny{
The Federal Administrative Court considers:
        1.
        1.1 Pursuant to Art. 31 VGG, the Federal Administrative Court assesses appeals against
        appeals against decisions pursuant to Art. 5 VwVG. The SEM is one of the
        authorities pursuant to Art. 33 VGG and is therefore a lower instance of the Federal
        administrative court. An exception relating to the subject matter within the meaning of
        of Art. 32 VGG does not apply. The Federal Administrative Court is therefore
        competent for the assessment of the present appeal and decides in the area of asylum in the
        usually decides in the area of asylum as well as in the present case finally
        (Art. 105 AsylG [SR 142.31]; Art. 83 let. d no. 1 BGG).
        1.2 The procedure is governed by the VwVG, the VGG and the BGG,
        unless the AsylG provides otherwise (Art. 37 VGG and Art. 6 AsylG).
        2. the cognizance of the Federal Administrative Court and the admissible r\"{a}ssige R\"{u}-
        are governed by Art. 106 para. 1 AsylA in the area of asylum and
        foreign nationals law according to Art. 49 VwVG (see BVGE 2014/26 E. 5),
        2.1 An appeal may be lodged against the violation of federal law (including
        abuse and exceeding of discretionary powers) as well as the incorrect and incomplete
        incorrect and incomplete determination of the legally relevant facts (Art. 106 para.
        (Art. 106 para. 1 AsylG).
        3.
        3.1 First of all, the question of procedural capacity as a prerequisite for a substantive judgement must be
        to be examined ex officio (FRITZ GYGI, Bundesverwaltungsrechtspflege,
        2nd ed. 1983, p. 73). As the procedural correlate of the capacity to act, it is
        ability to act, it is, according to the
}\\
\hline
\textbf{GPT-3.5-Turbo} (Fine-tuned)
&
\tiny{Translated into English}\\
\hline
\tiny{
    Das Bundesverwaltungsgericht zieht in Erw\"{a}gung: 1. 1.1 Gem\"{a}ss Art. 31 VGG beurteilt das Bundesverwaltungsgericht Be- schwerden gegen Verf\"{u}gungen nach Art. 5 VwVG. Das SEM geh\"{o}rt zu den Beh\"{o}rden nach Art. 33 VGG und ist daher eine Vorinstanz des Bundesver- waltungsgerichts. Eine das Sachgebiet betreffende Ausnahme im Sinne von Art. 32 VGG liegt nicht vor. Das Bundesverwaltungsgericht ist daher zust\"{a}ndig f\"{u}r die Beurteilung der vorliegenden Beschwerde, soweit diese auf dem Gebiet des Asyls (Art. 105 AsylG) zul\"{a}ssig (Art. 83 BGG) ist. 1.2 Beurteilt das Bundesverwaltungsgericht Beschwerden nach Art. 31 VGG, so wendet es unter Vorbehalt der Anwendung von staatsvertra- glichen Rechten materielles Bundesrecht an (Art. 37 Abs. 1 VGG). 
    1.3 Mit Beschwerde kann die Verletzung von Bundesrecht (einschliesslich die unrichtige oder unvollst\"{a}ndige Feststellung des rechtserheblichen Sach- verhalts [Art. 97 Abs. 1 BGG] und die Unangemessenheit [Art. 105 Abs. 2 BGG]), die Rechtsverletzung nach Art. 95 BGG und die unrichtige oder un- zweckm\"{a}ssige Sachverhaltsfeststellung (Art. 97 Abs. 1 BGG) ger\"{u}gt werden. Das Bundesverwaltungsgericht wendet nunmehr auch das Beschwerde- recht nach Art. 105 AsylG sowie die Staatsvertragsvorbehalte an, denen Rechnung zu tragen ist (Art. 37 Abs. 3 VGG). D-6530/2015 Seite 4 2. 2.1 Zur Beschwerde gegen Entscheide \"{u}ber Asylgesuche ist berechtigt, wer durch den angefochtenen Entscheid besonders ber\"{u}hrt ist und ein rechtlich gesch\"{u}tztes Inter
} 
&
\tiny{
    The Federal Administrative Court considers: 1. 1.1 In accordance with Art. 31 VGG, the Federal Administrative Court assesses appeals against decisions under Art. 5 VwVG. The SEM is one of the authorities under Art. 33 VGG and is therefore a lower instance of the Federal Administrative Court. There is no exception relating to the subject matter within the meaning of Art. 32 VGG. The Federal Administrative Court is therefore competent to judge the present appeal insofar as it is admissible in the area of asylum (Art. 105 AsylA) (Art. 83 BGG). 1.2 If the Federal Administrative Court assesses appeals in accordance with Art. 31 VGG, it shall apply substantive federal law, subject to the application of constitutional rights (Art. 37 para. 1 VGG).
    1.3 An appeal may be lodged for a violation of federal law (including incorrect or incomplete determination of the legally relevant facts [Art. 97 para. 1 FSCA] and inappropriateness [Art. 105 para. 2 FSCA]), violation of the law pursuant to Art. 95 FSCA and incorrect or inappropriate determination of the facts (Art. 97 para. 1 FSCA). The Federal Administrative Court now also applies the right of appeal in accordance with Art. 105 AsylA and the reservations of state treaties, which must be taken into account (Art. 37 para. 3 VGG). D-6530/2015 Page 4 2. 2.1 Anyone who is particularly affected by the contested decision and has a legally protected interest in the asylum application is entitled to appeal against decisions on asylum applications.
} \\
\hline
\multicolumn{2}{|p{\textwidth}|}{\textbf{Scores: }85.15, 80.08, 50.36, 57.45/41.46/57.45} \\
\hline
\textbf{LLaMA-2-13B Chat} (Fine-tuned)
&
\tiny{Translated into English}\\
\hline
\tiny{
    [Erw\"{a}gungen]: 1. 1.1 Gem\"{a}ss Art. 31 VGG beurteilt das Bundesverwaltungsgericht Be- schwerden gegen Verf\"{u}gungen nach Art. 5 VwVG. Das BFM geh\"{o}rt zu den Beh\"{o}rden nach Art. 33 VGG und ist daher eine Vorinstanz des Bundesver- waltungsgerichts. Eine das Sachgebiet betreffende Ausnahme im Sinne von Art. 32 VGG liegt nicht vor. Das Bundesverwaltungsgericht ist daher zust\"{a}ndig f\"{u}r die Beurteilung der vorliegenden Beschwerde und entschei- det auf dem Gebiet des Asyls endg\"{u}ltig, ausser bei Vorliegen eines Auslie- ferungsersuchens des Staates, vor welchem die beschwerdef\"{u}hrende Per- son Schutz sucht (Art. 105 AsylG [SR 142.31]; Art. 83 Bst. d Ziff. 1 BGG). Eine solche Ausnahme im Sinne von Art. 83 Bst. d Ziff. 1 BGG liegt nicht vor, weshalb das Bundesverwaltungsgericht endg\"{u}ltig entscheidet. 
    1.2 Das Verfahren richtet sich nach dem VwVG, dem VGG und dem BGG, soweit das AsylG nichts anderes bestimmt (Art. 37 VGG und Art. 6 AsylG). 1.3 Die Beschwerde ist frist- und formgerecht eingereicht. Der Beschwerdef\"{u}hrer hat am Verfahren vor der Vorinstanz teilgenommen, ist durch die angefochtene Verf\"{u}gung besonders ber\"{u}hrt und hat ein schutzw\"{u}rdiges Interesse an deren Aufhebung beziehungsweise \"{A}nde- rung. Er ist daher zur Einreichung der Beschwerde legitimiert (Art. 105 und Art. 108 Abs. 1 AsylG; Art. 48 Abs. 1 sowie Art. 52 Abs. 1 VwVG). Auf die Beschwerde ist einzutreten. 2. Mit Beschwerde kann die Verletzung von Bundesrecht
}
& 
\tiny{
    [Erw\"{a}gungen]: 1. 1.1 Pursuant to Art. 31 VGG, the Federal Administrative Court assesses appeals against decisions under Art. 5 VwVG. The FOM is one of the authorities under Art. 33 VGG and is therefore a lower instance of the Federal Administrative Court. There is no exception relating to the subject matter within the meaning of Art. 32 VGG. The Federal Administrative Court is therefore responsible for the assessment of the present appeal and makes a final decision in the area of asylum, except in the case of a request for expulsion from the state from which the appellant is seeking protection (Art. 105 AsylA [SR 142.31]; Art. 83 let. d no. 1 BGG). There is no such exception within the meaning of Art. 83 let. d no. 1 BGG, which is why the Federal Administrative Court makes a final decision. 
    1.2 The procedure is governed by the VwVG, the VGG and the BGG, unless the AsylG provides otherwise (Art. 37 VGG and Art. 6 AsylG). 1.3 The appeal has been submitted in due time and form. The appellant participated in the proceedings before the lower instance, is particularly affected by the contested decision and has a legitimate interest in its cancellation or amendment. He is therefore legitimised to lodge an appeal (Art. 105 and Art. 108 para. 1 AsylA; Art. 48 para. 1 and Art. 52 para. 1 VwVG). The appeal must be upheld. 2. the appeal may be lodged to challenge the violation of federal law
}\\
\hline
\multicolumn{2}{|p{\textwidth}|}{\textbf{Scores: }85.07, 78.41, 47.00, 60.56/51.46/60.56}\\
\hline
\end{tabular}
\end{table*}

\begin{table*}[h!]
\ContinuedFloat
\caption{continuation}
\begin{tabular}{|p{0.48\textwidth}|p{0.48\textwidth}|}
\hline
\textbf{Claude 2} (0-shot)&
\tiny{Translated into English}\\
\hline
\tiny{
    Hier meine Erw\"{a}gungen zum vorliegenden Sachverhalt: 1. Prozessvoraussetzungen Die sachliche Zust\"{a}ndigkeit des Bundesverwaltungsgerichts ergibt sich aus Art. 31 VGG. Gem\"{a}ss Art. 31 Abs. 1 VGG beurteilt das Bundesverwaltungsgericht Beschwerden gegen Verf\"{u}gungen nach Art. 5 AsylG. Die Vorinstanz hat am 5. Oktober 2015 eine Verf\"{u}gung nach Art. 5 AsylG erlassen. Gegen diese Verf\"{u}gung wurde frist- und formgerecht Beschwerde erhoben. Die Prozessvoraussetzungen sind damit erf\"{u}llt. 2. Materielle Beurteilung 2.1 Fl\"{u}chtlingseigenschaft Gem\"{a}ss Art. 3 AsylG gilt als Fl\"{u}chtling eine Person, die in ihrem Heimatstaat wegen ihrer Rasse, Religion, Nationalit\"{a}t, Zugeh\"{o}rigkeit zu einer bestimmten sozialen Gruppe oder wegen ihrer politischen Anschauungen ernsthaften Nachteilen ausgesetzt ist oder begr\"{u}ndete Furcht hat, solchen Nachteilen ausgesetzt zu werden.
    Der Beschwerdef\"{u}hrer macht geltend, er werde in seinem Heimatland Albanien zur Ehe mit einer Frau gezwungen, die er nicht heiraten m\"{o}chte. Eine Zwangsheirat kann eine Verletzung der Menschenrechte darstellen. Allerdings ist im vorliegenden Fall nicht ersichtlich, dass der Beschwerdef\"{u}hrer aufgrund eines in Art. 3 AsylG genannten Merkmals von Zwangsheirat bedroht w\"{a}re. Vielmehr scheint es sich um einen famili\"{a}ren Konflikt zwischen den Eltern des Beschwerdef\"{u}hrers und der Familie der jungen Frau zu handeln. Aus den Akten ergeben sich keine Hinweise, dass der Beschwerdef\"{u}hrer bei einer R\"{u}ckkehr nach Albanien staatlichen Verfolgungshandlungen ausgesetzt w\"{a}re. Auch liegen keine Beweise daf\"{u}r vor, dass er durch nichtstaat    
}
& 
\tiny{
    Here are my considerations on the facts of the case: 1. procedural requirements The Federal Administrative Court has subject-matter jurisdiction pursuant to Art. 31 VGG. Pursuant to Art. 31 para. 1 VGG, the Federal Administrative Court judges appeals against decisions under Art. 5 AsylG. On 5 October 2015, the lower instance issued an order pursuant to Art. 5 AsylA. An appeal against this decision was lodged in due time and form. The procedural requirements are therefore fulfilled. 2 Substantive assessment 2.1 Refugee status Pursuant to Art. 3 AsylA, a refugee is a person who is exposed to serious disadvantages in their home country because of their race, religion, nationality, membership of a particular social group or because of their political views, or who has a well-founded fear of being exposed to such disadvantages.
    The applicant claims that he is being forced to marry a woman in his home country of Albania whom he does not wish to marry. A forced marriage can constitute a violation of human rights. However, it is not apparent in this case that the applicant would be at risk of forced marriage due to one of the characteristics listed in Art. 3 AsylG. Rather, it appears to be a family conflict between the parents of the complainant and the young woman's family. There are no indications from the files that the complainant would be exposed to state persecution if he returned to Albania. There is also no evidence that he would be persecuted by non-governmental organisations.
}\\
\hline
\multicolumn{2}{|p{\textwidth}|}{\textbf{Scores: }66.96, 60.71, 26.59, 27.80/6.31/27.12}\\
\hline
\textbf{LLaMA-2-13B Chat} (0-shot)
&
\tiny{Translated into English}\\
\hline
\tiny{
    Erw\"{a}gungen Das Bundesverwaltungsgericht hat den Beschwerdef\"{u}hrer auf seine Erw\"{a}gungen hin geh\"{o}rt und bewertet die behaupteten Asylgr\"{u}nde. 1. Minderj\"{a}hrigkeit Der Beschwerdef\"{u}hrer gibt an, geboren zu sein und minderj\"{a}hrig zu sein. Die Vorinstanz hat jedoch festgestellt, dass er nicht minderj\"{a}hrig ist. Es ist jedoch unumstritten, dass der Beschwerdef\"{u}hrer eine Beziehung zu einer jungen Frau hatte, die von ihrem Vater verlangt wurde, diese zu heiraten. Der Beschwerdef\"{u}hrer hat jedoch keine Ehe mit ihr eingegangen und will nicht weitere Beziehung mit ihr haben. 2. Asylgr\"{u}nde Der Beschwerdef\"{u}hrer behauptet, dass er aus Albanien geflohen ist, weil er sich dort Schwierigkeiten because of his refusal to marry the young woman would have faced. However, the evidence presented does not support this claim. There is no indication that the young woman's father would have subjected the applicant to persecution or violence if he had not married his daughter.
    3. Fl\"{u}chtlingseigenschaft The applicant does not meet the criteria for refugee status under the Geneva Convention. He has not demonstrated that he is fleeing persecution or that his life or freedom would be threatened if he were to return to Albania. 4. Personalien The applicant has provided an identity card as proof of his personal details. However, there is no independent evidence to support his claims. The Vorinstanz has found that the applicant does not meet the criteria for refugee status, and therefore, his asylum application should be rejected. 5. Unzumutbarkeit des Wegweisungsvollzugs The applicant has requested that the court consider the unreasonableness of the deportation order. However, the court has found that the order is reasonable and in accordance with the law. 6. Allf\"{a}llige Wegweisung The court has considered the possibility of a conditional release of the applicant. 
    However, given the lack of evidence supporting the applicant's claims and the fact that he does not meet the criteria for refugee status, the court has decided not to grant conditional release. 7
}
&
\tiny{
    Considerations The Federal Administrative Court has examined the appellant's considerations and assesses the alleged grounds for asylum. 1. minority The appellant claims to have been born and to be a minor. However, the lower court found that he is not a minor. However, it is undisputed that the appellant had a relationship with a young woman who was required by her father to marry her. However, the appellant did not enter into a marriage with her and does not wish to have any further relations with her. 2. asylum seeker The complainant claims that he fled Albania because he faced difficulties [\textbf{From here, the model selects English as the output language.}] because of his refusal to marry the young woman would have faced. However, the evidence presented does not support this claim. There is no indication that the young woman's father would have subjected the applicant to persecution or violence if he had not married his daughter.
    3. Fl\"{u}chtlingseigenschaft The applicant does not meet the criteria for refugee status under the Geneva Convention. He has not demonstrated that he is fleeing persecution or that his life or freedom would be threatened if he were to return to Albania. 4. Personalien The applicant has provided an identity card as proof of his personal details. However, there is no independent evidence to support his claims. The Vorinstanz has found that the applicant does not meet the criteria for refugee status, and therefore, his asylum application should be rejected. 5. Unzumutbarkeit des Wegweisungsvollzugs The applicant has requested that the court consider the unreasonableness of the deportation order. However, the court has found that the order is reasonable and in accordance with the law. 6. Allf\"{a}llige Wegweisung The court has considered the possibility of a conditional release of the applicant. 
    However, given the lack of evidence supporting the applicant's claims and the fact that he does not meet the criteria for refugee status, the court has decided not to grant conditional release. 7
}
\\
\hline
\multicolumn{2}{|p{\textwidth}|}{\textbf{Scores: }62.23, 38.56, 11.97, 14.33/0.42/13.71} \\
\hline
\end{tabular}
\vspace{-2ex}
\end{table*}

\begin{table*}[ht]
\centering
\vspace{-3ex}
\caption{\tiny{Comparison of the generated considerations (\ac{CVG} task) from four \acp{LLM} in different settings. The scores of each example is stated below in the following order: BERTScore, BLEU-Score, METEOR, ROUGE-1 / ROUGE-2 / ROUGE-L. The first row contains the facts of the case which were given as input, the second row contains the the considerations, the target of this task. We provide an English translated version on the right.}}
\label{tab:cvg-analysis-fr}
\begin{tabular}{|p{0.48\textwidth}|p{0.48\textwidth}|}
\multicolumn{2}{p{1\textwidth}}{
\textbf{Generated Considerations - French} 
}\\
\hline
\tiny{Decision ID: \textit{8bb15591-be0b-5bc3-933a-e6f856a46fec}, Year: \textit{2017}, Language: French, Court: \textit{FR\_TC}, Source: \href{https://publicationtc.fr.ch/?locale=de}{102 2016 223}}
&
\tiny{Translated into English}\\
\hline
\tiny{
    \textbf{Input: }
    consid\'{e}rant en fait
    A. A.\_ et l'entreprise individuelle C.\_ ont conclu un contrat de travail d'une dur\'{e}e ind\'{e}termin\'{e}e, ayant pour objet l'engagement de A.\_ en qualit\'{e} de collaboratrice administrative, \`{a} un taux d'activit\'{e} de 40\%, d\`{e}s le 10 f\'{e}vrier 2014. Financ\'{e}e principalement par l'assurance ch\^{o}mage, respectivement par le SECO, C.\_ a pour but d'apporter son soutien aux jeunes en difficult\'{e} d'insertion professionnelle au moyen de cours, d'ateliers et de coaching et a notamment \'{e}t\'{e} mandat\'{e}e par l'association D.\_ pour apporter son soutien aux adolescents de langue allemande.
    D\`{e}s le 1er mars 2014, A.\_ \'{e}tait occup\'{e}e \`{a} 40\% en tant que collaboratrice administrative et \`{a} 10\% en tant qu'enseignante. Le contrat de travail, sign\'{e} le 10 juin 2014, pr\'{e}voyait un salaire mensuel brut de CHF 2'733.80 pour 16 heures d'activit\'{e} administrative hebdomadaire ainsi qu'un salaire horaire brut de CHF 50.20 pour 4 heures d'enseignement par semaine.
    D\`{e}s le 1er septembre 2014, A.\_ a augment\'{e} son taux d'enseignement \`{a} 40\%, \`{a} l'essai.
    En ao\^{u}t 2014, C.\_ a \'{e}t\'{e} transform\'{e}e en la soci\'{e}t\'{e} B.\_ GmbH, dont E.\_ est l'associ\'{e}e g\'{e}rante avec signature individuelle. B.\_ GmbH a repris tous les contrats de travail et a \'{e}t\'{e} inscrite au registre du commerce le 7 ao\^{u}t 2014.
    B. Le 13 octobre 2014 a eu lieu une s\'{e}ance entre E.\_, A.\_, l'enseignante F.\_ et l'apprenti G.\_. Au cours de celle-ci, E.\_ a signal\'{e} que A.\_ \'{e}tait manifestement d\'{e}bord\'{e}e par l'activit\'{e} suppl\'{e}mentaire d'enseignement, que l'essai n'avait ainsi pas \'{e}t\'{e} probant, qu'il fallait donc l'arr\^{e}ter et qu'il fallait engager quelqu'un pour assumer ce volet de travail. \`{A} partir du 15 octobre 2014, A.\_ a \'{e}t\'{e} mise en arr\^{e}t de travail \`{a} 100 \% jusqu'au 31 octobre 2014 pour des raisons de sant\'{e}.
    En date du 30 octobre 2014, A.\_ a inform\'{e} E.\_ qu'elle reprendrait son activit\'{e} le 3 novembre 2014. N\'{e}anmoins, le 2 novembre 2014, A.\_ a \'{e}t\'{e} hospitalis\'{e}e pour des douleurs dorsales. Elle a alors \'{e}t\'{e} mise en arr\^{e}t de travail \`{a} 100\% du 2 au 4 novembre 2014. Le 4 novembre 2014, A.\_ a indiqu\'{e} \`{a} son employeur qu'elle se pr\'{e}senterait \`{a} son travail le lendemain, mais qu'un rendez-vous chez son m\'{e}decin traitant \'{e}tait pr\'{e}vu le m\^{e}me jour. De plus, elle a demand\'{e} \`{a} E.\_ de lui consacrer un peu de temps pour un entretien. Tribunal cantonal TC Page 3 de 9 E.\_ et A.\_ se sont entretenues le 5 novembre 2014. Au cours de cet entretien, A.\_ a d\'{e}clar\'{e} son souhait de poursuivre son activit\'{e} au sein de l'entreprise. [...]
}
& 
\tiny{
in fact
    A. A.\_ and the sole proprietorship C.\_ entered into an employment contract for an indefinite period, the purpose of which was to employ A.\_ as an administrative assistant, with an activity rate of 40\%, from 10 February 2014. Funded mainly by the unemployment insurance scheme, respectively by the SECO, C.\_ aims to provide support for young people experiencing difficulties in finding employment by means of courses, workshops and coaching, and has in particular been mandated by the D.\_ association to provide support for German-speaking teenagers.
    Since 1 March 2014, A.\_ has been employed as an administrative assistant (40 \%) and a teacher (10 \%). The employment contract, signed on 10 June 2014, provided for a gross monthly salary of CHF 2,733.80 for 16 hours of administrative work per week and a gross hourly salary of CHF 50.20 for 4 hours of teaching per week.
    From 1 September 2014, A.\ increased her teaching rate by 40 per cent, on a trial basis.
    In August 2014, C.\_ transformed into the company B.\_ GmbH, of which E.\_ is the managing partner with individual signature. B.\_ GmbH took over all employment contracts and was entered in the commercial register on 7 August 2014.
    B. On 13 October 2014 a meeting took place between E.\_, A.\_, teacher F.\_ and apprentice G.\_. During this meeting, E.\_ pointed out that A.\_ was clearly overwhelmed by the additional teaching activity, that the trial had therefore not been successful, that it should therefore be stopped and that someone should be hired to take on this part of the work. From 15 October 2014, A.\_ was placed on 100\% work stoppage until 31 October 2014 for health reasons.
    On 30 October 2014, A.\_ informed E.\_ that she would resume her activity on 3 November 2014. Nevertheless, on 2 November 2014, A.\_ was hospitalised for back pain. She was then placed on 100 \% sick leave from 2 to 4 November 2014. On 4 November 2014, A.\ indicated to her employer that she would report to work the following day, but that an appointment with her GP was scheduled for the second day. In addition, she asked E.\ to give her some time for an interview. Cantonal Court TC Page 3 of 9 E.\_ and A.\_ spoke on 5 November 2014. During this interview, A.\_ declared her wish to continue working for the company. [...]
}\\
\hline
\tiny{
    \textbf{Target: }
    en droit
    1. a) La d\'{e}cision attaqu\'{e}e constitue une d\'{e}cision finale de premi\`{e}re instance au sens des art. 308 al. 1 et 236 CPC. La voie de droit ouverte contre une telle d\'{e}cision est l'appel (art. 308 al. 1 let. a CPC), sauf si la valeur litigieuse est inf\'{e}rieure \`{a} CHF 10'000.- (art. 308 al. 2 CPC), auquel cas la d\'{e}cision ne peut faire l'objet que d'un recours (art. 319 let. a CPC). Dans le cas d'esp\`{e}ce, la valeur litigieuse au dernier \'{e}tat des conclusions est sup\'{e}rieure \`{a} CHF 10'000.- (cf. demande, ch. 1 et 2; r\'{e}ponse, ad. 1), de sorte que la voie de l'appel est ouverte. La valeur litigieuse devant la Cour est n\'{e}anmoins inf\'{e}rieure \`{a} CHF 15'000.- (art. 74 al. 1 let. a et art. 51 al. 1 let. a LTF). Par cons\'{e}quent, cet arr\^{e}t ne peut faire l'objet que d'un recours constitutionnel subsidiaire (art. 113 ss LTF).
    b) L'appel du 21 octobre 2016 a \'{e}t\'{e} d\'{e}pos\'{e} en temps utile, soit dans le d\'{e}lai l\'{e}gal de 30 jours \`{a} compter de la notification de la d\'{e}cision attaqu\'{e}e survenue le 21 septembre 2016 (art. 311 al. 1 CPC). Il en va de m\^{e}me de l'appel joint d\'{e}pos\'{e} le 13 d\'{e}cembre 2016, soit dans le d\'{e}lai l\'{e}gal de 30 jours \`{a} compter de la notification de l'appel survenue le 14 novembre 2016 (art. 312 al. 2 CPC).
    c) D\^{u}ment motiv\'{e}s et dot\'{e}s de conclusions, l'appel et l'appel joint sont pour le surplus recevables en la forme.
    d) La cognition de la Cour est pleine et enti\`{e}re en fait comme en droit (art. 310 CPC).
    2. Dans son appel, l'appelante reproche aux premiers juges de ne pas avoir reconnu le caract\`{e}re abusif de la r\'{e}siliation intervenue le 15 d\'{e}cembre 201
}
&
\tiny{
    in law
    1. a) The contested decision constitutes a final decision of first instance within the meaning of articles 308 al. 1 and 236 CPC. The legal remedy available against such a decision is an appeal (art. 308 al. 1 let. a CPC), unless the amount in dispute is less than CHF 10,000 (art. 308 al. 2 CPC), in which case the decision can only be appealed (art. 319 let. a CPC). In the case in point, the amount in dispute at the last stage of the pleadings is greater than CHF 10,000 (see the request, points 1 and 2; the response, point 1), so that the appeal procedure is open. The amount in dispute before the Court is nevertheless less than CHF 15,000 (art. 74 para. 1 let. a and art. 51 para. 1 let. a FSCA). Consequently, this judgment can only be the subject of a subsidiary constitutional appeal (art. 113 ff LTF).
    b) The appeal of 21 October 2016 was lodged in good time, i.e. within the legal time limit of 30 days from notification of the contested decision on 21 September 2016 (art. 311 para. 1 CPC). The same applies to the joint appeal filed on 13 December 2016, i.e. within the legal time limit of 30 days from notification of the appeal on 14 November 2016 (art. 312 para. 2 CPC).
    c) Duly motivated and endowed with submissions, the appeal and the joint appeal are otherwise admissible in form.
   d) The Court has full jurisdiction in fact and in law (art. 310 CPC).
    2. In her appeal, the appellant criticised the lower courts for failing to recognise the abusive nature of the termination on 15 December 201
}\\
\hline
\end{tabular}
\end{table*}

\begin{table*}[h!]
\ContinuedFloat
\caption{continuation}
\begin{tabular}{|p{0.48\textwidth}|p{0.48\textwidth}|}
\hline
\textbf{GPT-3.5-Turbo} (Fine-tuned)
&
\tiny{Translated into English}\\
\hline
\tiny{
en droit 1. En vertu de l'art. 57C al. 3 let. b de la loi du 11 avril 1991 sur l'organisation judiciaire (LOJ ; RS 173.01), le Tribunal cantonal conna\^{\i}t seul des causes portant sur la l\'{e}galit\'{e} de certaines d\'{e}cisions des tribunaux cantonaux. Dans le cas particulier, interjet\'{e} contre une d\'{e}cision finale prise par la juridiction inf\'{e}rieure sur la base de l'art. 64A al. 1 let. b de la loi sur la proc\'{e}dure civile du 19 novembre 2010 (CPC ; RS 270.11), le recours, d\'{e}pos\'{e} dans le d\'{e}lai (art. 48A al. 3 CPC) et la forme pr\'{e}vus par la loi (art. 79 al. 1 LOJ et 321-327 CPC), n'est en principe pas recevable, en vertu de l'art. 57C al. 2 LOJ, sauf s'il doit \^{e}tre trait\'{e} comme un recours de droit public (cf. art. 57D LOJ). 2. D\'{e}pos\'{e} dans le d\'{e}lai et les formes prescrits par la loi \`{a} l'encontre d'une d\'{e}cision finale prise par un tribunal cantonal, le recours doit \^{e}tre trait\'{e} comme un recours de droit public, \'{e}tant d\`{e}s lors admissible sur le principe (art. 57C al. 4 LOJ). 3. En droit public fribourgeois, le TRPC appr\'{e}cie librement la recevabilit\'{e} et la wellfondation des recours de droit public cantonal, interpr\`{e}te et applique le droit cantonal de mani\`{e}re ind\'{e}pendante. En cas de violation du droit f\'{e}d\'{e}ral, le recours est, quant \`{a} lui, soumis au Tribunal f\'{e}d\'{e}ral. Celui-ci contr\^{o}le la violation du droit f\'{e}d\'{e}ral (selon les principes applicables au recours en mati\`{e}re de droit public, cf. \`{a} ce propos les art. 82 ss LTF), le droit constitutionnel et, le cas \'{e}ch\'{e}ant, l'usage du pouvoir d'appr\'{e}ciation cantonal selon la maxime de l'arbitraire (Art. 49 al. 1 Cst. et 106 al. 2 LTF
} 
&
\tiny{
in law 1. By virtue of art. 57C para. 3 let. b of the law of 11th April 1991 on the organisation of the judiciary (LOJ; RS 173.01), the Tribunal cantonal alone hears cases relating to the legality of certain decisions of the cantonal courts. In this particular case, brought against a final decision taken by the lower court on the basis of art. 64A para. 1 let. b of the law on civil procedure of 19 November 2010 (CPC; RS 270.11), the appeal, lodged within the time limit (art. 48A al. 3 CPC) and in the form prescribed by law (art. 79 al. 1 LOJ and 321-327 CPC), is in principle inadmissible, by virtue of art. 57C al. 2 LOJ, unless it is to be treated as an appeal under public law (cf. art. 57D LOJ). 2. Filed within the time limit and in the form prescribed by law against a final decision taken by a cantonal court, the appeal must be treated as an appeal under public law, and is therefore admissible in principle (art. 57C al. 4 LOJ). 3. Under Fribourg public law, the TRPC freely assesses the admissibility and merits of cantonal public law appeals, and interprets and applies cantonal law independently. In the event of a breach of federal law, the appeal is submitted to the Federal Court. This counteracts the violation of federal law (according to the principles applicable to appeals in matters of public law, cf. in this respect Art. 82 ff LTF), constitutional law and, where applicable, the use of the cantonal power of appraisal according to the maxim of arbitrariness (Art. 49 al. 1 Cst. and 106 al. 2 LTF).
} \\
\hline
\multicolumn{2}{|p{\textwidth}|}{\textbf{Scores: }73.33, 64.17, 28.71, 24.58/6.34/24.58} \\
\hline
\textbf{LLaMA-2-13B Chat} (Fine-tuned)
&
\tiny{Translated into English}\\
\hline
\tiny{
    [Consid\'{e}rations]: En droit, les premiers juges ont consid\'{e}r\'{e} qu'au vu des faits retenus, il fallait admettre le licenciement abusif. En effet, le lien de confiance qui unissait les parties avait \'{e}t\'{e} irr\'{e}m\'{e}diablement rompu d\`{e}s le 5 novembre 2014. A.\_ avait refus\'{e} de signer le avertissement et la lettre de r\'{e}siliation de son contrat de travail, ce qui avait pour cons\'{e}quence qu'elle ne pouvait plus \^{e}tre tenue de respecter les conditions de la r\'{e}siliation. Par ailleurs, le fait que la demanderesse ait \'{e}t\'{e} mise en arr\^{e}t de travail \`{a} 100 \% du 2 au 11 novembre 2014, puis \`{a} 100 \% d\`{e}s le 12 novembre 2014 jusqu'au 30 novembre 2014, puis \`{a} 70 \% du 1er au 14 d\'{e}cembre 2014, puis \`{a} 50 \% du 15 d\'{e}cembre 2014 au 15 janvier 2015 et enfin \`{a} 25 \% du 16 au 29 janvier 2015 ne pouvait \^{e}tre interpr\'{e}t\'{e} comme une r\'{e}siliation. En effet, la demanderesse ne pouvait pas, en tant que collaboratrice \`{a} 40\%, assumer son travail \`{a} 100\% d\`{e}s le 2 novembre 2014, ce qu'elle aurait d\^{u} faire en tant qu'enseignante si l'on tient compte de la situation m\'{e}dicale de cette derni\`{e}re. Par ailleurs, le fait que l'employeur ait licenci\'{e} la demanderesse le 7 novembre 2014, soit durant une p\'{e}riode de protection, ne pouvait pas \^{e}tre consid\'{e}r\'{e} comme un licenciement abusif. En effet, l'employeur n'avait pas connaissance du fait que la demanderesse \'{e}tait en arr\^{e}t de travail \`{a} 100 \% du 2 au 11 novembre 2014. Enfin, l'employeur avait mis un terme au contrat de travail par courrier du 10 d\'{e}cembre 2014, soit \`{a} la fin de la p\'{e}riode de protection, ce qui \'{e}tait r\'{e}gulier.
}
& 
\tiny{
[Considerations]: In law, the first judges considered that, in view of the facts, unfair dismissal should be accepted. Indeed, the bond of trust between the parties had been irretrievably broken on 5 November 2014. A.\_ had refused to sign the warning and the letter terminating her employment contract, with the result that she could no longer be required to comply with the terms of the termination. In addition, the fact that the plaintiff was placed on 100\% work stoppage from 2 to 11 November 2014, then on 100\% work stoppage from 12 November 2014 until 30 november 2014, then 70\% from 1 to 14 December 2014, then 50\% from 15 December 2014 to 15 January 2015 and finally 25\% from 16 to 29 January 2015 could not be construed as a termination. Indeed, the plaintiff could not, as an employee of the company for 40 years, assume her work for 100 years from 2 November 2014, which she would have had to do as a teacher if one takes into account the medical situation of the latter. Moreover, the fact that the employer had dismissed the plaintiff on 7 November 2014, i.e. during a period of protection, could not be considered unfair dismissal. In fact, the employer was not aware that the plaintiff was on 100\% sick leave from 2 to 11 November 2014. Finally, the employer had terminated the employment contract by letter dated 10 December 2014, i.e. the end of the protection period, which was unusual.
}\\
\hline
\multicolumn{2}{|p{\textwidth}|}{\textbf{Scores: }63.70, 57.75, 22.96, 23.57/2.20/22.14}\\
\hline
\end{tabular}
\end{table*}

\begin{table*}[h!]
\ContinuedFloat
\caption{continuation}
\begin{tabular}{|p{0.48\textwidth}|p{0.48\textwidth}|}
\hline
\textbf{Claude 2} (0-shot)
&
\tiny{Translated into English}\\
\hline
\tiny{
    I. Conditions de proc\'{e}dure
    1. Selon l'art. 343 al. 1 ch. 1 CPC, le tribunal \'{e}tablit sa comp\'{e}tence d'office et examine s'il a la comp\'{e}tence mat\'{e}rielle, fonctionnelle et locale pour conna\^{\i}tre de la demande. 
    En l'esp\`{e}ce, le litige porte sur une demande en paiement pour licenciement abusif fond\'{e}e sur un contrat de travail. Il s'agit donc d'un litige relevant du droit du travail, pour lequel le Tribunal des prud'hommes de l'arrondissement de la Sarine est comp\'{e}tent ratione materiae (art. 8 CPC). De plus, la comp\'{e}tence territoriale est \'{e}tablie \`{a} l'art. 34 CPC. La comp\'{e}tence fonctionnelle n'est pas contest\'{e}e.
    Les conditions de proc\'{e}dure sont ainsi remplies. 
    II. En droit
    2. La demanderesse a \'{e}t\'{e} engag\'{e}e par l'entreprise individuelle C.\_ \`{a} un taux d'activit\'{e} de 40\% d\`{e}s le 10 f\'{e}vrier 2014. Suite \`{a} la transformation de C.\_ en la soci\'{e}t\'{e} B.\_ GmbH en ao\^{u}t 2014, cette derni\`{e}re a repris le contrat de travail. La demanderesse a augment\'{e} son taux d'activit\'{e} \`{a} 50\% d\`{e}s le 1er septembre 2014, \`{a} l'essai. 
    3. Le 15 octobre 2014, la demanderesse a \'{e}t\'{e} mise en arr\^{e}t maladie jusqu'au 31 octobre 2014. Le m\^{e}me jour, elle a adress\'{e} un courrier au Service public de l'emploi dans lequel elle mettait en cause le professionnalisme de l'intim\'{e}e. 
    4. Le 30 octobre 2014, la demanderesse a inform\'{e} l'intim\'{e}e de sa reprise de travail le 3 novembre 2014. Toutefois, elle a \'{e}t\'{e} hospitalis\'{e}e du 2 au 4 novembre 2014 et mise \`{a} nouveau en arr\^{e}t maladie. Le 5 novembre 2014, l'intim\'{e}e lui a signifi\'{e} un avertissement ainsi qu'une r\'{e}siliation ordinaire de son
}
& 
\tiny{
    I. Procedural conditions
    1. According to art. 343 al. 1 ch. 1 CPC, the court establishes its jurisdiction ex officio and examines whether it has the material, functional and local jurisdiction to hear the claim. 
    In this case, the dispute concerns a claim for payment for unfair dismissal based on an employment contract. It is therefore an employment law dispute, for which the Tribunal des prud'hommes de l'arrondissement de la Sarine has jurisdiction ratione materiae (art. 8 CPC). In addition, territorial jurisdiction is established by art. 34 CPC. Functional jurisdiction is not contested.
    The procedural requirements are therefore met.
    II. In law
    2. The plaintiff was engaged by the sole proprietorship C.\_ with an activity rate of 40 \% on 10 February 2014. Following the transformation of C.\_ into the company B.\_ GmbH in August 2014, the latter took over the employment contract. The plaintiff increased her rate of activity by 50\% from 1 September 2014, on a trial basis. 
    3. On 15 October 2014, the claimant was placed on sick leave until 31 October 2014. On the same day, she sent a letter to the Public Employment Service in which she questioned the Respondent's professionalism. 
    4. On 30 October 2014, the claimant informed the respondent that she would be returning to work on 3 November 2014. However, she was hospitalised from 2 to 4 November 2014 and placed on sick leave again. On 5 November 2014, the respondent served her with a warning and an ordinary termination of her employment.
}
 \\
\hline
\multicolumn{2}{|p{\textwidth}|}{\textbf{Scores: }70.04, 61.20, 23.83, 24.75/2.92/24.75}\\
\hline
\textbf{LLaMA-2-13B Chat} (0-shot)
&
\tiny{Translated into English}\\
\hline
\tiny{
    Consid\'{e}rant les faits all\'{e}gu\'{e}s et les demandes des parties, le tribunal est appel\'{e} \`{a} \'{e}valuer les dispositions juridiques pertinentes et \`{a} appr\'{e}cier les preuves pour d\'{e}terminer si le licenciement d'A.\_ a \'{e}t\'{e} abusif.
    Tout d'abord, il est important de noter que le contrat de travail entre A.\_ et B.\_ GmbH a \'{e}t\'{e} r\'{e}sili\'{e} pendant une p\'{e}riode de protection, ce qui signifie que la r\'{e}siliation \'{e}tait nulle et non avenue. A.\_ a \'{e}galement signal\'{e} des manquements de l'entreprise et a exprim\'{e} des critiques quant \`{a} la gestion d'E.\_, ce qui peut \^{e}tre consid\'{e}r\'{e} comme une raison valable pour mettre fin au contrat. Cependant, il est important de noter que ces manquements et ces critiques n'ont pas \'{e}t\'{e} formellement \'{e}tay\'{e}s par des preuves solides.
    Ensuite, il est important de consid\'{e}rer le lien de confiance qui existe entre A.\_ et E.\_. Bien que E.\_ ait signal\'{e} que le lien de confiance avait \'{e}t\'{e} rompu en raison des critiques formul\'{e}es par A.\_, il est possible de consid\'{e}rer que le lien de confiance n'a pas \'{e}t\'{e} rompu de mani\`{e}re formelle et que les rapports de travail pouvaient continuer.
    En outre, il est important de prendre en compte la situation de sant\'{e} d'A.\_, qui a \'{e}t\'{e} hospitalis\'{e}e pour des douleurs dorsales et a \'{e}t\'{e} mise en arr\^{e}t de travail \`{a} plusieurs reprises. Bien que les pr\'{e}judices pour la sant\'{e} d'A.\_ soient acknowledgeables, il est important de noter que ces pr\'{e}judices n'ont pas \'{e}t\'{e} directement caus\'{e}s par le licenciement d'A.\_.
    Enfin, il est important de consid\'{e}rer les dispositions juridiques pertinentes. Le Code des obligations suisse (CO) pr\'{e}voit que le licenciement d'un employ\'{e} doit \^{e}tre justifi\'{e} et que le employeur doit fournir des preuves solides pour \'{e}tayer la r\'{e}siliation. Le CO \'{e}galement pr\'{e}voit que le licenciement abusif est une ground for dismissal.
    En r\'{e}sum\'{e}, le tribunal doit \'{e}valuer les dispositions jur
}
&
\tiny{
    Considering the facts of the case and the claims of the parties, the Tribunal is called upon to evaluate the relevant legal provisions and assess the evidence to determine whether A's dismissal was unfair.
    First of all, it is important to note that the employment contract between A.\_ and B.\_ GmbH was terminated during a period of protection, which means that the termination was null and void. A.\_ also pointed out shortcomings in the company and expressed criticism of E.\_'s management, which could be considered a valid reason for terminating the contract. However, it is important to note that these failings and criticisms have not been formally \'{e}tayed'{e} by solid evidence.
    Secondly, it is important to consider the bond of trust that exists between A.\_ and E.\_. Although E.\_ reported that the bond of trust had been broken because of A.\_'s criticisms, it is possible to consider that the bond of trust had not been formally broken and that the working relationship could continue.
    In addition, it is important to take into account the health situation of A.\_, who has been hospitalised for back pain and has been off work on several occasions. Although the harm to A.e's health is acknowledged, it is important to note that this harm was not directly caused by A.e's dismissal.
    Finally, it is important to consider the relevant legal provisions. The Swiss Code of Obligations (CO) provides that the dismissal of an employee must be justified and that the employer must provide solid evidence to support the dismissal. The CO also provides that unfair dismissal is a ground for dismissal.
    In summing up, the court must assess the legal provisions of the case.
}
\\
\hline
\multicolumn{2}{|p{\textwidth}|}{\textbf{Scores: }63.92, 49.30, 13.62, 14.63/1.26/13.94} \\
\hline
\end{tabular}
\vspace{-2ex}
\end{table*}

\clearpage
\subsection{\acf{LDS}}
\label{app:lds_error_analysis}
\autoref{tab:lds-analysis} shows a comparison of generated text from GPT-3.5-Turbo-16K, GPT-4, Claude Instant and Claude 2.

\begin{table*}[ht]
\centering
\caption{Comparison of the generated regeste (\ac{LDS} task) from four \acp{LLM} in different settings. The scores of each example is stated below in the following order: BERTScore, BLEU-Score, METEOR, ROUGE-1 / ROUGE-2 / ROUGE-L. Due to the size of the input, only the target is displayed. We provide an English translated version on the right.} 
\label{tab:lds-analysis}
\begin{tabular}{|p{0.48\textwidth}|p{0.48\textwidth}|}
\multicolumn{2}{p{1\textwidth}}{
\textbf{Generated Regeste} 
}\\
\hline
\tiny{Decision ID: 3e70603c-bbcd-47de-a066-23124945fcc1, Year: 2017, Language: German, Court: CH\_BGE, Source: \href{https://www.bger.ch/ext/eurospider/live/de/php/clir/http/index.php?highlight_docid=atf%3A%2F%2F143-III-233%3Ade&lang=de&zoom=&type=show_document}{BGE 143 III 233}}
& 
\tiny{Translated into English} \\
\hline
\tiny{\textbf{Input: } Sachverhalt ab Seite 234 BGE 143 III 233 S. 234 A. Mit Eingabe vom 22. März 2013 begehrte B. (Ehemann) beim Zivilgericht Basel-Stadt die Scheidung der Ehe mit A. (Ehefrau). Mit Entscheid vom 1. Mai 2013 wurde er verpflichtet, der Ehefrau für die Dauer der Aufhebung des gemeinsamen Haushalts einen monatlichen Unterhaltsbeitrag von Fr. 1'520.- bis 31. August 2013 und von Fr. 3'000.- ab September 2013 zu entrichten. B. B.a Am 2. August 2015 ersuchte der Ehemann um Reduktion des Unterhaltsbeitrages für die Ehefrau auf monatlich Fr. 1'500.- resp. Fr. 500.-. Die Ehefrau widersetzte sich diesem Begehren. Mit Entscheid vom 10. November 2015 hob der Instruktionsrichter des Zivilgerichts Basel-Stadt den monatlichen Unterhaltsbeitrag für die Ehefrau für die Dauer des Scheidungsverfahrens rückwirkend per 1. Juli 2015 auf. B.b Am 1. März 2016 hiess der Ausschuss des Appellationsgerichts des Kantons Basel-Stadt (nachfolgend: Appellationsgericht) die Berufung der Ehefrau teilweise gut und änderte die erstinstanzliche Unterhaltsregelung für die Ehefrau ab. Der vom Ehemann an den Unterhalt der Ehefrau zu bezahlende Unterhaltsbeitrag von Fr. 3'000.- wurde per 1. August 2015 auf Fr. 500.- reduziert und mit Wirkung per 1. September 2015 aufgehoben. C. Die Ehefrau (nachfolgend: Beschwerdeführerin) hat am 22. April 2016 (Postaufgabe) gegen den vorgenannten Entscheid des Appellationsgerichts beim Bundesgericht Beschwerde erhoben. Sie beantragt, den angefochtenen Entscheid aufzuheben und das Begehren des Ehemannes (nachfolgend: Beschwerdegegner) um Aufhebung des Ehegattenunterhalts für die Dauer des Scheidungsverfahrens abzuweisen. Der Beschwerdegegner hat sich am 16. Februar 2017 vernehmen lassen. Er schliesst auf Abweisung der Beschwerde. Das Bundesgericht hebt die kantonalen Entscheide auf und weist das Gesuch des Beschwerdegegners um Abänderung des Ehegattenunterhalts für die Dauer des Scheidungsverfahrens ab. 
(Zusammenfassung) BGE 143 III 233 S. 235 Erwägungen Aus den Erwägungen: 3. 3.1 Abänderungsgrund bildet die veränderte Einkommenslage beim Beschwerdegegner. Die erste Instanz hat dessen seit Januar 2015 bestehende Arbeitslosigkeit berücksichtigt und dabei angenommen, der Beschwerdegegner habe seine Arbeitsstelle nicht freiwillig, sondern einzig deshalb gekündigt, weil ihm die Auflösung des Arbeitsverhältnisses von seiner Arbeitgeberin nahegelegt worden sei. Das Appellationsgericht hat geprüft, ob die Kündigung als eigenmächtiges, mithin rechtsmissbräuchliches Verhalten zu werten sei, das einer Abänderung der Eheschutzmassnahmen entgegensteht.
Als Ergebnis seiner Würdigung der tatsächlichen Gegebenheiten ist es zum Schluss gelangt, die Beschwerdeführerin habe nicht glaubhaft gemacht, dass die Kündigung des Arbeitsverhältnisses durch den Beschwerdegegner ausschliesslich zu ihrer Schädigung erfolgt sei. Das Appellationsgericht hat daher - wie die erste Instanz - dem Beschwerdegegner kein hypothetisches Einkommen angerechnet. 3.2 Bei der Bemessung des Unterhaltsbeitrages ist grundsätzlich vom tatsächlich erzielten Einkommen des Unterhaltspflichtigen auszugehen. Soweit dieses Einkommen allerdings nicht ausreicht, um den ausgewiesenen Bedarf zu decken, kann ein hypothetisches Einkommen angerechnet werden, sofern dieses zu erreichen zumutbar und möglich ist ( BGE 137 III 118 E. 2.3). [...] 
}
& 
\tiny{
    Facts from page 234 BGE 143 III 233 p. 234 A. By petition dated 22 March 2013, B. (husband) applied to the Civil Court of Basel-Stadt for a divorce from his marriage to A. (wife). By decision of 1 May 2013, he was ordered to pay the wife a monthly maintenance contribution of CHF 1,520 until 31 August 2013 and CHF 3,000 from September 2013 for the duration of the dissolution of the joint household. B. B.a On 2 August 2015, the husband requested that the maintenance contribution for the wife be reduced to CHF 1,500 per month and CHF 500 per month respectively. The wife opposed this request. In a decision dated 10 November 2015, the instructing judge of the Civil Court of Basel-Stadt cancelled the monthly maintenance contribution for the wife for the duration of the divorce proceedings with retroactive effect from 1 July 2015. B.b On 1 March 2016, the Committee of the Court of Appeal of the Canton of Basel-Stadt (hereinafter: Court of Appeal) partially upheld the wife's appeal and modified the first-instance maintenance arrangement for the wife. The maintenance contribution of CHF 3,000 to be paid by the husband to the wife was reduced to CHF 500 with effect from 1 August 2015 and cancelled with effect from 1 September 2015. 
    C. On 22 April 2016 (date of posting), the wife (hereinafter: complainant) lodged an appeal with the Federal Supreme Court against the aforementioned decision of the Court of Appeal. She requested that the contested decision be set aside and that the husband's (hereinafter: respondent) request for cancellation of spousal maintenance for the duration of the divorce proceedings be dismissed. The respondent was heard on 16 February 2017. He concludes that the appeal should be dismissed. The Federal Supreme Court overturns the cantonal decisions and rejects the respondent's request for a modification of spousal maintenance for the duration of the divorce proceedings. 
    (Summary) BGE 143 III 233 p. 235 Considerations From the considerations: 3. 3.1 The reason for the modification is the change in the respondent's income situation. The first instance took into account his unemployment since January 2015 and assumed that the respondent had not resigned from his job voluntarily, but only because he had been advised by his employer to terminate the employment relationship. The Court of Appeal examined whether the termination was to be regarded as unauthorised, and therefore abusive, behaviour that precluded a modification of the matrimonial protection measures.
    As a result of its assessment of the factual circumstances, it concluded that the complainant had not credibly demonstrated that the respondent had terminated the employment relationship solely to her detriment. The Court of Appeal therefore - like the first instance - did not impute any hypothetical income to the respondent. 3.2 When calculating the maintenance contribution, the actual income of the maintenance debtor must be taken as a basis. However, if this income is not sufficient to cover the stated needs, a hypothetical income can be taken into account, provided that it is reasonable and possible to achieve this ( BGE 137 III 118 E. 2.3). [...]
}
\\
\hline
\end{tabular}
\end{table*}

\begin{table*}[h!]
\ContinuedFloat
\caption{continuation}
\begin{tabular}{|p{0.48\textwidth}|p{0.48\textwidth}|}
\hline
\tiny{\textbf{Target: }
    Regeste
    Abänderung des Unterhaltsbeitrages für die Dauer des Scheidungsverfahrens infolge veränderter Verhältnisse; Rechtsmissbrauch (
    Art. 179 Abs. 1,
    Art. 2 Abs. 2 ZGB
    ).
    Vermindert der Unterhaltspflichtige sein Einkommen in Schädigungsabsicht, ist eine Abänderung der Unterhaltsleistung selbst dann ausgeschlossen, wenn die Einkommensreduktion nicht mehr rückgängig gemacht werden kann (Änderung der Rechtsprechung; E. 3).}
& 
\tiny{
    Rulings
    Modification of the maintenance contribution for the duration of the divorce proceedings due to changed circumstances; abuse of rights (
    Art. 179 para. 1,
    Art. 2 para. 2 ZGB
    ).
    If the maintenance debtor reduces his or her income with the intention of causing damage, a modification of the maintenance payment is excluded even if the reduction in income can no longer be reversed (change in case law; E. 3).}\\
\hline
\multicolumn{2}{|p{1\textwidth}|}{
    \tiny{Decision ID: \textit{3e70603c-bbcd-47de-a066-23124945fcc1}, Year: \textit{2017}, 
    Language: German, Court: \textit{CH\_BGE}}}\\
\hline
\textbf{GPT-4} (1-shot)
& 
\tiny{Translated into English} \\
\hline
    \tiny{Regeste
    Ab\"{a}nderung von Unterhaltsbeitr\"{a}gen bei Reduktion des Einkommens (BGE 137 III 118).
    Bei der Bemessung des Unterhaltsbeitrages ist grunds\"{a}tzlich vom tats\"{a}chlich erzielten Einkommen des Unterhaltspflichtigen auszugehen. Eine Ab\"{a}nderung der Unterhaltsleistung ist jedoch auszuschliessen, wenn der Unterhaltspflichtige sein Einkommen in Sch\"{a}digungsabsicht vermindert hat, selbst dann wenn die Einkommensverminderung nicht mehr r\"{u}ckg\"{a}ngig gemacht werden kann (E. 3.4).}
&
    \tiny{Rulings
    Modification of maintenance contributions in the event of a reduction in income (BGE 137 III 118).
    When calculating the maintenance contribution, the actual income of the maintenance debtor must be taken as a basis. However, a reduction in the maintenance payment must be ruled out if the maintenance debtor has reduced his or her income with the intention to pay maintenance, even if the reduction in income can no longer be attributed to him or her (E. 3.4).}
\\
\hline
\multicolumn{2}{|p{1\textwidth}|}{79.82, 68.93, 55.47, 45.16/28.04/43.01}\\
\hline
\textbf{Claude 2} (1-shot)
& 
\tiny{Translated into English} \\
\hline
\tiny{Regeste
Art. 137 ZGB (Unterhaltsab\"{a}nderung). Ab\"{a}nderung des Ehegattenunterhalts bei ver\"{a}nderter Einkommenslage. Bei freiwilliger Sch\"{a}digung ist eine Ab\"{a}nderung selbst dann ausgeschlossen, wenn die Einkommensverminderung nicht mehr r\"{u}ckg\"{a}ngig gemacht werden kann (E. 3).}
&
\tiny{
    Regulations
Art. 137 CC (modification of maintenance). Modification of spousal maintenance in the event of a change in income. In the case of voluntary termination, a modification is excluded even if the reduction in income can no longer be reversed (E. 3).
}
\\
\hline
\multicolumn{2}{|p{1\textwidth}|}{83.88, 47.53, 42.44, 53.33/29.27/53.33 }\\
\hline
\textbf{GPT-3.5-Turbo-16K} (0-shot)
& 
\tiny{Translated into English} \\
\hline
\tiny{BGE 143 III 233 S. 235 - Ehegattenunterhalt: Ab\"{a}nderung aufgrund ver\"{a}nderter Einkommenslage des Unterhaltspflichtigen; Anrechnung eines hypothetischen Einkommens; Rechtsmissbr\"{a}uchliches Verhalten des Unterhaltspflichtigen. 
Der unterhaltspflichtige Ehemann beantragte eine Ab\"{a}nderung des Ehegattenunterhalts aufgrund seiner ver\"{a}nderten Einkommenslage. Das Gericht stellte fest, dass eine Ab\"{a}nderung ausgeschlossen ist, wenn die ver\"{a}nderte Einkommenslage durch eigenm\"{a}chtiges, widerrechtliches oder rechtsmissbr\"{a}uchliches Verhalten des Unterhaltspflichtigen herbeigef\"{u}hrt wurde. Es wurde entschieden, dass in diesem Fall kein hypothetisches Einkommen angerechnet werden kann, selbst wenn die Einkommensverminderung nicht mehr r\"{u}ckg\"{a}ngig gemacht werden kann.}
& 
\tiny{
    BGE 143 III 233 p. 235 - Spousal maintenance: Modification due to a change in the debtor's income situation; imputation of a hypothetical income; abuse of law by the debtor. 
The husband, who was obliged to pay maintenance, applied for a modification of the spousal maintenance on the basis of his changed income situation. The court found that a modification is excluded if the changed income situation was brought about by unlawful, wrongful or abusive behaviour on the part of the maintenance debtor. It was decided that no hypothetical income can be taken into account in this case, even if the reduction in income can no longer be attributed.
}
 \\
\hline
\multicolumn{2}{|p{1\textwidth}|}{72.21, 43.71, 44.39, 27.52/12.60/23.85}\\
\hline
\textbf{Claude Instant} (0-shot)
& 
\tiny{Translated into English} \\
\hline
\tiny{Regeste:
a) Art. 107ff ZGB, 128 III 4 E. 4 
b) Ehegattenunterhalt, Ab\"{a}nderungsverfahren, hypothetisches Einkommen
c) Das Bundesgericht hielt fest, dass dem Unterhaltspflichtigen, der sein Einkommen in Sch\"{a}digungsabsicht vermindert hat, selbst dann ein hypothetisches Einkommen anzurechnen ist, wenn die Einkommensverminderung nicht mehr r\"{u}ckg\"{a}ngig gemacht werden kann. Somit ist eine Ab\"{a}nderung der Unterhaltsleistung in einem solchen Fall ausgeschlossen.}
&
\tiny{
    Regeste:
a) Art. 107ff ZGB, 128 III 4 E. 4 
b) Spousal maintenance, modification procedure, hypothetical income
c) The Federal Supreme Court held that a hypothetical income must be credited to the maintenance debtor who has reduced his or her income with the intention to pay maintenance, even if the reduction in income can no longer be attributed to him or her. A modification of the maintenance payment is therefore excluded in such a case.
}
 \\
\hline
\multicolumn{2}{|p{1\textwidth}|}{76.91, 67.20, 49.20, 41.67/24.30/41.67} \\
\hline
\end{tabular}
\end{table*}

\clearpage\section{Prompts}
\label{app:prompts}

In the following, we provide the prompts that we used for experiments with LLMs over 2B parameters.

\subsection{Text Classification}
We created our prompts in a style similar to that used in \citet{Chalkidis2023ChatGPTMP}. First, we introduce the document, e.g., the court decision. Then, we provide an explanation of the labels and the task in general. Finally, we present the labels from which the model can choose. To limit the model from producing excessive output, we conclude the prompt with a cloze-style sentence, such as 'The relevant label in this case is:', which the model simply needs to complete. 
\subsubsection{Criticality Prediction}

\textbf{\acf{CP} LD Facts}\\
Given the facts from the following Swiss Federal Supreme Court Decision:\\\{INPUT FROM THE VALIDATION SET\}\\Federal Supreme Court Decisions in Switzerland that are published additionally get the label critical, those Federal Supreme Court Decisions that are not published additionally, get the label non-critical. Therefore, there are two labels to choose from:\\
- critical\\- non-critical\\The relevant label in this case is:

\textbf{\acf{CP} LD Considerations}\\
Given the considerations from the following Swiss Federal Supreme Court Decision:\\\{INPUT FROM THE VALIDATION SET\}\\Federal Supreme Court Decisions in Switzerland that are published additionally get the label critical, those Federal Supreme Court Decisions that are not published additionally, get the label non-critical. Therefore, there are two labels to choose from:\\
- critical\\- non-critical\\The relevant label in this case is:

\textbf{\acf{CP} Citation Facts}\\
Given the facts from the following Swiss Federal Supreme Court Decision:\\\{INPUT FROM THE VALIDATION SET\}\\How likely is it that this Swiss Federal Supreme Court Decision gets cited. Choose between one of the following labels (a bigger number in the label means that the court decision is more likely to be cited):\\
- critical-1\\- critical-2\\- critical-3\\- critical-4\\The relevant label in this case is:

\textbf{\acf{CP} Citation Considerations}\\Given the considerations from the following Swiss Federal Supreme Court Decision:\\\{INPUT FROM THE VALIDATION SET\}\\How likely is it that this Swiss Federal Supreme Court Decision gets cited. Choose between one of the following labels (a bigger number in the label means that the court decision is more likely to be cited): 
- critical-1\\- critical-2\\- critical-3\\- critical-4\\The relevant label in this case is:

\subsubsection{Judgment Prediction}

\textbf{\acf{JP} Facts}\\
Given the facts from the following court decision:\\\{INPUT FROM THE VALIDATION SET\}\\Will this court decision get approved or dismissed? There are two labels to choose from:\\
- dismissal\\- approval\\The relevant label in this case is:

\textbf{\acf{JP} Considerations}\\
Given the considerations from the following court decision:\\\{INPUT FROM THE VALIDATION SET\}\\Will this court decision get approved or dismissed? There are two labels to choose from:\\
- dismissal\\- approval\\The relevant label in this case is:

\subsubsection{(Sub) Law Area Prediction}
\textbf{\acf{LAP} Facts}\\
Given the facts from the following court decision:\\\{INPUT FROM THE VALIDATION SET\}\\Which topic/law area is relevant out of the following options:\\
- Civil\\- Public\\- Criminal\\- Social\\The relevant option is:

\textbf{\acf{LAP} Considerations}\\
Given the considerations from the following court decision:\\\{INPUT FROM THE VALIDATION SET\}\\Which topic/law area is relevant out of the following options:\\
- Civil\\- Public\\- Criminal\\- Social\\The relevant option is:

\textbf{\acf{SLAP} Facts}\\
Given the facts from the following court decision:\\\{INPUT FROM THE VALIDATION SET\}\\Which topic/law area is relevant out of the following options:\\
- Rental and Lease\\- Employment Contract\\- Bankruptcy\\- Family\\- Competition and Antitrust\\- Intellectual Property\\- Substantive Criminal\\- Criminal Procedure\\- Tax\\- Urban Planning and Environmental\\- Expropriation\\- Public Administration\\- Other Fiscal\\The relevant option is:

\textbf{\acf{SLAP} Considerations}\\
Given the considerations from the following court decision:\\\{INPUT FROM THE VALIDATION SET\}\\Which topic/law area is relevant out of the following options:\\
- Rental and Lease\\- Employment Contract\\- Bankruptcy\\- Family\\- Competition and Antitrust\\- Intellectual Property\\- Substantive Criminal\\- Criminal Procedure\\- Tax\\- Urban Planning and Environmental\\- Expropriation\\- Public Administration\\- Other Fiscal\\The relevant option is:

\subsection{Text Generation}
In the subsequent sections, we detail the prompts employed for the 0-shot setup. For the 1-shot experiments, we consistently appended the same example in the respective language to the 0-shot instruction.

\subsubsection{\acf{CVG}}
\textbf{\acf{CVG} in German}\\
'Ziel: Generiere Erw\"{a}gungen basierend auf dem gegebenen Sachverhalt eines Schweizer Gerichtsurteils.
\\Hintergrund: Ein Schweizer Gerichtsurteil besteht aus Rubrum, Sachverhalt, Erw\"{a}gungen, Dispositiv (Urteilsformel) und Unterschrift. Die Erw\"{a}gungen sind die rechtliche W\"{u}rdigung des Geschehens durch das Gericht.
\\Anweisung:
\\-Sachverhalt Verstehen: Der gegebene Sachverhalt enth\"{a}lt bestrittene und unbestrittene Fakten, die Begehren der Parteien, das Beweisverfahren und die Prozessgeschichte.
\\-Beginne mit Prozessvoraussetzungen: Pr\"{u}fe zun\"{a}chst, ob die Prozessvoraussetzungen (z.B. Zust\"{a}ndigkeit des Gerichts) erf\"{u}llt sind. Wenn nicht strittig, reicht es aus zu best\"{a}tigen, dass die Voraussetzungen erf\"{u}llt sind.
\\-Rechtliche W\"{u}rdigung: Eruiere relevante Rechtss\"{a}tze basierend auf den behaupteten und rechtlich relevanten Tatsachen. \\-Setze dich mit den rechtlichen Standpunkten der Parteien auseinander. \\-Beachte die Beweislastverteilung und w\"{u}rdige die Beweise frei, aber ber\"{u}cksichtige relevante gesetzliche Beweisregeln. \\-Iura novit curia: Deine rechtliche W\"{u}rdigung muss nicht zwangsl\"{a}ufig dem rechtlichen Vorbringen der Parteien entsprechen. Ber\"{u}cksichtige andere m\"{o}gliche Argumentationslinien. \\-Zusammenfassung: Fasse am Ende deine Erw\"{a}gungen, das Ergebnis Ihrer rechtlichen W\"{u}rdigung, zusammen. \\-Output: Der generierte Text sollte strukturiert, klar und in der Form von typischen Erw\"{a}gungen eines Schweizer Gerichtsurteils sein. \\\\
\{Sachverhalt des Schweizer Gerichtsurteils\}: \\
\{INPUT FROM THE VALIDATION SET\}\\\\
\{Erwägungen\}: \\

\textbf{\acf{CVG} in French}\\
But: G\'{e}n\`{e}re des consid\'{e}rations bas\'{e}es sur les faits donn\'{e}s d'un jugement suisse.
\\Contexte: Un jugement suisse est compos\'{e} du rubrum, des faits, des consid\'{e}rations, du dispositif (formule du jugement) et de la signature. Les consid\'{e}rations sont l'appr\'{e}ciation juridique des \'{e}v\'{e}nements par le tribunal.
\\Instructions:
\\- Comprends les faits: Les faits donn\'{e}s contiennent des faits contest\'{e}s et non contest\'{e}s, les demandes des parties, la proc\'{e}dure de preuve et l'historique du proc\`{e}s.
\\- Commence par les conditions de procédure: Vérifie d'abord si les conditions de procédure (par exemple, la compétence du tribunal) sont remplies. Si cela n'est pas contesté, il suffit de confirmer que les conditions sont remplies. \\- Appréciation juridique: Évalue les dispositions juridiques pertinentes basées sur les faits allégués et juridiquement pertinents. \\- Confronte-toi aux points de vue juridiques des parties. \\- Tiens compte de la répartition de la charge de la preuve et évalue les preuves librement, mais tiens compte des règles légales de preuve pertinentes. \\- Iura novit curia: Ton appréciation juridique ne doit pas nécessairement correspondre aux arguments juridiques présentés par les parties. Considère d'autres lignes d'argumentation possibles. \\- Résumé: Résume à la fin de tes considérations le résultat de ton appréciation juridique. \\- Résultat: Le texte généré devrait être structuré, clair et sous la forme de considérations typiques d'un jugement suisse.
\\\\\{Faits du jugement suisse\}:\\
\{INPUT FROM THE VALIDATION SET\}\\\\
\{Consid\'{e}rations\}:

\textbf{\acf{CVG} in Italian}\\
Obiettivo: Genera considerazioni basate sui fatti presentati in una sentenza svizzera.
\\Contesto: Una sentenza svizzera si compone di rubrum, fatti, considerazioni, dispositivo (formula della sentenza) e firma. Le considerazioni rappresentano la valutazione giuridica degli eventi da parte del tribunale.
\\Istruzioni:
\\- Comprendi i fatti: I fatti presentati includono fatti contestati e non contestati, le richieste delle parti, la procedura probatoria e la storia del processo.
\\- Inizia con le condizioni processuali: Verifica prima di tutto se le condizioni processuali (ad es. la competenza del tribunale) sono soddisfatte. Se non contestate, basta confermare che le condizioni sono state soddisfatte.
\\- Valutazione giuridica: Valuta le norme giuridiche rilevanti in base ai fatti affermati e giuridicamente rilevanti. \\- Confrontati con i punti di vista giuridici delle parti. \\- Tieni conto della distribuzione dell'onere della prova e valuta le prove liberamente, ma considera le regole di prova legalmente rilevanti. - Iura novit curia: La tua valutazione giuridica non deve necessariamente corrispondere alle argomentazioni giuridiche delle parti. Considera altre possibili linee di argomentazione. \\- Riassunto: Riassumi alla fine delle tue considerazioni il risultato della tua valutazione giuridica. \\- Risultato: Il testo generato dovrebbe essere strutturato, chiaro e nella forma di considerazioni tipiche di una sentenza svizzera."
\\\\\{Fatti della sentenza svizzera\}:\\
\{INPUT FROM THE VALIDATION SET\}\\\\
\{Considerazioni\}:

\subsubsection{\acf{LDS}}
In the \ac{LDS} task, we used only one German prompt because the output (the regeste) is always in German in our dataset. Despite the input being multilingual, models tend to generate in the prompt's language, regardless of the input's language.
\\\\\textbf{\acf{LDS}}\\
Ziel: Generiere eine Regeste basierend auf einem Schweizer Gerichtsurteils.
\\Hintergrund: Ein Schweizer Gerichtsurteil setzt sich aus Sachverhalt, Erw\"{a}gungen und Dispositiv zusammen. Die Regeste dient als Kurzzusammenfassung und beinhaltet Leits\"{a}tze des Urteils. Nur Leitentscheide haben eine Regeste.
\\Anweisung:
\\1. Sachverhalt: Lies und verstehe den gegebenen Sachverhalt.
\\2. Erw\"{a}gungen: Analysiere die Erw\"{a}gungen, um die Hauptargumente und Gr\"{u}nde zu identifizieren.
\\3. Dispositiv: Beachte das Dispositiv, da es das endg\"{u}ltige Urteil enth\"{a}lt.
\\4. Erstelle die Regeste: Die Regeste sollte aus drei sehr kurzen Teilen bestehen: a. Zitiere die wichtigsten relevanten Artikelziffern (ohne den Artikeltitel). b. Nenne kurze, relevante, deskriptive Keywords, \"{u}ber die Thematik des Falls. c. Formuliere einen sehr kurzen Fliesstext, der die wichtigsten Erw\"{a}gungen zitiert und kurz zusammenfasst.
\\Output: Die Regeste sollte eine klare und strukturierte Kurzzusammenfassung des Urteils bieten, die aus zitierten Artikeln, Keywords und einem sehr kurzen Fliesstext besteht.
\\\\
\{Gegebener Sachverhalt, Erwägungen und Dispositiv\}:\\
\{INPUT FROM THE VALIDATION SET\}
\\\\\{Regeste auf Deutsch\}:

\clearpage\section{Example Generations}
\label{app:example_generations}


 \autoref{tab:examples_court_view} and \autoref{tab:examples_lds} show excerpts of examples produced by the best model for the CVG and LDS tasks, respectively.

\onecolumn
{\tiny
\begin{longtable}{l|p{13cm}}
\caption{Nine examples of generated considerations by mT5\textsubscript{Large} with input length 2048 across three languages, showcasing high, average, and low scored outputs in \ac{CVG} Task}\\
\hline
Target & considerations: Erwägungen: 1. 1.1 Der angefochtene Entscheid ist in Anwendung von Sozialversicherungsrecht ergangen. Die Sozialversicherungsrechtliche Abteilung des Verwaltungsgerichts beurteilt gemäss Art. 57 des Bundesgesetzes vom 6. Oktober 2000 über den Allgemeinen Teil des Sozialversicherungsrechts (ATSG; SR 830.1) i.V.m. Art. 54 Abs. 1 lit. a des kantonalen Gesetzes vom 11. Juni 2009 über die Organisation der Gerichtsbehörden und der Staatsanwaltschaft (GSOG; BSG 161.1) Beschwerden gegen solche Entscheide. Die Beschwerdeführerin ist im vorinstanzlichen Verfahren mit ihren Anträgen nicht durchgedrungen, durch den angefochtenen Entscheid berührt und hat ein schutzwürdiges Interesse an dessen Aufhebung, weshalb sie zur Beschwerde befugt ist (Art. 59 ATSG). Die örtliche Zuständigkeit ist gegeben (Art. 58 ATSG). Da auch die Bestimmungen über Frist (Art. 60 ATSG) sowie Form (Art. 61 lit. b ATSG; Art. 81 Abs. 1 i.V.m. Art. 32 des kantonalen Gesetzes vom 23. Mai 1989 Urteil des Verwaltungsgerichts des Kantons Bern vom 6. Dez. 2022, UV/21/844, Seite 4 über die Verwaltungsrechtspflege [VRPG; BSG 155.21]) eingehalten sind, ist auf die Beschwerde einzutreten. 1.2 Anfechtungsobjekt bildet der Einspracheentscheid vom 5. November 2021 (act. II 45). Streitig und zu prüfen ist, ob die Beschwerdegegnerin ihre Leistungen im Zusammenhang mit dem Ereignis vom 19. Juni 2020 zu Recht per 30. September 2020 eingestellt hat. 1.3 Die Abteilungen urteilen gewöhnlich in einer Kammer bestehend aus drei Richterinnen oder Richtern (Art. 56 Abs. 1 GSOG). 1.4 Das Gericht überprüft den angefochtenen Entscheid frei und ist an die Begehren der Parteien nicht gebunden (Art. 61 lit. c und d ATSG; Art. 80 lit. c Ziff. 1 und Art. 84 Abs. 3 VRPG). 2. 2.1 Die Zusprechung von Leistungen der obligatorischen Unfallversicherung setzt grundsätzlich das Vorliegen eines Berufsunfalles, eines Nichtberufsunfalles oder einer Berufskrankheit (Art. 6 Abs\\ 
\hline
Predicted & considerations: Erwägungen: 1. 1.1 Der angefochtene Entscheid ist in Anwendung von Sozialversicherungsrecht ergangen. Die Sozialversicherungsrechtliche Abteilung des Verwaltungsgerichts beurteilt gemäss Art. 57 des Bundesgesetzes über den Allgemeinen Teil des Sozialversicherungsrechts vom 6. Oktober 2000 (ATSG; SR 830.1) i.V.m. Art. 54 Abs. 1 lit. a des kantonalen Gesetzes über die Organisation der Gerichtsbehörden und der Staatsanwaltschaft vom 11. Juni 2009 (GSOG; BSG 161.1) Beschwerden gegen solche Entscheide. Die Beschwerdeführerin ist im vorinstanzlichen Verfahren mit ihren Anträgen nicht durchgedrungen, durch den angefochtenen Entscheid berührt und hat ein schutzwürdiges Interesse an dessen Aufhebung, weshalb sie zur Beschwerde befugt ist (Art. 59 ATSG). Die örtliche Zuständigkeit ist gegeben (Art. 69 Abs. 1 lit. a des Bundesgesetzes über die Unfallversicherung vom 20. Dezember 1946 [UVG; SR 832.20]). Da auch die Bestimmungen über Frist (Art. 60 ATSG) sowie Form (Art. 61 lit. b ATSG; Art. 81 Abs. 1 i.V.m. Art. 32 des kantonalen Gesetzes über die Verwaltungsrechtspflege vom 23. Mai 1989 [VRPG; BSG 155.21]) eingehalten sind, ist auf die Beschwerde einzutreten. 1.2 Das Gericht überprüft den angefochtenen Entscheid frei und ist an die Begehren der Parteien nicht gebunden (Art. 61 lit. c und d ATSG; Art. 80 lit. c Ziff. 1 und Art. 84 Abs. 3 VRPG). 1.3 Der Beschwerdeführer hat am vorinstanzlichen Verfahren teilgenommen, ist durch den angefochtenen Entscheid besonders berührt und hat ein schutzwürdiges Interesse an dessen Aufhebung oder Änderung. Auf die Beschwerde ist einzutreten. 2. 2.1 Die Beschwerdegegnerin hat die Bestimmungen auf den Anspruch auf Heilbehandlung und Taggeld in Form von Heilbehandlung zufolge ihrer im August 2021 (act. II 42 und 46) eingereichten medizinischen Berichte zutreffend wiedergegeben. Darauf wird verwiesen. 2.2 Nach der Rechtsprechung ist der Beschwerdeführer durch den angefochtenen Entscheid berührt und haben ein schutzwürdiges\\ 
\hline
Scores & BERT: 91.69, BLEU: 84.17, METEOR: 62.43, ROUGE-1/2/L: 75.49 / 65.78 / 75.49 \\ 
\hline
\end{longtable}
}
\onecolumn
{\tiny
\begin{longtable}{l|p{13cm}}
\hline
Target & considerations: EN DROIT 1. Conformément à l'art. 134 al. 1 let. a ch. 2 de la loi sur l'organisation judiciaire, du 26 septembre 2010 (LOJ - E 2 05 ) en vigueur dès le 1 er janvier 2011, la chambre des assurances sociales de la Cour de justice connaît, en instance unique, des contestations prévues à l'art. 56 de la loi fédérale sur la partie générale du droit des assurances sociales, du 6 octobre 2000 (LPGA - RS 830.1 ) relatives à la loi fédérale sur l’assurance-invalidité du 19 juin 1959 (LAI - RS 831.20 ).![endif]>![if> Sa compétence pour juger du cas d’espèce est ainsi établie. 2. À teneur de l’art. 1 al. 1 LAI, les dispositions de la LPGA s'appliquent à l’assurance-invalidité (ci-après : AI), à moins que la loi n’y déroge expressément.![endif]>![if> Toutefois, les modifications légales contenues dans la LPGA constituent, en règle générale, une version formalisée dans la loi de la jurisprudence relative aux notions correspondantes avant l’entrée en vigueur de la LPGA ; il n’en découle aucune modification du point de vue de leur contenu, de sorte que la jurisprudence développée à leur propos peut être reprise et appliquée (ATF 130 V 343 consid. 3). 3. Les modifications de la loi fédérale sur l’assurance-invalidité du 21 mars 2003 (4 ème révision), du 6 octobre 2006 (5 ème révision) et du 18 mars 2011 (révision 6a), entrées en vigueur le 1 er janvier 2004, respectivement, le 1 er janvier 2008 et le 1 er janvier 2012, entraînent la modification de nombreuses dispositions légales dans le domaine de l’assurance-invalidité.\\ 
\hline
Predicted & considerations: EN DROIT 1. Conformément à l'art. 134 al. 1 let. a ch. 4 de la loi sur l'organisation judiciaire, du 26 septembre 2010 (LOJ; RS E 2 05 ) en vigueur dès le 1 er janvier 2011, la Chambre des assurances sociales de la Cour de justice connaît, en instance unique, des contestations prévues à l'art. 56 de la loi fédérale sur la partie générale du droit des assurances sociales, du 6 octobre 2000 (LPGA; RS 830.1 ) relatives à la loi fédérale sur l’assurance-invalidité du 19 juin 1959 (LAI; RS 831.20 ).![endif]>![if> Sa compétence pour juger du cas d’espèce est ainsi établie.![endif]>![if> 2. Les dispositions de la LPGA s'appliquent à l'assurance-invalidité, à moins que la loi n'y déroge expressément.![endif]>![if> Toutefois, les modifications légales contenues dans la LPGA constituent, en règle générale, une version formalisée dans la loi de la jurisprudence relative aux notions correspondantes avant l'entrée en vigueur de la LPGA; il n'en découle aucune modification du point de vue de leur contenu, de sorte que la jurisprudence développée à leur propos peut être reprise et appliquée (ATF 130 V 230 consid. 1.1; 335 consid. 1.2; ATF 129 V 4 consid. 1.2; ATF 129 V 4 consid. 1.2; ATF 127 V 467 consid. 1, 126 V 136 consid. 4b et les références). 3. A teneur de l'art. 17 al. 1 de la loi fédérale sur la partie générale du droit des assurances sociales, du 6 octobre 2000 (LPGA; RS 830.1 ), les modifications légales contenues dans la LPGA constituent, en règle générale, une version formalisée dans la LAI.\\ 
\hline
Scores & BERT: 91.27, BLEU: 85.8, METEOR: 68.12, ROUGE-1/2/L: 78.29 / 66.82 / 78.29 \\
\hline
\end{longtable}
}
\onecolumn
{\tiny
\begin{longtable}{l|p{13cm}}
\hline
Target & considerations: in diritto In ordine 2.1. La presente vertenza non pone questioni giuridiche di principio e non è di rilevante importanza (ad esempio per la difficoltà dell’istruttoria o della valutazione delle prove). Il TCA può dunque decidere nella composizione di un Giudice unico ai sensi dell'articolo 49 cpv. 2 LOG (STF 9C\_699/2014 del 31 agosto 2015; 8C\_855/2010 dell’11 luglio 2011; 9C\_211/2010 del 18 febbraio 2011). Nel merito 2.2. Oggetto del contendere è sapere se a ragione l’Ufficio AI ha soppresso all’assicurato il diritto ad una rendita intera con effetto dal 1. settembre 2015 (cfr. consid. 1.3). L’insorgente postula l’annullamento della decisione impugnata del 10 maggio 2017 e quindi il mantenimento del diritto alla rendita intera. 2.3. Secondo l’art. 4 cpv. 1 LAI in relazione con gli art. 7 e 8 della LPGA, con invalidità s’intende l'incapacità al guadagno presunta permanente o di rilevante durata, cagionata da un danno alla salute fisica o psichica, conseguente ad infermità congenita, malattia o infortunio. Gli elementi fondamentali dell'invalidità, secondo la surriferita definizione, sono quindi un danno alla salute fisica o psichica conseguente a infermità congenita, malattia o infortunio, e la conseguente incapacità di guadagno. Occorre quindi che il danno alla salute abbia cagionato una diminuzione della capacità di guadagno, perché il caso possa essere sottoposto all'assicurazione per l'invalidità ( Duc, L’assurance invalidité, in: Meyer (Hrsg.), Schweizerisches Bundesverwaltungsrecht, Band XIV, Soziale Sicherheit, 2a ed., Basilea/Ginevra/Monaco di Baviera 2007, pag. 1411, n. 46 ). Giusta l'art. 28 cpv.\\ 
\hline
Predicted & considerations: in diritto In ordine 2.1. La presente vertenza non pone questioni giuridiche di principio e non è di rilevante importanza (ad esempio per la difficoltà dell’istruttoria o della valutazione delle prove). Il TCA può dunque decidere nella composizione di un Giudice unico ai sensi degli articoli 26 c cpv. 2 della Legge organica giudiziaria civile e penale e 2 cpv. 1 della Legge di procedura per le cause davanti al Tribunale delle assicurazioni (cfr. STF 9C\_211/2010 del 18 febbraio 2011; STF 9C\_792/2007 del 7 novembre 2008; STF H 180/06 e H 183/06 del 21 dicembre 2007; STFA I 707/00 del 21 luglio 2003; STFA H 335/00 del 18 febbraio 2002; STFA U 347/98 del 10 ottobre 2001, pubblicata in RDAT I-2002 pag. 190 seg.; STFA H 304/99 del 22 dicembre 2000; STFA I 623/98 del 26 ottobre 1999). Nel merito 2.2. Oggetto della lite è sapere se l’Ufficio AI ha diritto alla rendita intera intera dal 1. settembre 2015 (doc. AI 122/445-450), a causa dell’invalidità della persona assicurata. Secondo l'art. 4 cpv. 1 LAI in relazione con gli art. 7 e 8 della LPGA, con invalidità s’intende l'incapacità al guadagno presunta permanente o di rilevante durata, cagionata da un danno alla salute fisica o psichica, conseguente a infermità congenita, malattia o infortunio. Gli elementi fondamentali dell'invalidità, secondo la surriferita definizione, sono quindi un danno alla salute fisica o psichica conseguente a infermità congenita, malattia o infortunio, e la conseguente incapacità di guadagno. Occorre quindi che il danno alla salute abbia cagionato una diminuzione della capacità di guadagno, perché il caso possa \\ 
\hline
Scores & BERT: 86.57, BLEU: 81.39, METEOR: 65.76, ROUGE-1/2/L: 68.54 / 58.87 / 66.85 \\
\hline
\end{longtable}
}
\onecolumn
{\tiny
\begin{longtable}{l|p{13cm}}
\hline
Target & considerations: Das Bundesverwaltungsgericht zieht in Erwägung: 1. Nach Art. 53 Abs. 1 KVG kann gegen Beschlüsse der Kantonsregierungen nach Art. 39 KVG beim Bundesverwaltungsgericht Beschwerde geführt werden. Der angefochtene Beschluss Nr. 617/2019 des Regierungsrats des Kantons Zürich vom 26. Juni 2019 wurde gestützt auf Art. 39 KVG er- lassen. Das Bundesverwaltungsgericht ist deshalb zur Beurteilung der Be- schwerde zuständig (vgl. auch Art. 90a Abs. 2 KVG). Die Beschwerdefüh- rerin hat am vorinstanzlichen Verfahren teilgenommen, ist als Adressatin durch den angefochtenen Regierungsratsbeschluss besonders berührt und hat insoweit an dessen Aufhebung beziehungsweise Abänderung ein schutzwürdiges Interesse (Art. 48 Abs. 1 VwVG). Sie ist daher zur Be- schwerde legitimiert. Auf die frist- und formgerecht erhobene Beschwerde ist, nachdem auch der Kostenvorschuss rechtzeitig geleistet wurde, einzu- treten (Art. 50 Abs. 1, Art. 52 Abs. 1 und Art. 63 Abs. 4 VwVG). Das Haupt- rechtsbegehren der Beschwerdeführerin richtet sich auf Aufhebung der Zif- fer I des Dispositivs des angefochtenen Beschlusses. Soweit sie dabei Be- zug auf «C. Neue Regelung der Verlegungen ab Geburtshaus» nimmt, ist das im Lichte der Begründung der Beschwerde nicht so zu verstehen, dass sie auch die Aufhebung eines Teils der Begründung des angefochtenen Beschlusses beantragt. Der Antrag der Vorinstanz, auf die Beschwerde sei teilweise nicht einzutreten, weil sie sich auch gegen die Begründung des angefochtenen RRB richte, ist daher unbegründet. 2. 2.1 Mit Beschwerde gegen einen Spitallistenentscheid im Sinne von Art. 39 Abs. 1 KVG kann die Verletzung von Bundesrecht unter Einschluss des Missbrauchs oder der Überschreitung des Ermessens und die unrichtige oder unvollständige Feststellung des rechtserheblichen Sachverhalts ge- rügt werden. Nicht zulässig ist hingegen die Rüge der\\ 
\hline
Predicted & considerations: Das Bundesverwaltungsgericht zieht in Erwägung: 1. 1.1 Gemäss Art. 31 des Verwaltungsgerichtsgesetzes vom 17. Juni 2005 (VGG, SR 173.32) beurteilt das Bundesverwaltungsgericht Beschwerden gegen Verfügungen nach Art. 5 des Verwaltungsverfahrensgesetzes vom 20. Dezember 1968 (VwVG, SR 172.021), sofern keine Ausnahme nach Art. 32 VGG vorliegt. Als Vorinstanzen gelten die in Art. 33 und 34 VGG genannten Behörden. Dazu gehören u.a. Verfügungen im Sinne von Art. 5 VwVG. Das Bundesverwaltungsgericht ist somit zuständig für die Beurteilung der vorliegenden Beschwerde und entscheidet auf dem Gebiet des Gesundheits- schutzes endgültig, ausser bei Eintreten auf einen Entscheid, der eine Verfügung im Sinne von Art. 5 VwVG betrifft. Eine Ausnahme nach Art. 32 VGG liegt nicht vor. 1.2 Das Bundesverwaltungsgericht ist zur Beurteilung der vorliegenden Beschwerde zuständig. 1.3. Gegen Entscheide des Regierungsrats ist ein Beschwerdeverfahren zulässig, wenn dieses eine Rechtsmittelinstanz ist und grundsätzlich der Rechts- schutzinteresse der Beschwerde- berechtigten gegeben ist (vgl. art. 32 VGG; BGE 138 II 274 E. 5.1 S. 280). 1.4 Das Bundesverwaltungsgericht wendet im Beschwerde- verfahren das Bundesrecht von Amtes wegen an (Art. 106 Abs. 1 des Bundesgesetzes vom 20. Dezember 1968 über das Verwaltungsverfahren [VwVG, SR 172.021]). Es wird gemäss Art. 62 Abs. 4 VwVG zur Beurteilung von Beschwerden in Zivilsachen gemäss Art. 82 Abs. 1 und Art. 86 Abs. 1 Bst. d bis Bst. e VwVG grundsätzlich nur geprüft, wenn diese vor Bundesgericht nicht mehr vorgetragen wurden und keine Verletzung des Op- fers gegeben wurde. Das Bundesverwal- tungsgericht wendet im Beschwerdeverfahren das Bundesrecht von Amtes wegen an. Es ist gemäss Art. 62 Abs. 4 VwVG an die Begründung der Begehren nicht gebunden und kann die Beschwerde auch aus anderen als den geltend gemachten Gründen gutheissen \\ 
\hline
Scores & BERT: 75.72, BLEU: 67.17, METEOR: 31.26, ROUGE-1/2/L: 26.43 / 8.84 / 25.23 \\
\hline
\end{longtable}
}
\onecolumn
{\tiny
\begin{longtable}{l|p{13cm}}
\hline
Target & considerations: Considérant en droit: 1. Selon l'art. 5 al. 1 des dispositions du RPAC relatives à la Commission de recours individuel, la décision rendue par cette autorité peut faire l'objet d'un recours de droit administratif auprès du Tribunal cantonal dans les trente jours suivant la communication de la décision motivée, conformément à la loi cantonale du 28 octobre 2008 sur la procédure administrative (LPA-VD; BLV 173.36). D’après l’art. 92 al. 1 LPA-VD, le Tribunal cantonal connaît des recours contre les décisions et décisions sur recours rendues par les autorités administratives, lorsque la loi ne prévoit aucune autre autorité pour en connaître. En l'espèce, le recours a été déposé en temps utile (art. 5 al. 1 RPAC) et il satisfait aux autres conditions formelles de recevabilité (art. 75 al. 1 let. a, 79 al. 1 et 99 LPA-VD, applicables par renvoi de l'art. 5 al. 1 RPAC), si bien qu’il convient d’entrer en matière. 2. a) La classification salariale du poste qu’occupe le recourant est litigieuse en l’occurrence. b) L’organisation de l’administration fait partie des tâches propres des autorités communales (art. 2 de la loi vaudoise du 28 février 1956 sur les communes [LC; BLV 175.11]). Selon cette loi, il incombe au Conseil général ou communal de définir le statut des collaborateurs communaux et la base de leur rémunération (art. 4 al. 1 ch. 9 LC), la municipalité ayant la compétence de nommer les collaborateurs et employés de la commune, de fixer leur traitement et d'exercer le pouvoir disciplinaire (art. 42 al. 1 ch. 3 LC). Conformément à l'art. 33 al. 1 RPAC, le traitement du fonctionnaire comprend le traitement de base (let. a), les allocations\\ 
\hline
Predicted & considerations: Considérant en droit: 1. 1.1 Le personnel doit maintenir, par contrat de droit privé, une commune, un partenaire public, pour assurer la bonne exécution de leurs missions spéciales, ainsi qu’un minimum d’administration (art. 10 al. 1 du RPAC). L’art. 10 al. 2 RPAC prévoit que les fonctionnaires dans toute la ville de Lausanne sont soumis à la qualification communale et leur remplacement a lieu par des fonctions communales cantonales, comme applicable en la matière. Elle constitue en revanche une décision sujette à recours auprès de la juridiction compétente (art. 20 al. 2 et art. 30 al. 1 de la loi sur le personnel des communes du 15 novembre 1976 (LPAC). 1.2 Selon la jurisprudence relative aux dispositions transitoires applicables au moment de la notification de la décision pour connaître du jugement du service, la décision de rémunération fait apparaître un changement de système de rémunération selon le RPAC. Cette disposition, qui comprend ainsi des critères objectifs et suffisants, permet au fonctionnaire de connaître de tous les aspects de la classification (Tappy, CPC, n. 1 ad art. 10 LPAC), ainsi que par le service de réduire son salaire. En l'espèce, l'autorité intimée a constaté que l'établissement des faits importants du dossier et de la satisfaction exigée sur le plan d'administration devait être complété dans la mesure où cela n’était pas possible, ce qui constituait effectivement un changement de système (Tappy, op. cit., n. 13 ad art. 10 LPAC). Il a constaté que le Service du personnel avait fixé le niveau de fonction attribué au fonctionnaire pour son titre de travail. Il a en outre estimé que cette classification était une décision de mise en œuvre déléguée à celui qui a entrepris une modification légale ou réglementaire du service. Parcon\\ 
\hline
Scores & BERT: 75.74, BLEU: 66.19, METEOR: 31.69, ROUGE-1/2/L: 28.32 / 4.21 / 26.01 \\
\hline
\end{longtable}
}
\onecolumn
{\tiny
\begin{longtable}{l|p{13cm}}
\hline
Target & considerations: Diritto: 1. ll ricorso in materia di diritto pubblico può essere presentato per violazione del diritto, così come stabilito dagli art. 95 e 96 LTF. Il Tribunale federale fonda la sua sentenza sui fatti accertati dall'autorità inferiore (art. 105 cpv. 1 LTF). Può scostarsi da questo accertamento solo qualora esso sia avvenuto in modo manifestamente inesatto, ovvero arbitrario (DTF 134 V 53 consid. 4.3 pag. 62), oppure in violazione del diritto ai sensi dell'art. 95 LTF (art. 105 cpv. 2 LTF). Salvo i casi in cui tale inesattezza sia lampante, la parte ricorrente che intende contestare i fatti accertati dall'autorità inferiore deve spiegare, in maniera circostanziata, per quale motivo ritiene che le condizioni di una delle eccezioni previste dall'art. 105 cpv. 2 LTF sarebbero realizzate (cfr. DTF 133 II 249 consid. 1.4.3 pag. 254 con riferimento). 2. Oggetto del contendere è sapere se sia lesivo del diritto federale il giudizio cantonale, che conferma la decisione su opposizione, la quale a sua volta nega all'assicurato le indennità per insolvenza per non avere rivendicato tempestivamente le proprie pretese salariali. 3. 3.1. Il Tribunale cantonale delle assicurazioni, esposto lo svolgimento del processo, le norme e la prassi ritenute applicabili, ha accertato che il ricorrente è stato legato da un contratto di lavoro con la ditta B.\_ SA, dichiarata fallita il 15 maggio 2018, dal 15 aprile 2017 al 18 settembre 2017 come informatico e con un salario di fr. 3'360.- mensili. L'ultimo giorno di lavoro effettuato è stato il 1° settembre 2017. Il salario è stato saldato dal 15 al 30 aprile 2017. Per contro, non è stato pagato per i mesi di maggio, giugno, luglio e agosto, nonché per il 1° settembre 2017. Il\\ 
\hline
Predicted & considerations: Diritto: 1. 1.1. La procedura ricorsuale in materia di diritto pubblico è disciplinata dal diritto federale (art. 82 lett. a LTF) e dalla legge federale, in vigore dal 1° gennaio 2003, sull'assicurazione contro la disoccupazione: si tratta infatti di una causa di diritto pubblico contro una decisione finale adottata dalla Corte cantonale delle assicurazioni, entrata in vigore il 1° gennaio 2003. Il ricorso presentato dall'assoggettato al diritto di ricorrere (art. 89 cpv. 1 LTF) è di massima ammissibile dinanzi al Tribunale federale (art. 89 cpv. 1 LTF). 1.2. Il ricorso in materia di diritto pubblico può essere presentato per violazione del diritto, così come stabilito dagli art. 95 e 96 LTF, oltre che per violazione del diritto ai sensi dell'art. 95 LTF. Il Tribunale federale applica il diritto federale (art. 106 cpv. 1 LTF), senza essere vincolato né dai motivi addotti nel ricorso (art. 106 cpv. 2 LTF). Per contro, nel ricorso in materia di diritto pubblico il Tribunale federale esamina d'ufficio e con piena cognizione l'ammissibilità dei gravami che gli vengono sottoposti (DTF 133 III 439 consid. 1.3). 1.3. La critica del giudizio impugnato esplica degli effetti (art. 105 cpv. 1 LTF). Non è ammissibile che i ricorsi in materia di diritto pubblico possano essere decisi in base al diritto federale, ai sensi dell'art. 95 LTF, senza istruttoria (art. 97 cpv. 1 LTF, Art. 105 cpv. 2 LTF). 1.3. Il Tribunale federale esamina d'ufficio e con piena cognizione l'ammissibilità dei ricorsi che gli vengono sottoposti (DTF 133 III 439 consid. 1.3). 1.4. Con il ricorso in materia di pubblico contro la decisione di primo grado, il Tribunale cantonale ha emesso\\ 
\hline
Scores & BERT: 75.79, BLEU: 66.28, METEOR: 30.29, ROUGE-1/2/L: 37.74 / 20.43 / 36.48 \\
\hline
\end{longtable}
}
\onecolumn
{\tiny
\begin{longtable}{l|p{13cm}}
\hline
Target & considerations: Das Versicherungsgericht zieht in Erwägung: 1. Streitig und zu prüfen ist der Rentenanspruch der Beschwerdeführerin. - 3 - 2. Am 1. Januar 2022 sind die Änderungen betreffend Weiterentwicklung der IV (WEIV) in Kraft getreten. Weder dem IVG noch der IVV sind besondere Übergangsbestimmungen betreffend die Anwendbarkeit dieser Änderun- gen im Hinblick auf nach dem 1. Januar 2022 beurteilte mögliche Ansprü- che des Zeitraums bis zum 31. Dezember 2021 zu entnehmen. Es sind daher nach den allgemeinen übergangsrechtlichen Grundsätzen jene Be- stimmungen anzuwenden, die bei der Erfüllung des rechtlich zu ordnenden oder zu Rechtsfolgen führenden Tatbestands Geltung haben beziehungs- weise hatten (vgl. Urteil des Bundesgerichts 8C\_136/2021 vom 7. April 2022 E. 3.2.1 mit Hinweis unter anderem auf BGE 144 V 210 E. 4.3.1 S. 213). Da vorliegend Leistungen mit Anspruchsbeginn vor dem 1. Januar 2022 streitig sind, ist für deren Beurteilung die bis zum 31. Dezember 2021 geltende Rechtslage massgebend. 3. 3.1. Die Beschwerdegegnerin stützte sich in der Verfügung vom 14. Juni 2022 (Vernehmlassungsbeilage [VB] 162) in medizinischer Hinsicht im Wesent- lichen auf das bidisziplinäre Gutachten der SMAB vom 3. August 2021 (VB 139; Fachdisziplinen: Neurologie und Psychiatrie). 3.2. Die SMAB-Gutachterinnen stellten in der interdisziplinären Gesamtbeurtei- lung die folgenden Diagnosen mit Auswirkungen auf die Arbeitsfähigkeit (VB 139.1 S. 6): "1. Neuropathische Schmerzen an der Innenseite des Mittelfingers links und der Handinnenfläche links (Schnittverletzung am Mittelfinger links mit initialer Nervenverletzung 19.11.2014, epineurale und weitere mehrfache operative Eingriffe 2014-2018), 2. Chronische Schmerzstörung mit somatischen und psychischen (ICD-10: F45.41)". Ohne Auswirkung auf die Arbeitsfähigkeit sei eine rezidivierende depres- sive Störung, gegenwärtig remittiert. Die Gutachterinnen hielten fest, ins- gesamt bestehe aufgrund der psychiatrischen und neurologi\\ 
\hline
Predicted & considerations: Das Gericht zieht in Erwägung: 1. Gemäss Art. 87 Abs. 2 des Bundesgesetzes über die Unfallversicherung (UVG) besteht Anspruch auf eine ganze Rente, wenn die versicherte Person mindestens zu 66 2/3 Prozent, auf eine halbe Rente, wenn sie mindestens zu 50 Prozent oder auf eine Viertelsrente, wenn sie mindestens zu 40 Prozent invalid ist. In Härtefällen besteht gemäss Art. 87 Abs. 2 UVG auch ein Anspruch auf eine halbe Rente, wenn sie mindestens zu 50 Prozent, wenn sie mindestens zu 40 Prozent invalid ist (Art. 8 Abs. 1 UVG). Anspruch auf eine Rente haben gemäss Art. 28 Abs. 2 des Bundes- gesetzes über die Invalidenversicherung (IVG) Versicherte, die im Sinne von Art. 16 Abs. 1 IVG in Verbindung mit Art. 28 bis Abs. 1 IVG in Verbindung mit Abs. 1 und 2 IVG auch dann Anspruch auf eine Viertelsrente, wenn sie mindestens zu 40 Prozent invalid sind (Art. 28 Abs. 2 IVG i.V.m. Art. 16 ATSG). 2. 2.1 Anspruch auf eine Rente haben gemäss Art. 28 Abs. 1 IVG Versicherte, die: a. ihre Erwerbsfähigkeit oder die Fähigkeit, sich im Aufgabenbereich zu betä- tigen, nicht durch zumutbare Eingliederungsmassnahmen wieder herstellen, erhalten oder verbessern können; b. während eines Jahres ohne wesentlichen Unterbruch durchschnittlich min- destens 40 Prozent arbeitsunfähig (Art. 6 ATSG) gewesen sind; und c. nach Ablauf dieses Jahres zu mindestens 40 Prozent invalid (Art. 8 ATSG) sind. 2.2 Die Beschwerdeführerin bringt vor, die ärztlich eingeholten ärztlichen Berichte seien als diagnostisch zu qualifizieren. Das trifft vorliegend nicht zu. Ihr Gesundheitszustand sei gemäss Abklärungen vom RAD mit einer Invalidität von mindestens 40 Prozent zu vereinbaren. Die Leistungsfähigkeit sei in Art. 16 ATSG eingetreten. 2.2 Die Leistungsfähigkeit sei in Art. 16 ATSG i.V.m. Art. 28 Abs. 1 IVG \\ 
\hline
Scores & BERT: 62.42, BLEU: 51.67, METEOR: 19.69, ROUGE-1/2/L: 17.58 / 2.56 / 17.58 \\
\hline
\end{longtable}
}
\onecolumn
{\tiny
\begin{longtable}{l|p{13cm}}
\hline
Target & considerations: le conseil de B\_ a rappelé que celle-ci contestait que son frère aurait été en droit de loger gratuitement au 1\_ dès le mois d'avril 2015. Afin d'éviter la notification d'une poursuite interruptive de la prescription, A\_ était invité à signer une déclaration de renonciation à la prescription concernant les loyers qui, selon B\_, étaient dus à la succession de leurs défunts parents. Etait annexée à ce courrier une déclaration de renonciation à invoquer la prescription libellée en ces termes : " Par la présente, A\_ déclare renoncer à invoquer toute exception de prescription vis-à-vis des successions de [feu D\_ et feu C\_], eu égard à toutes prétentions éventuelles de loyers et charges alors qu'il logeait chemin 1\_ à \_ Fribourg. Cette déclaration de renonciation est valable jusqu'au 31 décembre 2020. La présente déclaration ne vaut en aucun cas reconnaissance de dette ou de responsabilité quelconque. La présente déclaration n'a d'effet que si la prescription n'est pas déjà acquise à ce jour ". A\_ n'a pas donné suite à ce courrier. f. Par réquisition de poursuite du 7 avril 2020, reçue le lendemain par l'Office cantonal des poursuites de Genève (ci-après : l'Office), Me H\_, agissant en qualité de représentant du créancier, à savoir " Pour l'hoirie de feu C\_, décédée le \_ 2019, Madame B\_, Rte 2\_, \_ Fribourg ", a requis la poursuite de A\_, domicilié avenue 3\_, \_ Genève, pour la somme de 112'500 fr. avec intérêts à 5\% dès le 1 er mai 2017, réclamée à titre de " Loyers et charges dus pour le logement du 1\_ à \_ Fribourg d'avril 2015 à mai 2019 (50 x\\ 
\hline
Predicted & considerations: EN DROIT 1. 1.1 Le litige porte sur la désignation d'un représentant de la communauté héréditaire et la désignation d'un administrateur de la régie G\_ SA (art. 602 al. 3 CC). Le motif de la désignation du représentant est incontestable. Selon l'art. 602 al. 5 CC, la désignation d'un représentant de la communauté héréditaire est donc une décision de dernière instance cantonale au sens des art. 80 al. 1 et 82 al. 1 CPC, qui doit être notifiée à l'unique partie plaignante qui a pris part à la procédure (ATF 110 IV 92 consid. 1a; arrêt du Tribunal fédéral X. du 27 juillet 2004 ; ATF 99 III 58 consid. 1; ATF 118 Ia 397 consid. 1b; Tappy, CPC-VD, n. 3 ad art. 602 CC ; Replin, Le représentant de la communauté héréditaire, 5 ème éd., Lausanne 2013, p. 569; ATF 118 IV 286 consid. 2a; TF 6B\_211/2007 du 29 janvier 2008, consid. 5.3; ATF 117 IV 29 consid. 3b; TF 8B\_44/2007 du 15 août 2007, consid. 3.2; Tappy, Procédure civile, tome II, ad art. 602; Piquerez, in : Kuhn/Jeanneret [éd.], Basler Kommentar, n. 6 ad art. 602; TF 7B\_51/2007 du 1 er janvier 2008, consid. 3.2; ATF 130 III 136 consid. 1.2.1; TF 9C\_438/2007 du 30 septembre 2007, consid. 3.1; ATF 134 III 102 consid. 3.1; TF 6B\_71/2007 du 24 août 2007, consid. 5b; TF 9C\_792/2007 du 11 août 2007, consid. 4.2). 1.2 La désignation d'un représentant et de l'administration régulière de l'ensemble de la succession ont été rejetées. 2. 2.1. L'art. 602 al. 3 CC ouvre un recours au Tribunal cantonal, sans être lié par l\\ 
\hline
Scores & BERT: 63.90, BLEU: 46.65, METEOR: 19.12, ROUGE-1/2/L: 11.52 / 1.64 / 10.91 \\
\hline
\end{longtable}
}
\onecolumn
{\tiny
\begin{longtable}{l|p{13cm}}
\hline
Target & considerations: A.c. Con progetto di decisione dell'8 gennaio 2014 l'UAI ha concluso, in via di riconsiderazione, per la soppressione della mezza rendita d'invalidità con effetto retroattivo al 30 aprile 2011. In sede di audizione, l'UAI ha esperito nuovi accertamenti, in particolare la perizia pluridisciplinare dell'8 aprile 2016. Con decisione del 14 luglio 2016 l'UAI ha riconosciuto che al momento dell'assegnazione della mezza rendita d'invalidità è incorso in un errore manifesto avendo applicato il "Prozentvergleich" in luogo del metodo ordinario di confronto dei redditi, da cui risulta un grado d'invalidità del 26 \%. Ritenuta l'importanza della sua correzione, l'UAI, riconsiderando le decisioni 14 dicembre 2006 e 14 maggio 2007 nonché la comunicazione del 20 marzo 2009, ha predisposto la soppressione del diritto alla mezza rendita invalidità con effetto retroattivo al 30 aprile 2011. B. A.\_ si è aggravato il 13 settembre 2016 al Tribunale delle assicurazioni del Cantone Ticino, che con giudizio del 17 maggio 2017 ha respinto il gravame. C. Il 19 giugno 2017 (timbro postale) A.\_ inoltra un ricorso in materia di diritto pubblico al Tribunale federale, cui chiede di ammettere il ricorso e riformare il giudizio cantonale nel senso di confermare il diritto ad almeno una mezza rendita d'invalidità e ordinare all'UAI di versare tutte le prestazioni trattenute dal momento della revisione ad oggi. L'UAI e l'Ufficio federale delle assicurazioni sociali rinunciano a determinarsi. Diritto: 1. Il ricorso in materia di diritto pubblico può essere presentato per violazione del diritto così come stabilito dagli art. 95 e 96 LTF. Il Tribunale federale esamina d'ufficio l'applicazione de l diritto federale (art. 106 cpv. 1 L\\ 
\hline
Predicted & considerations: Diritto: 1. Il Tribunale delle assicurazioni esamina d'ufficio e con piena cognizione l'ammissibilità dei gravami che gli vengono sottoposti (DTF 134 V 131 consid. 1.1 pag. 131; 133 V 477 consid. 1.1 pag. 481). In particolare, i gravami inoltrati già in tempo utile contro le decisioni di prima istanza rese dall'UAI in materia di assicurazioni sociali possono essere decisi sulla scorta degli atti, senza istruttoria (DTF 133 V 477 consid. 1.2 pag. 481). Il giudizio può essere reso sulla base degli atti, senza istruttoria (DTF 133 V 477 consid. 1.2 pag. 481; 133 V 477 consid. 1.2 pag. 481). 2. Il Tribunale delle assicurazioni esamina d'ufficio e con piena cognizione l'ammissibilità dei gravami che gli vengono sottoposti (DTF 133 V 477 consid. 1.2 pag. 481). 3. 3.1. Giusta l'art. 4 cpv. 1 LAI in relazione con gli art. 7 e 8 della LPGA, con invalidità s’intende l'incapacità al guadagno presunta permanente o di rilevante durata, cagionata da un danno alla salute fisica o psichica, conseguente a infermità congenita, malattia o infortunio. Gli elementi fondamentali dell'invalidità, secondo la surriferita definizione, sono quindi un danno alla salute fisica o psichica conseguente a infermità congenita, malattia o infortunio, e la conseguente incapacità di guadagno. Occorre quindi che il danno alla salute abbia cagionato una diminuzione della capacità di guadagno, perché il caso possa essere sottoposto all'assicurazione per l'invalidità (DTF 134 V 329 consid. 3.2.3 pag. 331; 133 V 477 consid. 3.3.1 pag. 481; 129 V 4 consid. 1.2 pag. 5 con rinvii). 3.2. Giusta\\ 
\hline
Scores & BERT: 67.86, BLEU: 55.76, METEOR: 25.41, ROUGE-1/2/L: 24.32 / 5.59 / 23.65 \\

\label{tab:examples_court_view}
\end{longtable}
}

\clearpage


\begin{longtable}{l|p{13cm}}
\caption{Three examples of generated summaries by mT5\textsubscript{Base} with input length 4096, showcasing high, average, and low scored outputs in \ac{LDS} Task}\\
\hline
Target & regeste: Regeste Teilklage und negative Feststellungswiderklage ( Art. 86 und 224 ZPO ). Die Ausnahme vom Erfordernis der gleichen Verfahrensart gemäss Art. 224 Abs. 1 ZPO für negative Feststellungswiderklagen ist nicht auf den Fall beschränkt, dass es sich bei der Hauptklage um eine sogenannte echte Teilklage handelt, sondern gilt allgemein dann, wenn die Teilklage eine Ungewissheit zur Folge hat, die es rechtfertigt, die Feststellung des Nichtbestands einer Forderung oder eines Rechtsverhältnisses zu verlangen (E. 2).\\ 
\hline
Predicted & regeste: Regeste Voraussetzungen, unter denen eine Widerklage erhoben werden darf, wenn eine echte Teilklage beurteilt wird (E. 2.1). Zur Anwendung von Art. 224 Abs. 1 ZPO ist die Ausnahme vom Erfordernis der gleichen Verfahrensart gemäss Art. 224 Abs. 1 ZPO nicht auf diesen Fall beschränkt, sondern gilt allgemein dann, wenn die Teilklage eine Ungewissheit zur Folge hat, die es rechtfertigt, im Sinne von Art. 88 ZPO die Feststellung des Nichtbestands einer Forderung oder eines Rechtsverhältnisses zu verlangen (E. 2.2).\\ 
\hline
Scores & BERT: 88.03, BLEU: 30.45, METEOR: 43.78, ROUGE-1/2/L: 72.44 / 54.05 / 70.87 \\

\hline \hline
Target & regeste: Regeste Uferschutzplanung Wohlensee; Gesamtinteressenabw\"{a}gung. Die Beh\"{o}rden nehmen bei der Genehmigung einer Sondernutzungsplanung eine umfassende Interessenabw\"{a}gung vor (vgl. Art. 3 RPV ; E. 3.2). W\"{u}rdigung des Gutachtens der Vogelwarte Sempach und der Stellungnahme des BAFU als Fachbeh\"{o}rde (E. 5.5, 6.3 und 6.4). Im zu beurteilenden Fall wird das \"{o}ffentliche Interesse an einer ufernahen Wegf\"{u}hrung (vgl. Art. 3 Abs. 2 lit. c RPG ) relativiert, weil ein durchgehender Sichtschutz zum Schutz der im Gebiet vorkommenden Vogelarten erforderlich w\"{a}re, was Spazierg\"{a}ngern das Erleben der Uferlandschaft nur sehr beschr\"{a}nkt erm\"{o}glichen w\"{u}rde (E. 6.5 sowie 3.3). Das Gebiet des Wohlensees stellt eines von insgesamt 25 Wasser- und Zugvogelreservaten von nationaler Bedeutung dar (vgl. Art. 11 Abs. 2 JSG, Art. 2 Abs. 1 i.V.m. Anhang 1 WZVV). \\

\hline
Predicted & regeste: Regeste Art. 3 Abs. 2 lit. c RPG ; schweizerische Vogelwarte Sempach; Umweltschutz. Gutachten im Bereich des Umweltrechts, insbesondere von dem UVG/USG ergangenen Streitwert (E. 3.1). Ein öffentlicher Weg, der über Privatgrundstücke führt und mit dem Enteignungsrecht zugunsten des Gemeinwesens verbunden ist, stellt eine öffentlich-rechtliche Eigentumsbeschränkung dar (E. 3.3). Interessenabwägungen sind grundsätzlich frei (E. 3.4 und 3.5). Umfang des geplanten Uferweges; ein ausweisenes öffentliches Interesse an einem direkt dem Ufer entlangführenden Weg sowie an einer zumindest ufernahen Wegführung (E. 3.5).\\
\hline
Scores & BERT: 73.91, BLEU: 47.21, METEOR: 30.91, ROUGE-1/2/L: 32.75 / 10.63 / 32.75 \\

\hline \hline
Target & regeste: Regeste a Schwere Bedrängnis, grosse seelische Belastung und heftige Gemütsbewegung; Art. 48 lit. a Ziff. 2 und Art. 48 lit. c StGB. Abgrenzung dieser Strafmilderungsgründe untereinander (E. 2.1-2.3). Regeste b Mord, heftige Gemütsbewegung und grosse seelische Belastung; Art. 48 lit. c und Art. 112 StGB. Die für den Mord typische Skrupellosigkeit schliesst die Bejahung der Strafmilderungsgründe der heftigen Gemütsbewegung und der grossen seelischen Belastung aus (E. 2.5).\\
\hline
Predicted & regeste: Regeste Art. 112 und 49 Abs. 2 StGB ; tiefe Gewalt, Schwere Gewalt, Mord. Die Beeinträchtigung oder der schwere Gewalt ist einen psychischen Klumpen im Sinne des Art. 48 StGB, wenn beide Personen einen Beziehungskonflikt bilden und sich ein bestimmtes Verhältnis zwischen einseitigen Umständen des Tatvergehens und der Begründung der Tatbegehren haben (E. 2 und 3).\\
\hline
Scores & BERT: 69.3, BLEU: 12.14, METEOR: 14.72, ROUGE-1/2/L: 30.43 / 10.26 / 30.43 \\

\label{tab:examples_lds}
\end{longtable}


\clearpage

\bibliography{custom,references}


\begin{thebibliography}{138}
\ifx \bisbn   \undefined \def \bisbn  #1{ISBN #1}\fi
\ifx \binits  \undefined \def \binits#1{#1}\fi
\ifx \bauthor  \undefined \def \bauthor#1{#1}\fi
\ifx \batitle  \undefined \def \batitle#1{#1}\fi
\ifx \bjtitle  \undefined \def \bjtitle#1{#1}\fi
\ifx \bvolume  \undefined \def \bvolume#1{\textbf{#1}}\fi
\ifx \byear  \undefined \def \byear#1{#1}\fi
\ifx \bissue  \undefined \def \bissue#1{#1}\fi
\ifx \bfpage  \undefined \def \bfpage#1{#1}\fi
\ifx \blpage  \undefined \def \blpage #1{#1}\fi
\ifx \burl  \undefined \def \burl#1{\textsf{#1}}\fi
\ifx \doiurl  \undefined \def \doiurl#1{\url{https://doi.org/#1}}\fi
\ifx \betal  \undefined \def \betal{\textit{et al.}}\fi
\ifx \binstitute  \undefined \def \binstitute#1{#1}\fi
\ifx \binstitutionaled  \undefined \def \binstitutionaled#1{#1}\fi
\ifx \bctitle  \undefined \def \bctitle#1{#1}\fi
\ifx \beditor  \undefined \def \beditor#1{#1}\fi
\ifx \bpublisher  \undefined \def \bpublisher#1{#1}\fi
\ifx \bbtitle  \undefined \def \bbtitle#1{#1}\fi
\ifx \bedition  \undefined \def \bedition#1{#1}\fi
\ifx \bseriesno  \undefined \def \bseriesno#1{#1}\fi
\ifx \blocation  \undefined \def \blocation#1{#1}\fi
\ifx \bsertitle  \undefined \def \bsertitle#1{#1}\fi
\ifx \bsnm \undefined \def \bsnm#1{#1}\fi
\ifx \bsuffix \undefined \def \bsuffix#1{#1}\fi
\ifx \bparticle \undefined \def \bparticle#1{#1}\fi
\ifx \barticle \undefined \def \barticle#1{#1}\fi
\bibcommenthead
\ifx \bconfdate \undefined \def \bconfdate #1{#1}\fi
\ifx \botherref \undefined \def \botherref #1{#1}\fi
\ifx \url \undefined \def \url#1{\textsf{#1}}\fi
\ifx \bchapter \undefined \def \bchapter#1{#1}\fi
\ifx \bbook \undefined \def \bbook#1{#1}\fi
\ifx \bcomment \undefined \def \bcomment#1{#1}\fi
\ifx \oauthor \undefined \def \oauthor#1{#1}\fi
\ifx \citeauthoryear \undefined \def \citeauthoryear#1{#1}\fi
\ifx \endbibitem  \undefined \def \endbibitem {}\fi
\ifx \bconflocation  \undefined \def \bconflocation#1{#1}\fi
\ifx \arxivurl  \undefined \def \arxivurl#1{\textsf{#1}}\fi
\csname PreBibitemsHook\endcsname

\bibitem[\protect\citeauthoryear{Ashley}{2017}]{ashley_2017}
\begin{bbook}
\bauthor{\bsnm{Ashley}, \binits{K.D.}}:
\bbtitle{Artificial Intelligence and Legal Analytics: New Tools for Law Practice in the Digital Age}.
\bpublisher{Cambridge University Press}, \blocation{???}
(\byear{2017}).
\doiurl{10.1017/9781316761380}
\end{bbook}
\endbibitem

\bibitem[\protect\citeauthoryear{Katz et~al.}{2023}]{katz2023natural}
\begin{botherref}
\oauthor{\bsnm{Katz}, \binits{D.M.}},
\oauthor{\bsnm{Hartung}, \binits{D.}},
\oauthor{\bsnm{Gerlach}, \binits{L.}},
\oauthor{\bsnm{Jana}, \binits{A.}},
\oauthor{\bsnm{au2}, \binits{M.J.B.I.}}:
Natural Language Processing in the Legal Domain
(2023)
\end{botherref}
\endbibitem

\bibitem[\protect\citeauthoryear{Paul et~al.}{2021}]{paul2021lesicin}
\begin{botherref}
\oauthor{\bsnm{Paul}, \binits{S.}},
\oauthor{\bsnm{Goyal}, \binits{P.}},
\oauthor{\bsnm{Ghosh}, \binits{S.}}:
LeSICiN: A Heterogeneous Graph-based Approach for Automatic Legal Statute Identification from Indian Legal Documents
(2021)
\end{botherref}
\endbibitem

\bibitem[\protect\citeauthoryear{Chalkidis et~al.}{2019}]{chalkidis_extreme_2019}
\begin{bchapter}
\bauthor{\bsnm{Chalkidis}, \binits{I.}},
\bauthor{\bsnm{Fergadiotis}, \binits{E.}},
\bauthor{\bsnm{Malakasiotis}, \binits{P.}},
\bauthor{\bsnm{Aletras}, \binits{N.}},
\bauthor{\bsnm{Androutsopoulos}, \binits{I.}}:
\bctitle{Extreme {Multi}-{Label} {Legal} {Text} {Classification}: {A} {Case} {Study} in {EU} {Legislation}}.
In: \bbtitle{Proceedings of the {Natural} {Legal} {Language} {Processing} {Workshop} 2019},
pp. \bfpage{78}--\blpage{87}.
\bpublisher{Association for Computational Linguistics},
\blocation{Minneapolis, Minnesota}
(\byear{2019}).
\doiurl{10.18653/v1/W19-2209} .
\burl{https://www.aclweb.org/anthology/W19-2209}
Accessed 2021-03-02
\end{bchapter}
\endbibitem

\bibitem[\protect\citeauthoryear{Hendrycks et~al.}{2021}]{hendrycks2021cuad}
\begin{botherref}
\oauthor{\bsnm{Hendrycks}, \binits{D.}},
\oauthor{\bsnm{Burns}, \binits{C.}},
\oauthor{\bsnm{Chen}, \binits{A.}},
\oauthor{\bsnm{Ball}, \binits{S.}}:
CUAD: An Expert-Annotated NLP Dataset for Legal Contract Review
(2021)
\end{botherref}
\endbibitem

\bibitem[\protect\citeauthoryear{Li and Zhang}{2021}]{Li2021CourtOG}
\begin{bchapter}
\bauthor{\bsnm{Li}, \binits{Q.}},
\bauthor{\bsnm{Zhang}, \binits{Q.}}:
\bctitle{Court opinion generation from case fact description with legal basis}.
In: \bbtitle{AAAI Conference on Artificial Intelligence}
(\byear{2021})
\end{bchapter}
\endbibitem

\bibitem[\protect\citeauthoryear{Semo et~al.}{2022}]{semo_classactionprediction_2022}
\begin{bchapter}
\bauthor{\bsnm{Semo}, \binits{G.}},
\bauthor{\bsnm{Bernsohn}, \binits{D.}},
\bauthor{\bsnm{Hagag}, \binits{B.}},
\bauthor{\bsnm{Hayat}, \binits{G.}},
\bauthor{\bsnm{Niklaus}, \binits{J.}}:
\bctitle{{ClassActionPrediction}: {A} {Challenging} {Benchmark} for {Legal} {Judgment} {Prediction} of {Class} {Action} {Cases} in the {US}}.
In: \bbtitle{Proceedings of the {Natural} {Legal} {Language} {Processing} {Workshop} 2022},
pp. \bfpage{31}--\blpage{46}.
\bpublisher{Association for Computational Linguistics},
\blocation{Abu Dhabi, United Arab Emirates (Hybrid)}
(\byear{2022}).
\burl{https://aclanthology.org/2022.nllp-1.3}
Accessed 2023-04-17
\end{bchapter}
\endbibitem

\bibitem[\protect\citeauthoryear{Brugger et~al.}{2023}]{brugger_multilegalsbd_2023}
\begin{bchapter}
\bauthor{\bsnm{Brugger}, \binits{T.}},
\bauthor{\bsnm{Stürmer}, \binits{M.}},
\bauthor{\bsnm{Niklaus}, \binits{J.}}:
\bctitle{{MultiLegalSBD}: {A} {Multilingual} {Legal} {Sentence} {Boundary} {Detection} {Dataset}}.
In: \bbtitle{Proceedings of the {Nineteenth} {International} {Conference} on {Artificial} {Intelligence} and {Law}}.
\bsertitle{{ICAIL} '23},
pp. \bfpage{42}--\blpage{51}.
\bpublisher{Association for Computing Machinery},
\blocation{New York, NY, USA}
(\byear{2023}).
\doiurl{10.1145/3594536.3595132} .
\burl{https://dl.acm.org/doi/10.1145/3594536.3595132}
Accessed 2023-10-21
\end{bchapter}
\endbibitem

\bibitem[\protect\citeauthoryear{Hwang et~al.}{2022}]{hwang_multi-task_2022}
\begin{botherref}
\oauthor{\bsnm{Hwang}, \binits{W.}},
\oauthor{\bsnm{Lee}, \binits{D.}},
\oauthor{\bsnm{Cho}, \binits{K.}},
\oauthor{\bsnm{Lee}, \binits{H.}},
\oauthor{\bsnm{Seo}, \binits{M.}}:
A {Multi}-{Task} {Benchmark} for {Korean} {Legal} {Language} {Understanding} and {Judgement} {Prediction}.
arXiv.
arXiv:2206.05224 [cs]
(2022).
\url{http://arxiv.org/abs/2206.05224}
Accessed 2023-04-28
\end{botherref}
\endbibitem

\bibitem[\protect\citeauthoryear{Niklaus et~al.}{2023}]{niklaus2023lextreme}
\begin{botherref}
\oauthor{\bsnm{Niklaus}, \binits{J.}},
\oauthor{\bsnm{Matoshi}, \binits{V.}},
\oauthor{\bsnm{Rani}, \binits{P.}},
\oauthor{\bsnm{Galassi}, \binits{A.}},
\oauthor{\bsnm{Stürmer}, \binits{M.}},
\oauthor{\bsnm{Chalkidis}, \binits{I.}}:
LEXTREME: A Multi-Lingual and Multi-Task Benchmark for the Legal Domain
(2023)
\end{botherref}
\endbibitem

\bibitem[\protect\citeauthoryear{Thakur et~al.}{2021}]{thakur_beir_2021}
\begin{botherref}
\oauthor{\bsnm{Thakur}, \binits{N.}},
\oauthor{\bsnm{Reimers}, \binits{N.}},
\oauthor{\bsnm{Rücklé}, \binits{A.}},
\oauthor{\bsnm{Srivastava}, \binits{A.}},
\oauthor{\bsnm{Gurevych}, \binits{I.}}:
{BEIR}: {A} {Heterogenous} {Benchmark} for {Zero}-shot {Evaluation} of {Information} {Retrieval} {Models}.
arXiv.
arXiv:2104.08663 [cs]
(2021).
\url{http://arxiv.org/abs/2104.08663}
Accessed 2023-04-24
\end{botherref}
\endbibitem

\bibitem[\protect\citeauthoryear{Chen et~al.}{2022}]{chen_mtg_2022}
\begin{bchapter}
\bauthor{\bsnm{Chen}, \binits{Y.}},
\bauthor{\bsnm{Song}, \binits{Z.}},
\bauthor{\bsnm{Wu}, \binits{X.}},
\bauthor{\bsnm{Wang}, \binits{D.}},
\bauthor{\bsnm{Xu}, \binits{J.}},
\bauthor{\bsnm{Chen}, \binits{J.}},
\bauthor{\bsnm{Zhou}, \binits{H.}},
\bauthor{\bsnm{Li}, \binits{L.}}:
\bctitle{{MTG}: {A} {Benchmark} {Suite} for {Multilingual} {Text} {Generation}}.
In: \bbtitle{Findings of the {Association} for {Computational} {Linguistics}: {NAACL} 2022},
pp. \bfpage{2508}--\blpage{2527}.
\bpublisher{Association for Computational Linguistics},
\blocation{Seattle, United States}
(\byear{2022}).
\doiurl{10.18653/v1/2022.findings-naacl.192} .
\burl{https://aclanthology.org/2022.findings-naacl.192}
\end{bchapter}
\endbibitem

\bibitem[\protect\citeauthoryear{Guha et~al.}{2022}]{guha_legalbench_2022}
\begin{botherref}
\oauthor{\bsnm{Guha}, \binits{N.}},
\oauthor{\bsnm{Ho}, \binits{D.E.}},
\oauthor{\bsnm{Nyarko}, \binits{J.}},
\oauthor{\bsnm{Ré}, \binits{C.}}:
{LegalBench}: {Prototyping} a {Collaborative} {Benchmark} for {Legal} {Reasoning}.
arXiv.
arXiv:2209.06120 [cs]
(2022).
\url{http://arxiv.org/abs/2209.06120}
Accessed 2022-10-19
\end{botherref}
\endbibitem

\bibitem[\protect\citeauthoryear{Wang et~al.}{2019}]{wang_superglue_2019}
\begin{botherref}
\oauthor{\bsnm{Wang}, \binits{A.}},
\oauthor{\bsnm{Pruksachatkun}, \binits{Y.}},
\oauthor{\bsnm{Nangia}, \binits{N.}},
\oauthor{\bsnm{Singh}, \binits{A.}},
\oauthor{\bsnm{Michael}, \binits{J.}},
\oauthor{\bsnm{Hill}, \binits{F.}},
\oauthor{\bsnm{Levy}, \binits{O.}},
\oauthor{\bsnm{Bowman}, \binits{S.R.}}:
{SuperGLUE}: {A} {Stickier} {Benchmark} for {General}-{Purpose} {Language} {Understanding} {Systems},
30
(2019)
\end{botherref}
\endbibitem

\bibitem[\protect\citeauthoryear{Niklaus et~al.}{2021}]{niklaus_swiss-judgment-prediction_2021}
\begin{bchapter}
\bauthor{\bsnm{Niklaus}, \binits{J.}},
\bauthor{\bsnm{Chalkidis}, \binits{I.}},
\bauthor{\bsnm{Stürmer}, \binits{M.}}:
\bctitle{Swiss-{Judgment}-{Prediction}: {A} {Multilingual} {Legal} {Judgment} {Prediction} {Benchmark}}.
In: \bbtitle{Proceedings of the {Natural} {Legal} {Language} {Processing} {Workshop} 2021},
pp. \bfpage{19}--\blpage{35}.
\bpublisher{Association for Computational Linguistics},
\blocation{Punta Cana, Dominican Republic}
(\byear{2021}).
\burl{https://aclanthology.org/2021.nllp-1.3}
Accessed 2021-12-13
\end{bchapter}
\endbibitem

\bibitem[\protect\citeauthoryear{Niklaus et~al.}{2022}]{niklaus_empirical_2022}
\begin{bchapter}
\bauthor{\bsnm{Niklaus}, \binits{J.}},
\bauthor{\bsnm{Stürmer}, \binits{M.}},
\bauthor{\bsnm{Chalkidis}, \binits{I.}}:
\bctitle{An {Empirical} {Study} on {Cross}-{X} {Transfer} for {Legal} {Judgment} {Prediction}}.
In: \bbtitle{Proceedings of the 2nd {Conference} of the {Asia}-{Pacific} {Chapter} of the {Association} for {Computational} {Linguistics} and the 12th {International} {Joint} {Conference} on {Natural} {Language} {Processing} ({Volume} 1: {Long} {Papers})},
pp. \bfpage{32}--\blpage{46}.
\bpublisher{Association for Computational Linguistics},
\blocation{Online only}
(\byear{2022}).
\burl{https://aclanthology.org/2022.aacl-main.3}
Accessed 2023-01-27
\end{bchapter}
\endbibitem

\bibitem[\protect\citeauthoryear{Shaham et~al.}{2022}]{shaham2022scrolls}
\begin{botherref}
\oauthor{\bsnm{Shaham}, \binits{U.}},
\oauthor{\bsnm{Segal}, \binits{E.}},
\oauthor{\bsnm{Ivgi}, \binits{M.}},
\oauthor{\bsnm{Efrat}, \binits{A.}},
\oauthor{\bsnm{Yoran}, \binits{O.}},
\oauthor{\bsnm{Haviv}, \binits{A.}},
\oauthor{\bsnm{Gupta}, \binits{A.}},
\oauthor{\bsnm{Xiong}, \binits{W.}},
\oauthor{\bsnm{Geva}, \binits{M.}},
\oauthor{\bsnm{Berant}, \binits{J.}}, et al.:
Scrolls: Standardized comparison over long language sequences.
arXiv preprint arXiv:2201.03533
(2022)
\end{botherref}
\endbibitem

\bibitem[\protect\citeauthoryear{Hudson and Moubayed}{2022}]{hudson2022muld}
\begin{botherref}
\oauthor{\bsnm{Hudson}, \binits{G.T.}},
\oauthor{\bsnm{Moubayed}, \binits{N.A.}}:
Muld: The multitask long document benchmark.
arXiv preprint arXiv:2202.07362
(2022)
\end{botherref}
\endbibitem

\bibitem[\protect\citeauthoryear{Chalkidis et~al.}{2022}]{chalkidis2022lexglue}
\begin{bchapter}
\bauthor{\bsnm{Chalkidis}, \binits{I.}},
\bauthor{\bsnm{Jana}, \binits{A.}},
\bauthor{\bsnm{Hartung}, \binits{D.}},
\bauthor{\bsnm{Bommarito}, \binits{M.}},
\bauthor{\bsnm{Androutsopoulos}, \binits{I.}},
\bauthor{\bsnm{Katz}, \binits{D.}},
\bauthor{\bsnm{Aletras}, \binits{N.}}:
\bctitle{Lexglue: A benchmark dataset for legal language understanding in english}.
In: \bbtitle{Proceedings of the 60th Annual Meeting of the Association for Computational Linguistics (Volume 1: Long Papers)},
pp. \bfpage{4310}--\blpage{4330}
(\byear{2022})
\end{bchapter}
\endbibitem

\bibitem[\protect\citeauthoryear{Hu et~al.}{2020}]{hu_xtreme_2020}
\begin{botherref}
\oauthor{\bsnm{Hu}, \binits{J.}},
\oauthor{\bsnm{Ruder}, \binits{S.}},
\oauthor{\bsnm{Siddhant}, \binits{A.}},
\oauthor{\bsnm{Neubig}, \binits{G.}},
\oauthor{\bsnm{Firat}, \binits{O.}},
\oauthor{\bsnm{Johnson}, \binits{M.}}:
{XTREME}: {A} {Massively} {Multilingual} {Multi}-task {Benchmark} for {Evaluating} {Cross}-lingual {Generalization}.
arXiv.
arXiv:2003.11080 [cs]
(2020).
\doiurl{10.48550/arXiv.2003.11080} .
\url{http://arxiv.org/abs/2003.11080}
Accessed 2023-04-28
\end{botherref}
\endbibitem

\bibitem[\protect\citeauthoryear{Ruder et~al.}{2023}]{ruder2023xtremeup}
\begin{botherref}
\oauthor{\bsnm{Ruder}, \binits{S.}},
\oauthor{\bsnm{Clark}, \binits{J.H.}},
\oauthor{\bsnm{Gutkin}, \binits{A.}},
\oauthor{\bsnm{Kale}, \binits{M.}},
\oauthor{\bsnm{Ma}, \binits{M.}},
\oauthor{\bsnm{Nicosia}, \binits{M.}},
\oauthor{\bsnm{Rijhwani}, \binits{S.}},
\oauthor{\bsnm{Riley}, \binits{P.}},
\oauthor{\bsnm{Sarr}, \binits{J.-M.A.}},
\oauthor{\bsnm{Wang}, \binits{X.}},
\oauthor{\bsnm{Wieting}, \binits{J.}},
\oauthor{\bsnm{Gupta}, \binits{N.}},
\oauthor{\bsnm{Katanova}, \binits{A.}},
\oauthor{\bsnm{Kirov}, \binits{C.}},
\oauthor{\bsnm{Dickinson}, \binits{D.L.}},
\oauthor{\bsnm{Roark}, \binits{B.}},
\oauthor{\bsnm{Samanta}, \binits{B.}},
\oauthor{\bsnm{Tao}, \binits{C.}},
\oauthor{\bsnm{Adelani}, \binits{D.I.}},
\oauthor{\bsnm{Axelrod}, \binits{V.}},
\oauthor{\bsnm{Caswell}, \binits{I.}},
\oauthor{\bsnm{Cherry}, \binits{C.}},
\oauthor{\bsnm{Garrette}, \binits{D.}},
\oauthor{\bsnm{Ingle}, \binits{R.}},
\oauthor{\bsnm{Johnson}, \binits{M.}},
\oauthor{\bsnm{Panteleev}, \binits{D.}},
\oauthor{\bsnm{Talukdar}, \binits{P.}}:
XTREME-UP: A User-Centric Scarce-Data Benchmark for Under-Represented Languages
(2023)
\end{botherref}
\endbibitem

\bibitem[\protect\citeauthoryear{Wang et~al.}{2018}]{wang_glue_2018}
\begin{bchapter}
\bauthor{\bsnm{Wang}, \binits{A.}},
\bauthor{\bsnm{Singh}, \binits{A.}},
\bauthor{\bsnm{Michael}, \binits{J.}},
\bauthor{\bsnm{Hill}, \binits{F.}},
\bauthor{\bsnm{Levy}, \binits{O.}},
\bauthor{\bsnm{Bowman}, \binits{S.}}:
\bctitle{{GLUE}: {A} {Multi}-{Task} {Benchmark} and {Analysis} {Platform} for {Natural} {Language} {Understanding}}.
In: \bbtitle{Proceedings of the 2018 {EMNLP} {Workshop} {BlackboxNLP}: {Analyzing} and {Interpreting} {Neural} {Networks} for {NLP}},
pp. \bfpage{353}--\blpage{355}.
\bpublisher{Association for Computational Linguistics},
\blocation{Brussels, Belgium}
(\byear{2018}).
\doiurl{10.18653/v1/W18-5446} .
\burl{https://aclanthology.org/W18-5446}
Accessed 2021-08-19
\end{bchapter}
\endbibitem

\bibitem[\protect\citeauthoryear{Devlin et~al.}{2019}]{devlin_bert_2019}
\begin{bchapter}
\bauthor{\bsnm{Devlin}, \binits{J.}},
\bauthor{\bsnm{Chang}, \binits{M.-W.}},
\bauthor{\bsnm{Lee}, \binits{K.}},
\bauthor{\bsnm{Toutanova}, \binits{K.}}:
\bctitle{{BERT}: {Pre}-training of {Deep} {Bidirectional} {Transformers} for {Language} {Understanding}}.
In: \bbtitle{Proceedings of the 2019 {Conference} of the {North} {American} {Chapter} of the {Association} for {Computational} {Linguistics}: {Human} {Language} {Technologies}, {Volume} 1 ({Long} and {Short} {Papers})},
pp. \bfpage{4171}--\blpage{4186}.
\bpublisher{Association for Computational Linguistics},
\blocation{Minneapolis, Minnesota}
(\byear{2019}).
\doiurl{10.18653/v1/N19-1423} .
\burl{https://aclanthology.org/N19-1423}
\end{bchapter}
\endbibitem

\bibitem[\protect\citeauthoryear{Hendrycks et~al.}{2021}]{hendrycks_measuring_2021}
\begin{botherref}
\oauthor{\bsnm{Hendrycks}, \binits{D.}},
\oauthor{\bsnm{Burns}, \binits{C.}},
\oauthor{\bsnm{Basart}, \binits{S.}},
\oauthor{\bsnm{Zou}, \binits{A.}},
\oauthor{\bsnm{Mazeika}, \binits{M.}},
\oauthor{\bsnm{Song}, \binits{D.}},
\oauthor{\bsnm{Steinhardt}, \binits{J.}}:
Measuring {Massive} {Multitask} {Language} {Understanding}.
arXiv.
arXiv:2009.03300 [cs]
(2021).
\url{http://arxiv.org/abs/2009.03300}
Accessed 2022-10-20
\end{botherref}
\endbibitem

\bibitem[\protect\citeauthoryear{Srivastava et~al.}{2022}]{srivastava_beyond_2022}
\begin{botherref}
\oauthor{\bsnm{Srivastava}, \binits{A.}},
\oauthor{\bsnm{Rastogi}, \binits{A.}},
\oauthor{\bsnm{Rao}, \binits{A.}},
\oauthor{\bsnm{Shoeb}, \binits{A.A.M.}},
\oauthor{\bsnm{Abid}, \binits{A.}},
\oauthor{\bsnm{Fisch}, \binits{A.}},
\oauthor{\bsnm{Brown}, \binits{A.R.}},
\oauthor{\bsnm{Santoro}, \binits{A.}},
\oauthor{\bsnm{Gupta}, \binits{A.}},
\oauthor{\bsnm{Garriga-Alonso}, \binits{A.}},
\oauthor{\bsnm{Kluska}, \binits{A.}},
\oauthor{\bsnm{Lewkowycz}, \binits{A.}},
\oauthor{\bsnm{Agarwal}, \binits{A.}},
\oauthor{\bsnm{Power}, \binits{A.}},
\oauthor{\bsnm{Ray}, \binits{A.}},
\oauthor{\bsnm{Warstadt}, \binits{A.}},
\oauthor{\bsnm{Kocurek}, \binits{A.W.}},
\oauthor{\bsnm{Safaya}, \binits{A.}},
\oauthor{\bsnm{Tazarv}, \binits{A.}},
\oauthor{\bsnm{Xiang}, \binits{A.}},
\oauthor{\bsnm{Parrish}, \binits{A.}},
\oauthor{\bsnm{Nie}, \binits{A.}},
\oauthor{\bsnm{Hussain}, \binits{A.}},
\oauthor{\bsnm{Askell}, \binits{A.}},
\oauthor{\bsnm{Dsouza}, \binits{A.}},
\oauthor{\bsnm{Slone}, \binits{A.}},
\oauthor{\bsnm{Rahane}, \binits{A.}},
\oauthor{\bsnm{Iyer}, \binits{A.S.}},
\oauthor{\bsnm{Andreassen}, \binits{A.}},
\oauthor{\bsnm{Madotto}, \binits{A.}},
\oauthor{\bsnm{Santilli}, \binits{A.}},
\oauthor{\bsnm{Stuhlmüller}, \binits{A.}},
\oauthor{\bsnm{Dai}, \binits{A.}},
\oauthor{\bsnm{La}, \binits{A.}},
\oauthor{\bsnm{Lampinen}, \binits{A.}},
\oauthor{\bsnm{Zou}, \binits{A.}},
\oauthor{\bsnm{Jiang}, \binits{A.}},
\oauthor{\bsnm{Chen}, \binits{A.}},
\oauthor{\bsnm{Vuong}, \binits{A.}},
\oauthor{\bsnm{Gupta}, \binits{A.}},
\oauthor{\bsnm{Gottardi}, \binits{A.}},
\oauthor{\bsnm{Norelli}, \binits{A.}},
\oauthor{\bsnm{Venkatesh}, \binits{A.}},
\oauthor{\bsnm{Gholamidavoodi}, \binits{A.}},
\oauthor{\bsnm{Tabassum}, \binits{A.}},
\oauthor{\bsnm{Menezes}, \binits{A.}},
\oauthor{\bsnm{Kirubarajan}, \binits{A.}},
\oauthor{\bsnm{Mullokandov}, \binits{A.}},
\oauthor{\bsnm{Sabharwal}, \binits{A.}},
\oauthor{\bsnm{Herrick}, \binits{A.}},
\oauthor{\bsnm{Efrat}, \binits{A.}},
\oauthor{\bsnm{Erdem}, \binits{A.}},
\oauthor{\bsnm{Karakaş}, \binits{A.}},
\oauthor{\bsnm{Roberts}, \binits{B.R.}},
\oauthor{\bsnm{Loe}, \binits{B.S.}},
\oauthor{\bsnm{Zoph}, \binits{B.}},
\oauthor{\bsnm{Bojanowski}, \binits{B.}},
\oauthor{\bsnm{Özyurt}, \binits{B.}},
\oauthor{\bsnm{Hedayatnia}, \binits{B.}},
\oauthor{\bsnm{Neyshabur}, \binits{B.}},
\oauthor{\bsnm{Inden}, \binits{B.}},
\oauthor{\bsnm{Stein}, \binits{B.}},
\oauthor{\bsnm{Ekmekci}, \binits{B.}},
\oauthor{\bsnm{Lin}, \binits{B.Y.}},
\oauthor{\bsnm{Howald}, \binits{B.}},
\oauthor{\bsnm{Diao}, \binits{C.}},
\oauthor{\bsnm{Dour}, \binits{C.}},
\oauthor{\bsnm{Stinson}, \binits{C.}},
\oauthor{\bsnm{Argueta}, \binits{C.}},
\oauthor{\bsnm{Ramírez}, \binits{C.F.}},
\oauthor{\bsnm{Singh}, \binits{C.}},
\oauthor{\bsnm{Rathkopf}, \binits{C.}},
\oauthor{\bsnm{Meng}, \binits{C.}},
\oauthor{\bsnm{Baral}, \binits{C.}},
\oauthor{\bsnm{Wu}, \binits{C.}},
\oauthor{\bsnm{Callison-Burch}, \binits{C.}},
\oauthor{\bsnm{Waites}, \binits{C.}},
\oauthor{\bsnm{Voigt}, \binits{C.}},
\oauthor{\bsnm{Manning}, \binits{C.D.}},
\oauthor{\bsnm{Potts}, \binits{C.}},
\oauthor{\bsnm{Ramirez}, \binits{C.}},
\oauthor{\bsnm{Rivera}, \binits{C.E.}},
\oauthor{\bsnm{Siro}, \binits{C.}},
\oauthor{\bsnm{Raffel}, \binits{C.}},
\oauthor{\bsnm{Ashcraft}, \binits{C.}},
\oauthor{\bsnm{Garbacea}, \binits{C.}},
\oauthor{\bsnm{Sileo}, \binits{D.}},
\oauthor{\bsnm{Garrette}, \binits{D.}},
\oauthor{\bsnm{Hendrycks}, \binits{D.}},
\oauthor{\bsnm{Kilman}, \binits{D.}},
\oauthor{\bsnm{Roth}, \binits{D.}},
\oauthor{\bsnm{Freeman}, \binits{D.}},
\oauthor{\bsnm{Khashabi}, \binits{D.}},
\oauthor{\bsnm{Levy}, \binits{D.}},
\oauthor{\bsnm{González}, \binits{D.M.}},
\oauthor{\bsnm{Perszyk}, \binits{D.}},
\oauthor{\bsnm{Hernandez}, \binits{D.}},
\oauthor{\bsnm{Chen}, \binits{D.}},
\oauthor{\bsnm{Ippolito}, \binits{D.}},
\oauthor{\bsnm{Gilboa}, \binits{D.}},
\oauthor{\bsnm{Dohan}, \binits{D.}},
\oauthor{\bsnm{Drakard}, \binits{D.}},
\oauthor{\bsnm{Jurgens}, \binits{D.}},
\oauthor{\bsnm{Datta}, \binits{D.}},
\oauthor{\bsnm{Ganguli}, \binits{D.}},
\oauthor{\bsnm{Emelin}, \binits{D.}},
\oauthor{\bsnm{Kleyko}, \binits{D.}},
\oauthor{\bsnm{Yuret}, \binits{D.}},
\oauthor{\bsnm{Chen}, \binits{D.}},
\oauthor{\bsnm{Tam}, \binits{D.}},
\oauthor{\bsnm{Hupkes}, \binits{D.}},
\oauthor{\bsnm{Misra}, \binits{D.}},
\oauthor{\bsnm{Buzan}, \binits{D.}},
\oauthor{\bsnm{Mollo}, \binits{D.C.}},
\oauthor{\bsnm{Yang}, \binits{D.}},
\oauthor{\bsnm{Lee}, \binits{D.-H.}},
\oauthor{\bsnm{Shutova}, \binits{E.}},
\oauthor{\bsnm{Cubuk}, \binits{E.D.}},
\oauthor{\bsnm{Segal}, \binits{E.}},
\oauthor{\bsnm{Hagerman}, \binits{E.}},
\oauthor{\bsnm{Barnes}, \binits{E.}},
\oauthor{\bsnm{Donoway}, \binits{E.}},
\oauthor{\bsnm{Pavlick}, \binits{E.}},
\oauthor{\bsnm{Rodola}, \binits{E.}},
\oauthor{\bsnm{Lam}, \binits{E.}},
\oauthor{\bsnm{Chu}, \binits{E.}},
\oauthor{\bsnm{Tang}, \binits{E.}},
\oauthor{\bsnm{Erdem}, \binits{E.}},
\oauthor{\bsnm{Chang}, \binits{E.}},
\oauthor{\bsnm{Chi}, \binits{E.A.}},
\oauthor{\bsnm{Dyer}, \binits{E.}},
\oauthor{\bsnm{Jerzak}, \binits{E.}},
\oauthor{\bsnm{Kim}, \binits{E.}},
\oauthor{\bsnm{Manyasi}, \binits{E.E.}},
\oauthor{\bsnm{Zheltonozhskii}, \binits{E.}},
\oauthor{\bsnm{Xia}, \binits{F.}},
\oauthor{\bsnm{Siar}, \binits{F.}},
\oauthor{\bsnm{Martínez-Plumed}, \binits{F.}},
\oauthor{\bsnm{Happé}, \binits{F.}},
\oauthor{\bsnm{Chollet}, \binits{F.}},
\oauthor{\bsnm{Rong}, \binits{F.}},
\oauthor{\bsnm{Mishra}, \binits{G.}},
\oauthor{\bsnm{Winata}, \binits{G.I.}},
\oauthor{\bsnm{Melo}, \binits{G.}},
\oauthor{\bsnm{Kruszewski}, \binits{G.}},
\oauthor{\bsnm{Parascandolo}, \binits{G.}},
\oauthor{\bsnm{Mariani}, \binits{G.}},
\oauthor{\bsnm{Wang}, \binits{G.}},
\oauthor{\bsnm{Jaimovitch-López}, \binits{G.}},
\oauthor{\bsnm{Betz}, \binits{G.}},
\oauthor{\bsnm{Gur-Ari}, \binits{G.}},
\oauthor{\bsnm{Galijasevic}, \binits{H.}},
\oauthor{\bsnm{Kim}, \binits{H.}},
\oauthor{\bsnm{Rashkin}, \binits{H.}},
\oauthor{\bsnm{Hajishirzi}, \binits{H.}},
\oauthor{\bsnm{Mehta}, \binits{H.}},
\oauthor{\bsnm{Bogar}, \binits{H.}},
\oauthor{\bsnm{Shevlin}, \binits{H.}},
\oauthor{\bsnm{Schütze}, \binits{H.}},
\oauthor{\bsnm{Yakura}, \binits{H.}},
\oauthor{\bsnm{Zhang}, \binits{H.}},
\oauthor{\bsnm{Wong}, \binits{H.M.}},
\oauthor{\bsnm{Ng}, \binits{I.}},
\oauthor{\bsnm{Noble}, \binits{I.}},
\oauthor{\bsnm{Jumelet}, \binits{J.}},
\oauthor{\bsnm{Geissinger}, \binits{J.}},
\oauthor{\bsnm{Kernion}, \binits{J.}},
\oauthor{\bsnm{Hilton}, \binits{J.}},
\oauthor{\bsnm{Lee}, \binits{J.}},
\oauthor{\bsnm{Fisac}, \binits{J.F.}},
\oauthor{\bsnm{Simon}, \binits{J.B.}},
\oauthor{\bsnm{Koppel}, \binits{J.}},
\oauthor{\bsnm{Zheng}, \binits{J.}},
\oauthor{\bsnm{Zou}, \binits{J.}},
\oauthor{\bsnm{Kocoń}, \binits{J.}},
\oauthor{\bsnm{Thompson}, \binits{J.}},
\oauthor{\bsnm{Kaplan}, \binits{J.}},
\oauthor{\bsnm{Radom}, \binits{J.}},
\oauthor{\bsnm{Sohl-Dickstein}, \binits{J.}},
\oauthor{\bsnm{Phang}, \binits{J.}},
\oauthor{\bsnm{Wei}, \binits{J.}},
\oauthor{\bsnm{Yosinski}, \binits{J.}},
\oauthor{\bsnm{Novikova}, \binits{J.}},
\oauthor{\bsnm{Bosscher}, \binits{J.}},
\oauthor{\bsnm{Marsh}, \binits{J.}},
\oauthor{\bsnm{Kim}, \binits{J.}},
\oauthor{\bsnm{Taal}, \binits{J.}},
\oauthor{\bsnm{Engel}, \binits{J.}},
\oauthor{\bsnm{Alabi}, \binits{J.}},
\oauthor{\bsnm{Xu}, \binits{J.}},
\oauthor{\bsnm{Song}, \binits{J.}},
\oauthor{\bsnm{Tang}, \binits{J.}},
\oauthor{\bsnm{Waweru}, \binits{J.}},
\oauthor{\bsnm{Burden}, \binits{J.}},
\oauthor{\bsnm{Miller}, \binits{J.}},
\oauthor{\bsnm{Balis}, \binits{J.U.}},
\oauthor{\bsnm{Berant}, \binits{J.}},
\oauthor{\bsnm{Frohberg}, \binits{J.}},
\oauthor{\bsnm{Rozen}, \binits{J.}},
\oauthor{\bsnm{Hernandez-Orallo}, \binits{J.}},
\oauthor{\bsnm{Boudeman}, \binits{J.}},
\oauthor{\bsnm{Jones}, \binits{J.}},
\oauthor{\bsnm{Tenenbaum}, \binits{J.B.}},
\oauthor{\bsnm{Rule}, \binits{J.S.}},
\oauthor{\bsnm{Chua}, \binits{J.}},
\oauthor{\bsnm{Kanclerz}, \binits{K.}},
\oauthor{\bsnm{Livescu}, \binits{K.}},
\oauthor{\bsnm{Krauth}, \binits{K.}},
\oauthor{\bsnm{Gopalakrishnan}, \binits{K.}},
\oauthor{\bsnm{Ignatyeva}, \binits{K.}},
\oauthor{\bsnm{Markert}, \binits{K.}},
\oauthor{\bsnm{Dhole}, \binits{K.D.}},
\oauthor{\bsnm{Gimpel}, \binits{K.}},
\oauthor{\bsnm{Omondi}, \binits{K.}},
\oauthor{\bsnm{Mathewson}, \binits{K.}},
\oauthor{\bsnm{Chiafullo}, \binits{K.}},
\oauthor{\bsnm{Shkaruta}, \binits{K.}},
\oauthor{\bsnm{Shridhar}, \binits{K.}},
\oauthor{\bsnm{McDonell}, \binits{K.}},
\oauthor{\bsnm{Richardson}, \binits{K.}},
\oauthor{\bsnm{Reynolds}, \binits{L.}},
\oauthor{\bsnm{Gao}, \binits{L.}},
\oauthor{\bsnm{Zhang}, \binits{L.}},
\oauthor{\bsnm{Dugan}, \binits{L.}},
\oauthor{\bsnm{Qin}, \binits{L.}},
\oauthor{\bsnm{Contreras-Ochando}, \binits{L.}},
\oauthor{\bsnm{Morency}, \binits{L.-P.}},
\oauthor{\bsnm{Moschella}, \binits{L.}},
\oauthor{\bsnm{Lam}, \binits{L.}},
\oauthor{\bsnm{Noble}, \binits{L.}},
\oauthor{\bsnm{Schmidt}, \binits{L.}},
\oauthor{\bsnm{He}, \binits{L.}},
\oauthor{\bsnm{Colón}, \binits{L.O.}},
\oauthor{\bsnm{Metz}, \binits{L.}},
\oauthor{\bsnm{Şenel}, \binits{L.K.}},
\oauthor{\bsnm{Bosma}, \binits{M.}},
\oauthor{\bsnm{Sap}, \binits{M.}},
\oauthor{\bsnm{Hoeve}, \binits{M.}},
\oauthor{\bsnm{Farooqi}, \binits{M.}},
\oauthor{\bsnm{Faruqui}, \binits{M.}},
\oauthor{\bsnm{Mazeika}, \binits{M.}},
\oauthor{\bsnm{Baturan}, \binits{M.}},
\oauthor{\bsnm{Marelli}, \binits{M.}},
\oauthor{\bsnm{Maru}, \binits{M.}},
\oauthor{\bsnm{Quintana}, \binits{M.J.R.}},
\oauthor{\bsnm{Tolkiehn}, \binits{M.}},
\oauthor{\bsnm{Giulianelli}, \binits{M.}},
\oauthor{\bsnm{Lewis}, \binits{M.}},
\oauthor{\bsnm{Potthast}, \binits{M.}},
\oauthor{\bsnm{Leavitt}, \binits{M.L.}},
\oauthor{\bsnm{Hagen}, \binits{M.}},
\oauthor{\bsnm{Schubert}, \binits{M.}},
\oauthor{\bsnm{Baitemirova}, \binits{M.O.}},
\oauthor{\bsnm{Arnaud}, \binits{M.}},
\oauthor{\bsnm{McElrath}, \binits{M.}},
\oauthor{\bsnm{Yee}, \binits{M.A.}},
\oauthor{\bsnm{Cohen}, \binits{M.}},
\oauthor{\bsnm{Gu}, \binits{M.}},
\oauthor{\bsnm{Ivanitskiy}, \binits{M.}},
\oauthor{\bsnm{Starritt}, \binits{M.}},
\oauthor{\bsnm{Strube}, \binits{M.}},
\oauthor{\bsnm{Swędrowski}, \binits{M.}},
\oauthor{\bsnm{Bevilacqua}, \binits{M.}},
\oauthor{\bsnm{Yasunaga}, \binits{M.}},
\oauthor{\bsnm{Kale}, \binits{M.}},
\oauthor{\bsnm{Cain}, \binits{M.}},
\oauthor{\bsnm{Xu}, \binits{M.}},
\oauthor{\bsnm{Suzgun}, \binits{M.}},
\oauthor{\bsnm{Tiwari}, \binits{M.}},
\oauthor{\bsnm{Bansal}, \binits{M.}},
\oauthor{\bsnm{Aminnaseri}, \binits{M.}},
\oauthor{\bsnm{Geva}, \binits{M.}},
\oauthor{\bsnm{Gheini}, \binits{M.}},
\oauthor{\bsnm{T}, \binits{M.V.}},
\oauthor{\bsnm{Peng}, \binits{N.}},
\oauthor{\bsnm{Chi}, \binits{N.}},
\oauthor{\bsnm{Lee}, \binits{N.}},
\oauthor{\bsnm{Krakover}, \binits{N.G.-A.}},
\oauthor{\bsnm{Cameron}, \binits{N.}},
\oauthor{\bsnm{Roberts}, \binits{N.}},
\oauthor{\bsnm{Doiron}, \binits{N.}},
\oauthor{\bsnm{Nangia}, \binits{N.}},
\oauthor{\bsnm{Deckers}, \binits{N.}},
\oauthor{\bsnm{Muennighoff}, \binits{N.}},
\oauthor{\bsnm{Keskar}, \binits{N.S.}},
\oauthor{\bsnm{Iyer}, \binits{N.S.}},
\oauthor{\bsnm{Constant}, \binits{N.}},
\oauthor{\bsnm{Fiedel}, \binits{N.}},
\oauthor{\bsnm{Wen}, \binits{N.}},
\oauthor{\bsnm{Zhang}, \binits{O.}},
\oauthor{\bsnm{Agha}, \binits{O.}},
\oauthor{\bsnm{Elbaghdadi}, \binits{O.}},
\oauthor{\bsnm{Levy}, \binits{O.}},
\oauthor{\bsnm{Evans}, \binits{O.}},
\oauthor{\bsnm{Casares}, \binits{P.A.M.}},
\oauthor{\bsnm{Doshi}, \binits{P.}},
\oauthor{\bsnm{Fung}, \binits{P.}},
\oauthor{\bsnm{Liang}, \binits{P.P.}},
\oauthor{\bsnm{Vicol}, \binits{P.}},
\oauthor{\bsnm{Alipoormolabashi}, \binits{P.}},
\oauthor{\bsnm{Liao}, \binits{P.}},
\oauthor{\bsnm{Liang}, \binits{P.}},
\oauthor{\bsnm{Chang}, \binits{P.}},
\oauthor{\bsnm{Eckersley}, \binits{P.}},
\oauthor{\bsnm{Htut}, \binits{P.M.}},
\oauthor{\bsnm{Hwang}, \binits{P.}},
\oauthor{\bsnm{Miłkowski}, \binits{P.}},
\oauthor{\bsnm{Patil}, \binits{P.}},
\oauthor{\bsnm{Pezeshkpour}, \binits{P.}},
\oauthor{\bsnm{Oli}, \binits{P.}},
\oauthor{\bsnm{Mei}, \binits{Q.}},
\oauthor{\bsnm{Lyu}, \binits{Q.}},
\oauthor{\bsnm{Chen}, \binits{Q.}},
\oauthor{\bsnm{Banjade}, \binits{R.}},
\oauthor{\bsnm{Rudolph}, \binits{R.E.}},
\oauthor{\bsnm{Gabriel}, \binits{R.}},
\oauthor{\bsnm{Habacker}, \binits{R.}},
\oauthor{\bsnm{Delgado}, \binits{R.R.}},
\oauthor{\bsnm{Millière}, \binits{R.}},
\oauthor{\bsnm{Garg}, \binits{R.}},
\oauthor{\bsnm{Barnes}, \binits{R.}},
\oauthor{\bsnm{Saurous}, \binits{R.A.}},
\oauthor{\bsnm{Arakawa}, \binits{R.}},
\oauthor{\bsnm{Raymaekers}, \binits{R.}},
\oauthor{\bsnm{Frank}, \binits{R.}},
\oauthor{\bsnm{Sikand}, \binits{R.}},
\oauthor{\bsnm{Novak}, \binits{R.}},
\oauthor{\bsnm{Sitelew}, \binits{R.}},
\oauthor{\bsnm{LeBras}, \binits{R.}},
\oauthor{\bsnm{Liu}, \binits{R.}},
\oauthor{\bsnm{Jacobs}, \binits{R.}},
\oauthor{\bsnm{Zhang}, \binits{R.}},
\oauthor{\bsnm{Salakhutdinov}, \binits{R.}},
\oauthor{\bsnm{Chi}, \binits{R.}},
\oauthor{\bsnm{Lee}, \binits{R.}},
\oauthor{\bsnm{Stovall}, \binits{R.}},
\oauthor{\bsnm{Teehan}, \binits{R.}},
\oauthor{\bsnm{Yang}, \binits{R.}},
\oauthor{\bsnm{Singh}, \binits{S.}},
\oauthor{\bsnm{Mohammad}, \binits{S.M.}},
\oauthor{\bsnm{Anand}, \binits{S.}},
\oauthor{\bsnm{Dillavou}, \binits{S.}},
\oauthor{\bsnm{Shleifer}, \binits{S.}},
\oauthor{\bsnm{Wiseman}, \binits{S.}},
\oauthor{\bsnm{Gruetter}, \binits{S.}},
\oauthor{\bsnm{Bowman}, \binits{S.R.}},
\oauthor{\bsnm{Schoenholz}, \binits{S.S.}},
\oauthor{\bsnm{Han}, \binits{S.}},
\oauthor{\bsnm{Kwatra}, \binits{S.}},
\oauthor{\bsnm{Rous}, \binits{S.A.}},
\oauthor{\bsnm{Ghazarian}, \binits{S.}},
\oauthor{\bsnm{Ghosh}, \binits{S.}},
\oauthor{\bsnm{Casey}, \binits{S.}},
\oauthor{\bsnm{Bischoff}, \binits{S.}},
\oauthor{\bsnm{Gehrmann}, \binits{S.}},
\oauthor{\bsnm{Schuster}, \binits{S.}},
\oauthor{\bsnm{Sadeghi}, \binits{S.}},
\oauthor{\bsnm{Hamdan}, \binits{S.}},
\oauthor{\bsnm{Zhou}, \binits{S.}},
\oauthor{\bsnm{Srivastava}, \binits{S.}},
\oauthor{\bsnm{Shi}, \binits{S.}},
\oauthor{\bsnm{Singh}, \binits{S.}},
\oauthor{\bsnm{Asaadi}, \binits{S.}},
\oauthor{\bsnm{Gu}, \binits{S.S.}},
\oauthor{\bsnm{Pachchigar}, \binits{S.}},
\oauthor{\bsnm{Toshniwal}, \binits{S.}},
\oauthor{\bsnm{Upadhyay}, \binits{S.}},
\oauthor{\bsnm{Shyamolima}},
\oauthor{\bsnm{Debnath}},
\oauthor{\bsnm{Shakeri}, \binits{S.}},
\oauthor{\bsnm{Thormeyer}, \binits{S.}},
\oauthor{\bsnm{Melzi}, \binits{S.}},
\oauthor{\bsnm{Reddy}, \binits{S.}},
\oauthor{\bsnm{Makini}, \binits{S.P.}},
\oauthor{\bsnm{Lee}, \binits{S.-H.}},
\oauthor{\bsnm{Torene}, \binits{S.}},
\oauthor{\bsnm{Hatwar}, \binits{S.}},
\oauthor{\bsnm{Dehaene}, \binits{S.}},
\oauthor{\bsnm{Divic}, \binits{S.}},
\oauthor{\bsnm{Ermon}, \binits{S.}},
\oauthor{\bsnm{Biderman}, \binits{S.}},
\oauthor{\bsnm{Lin}, \binits{S.}},
\oauthor{\bsnm{Prasad}, \binits{S.}},
\oauthor{\bsnm{Piantadosi}, \binits{S.T.}},
\oauthor{\bsnm{Shieber}, \binits{S.M.}},
\oauthor{\bsnm{Misherghi}, \binits{S.}},
\oauthor{\bsnm{Kiritchenko}, \binits{S.}},
\oauthor{\bsnm{Mishra}, \binits{S.}},
\oauthor{\bsnm{Linzen}, \binits{T.}},
\oauthor{\bsnm{Schuster}, \binits{T.}},
\oauthor{\bsnm{Li}, \binits{T.}},
\oauthor{\bsnm{Yu}, \binits{T.}},
\oauthor{\bsnm{Ali}, \binits{T.}},
\oauthor{\bsnm{Hashimoto}, \binits{T.}},
\oauthor{\bsnm{Wu}, \binits{T.-L.}},
\oauthor{\bsnm{Desbordes}, \binits{T.}},
\oauthor{\bsnm{Rothschild}, \binits{T.}},
\oauthor{\bsnm{Phan}, \binits{T.}},
\oauthor{\bsnm{Wang}, \binits{T.}},
\oauthor{\bsnm{Nkinyili}, \binits{T.}},
\oauthor{\bsnm{Schick}, \binits{T.}},
\oauthor{\bsnm{Kornev}, \binits{T.}},
\oauthor{\bsnm{Telleen-Lawton}, \binits{T.}},
\oauthor{\bsnm{Tunduny}, \binits{T.}},
\oauthor{\bsnm{Gerstenberg}, \binits{T.}},
\oauthor{\bsnm{Chang}, \binits{T.}},
\oauthor{\bsnm{Neeraj}, \binits{T.}},
\oauthor{\bsnm{Khot}, \binits{T.}},
\oauthor{\bsnm{Shultz}, \binits{T.}},
\oauthor{\bsnm{Shaham}, \binits{U.}},
\oauthor{\bsnm{Misra}, \binits{V.}},
\oauthor{\bsnm{Demberg}, \binits{V.}},
\oauthor{\bsnm{Nyamai}, \binits{V.}},
\oauthor{\bsnm{Raunak}, \binits{V.}},
\oauthor{\bsnm{Ramasesh}, \binits{V.}},
\oauthor{\bsnm{Prabhu}, \binits{V.U.}},
\oauthor{\bsnm{Padmakumar}, \binits{V.}},
\oauthor{\bsnm{Srikumar}, \binits{V.}},
\oauthor{\bsnm{Fedus}, \binits{W.}},
\oauthor{\bsnm{Saunders}, \binits{W.}},
\oauthor{\bsnm{Zhang}, \binits{W.}},
\oauthor{\bsnm{Vossen}, \binits{W.}},
\oauthor{\bsnm{Ren}, \binits{X.}},
\oauthor{\bsnm{Tong}, \binits{X.}},
\oauthor{\bsnm{Zhao}, \binits{X.}},
\oauthor{\bsnm{Wu}, \binits{X.}},
\oauthor{\bsnm{Shen}, \binits{X.}},
\oauthor{\bsnm{Yaghoobzadeh}, \binits{Y.}},
\oauthor{\bsnm{Lakretz}, \binits{Y.}},
\oauthor{\bsnm{Song}, \binits{Y.}},
\oauthor{\bsnm{Bahri}, \binits{Y.}},
\oauthor{\bsnm{Choi}, \binits{Y.}},
\oauthor{\bsnm{Yang}, \binits{Y.}},
\oauthor{\bsnm{Hao}, \binits{Y.}},
\oauthor{\bsnm{Chen}, \binits{Y.}},
\oauthor{\bsnm{Belinkov}, \binits{Y.}},
\oauthor{\bsnm{Hou}, \binits{Y.}},
\oauthor{\bsnm{Hou}, \binits{Y.}},
\oauthor{\bsnm{Bai}, \binits{Y.}},
\oauthor{\bsnm{Seid}, \binits{Z.}},
\oauthor{\bsnm{Zhao}, \binits{Z.}},
\oauthor{\bsnm{Wang}, \binits{Z.}},
\oauthor{\bsnm{Wang}, \binits{Z.J.}},
\oauthor{\bsnm{Wang}, \binits{Z.}},
\oauthor{\bsnm{Wu}, \binits{Z.}}:
Beyond the {Imitation} {Game}: {Quantifying} and extrapolating the capabilities of language models.
arXiv.
arXiv:2206.04615 [cs, stat]
(2022).
\doiurl{10.48550/arXiv.2206.04615} .
\url{http://arxiv.org/abs/2206.04615}
Accessed 2023-04-28
\end{botherref}
\endbibitem

\bibitem[\protect\citeauthoryear{Wei et~al.}{2022}]{Wei2022}
\begin{botherref}
\oauthor{\bsnm{Wei}, \binits{J.}},
\oauthor{\bsnm{Wang}, \binits{X.}},
\oauthor{\bsnm{Schuurmans}, \binits{D.}},
\oauthor{\bsnm{Bosma}, \binits{M.}},
\oauthor{\bsnm{Ichter}, \binits{B.}},
\oauthor{\bsnm{Xia}, \binits{F.}},
\oauthor{\bsnm{Chi}, \binits{E.}},
\oauthor{\bsnm{Le}, \binits{Q.}},
\oauthor{\bsnm{Zhou}, \binits{D.}}:
Chain-of-thought prompting elicits reasoning in large language models
(2022)
\end{botherref}
\endbibitem

\bibitem[\protect\citeauthoryear{Zhou et~al.}{2023}]{zhou_least--most_2023}
\begin{botherref}
\oauthor{\bsnm{Zhou}, \binits{D.}},
\oauthor{\bsnm{Schärli}, \binits{N.}},
\oauthor{\bsnm{Hou}, \binits{L.}},
\oauthor{\bsnm{Wei}, \binits{J.}},
\oauthor{\bsnm{Scales}, \binits{N.}},
\oauthor{\bsnm{Wang}, \binits{X.}},
\oauthor{\bsnm{Schuurmans}, \binits{D.}},
\oauthor{\bsnm{Cui}, \binits{C.}},
\oauthor{\bsnm{Bousquet}, \binits{O.}},
\oauthor{\bsnm{Le}, \binits{Q.}},
\oauthor{\bsnm{Chi}, \binits{E.}}:
Least-to-{Most} {Prompting} {Enables} {Complex} {Reasoning} in {Large} {Language} {Models}.
arXiv.
arXiv:2205.10625 [cs]
(2023).
\url{http://arxiv.org/abs/2205.10625}
Accessed 2023-11-28
\end{botherref}
\endbibitem

\bibitem[\protect\citeauthoryear{Wang et~al.}{2023}]{wang_self-consistency_2023}
\begin{botherref}
\oauthor{\bsnm{Wang}, \binits{X.}},
\oauthor{\bsnm{Wei}, \binits{J.}},
\oauthor{\bsnm{Schuurmans}, \binits{D.}},
\oauthor{\bsnm{Le}, \binits{Q.}},
\oauthor{\bsnm{Chi}, \binits{E.}},
\oauthor{\bsnm{Narang}, \binits{S.}},
\oauthor{\bsnm{Chowdhery}, \binits{A.}},
\oauthor{\bsnm{Zhou}, \binits{D.}}:
Self-{Consistency} {Improves} {Chain} of {Thought} {Reasoning} in {Language} {Models}.
arXiv.
arXiv:2203.11171 [cs]
(2023).
\doiurl{10.48550/arXiv.2203.11171} .
\url{http://arxiv.org/abs/2203.11171}
Accessed 2023-10-28
\end{botherref}
\endbibitem

\bibitem[\protect\citeauthoryear{Yao et~al.}{2023}]{yao_tree_2023}
\begin{botherref}
\oauthor{\bsnm{Yao}, \binits{S.}},
\oauthor{\bsnm{Yu}, \binits{D.}},
\oauthor{\bsnm{Zhao}, \binits{J.}},
\oauthor{\bsnm{Shafran}, \binits{I.}},
\oauthor{\bsnm{Griffiths}, \binits{T.L.}},
\oauthor{\bsnm{Cao}, \binits{Y.}},
\oauthor{\bsnm{Narasimhan}, \binits{K.}}:
Tree of {Thoughts}: {Deliberate} {Problem} {Solving} with {Large} {Language} {Models}.
arXiv.
arXiv:2305.10601 [cs]
(2023).
\doiurl{10.48550/arXiv.2305.10601} .
\url{http://arxiv.org/abs/2305.10601}
Accessed 2023-10-28
\end{botherref}
\endbibitem

\bibitem[\protect\citeauthoryear{Long}{2023}]{Long2023}
\begin{botherref}
\oauthor{\bsnm{Long}, \binits{J.}}:
{Large Language Model Guided Tree-of-Thought}
(2023)
{\href{https://arxiv.org/abs/2305.08291}{{arXiv:2305.08291}}}
\end{botherref}
\endbibitem

\bibitem[\protect\citeauthoryear{Besta et~al.}{2023}]{besta_graph_2023}
\begin{botherref}
\oauthor{\bsnm{Besta}, \binits{M.}},
\oauthor{\bsnm{Blach}, \binits{N.}},
\oauthor{\bsnm{Kubicek}, \binits{A.}},
\oauthor{\bsnm{Gerstenberger}, \binits{R.}},
\oauthor{\bsnm{Gianinazzi}, \binits{L.}},
\oauthor{\bsnm{Gajda}, \binits{J.}},
\oauthor{\bsnm{Lehmann}, \binits{T.}},
\oauthor{\bsnm{Podstawski}, \binits{M.}},
\oauthor{\bsnm{Niewiadomski}, \binits{H.}},
\oauthor{\bsnm{Nyczyk}, \binits{P.}},
\oauthor{\bsnm{Hoefler}, \binits{T.}}:
Graph of {Thoughts}: {Solving} {Elaborate} {Problems} with {Large} {Language} {Models}.
arXiv.
arXiv:2308.09687 [cs]
(2023).
\url{http://arxiv.org/abs/2308.09687}
Accessed 2023-11-28
\end{botherref}
\endbibitem

\bibitem[\protect\citeauthoryear{Guha et~al.}{2023}]{guha_legalbench_2023}
\begin{botherref}
\oauthor{\bsnm{Guha}, \binits{N.}},
\oauthor{\bsnm{Nyarko}, \binits{J.}},
\oauthor{\bsnm{Ho}, \binits{D.E.}},
\oauthor{\bsnm{Ré}, \binits{C.}},
\oauthor{\bsnm{Chilton}, \binits{A.}},
\oauthor{\bsnm{Narayana}, \binits{A.}},
\oauthor{\bsnm{Chohlas-Wood}, \binits{A.}},
\oauthor{\bsnm{Peters}, \binits{A.}},
\oauthor{\bsnm{Waldon}, \binits{B.}},
\oauthor{\bsnm{Rockmore}, \binits{D.N.}},
\oauthor{\bsnm{Zambrano}, \binits{D.}},
\oauthor{\bsnm{Talisman}, \binits{D.}},
\oauthor{\bsnm{Hoque}, \binits{E.}},
\oauthor{\bsnm{Surani}, \binits{F.}},
\oauthor{\bsnm{Fagan}, \binits{F.}},
\oauthor{\bsnm{Sarfaty}, \binits{G.}},
\oauthor{\bsnm{Dickinson}, \binits{G.M.}},
\oauthor{\bsnm{Porat}, \binits{H.}},
\oauthor{\bsnm{Hegland}, \binits{J.}},
\oauthor{\bsnm{Wu}, \binits{J.}},
\oauthor{\bsnm{Nudell}, \binits{J.}},
\oauthor{\bsnm{Niklaus}, \binits{J.}},
\oauthor{\bsnm{Nay}, \binits{J.}},
\oauthor{\bsnm{Choi}, \binits{J.H.}},
\oauthor{\bsnm{Tobia}, \binits{K.}},
\oauthor{\bsnm{Hagan}, \binits{M.}},
\oauthor{\bsnm{Ma}, \binits{M.}},
\oauthor{\bsnm{Livermore}, \binits{M.}},
\oauthor{\bsnm{Rasumov-Rahe}, \binits{N.}},
\oauthor{\bsnm{Holzenberger}, \binits{N.}},
\oauthor{\bsnm{Kolt}, \binits{N.}},
\oauthor{\bsnm{Henderson}, \binits{P.}},
\oauthor{\bsnm{Rehaag}, \binits{S.}},
\oauthor{\bsnm{Goel}, \binits{S.}},
\oauthor{\bsnm{Gao}, \binits{S.}},
\oauthor{\bsnm{Williams}, \binits{S.}},
\oauthor{\bsnm{Gandhi}, \binits{S.}},
\oauthor{\bsnm{Zur}, \binits{T.}},
\oauthor{\bsnm{Iyer}, \binits{V.}},
\oauthor{\bsnm{Li}, \binits{Z.}}:
{LegalBench}: {A} {Collaboratively} {Built} {Benchmark} for {Measuring} {Legal} {Reasoning} in {Large} {Language} {Models}.
arXiv.
arXiv:2308.11462 [cs]
(2023).
\doiurl{10.48550/arXiv.2308.11462} .
\url{http://arxiv.org/abs/2308.11462}
Accessed 2023-09-10
\end{botherref}
\endbibitem

\bibitem[\protect\citeauthoryear{Yu et~al.}{2022}]{Yu2022}
\begin{botherref}
\oauthor{\bsnm{Yu}, \binits{F.}},
\oauthor{\bsnm{Quartey}, \binits{L.}},
\oauthor{\bsnm{Schilder}, \binits{F.}}:
Legal prompting: Teaching a language model to think like a lawyer
(2022)
\end{botherref}
\endbibitem

\bibitem[\protect\citeauthoryear{Bundesgericht}{2019}]{SwissFederalSupremeCourt}
\begin{botherref}
\oauthor{\bsnm{Bundesgericht}, \binits{S.}}:
The Paths to the Swiss Federal Supreme Court.
\url{https://www.bger.ch/files/live/sites/bger/files/pdf/en/BG_Brosch%C3%BCreA5_E_Onl.pdf}.
Accessed: 2023-04-27
(2019)
\end{botherref}
\endbibitem

\bibitem[\protect\citeauthoryear{Brown et~al.}{2020}]{brown-etal-gpt3}
\begin{bchapter}
\bauthor{\bsnm{Brown}, \binits{T.}},
\bauthor{\bsnm{Mann}, \binits{B.}},
\bauthor{\bsnm{Ryder}, \binits{N.}},
\bauthor{\bsnm{Subbiah}, \binits{M.}},
\bauthor{\bsnm{Kaplan}, \binits{J.D.}},
\bauthor{\bsnm{Dhariwal}, \binits{P.}},
\bauthor{\bsnm{Neelakantan}, \binits{A.}},
\bauthor{\bsnm{Shyam}, \binits{P.}},
\bauthor{\bsnm{Sastry}, \binits{G.}},
\bauthor{\bsnm{Askell}, \binits{A.}},
\bauthor{\bsnm{Agarwal}, \binits{S.}},
\bauthor{\bsnm{Herbert-Voss}, \binits{A.}},
\bauthor{\bsnm{Krueger}, \binits{G.}},
\bauthor{\bsnm{Henighan}, \binits{T.}},
\bauthor{\bsnm{Child}, \binits{R.}},
\bauthor{\bsnm{Ramesh}, \binits{A.}},
\bauthor{\bsnm{Ziegler}, \binits{D.}},
\bauthor{\bsnm{Wu}, \binits{J.}},
\bauthor{\bsnm{Winter}, \binits{C.}},
\bauthor{\bsnm{Hesse}, \binits{C.}},
\bauthor{\bsnm{Chen}, \binits{M.}},
\bauthor{\bsnm{Sigler}, \binits{E.}},
\bauthor{\bsnm{Litwin}, \binits{M.}},
\bauthor{\bsnm{Gray}, \binits{S.}},
\bauthor{\bsnm{Chess}, \binits{B.}},
\bauthor{\bsnm{Clark}, \binits{J.}},
\bauthor{\bsnm{Berner}, \binits{C.}},
\bauthor{\bsnm{McCandlish}, \binits{S.}},
\bauthor{\bsnm{Radford}, \binits{A.}},
\bauthor{\bsnm{Sutskever}, \binits{I.}},
\bauthor{\bsnm{Amodei}, \binits{D.}}:
\bctitle{Language models are few-shot learners}.
In: \beditor{\bsnm{Larochelle}, \binits{H.}},
\beditor{\bsnm{Ranzato}, \binits{M.}},
\beditor{\bsnm{Hadsell}, \binits{R.}},
\beditor{\bsnm{Balcan}, \binits{M.F.}},
\beditor{\bsnm{Lin}, \binits{H.}} (eds.)
\bbtitle{Advances in Neural Information Processing Systems},
vol. \bseriesno{33},
pp. \bfpage{1877}--\blpage{1901}.
\bpublisher{Curran Associates, Inc.}, \blocation{???}
(\byear{2020}).
\burl{https://proceedings.neurips.cc/paper/2020/file/1457c0d6bfcb4967418bfb8ac142f64a-Paper.pdf}
\end{bchapter}
\endbibitem

\bibitem[\protect\citeauthoryear{Grave et~al.}{2018}]{grave2018learning}
\begin{bchapter}
\bauthor{\bsnm{Grave}, \binits{E.}},
\bauthor{\bsnm{Bojanowski}, \binits{P.}},
\bauthor{\bsnm{Gupta}, \binits{P.}},
\bauthor{\bsnm{Joulin}, \binits{A.}},
\bauthor{\bsnm{Mikolov}, \binits{T.}}:
\bctitle{Learning word vectors for 157 languages}.
In: \bbtitle{Proceedings of the International Conference on Language Resources and Evaluation (LREC 2018)}
(\byear{2018})
\end{bchapter}
\endbibitem

\bibitem[\protect\citeauthoryear{Chalkidis et~al.}{2021}]{chalkidis2021regulatory}
\begin{botherref}
\oauthor{\bsnm{Chalkidis}, \binits{I.}},
\oauthor{\bsnm{Fergadiotis}, \binits{M.}},
\oauthor{\bsnm{Manginas}, \binits{N.}},
\oauthor{\bsnm{Katakalou}, \binits{E.}},
\oauthor{\bsnm{Malakasiotis}, \binits{P.}}:
Regulatory Compliance through Doc2Doc Information Retrieval: A case study in EU/UK legislation where text similarity has limitations
(2021)
\end{botherref}
\endbibitem

\bibitem[\protect\citeauthoryear{Wang et~al.}{2020}]{wang_minilm_2020}
\begin{bchapter}
\bauthor{\bsnm{Wang}, \binits{W.}},
\bauthor{\bsnm{Wei}, \binits{F.}},
\bauthor{\bsnm{Dong}, \binits{L.}},
\bauthor{\bsnm{Bao}, \binits{H.}},
\bauthor{\bsnm{Yang}, \binits{N.}},
\bauthor{\bsnm{Zhou}, \binits{M.}}:
\bctitle{{MiniLM}: {Deep} {Self}-{Attention} {Distillation} for {Task}-{Agnostic} {Compression} of {Pre}-{Trained} {Transformers}}.
In: \bbtitle{Advances in {Neural} {Information} {Processing} {Systems}},
vol. \bseriesno{33},
pp. \bfpage{5776}--\blpage{5788}.
\bpublisher{Curran Associates, Inc.}, \blocation{???}
(\byear{2020}).
\burl{https://proceedings.neurips.cc/paper/2020/hash/3f5ee243547dee91fbd053c1c4a845aa-Abstract.html}
Accessed 2022-07-21
\end{bchapter}
\endbibitem

\bibitem[\protect\citeauthoryear{Taylor et~al.}{2022}]{taylor_galactica_2022}
\begin{botherref}
\oauthor{\bsnm{Taylor}, \binits{R.}},
\oauthor{\bsnm{Kardas}, \binits{M.}},
\oauthor{\bsnm{Cucurull}, \binits{G.}},
\oauthor{\bsnm{Scialom}, \binits{T.}},
\oauthor{\bsnm{Hartshorn}, \binits{A.}},
\oauthor{\bsnm{Saravia}, \binits{E.}},
\oauthor{\bsnm{Poulton}, \binits{A.}},
\oauthor{\bsnm{Kerkez}, \binits{V.}},
\oauthor{\bsnm{Stojnic}, \binits{R.}}:
Galactica: {A} {Large} {Language} {Model} for {Science}.
arXiv.
arXiv:2211.09085 [cs, stat]
(2022).
\doiurl{10.48550/arXiv.2211.09085} .
\url{http://arxiv.org/abs/2211.09085}
Accessed 2022-11-29
\end{botherref}
\endbibitem

\bibitem[\protect\citeauthoryear{Scao et~al.}{2022}]{scao_bloom_2022}
\begin{botherref}
\oauthor{\bsnm{Scao}, \binits{T.L.}},
\oauthor{\bsnm{Fan}, \binits{A.}},
\oauthor{\bsnm{Akiki}, \binits{C.}},
\oauthor{\bsnm{Pavlick}, \binits{E.}},
\oauthor{\bsnm{Ilić}, \binits{S.}},
\oauthor{\bsnm{Hesslow}, \binits{D.}},
\oauthor{\bsnm{Castagné}, \binits{R.}},
\oauthor{\bsnm{Luccioni}, \binits{A.S.}},
\oauthor{\bsnm{Yvon}, \binits{F.}},
\oauthor{\bsnm{Gallé}, \binits{M.}},
\oauthor{\bsnm{Tow}, \binits{J.}},
\oauthor{\bsnm{Rush}, \binits{A.M.}},
\oauthor{\bsnm{Biderman}, \binits{S.}},
\oauthor{\bsnm{Webson}, \binits{A.}},
\oauthor{\bsnm{Ammanamanchi}, \binits{P.S.}},
\oauthor{\bsnm{Wang}, \binits{T.}},
\oauthor{\bsnm{Sagot}, \binits{B.}},
\oauthor{\bsnm{Muennighoff}, \binits{N.}},
\oauthor{\bsnm{Moral}, \binits{A.V.}},
\oauthor{\bsnm{Ruwase}, \binits{O.}},
\oauthor{\bsnm{Bawden}, \binits{R.}},
\oauthor{\bsnm{Bekman}, \binits{S.}},
\oauthor{\bsnm{McMillan-Major}, \binits{A.}},
\oauthor{\bsnm{Beltagy}, \binits{I.}},
\oauthor{\bsnm{Nguyen}, \binits{H.}},
\oauthor{\bsnm{Saulnier}, \binits{L.}},
\oauthor{\bsnm{Tan}, \binits{S.}},
\oauthor{\bsnm{Suarez}, \binits{P.O.}},
\oauthor{\bsnm{Sanh}, \binits{V.}},
\oauthor{\bsnm{Laurençon}, \binits{H.}},
\oauthor{\bsnm{Jernite}, \binits{Y.}},
\oauthor{\bsnm{Launay}, \binits{J.}},
\oauthor{\bsnm{Mitchell}, \binits{M.}},
\oauthor{\bsnm{Raffel}, \binits{C.}},
\oauthor{\bsnm{Gokaslan}, \binits{A.}},
\oauthor{\bsnm{Simhi}, \binits{A.}},
\oauthor{\bsnm{Soroa}, \binits{A.}},
\oauthor{\bsnm{Aji}, \binits{A.F.}},
\oauthor{\bsnm{Alfassy}, \binits{A.}},
\oauthor{\bsnm{Rogers}, \binits{A.}},
\oauthor{\bsnm{Nitzav}, \binits{A.K.}},
\oauthor{\bsnm{Xu}, \binits{C.}},
\oauthor{\bsnm{Mou}, \binits{C.}},
\oauthor{\bsnm{Emezue}, \binits{C.}},
\oauthor{\bsnm{Klamm}, \binits{C.}},
\oauthor{\bsnm{Leong}, \binits{C.}},
\oauthor{\bsnm{Strien}, \binits{D.}},
\oauthor{\bsnm{Adelani}, \binits{D.I.}},
\oauthor{\bsnm{Radev}, \binits{D.}},
\oauthor{\bsnm{Ponferrada}, \binits{E.G.}},
\oauthor{\bsnm{Levkovizh}, \binits{E.}},
\oauthor{\bsnm{Kim}, \binits{E.}},
\oauthor{\bsnm{Natan}, \binits{E.B.}},
\oauthor{\bsnm{De~Toni}, \binits{F.}},
\oauthor{\bsnm{Dupont}, \binits{G.}},
\oauthor{\bsnm{Kruszewski}, \binits{G.}},
\oauthor{\bsnm{Pistilli}, \binits{G.}},
\oauthor{\bsnm{Elsahar}, \binits{H.}},
\oauthor{\bsnm{Benyamina}, \binits{H.}},
\oauthor{\bsnm{Tran}, \binits{H.}},
\oauthor{\bsnm{Yu}, \binits{I.}},
\oauthor{\bsnm{Abdulmumin}, \binits{I.}},
\oauthor{\bsnm{Johnson}, \binits{I.}},
\oauthor{\bsnm{Gonzalez-Dios}, \binits{I.}},
\oauthor{\bsnm{Rosa}, \binits{J.}},
\oauthor{\bsnm{Chim}, \binits{J.}},
\oauthor{\bsnm{Dodge}, \binits{J.}},
\oauthor{\bsnm{Zhu}, \binits{J.}},
\oauthor{\bsnm{Chang}, \binits{J.}},
\oauthor{\bsnm{Frohberg}, \binits{J.}},
\oauthor{\bsnm{Tobing}, \binits{J.}},
\oauthor{\bsnm{Bhattacharjee}, \binits{J.}},
\oauthor{\bsnm{Almubarak}, \binits{K.}},
\oauthor{\bsnm{Chen}, \binits{K.}},
\oauthor{\bsnm{Lo}, \binits{K.}},
\oauthor{\bsnm{Von~Werra}, \binits{L.}},
\oauthor{\bsnm{Weber}, \binits{L.}},
\oauthor{\bsnm{Phan}, \binits{L.}},
\oauthor{\bsnm{allal}, \binits{L.B.}},
\oauthor{\bsnm{Tanguy}, \binits{L.}},
\oauthor{\bsnm{Dey}, \binits{M.}},
\oauthor{\bsnm{Muñoz}, \binits{M.R.}},
\oauthor{\bsnm{Masoud}, \binits{M.}},
\oauthor{\bsnm{Grandury}, \binits{M.}},
\oauthor{\bsnm{Šaško}, \binits{M.}},
\oauthor{\bsnm{Huang}, \binits{M.}},
\oauthor{\bsnm{Coavoux}, \binits{M.}},
\oauthor{\bsnm{Singh}, \binits{M.}},
\oauthor{\bsnm{Jiang}, \binits{M.T.-J.}},
\oauthor{\bsnm{Vu}, \binits{M.C.}},
\oauthor{\bsnm{Jauhar}, \binits{M.A.}},
\oauthor{\bsnm{Ghaleb}, \binits{M.}},
\oauthor{\bsnm{Subramani}, \binits{N.}},
\oauthor{\bsnm{Kassner}, \binits{N.}},
\oauthor{\bsnm{Khamis}, \binits{N.}},
\oauthor{\bsnm{Nguyen}, \binits{O.}},
\oauthor{\bsnm{Espejel}, \binits{O.}},
\oauthor{\bsnm{Gibert}, \binits{O.}},
\oauthor{\bsnm{Villegas}, \binits{P.}},
\oauthor{\bsnm{Henderson}, \binits{P.}},
\oauthor{\bsnm{Colombo}, \binits{P.}},
\oauthor{\bsnm{Amuok}, \binits{P.}},
\oauthor{\bsnm{Lhoest}, \binits{Q.}},
\oauthor{\bsnm{Harliman}, \binits{R.}},
\oauthor{\bsnm{Bommasani}, \binits{R.}},
\oauthor{\bsnm{López}, \binits{R.L.}},
\oauthor{\bsnm{Ribeiro}, \binits{R.}},
\oauthor{\bsnm{Osei}, \binits{S.}},
\oauthor{\bsnm{Pyysalo}, \binits{S.}},
\oauthor{\bsnm{Nagel}, \binits{S.}},
\oauthor{\bsnm{Bose}, \binits{S.}},
\oauthor{\bsnm{Muhammad}, \binits{S.H.}},
\oauthor{\bsnm{Sharma}, \binits{S.}},
\oauthor{\bsnm{Longpre}, \binits{S.}},
\oauthor{\bsnm{Nikpoor}, \binits{S.}},
\oauthor{\bsnm{Silberberg}, \binits{S.}},
\oauthor{\bsnm{Pai}, \binits{S.}},
\oauthor{\bsnm{Zink}, \binits{S.}},
\oauthor{\bsnm{Torrent}, \binits{T.T.}},
\oauthor{\bsnm{Schick}, \binits{T.}},
\oauthor{\bsnm{Thrush}, \binits{T.}},
\oauthor{\bsnm{Danchev}, \binits{V.}},
\oauthor{\bsnm{Nikoulina}, \binits{V.}},
\oauthor{\bsnm{Laippala}, \binits{V.}},
\oauthor{\bsnm{Lepercq}, \binits{V.}},
\oauthor{\bsnm{Prabhu}, \binits{V.}},
\oauthor{\bsnm{Alyafeai}, \binits{Z.}},
\oauthor{\bsnm{Talat}, \binits{Z.}},
\oauthor{\bsnm{Raja}, \binits{A.}},
\oauthor{\bsnm{Heinzerling}, \binits{B.}},
\oauthor{\bsnm{Si}, \binits{C.}},
\oauthor{\bsnm{Salesky}, \binits{E.}},
\oauthor{\bsnm{Mielke}, \binits{S.J.}},
\oauthor{\bsnm{Lee}, \binits{W.Y.}},
\oauthor{\bsnm{Sharma}, \binits{A.}},
\oauthor{\bsnm{Santilli}, \binits{A.}},
\oauthor{\bsnm{Chaffin}, \binits{A.}},
\oauthor{\bsnm{Stiegler}, \binits{A.}},
\oauthor{\bsnm{Datta}, \binits{D.}},
\oauthor{\bsnm{Szczechla}, \binits{E.}},
\oauthor{\bsnm{Chhablani}, \binits{G.}},
\oauthor{\bsnm{Wang}, \binits{H.}},
\oauthor{\bsnm{Pandey}, \binits{H.}},
\oauthor{\bsnm{Strobelt}, \binits{H.}},
\oauthor{\bsnm{Fries}, \binits{J.A.}},
\oauthor{\bsnm{Rozen}, \binits{J.}},
\oauthor{\bsnm{Gao}, \binits{L.}},
\oauthor{\bsnm{Sutawika}, \binits{L.}},
\oauthor{\bsnm{Bari}, \binits{M.S.}},
\oauthor{\bsnm{Al-shaibani}, \binits{M.S.}},
\oauthor{\bsnm{Manica}, \binits{M.}},
\oauthor{\bsnm{Nayak}, \binits{N.}},
\oauthor{\bsnm{Teehan}, \binits{R.}},
\oauthor{\bsnm{Albanie}, \binits{S.}},
\oauthor{\bsnm{Shen}, \binits{S.}},
\oauthor{\bsnm{Ben-David}, \binits{S.}},
\oauthor{\bsnm{Bach}, \binits{S.H.}},
\oauthor{\bsnm{Kim}, \binits{T.}},
\oauthor{\bsnm{Bers}, \binits{T.}},
\oauthor{\bsnm{Fevry}, \binits{T.}},
\oauthor{\bsnm{Neeraj}, \binits{T.}},
\oauthor{\bsnm{Thakker}, \binits{U.}},
\oauthor{\bsnm{Raunak}, \binits{V.}},
\oauthor{\bsnm{Tang}, \binits{X.}},
\oauthor{\bsnm{Yong}, \binits{Z.-X.}},
\oauthor{\bsnm{Sun}, \binits{Z.}},
\oauthor{\bsnm{Brody}, \binits{S.}},
\oauthor{\bsnm{Uri}, \binits{Y.}},
\oauthor{\bsnm{Tojarieh}, \binits{H.}},
\oauthor{\bsnm{Roberts}, \binits{A.}},
\oauthor{\bsnm{Chung}, \binits{H.W.}},
\oauthor{\bsnm{Tae}, \binits{J.}},
\oauthor{\bsnm{Phang}, \binits{J.}},
\oauthor{\bsnm{Press}, \binits{O.}},
\oauthor{\bsnm{Li}, \binits{C.}},
\oauthor{\bsnm{Narayanan}, \binits{D.}},
\oauthor{\bsnm{Bourfoune}, \binits{H.}},
\oauthor{\bsnm{Casper}, \binits{J.}},
\oauthor{\bsnm{Rasley}, \binits{J.}},
\oauthor{\bsnm{Ryabinin}, \binits{M.}},
\oauthor{\bsnm{Mishra}, \binits{M.}},
\oauthor{\bsnm{Zhang}, \binits{M.}},
\oauthor{\bsnm{Shoeybi}, \binits{M.}},
\oauthor{\bsnm{Peyrounette}, \binits{M.}},
\oauthor{\bsnm{Patry}, \binits{N.}},
\oauthor{\bsnm{Tazi}, \binits{N.}},
\oauthor{\bsnm{Sanseviero}, \binits{O.}},
\oauthor{\bsnm{Platen}, \binits{P.}},
\oauthor{\bsnm{Cornette}, \binits{P.}},
\oauthor{\bsnm{Lavallée}, \binits{P.F.}},
\oauthor{\bsnm{Lacroix}, \binits{R.}},
\oauthor{\bsnm{Rajbhandari}, \binits{S.}},
\oauthor{\bsnm{Gandhi}, \binits{S.}},
\oauthor{\bsnm{Smith}, \binits{S.}},
\oauthor{\bsnm{Requena}, \binits{S.}},
\oauthor{\bsnm{Patil}, \binits{S.}},
\oauthor{\bsnm{Dettmers}, \binits{T.}},
\oauthor{\bsnm{Baruwa}, \binits{A.}},
\oauthor{\bsnm{Singh}, \binits{A.}},
\oauthor{\bsnm{Cheveleva}, \binits{A.}},
\oauthor{\bsnm{Ligozat}, \binits{A.-L.}},
\oauthor{\bsnm{Subramonian}, \binits{A.}},
\oauthor{\bsnm{Névéol}, \binits{A.}},
\oauthor{\bsnm{Lovering}, \binits{C.}},
\oauthor{\bsnm{Garrette}, \binits{D.}},
\oauthor{\bsnm{Tunuguntla}, \binits{D.}},
\oauthor{\bsnm{Reiter}, \binits{E.}},
\oauthor{\bsnm{Taktasheva}, \binits{E.}},
\oauthor{\bsnm{Voloshina}, \binits{E.}},
\oauthor{\bsnm{Bogdanov}, \binits{E.}},
\oauthor{\bsnm{Winata}, \binits{G.I.}},
\oauthor{\bsnm{Schoelkopf}, \binits{H.}},
\oauthor{\bsnm{Kalo}, \binits{J.-C.}},
\oauthor{\bsnm{Novikova}, \binits{J.}},
\oauthor{\bsnm{Forde}, \binits{J.Z.}},
\oauthor{\bsnm{Clive}, \binits{J.}},
\oauthor{\bsnm{Kasai}, \binits{J.}},
\oauthor{\bsnm{Kawamura}, \binits{K.}},
\oauthor{\bsnm{Hazan}, \binits{L.}},
\oauthor{\bsnm{Carpuat}, \binits{M.}},
\oauthor{\bsnm{Clinciu}, \binits{M.}},
\oauthor{\bsnm{Kim}, \binits{N.}},
\oauthor{\bsnm{Cheng}, \binits{N.}},
\oauthor{\bsnm{Serikov}, \binits{O.}},
\oauthor{\bsnm{Antverg}, \binits{O.}},
\oauthor{\bsnm{Wal}, \binits{O.}},
\oauthor{\bsnm{Zhang}, \binits{R.}},
\oauthor{\bsnm{Zhang}, \binits{R.}},
\oauthor{\bsnm{Gehrmann}, \binits{S.}},
\oauthor{\bsnm{Pais}, \binits{S.}},
\oauthor{\bsnm{Shavrina}, \binits{T.}},
\oauthor{\bsnm{Scialom}, \binits{T.}},
\oauthor{\bsnm{Yun}, \binits{T.}},
\oauthor{\bsnm{Limisiewicz}, \binits{T.}},
\oauthor{\bsnm{Rieser}, \binits{V.}},
\oauthor{\bsnm{Protasov}, \binits{V.}},
\oauthor{\bsnm{Mikhailov}, \binits{V.}},
\oauthor{\bsnm{Pruksachatkun}, \binits{Y.}},
\oauthor{\bsnm{Belinkov}, \binits{Y.}},
\oauthor{\bsnm{Bamberger}, \binits{Z.}},
\oauthor{\bsnm{Kasner}, \binits{Z.}},
\oauthor{\bsnm{Rueda}, \binits{A.}},
\oauthor{\bsnm{Pestana}, \binits{A.}},
\oauthor{\bsnm{Feizpour}, \binits{A.}},
\oauthor{\bsnm{Khan}, \binits{A.}},
\oauthor{\bsnm{Faranak}, \binits{A.}},
\oauthor{\bsnm{Santos}, \binits{A.}},
\oauthor{\bsnm{Hevia}, \binits{A.}},
\oauthor{\bsnm{Unldreaj}, \binits{A.}},
\oauthor{\bsnm{Aghagol}, \binits{A.}},
\oauthor{\bsnm{Abdollahi}, \binits{A.}},
\oauthor{\bsnm{Tammour}, \binits{A.}},
\oauthor{\bsnm{HajiHosseini}, \binits{A.}},
\oauthor{\bsnm{Behroozi}, \binits{B.}},
\oauthor{\bsnm{Ajibade}, \binits{B.}},
\oauthor{\bsnm{Saxena}, \binits{B.}},
\oauthor{\bsnm{Ferrandis}, \binits{C.M.}},
\oauthor{\bsnm{Contractor}, \binits{D.}},
\oauthor{\bsnm{Lansky}, \binits{D.}},
\oauthor{\bsnm{David}, \binits{D.}},
\oauthor{\bsnm{Kiela}, \binits{D.}},
\oauthor{\bsnm{Nguyen}, \binits{D.A.}},
\oauthor{\bsnm{Tan}, \binits{E.}},
\oauthor{\bsnm{Baylor}, \binits{E.}},
\oauthor{\bsnm{Ozoani}, \binits{E.}},
\oauthor{\bsnm{Mirza}, \binits{F.}},
\oauthor{\bsnm{Ononiwu}, \binits{F.}},
\oauthor{\bsnm{Rezanejad}, \binits{H.}},
\oauthor{\bsnm{Jones}, \binits{H.}},
\oauthor{\bsnm{Bhattacharya}, \binits{I.}},
\oauthor{\bsnm{Solaiman}, \binits{I.}},
\oauthor{\bsnm{Sedenko}, \binits{I.}},
\oauthor{\bsnm{Nejadgholi}, \binits{I.}},
\oauthor{\bsnm{Passmore}, \binits{J.}},
\oauthor{\bsnm{Seltzer}, \binits{J.}},
\oauthor{\bsnm{Sanz}, \binits{J.B.}},
\oauthor{\bsnm{Fort}, \binits{K.}},
\oauthor{\bsnm{Dutra}, \binits{L.}},
\oauthor{\bsnm{Samagaio}, \binits{M.}},
\oauthor{\bsnm{Elbadri}, \binits{M.}},
\oauthor{\bsnm{Mieskes}, \binits{M.}},
\oauthor{\bsnm{Gerchick}, \binits{M.}},
\oauthor{\bsnm{Akinlolu}, \binits{M.}},
\oauthor{\bsnm{McKenna}, \binits{M.}},
\oauthor{\bsnm{Qiu}, \binits{M.}},
\oauthor{\bsnm{Ghauri}, \binits{M.}},
\oauthor{\bsnm{Burynok}, \binits{M.}},
\oauthor{\bsnm{Abrar}, \binits{N.}},
\oauthor{\bsnm{Rajani}, \binits{N.}},
\oauthor{\bsnm{Elkott}, \binits{N.}},
\oauthor{\bsnm{Fahmy}, \binits{N.}},
\oauthor{\bsnm{Samuel}, \binits{O.}},
\oauthor{\bsnm{An}, \binits{R.}},
\oauthor{\bsnm{Kromann}, \binits{R.}},
\oauthor{\bsnm{Hao}, \binits{R.}},
\oauthor{\bsnm{Alizadeh}, \binits{S.}},
\oauthor{\bsnm{Shubber}, \binits{S.}},
\oauthor{\bsnm{Wang}, \binits{S.}},
\oauthor{\bsnm{Roy}, \binits{S.}},
\oauthor{\bsnm{Viguier}, \binits{S.}},
\oauthor{\bsnm{Le}, \binits{T.}},
\oauthor{\bsnm{Oyebade}, \binits{T.}},
\oauthor{\bsnm{Le}, \binits{T.}},
\oauthor{\bsnm{Yang}, \binits{Y.}},
\oauthor{\bsnm{Nguyen}, \binits{Z.}},
\oauthor{\bsnm{Kashyap}, \binits{A.R.}},
\oauthor{\bsnm{Palasciano}, \binits{A.}},
\oauthor{\bsnm{Callahan}, \binits{A.}},
\oauthor{\bsnm{Shukla}, \binits{A.}},
\oauthor{\bsnm{Miranda-Escalada}, \binits{A.}},
\oauthor{\bsnm{Singh}, \binits{A.}},
\oauthor{\bsnm{Beilharz}, \binits{B.}},
\oauthor{\bsnm{Wang}, \binits{B.}},
\oauthor{\bsnm{Brito}, \binits{C.}},
\oauthor{\bsnm{Zhou}, \binits{C.}},
\oauthor{\bsnm{Jain}, \binits{C.}},
\oauthor{\bsnm{Xu}, \binits{C.}},
\oauthor{\bsnm{Fourrier}, \binits{C.}},
\oauthor{\bsnm{Periñán}, \binits{D.L.}},
\oauthor{\bsnm{Molano}, \binits{D.}},
\oauthor{\bsnm{Yu}, \binits{D.}},
\oauthor{\bsnm{Manjavacas}, \binits{E.}},
\oauthor{\bsnm{Barth}, \binits{F.}},
\oauthor{\bsnm{Fuhrimann}, \binits{F.}},
\oauthor{\bsnm{Altay}, \binits{G.}},
\oauthor{\bsnm{Bayrak}, \binits{G.}},
\oauthor{\bsnm{Burns}, \binits{G.}},
\oauthor{\bsnm{Vrabec}, \binits{H.U.}},
\oauthor{\bsnm{Bello}, \binits{I.}},
\oauthor{\bsnm{Dash}, \binits{I.}},
\oauthor{\bsnm{Kang}, \binits{J.}},
\oauthor{\bsnm{Giorgi}, \binits{J.}},
\oauthor{\bsnm{Golde}, \binits{J.}},
\oauthor{\bsnm{Posada}, \binits{J.D.}},
\oauthor{\bsnm{Sivaraman}, \binits{K.R.}},
\oauthor{\bsnm{Bulchandani}, \binits{L.}},
\oauthor{\bsnm{Liu}, \binits{L.}},
\oauthor{\bsnm{Shinzato}, \binits{L.}},
\oauthor{\bsnm{Bykhovetz}, \binits{M.H.}},
\oauthor{\bsnm{Takeuchi}, \binits{M.}},
\oauthor{\bsnm{Pàmies}, \binits{M.}},
\oauthor{\bsnm{Castillo}, \binits{M.A.}},
\oauthor{\bsnm{Nezhurina}, \binits{M.}},
\oauthor{\bsnm{Sänger}, \binits{M.}},
\oauthor{\bsnm{Samwald}, \binits{M.}},
\oauthor{\bsnm{Cullan}, \binits{M.}},
\oauthor{\bsnm{Weinberg}, \binits{M.}},
\oauthor{\bsnm{De~Wolf}, \binits{M.}},
\oauthor{\bsnm{Mihaljcic}, \binits{M.}},
\oauthor{\bsnm{Liu}, \binits{M.}},
\oauthor{\bsnm{Freidank}, \binits{M.}},
\oauthor{\bsnm{Kang}, \binits{M.}},
\oauthor{\bsnm{Seelam}, \binits{N.}},
\oauthor{\bsnm{Dahlberg}, \binits{N.}},
\oauthor{\bsnm{Broad}, \binits{N.M.}},
\oauthor{\bsnm{Muellner}, \binits{N.}},
\oauthor{\bsnm{Fung}, \binits{P.}},
\oauthor{\bsnm{Haller}, \binits{P.}},
\oauthor{\bsnm{Chandrasekhar}, \binits{R.}},
\oauthor{\bsnm{Eisenberg}, \binits{R.}},
\oauthor{\bsnm{Martin}, \binits{R.}},
\oauthor{\bsnm{Canalli}, \binits{R.}},
\oauthor{\bsnm{Su}, \binits{R.}},
\oauthor{\bsnm{Su}, \binits{R.}},
\oauthor{\bsnm{Cahyawijaya}, \binits{S.}},
\oauthor{\bsnm{Garda}, \binits{S.}},
\oauthor{\bsnm{Deshmukh}, \binits{S.S.}},
\oauthor{\bsnm{Mishra}, \binits{S.}},
\oauthor{\bsnm{Kiblawi}, \binits{S.}},
\oauthor{\bsnm{Ott}, \binits{S.}},
\oauthor{\bsnm{Sang-aroonsiri}, \binits{S.}},
\oauthor{\bsnm{Kumar}, \binits{S.}},
\oauthor{\bsnm{Schweter}, \binits{S.}},
\oauthor{\bsnm{Bharati}, \binits{S.}},
\oauthor{\bsnm{Laud}, \binits{T.}},
\oauthor{\bsnm{Gigant}, \binits{T.}},
\oauthor{\bsnm{Kainuma}, \binits{T.}},
\oauthor{\bsnm{Kusa}, \binits{W.}},
\oauthor{\bsnm{Labrak}, \binits{Y.}},
\oauthor{\bsnm{Bajaj}, \binits{Y.S.}},
\oauthor{\bsnm{Venkatraman}, \binits{Y.}},
\oauthor{\bsnm{Xu}, \binits{Y.}},
\oauthor{\bsnm{Xu}, \binits{Y.}},
\oauthor{\bsnm{Xu}, \binits{Y.}},
\oauthor{\bsnm{Tan}, \binits{Z.}},
\oauthor{\bsnm{Xie}, \binits{Z.}},
\oauthor{\bsnm{Ye}, \binits{Z.}},
\oauthor{\bsnm{Bras}, \binits{M.}},
\oauthor{\bsnm{Belkada}, \binits{Y.}},
\oauthor{\bsnm{Wolf}, \binits{T.}}:
{BLOOM}: {A} {176B}-{Parameter} {Open}-{Access} {Multilingual} {Language} {Model}.
arXiv.
arXiv:2211.05100 [cs]
(2022).
\doiurl{10.48550/arXiv.2211.05100} .
\url{http://arxiv.org/abs/2211.05100}
Accessed 2022-11-15
\end{botherref}
\endbibitem

\bibitem[\protect\citeauthoryear{Xue et~al.}{2021}]{xue_mt5_2021}
\begin{botherref}
\oauthor{\bsnm{Xue}, \binits{L.}},
\oauthor{\bsnm{Constant}, \binits{N.}},
\oauthor{\bsnm{Roberts}, \binits{A.}},
\oauthor{\bsnm{Kale}, \binits{M.}},
\oauthor{\bsnm{Al-Rfou}, \binits{R.}},
\oauthor{\bsnm{Siddhant}, \binits{A.}},
\oauthor{\bsnm{Barua}, \binits{A.}},
\oauthor{\bsnm{Raffel}, \binits{C.}}:
{mT5}: {A} massively multilingual pre-trained text-to-text transformer.
arXiv:2010.11934 [cs]
(2021).
arXiv: 2010.11934.
Accessed 2021-11-12
\end{botherref}
\endbibitem

\bibitem[\protect\citeauthoryear{Touvron et~al.}{2023}]{touvron_llama_2023-1}
\begin{botherref}
\oauthor{\bsnm{Touvron}, \binits{H.}},
\oauthor{\bsnm{Martin}, \binits{L.}},
\oauthor{\bsnm{Stone}, \binits{K.}},
\oauthor{\bsnm{Albert}, \binits{P.}},
\oauthor{\bsnm{Almahairi}, \binits{A.}},
\oauthor{\bsnm{Babaei}, \binits{Y.}},
\oauthor{\bsnm{Bashlykov}, \binits{N.}},
\oauthor{\bsnm{Batra}, \binits{S.}},
\oauthor{\bsnm{Bhargava}, \binits{P.}},
\oauthor{\bsnm{Bhosale}, \binits{S.}},
\oauthor{\bsnm{Bikel}, \binits{D.}},
\oauthor{\bsnm{Blecher}, \binits{L.}},
\oauthor{\bsnm{Ferrer}, \binits{C.C.}},
\oauthor{\bsnm{Chen}, \binits{M.}},
\oauthor{\bsnm{Cucurull}, \binits{G.}},
\oauthor{\bsnm{Esiobu}, \binits{D.}},
\oauthor{\bsnm{Fernandes}, \binits{J.}},
\oauthor{\bsnm{Fu}, \binits{J.}},
\oauthor{\bsnm{Fu}, \binits{W.}},
\oauthor{\bsnm{Fuller}, \binits{B.}},
\oauthor{\bsnm{Gao}, \binits{C.}},
\oauthor{\bsnm{Goswami}, \binits{V.}},
\oauthor{\bsnm{Goyal}, \binits{N.}},
\oauthor{\bsnm{Hartshorn}, \binits{A.}},
\oauthor{\bsnm{Hosseini}, \binits{S.}},
\oauthor{\bsnm{Hou}, \binits{R.}},
\oauthor{\bsnm{Inan}, \binits{H.}},
\oauthor{\bsnm{Kardas}, \binits{M.}},
\oauthor{\bsnm{Kerkez}, \binits{V.}},
\oauthor{\bsnm{Khabsa}, \binits{M.}},
\oauthor{\bsnm{Kloumann}, \binits{I.}},
\oauthor{\bsnm{Korenev}, \binits{A.}},
\oauthor{\bsnm{Koura}, \binits{P.S.}},
\oauthor{\bsnm{Lachaux}, \binits{M.-A.}},
\oauthor{\bsnm{Lavril}, \binits{T.}},
\oauthor{\bsnm{Lee}, \binits{J.}},
\oauthor{\bsnm{Liskovich}, \binits{D.}},
\oauthor{\bsnm{Lu}, \binits{Y.}},
\oauthor{\bsnm{Mao}, \binits{Y.}},
\oauthor{\bsnm{Martinet}, \binits{X.}},
\oauthor{\bsnm{Mihaylov}, \binits{T.}},
\oauthor{\bsnm{Mishra}, \binits{P.}},
\oauthor{\bsnm{Molybog}, \binits{I.}},
\oauthor{\bsnm{Nie}, \binits{Y.}},
\oauthor{\bsnm{Poulton}, \binits{A.}},
\oauthor{\bsnm{Reizenstein}, \binits{J.}},
\oauthor{\bsnm{Rungta}, \binits{R.}},
\oauthor{\bsnm{Saladi}, \binits{K.}},
\oauthor{\bsnm{Schelten}, \binits{A.}},
\oauthor{\bsnm{Silva}, \binits{R.}},
\oauthor{\bsnm{Smith}, \binits{E.M.}},
\oauthor{\bsnm{Subramanian}, \binits{R.}},
\oauthor{\bsnm{Tan}, \binits{X.E.}},
\oauthor{\bsnm{Tang}, \binits{B.}},
\oauthor{\bsnm{Taylor}, \binits{R.}},
\oauthor{\bsnm{Williams}, \binits{A.}},
\oauthor{\bsnm{Kuan}, \binits{J.X.}},
\oauthor{\bsnm{Xu}, \binits{P.}},
\oauthor{\bsnm{Yan}, \binits{Z.}},
\oauthor{\bsnm{Zarov}, \binits{I.}},
\oauthor{\bsnm{Zhang}, \binits{Y.}},
\oauthor{\bsnm{Fan}, \binits{A.}},
\oauthor{\bsnm{Kambadur}, \binits{M.}},
\oauthor{\bsnm{Narang}, \binits{S.}},
\oauthor{\bsnm{Rodriguez}, \binits{A.}},
\oauthor{\bsnm{Stojnic}, \binits{R.}},
\oauthor{\bsnm{Edunov}, \binits{S.}},
\oauthor{\bsnm{Scialom}, \binits{T.}}:
Llama 2: {Open} {Foundation} and {Fine}-{Tuned} {Chat} {Models}.
arXiv.
arXiv:2307.09288 [cs]
(2023).
\doiurl{10.48550/arXiv.2307.09288} .
\url{http://arxiv.org/abs/2307.09288}
Accessed 2023-08-06
\end{botherref}
\endbibitem

\bibitem[\protect\citeauthoryear{Conneau et~al.}{2020}]{conneau_unsupervised_2020}
\begin{botherref}
\oauthor{\bsnm{Conneau}, \binits{A.}},
\oauthor{\bsnm{Khandelwal}, \binits{K.}},
\oauthor{\bsnm{Goyal}, \binits{N.}},
\oauthor{\bsnm{Chaudhary}, \binits{V.}},
\oauthor{\bsnm{Wenzek}, \binits{G.}},
\oauthor{\bsnm{Guzmán}, \binits{F.}},
\oauthor{\bsnm{Grave}, \binits{E.}},
\oauthor{\bsnm{Ott}, \binits{M.}},
\oauthor{\bsnm{Zettlemoyer}, \binits{L.}},
\oauthor{\bsnm{Stoyanov}, \binits{V.}}:
Unsupervised {Cross}-lingual {Representation} {Learning} at {Scale}.
arXiv:1911.02116 [cs]
(2020).
arXiv: 1911.02116.
Accessed 2021-10-05
\end{botherref}
\endbibitem

\bibitem[\protect\citeauthoryear{Sanh et~al.}{2020}]{sanh_distilbert_2020}
\begin{botherref}
\oauthor{\bsnm{Sanh}, \binits{V.}},
\oauthor{\bsnm{Debut}, \binits{L.}},
\oauthor{\bsnm{Chaumond}, \binits{J.}},
\oauthor{\bsnm{Wolf}, \binits{T.}}:
{DistilBERT}, a distilled version of {BERT}: smaller, faster, cheaper and lighter.
arXiv:1910.01108 [cs]
(2020).
arXiv: 1910.01108.
Accessed 2020-08-25
\end{botherref}
\endbibitem

\bibitem[\protect\citeauthoryear{He et~al.}{2021a}]{he_deberta_2021}
\begin{botherref}
\oauthor{\bsnm{He}, \binits{P.}},
\oauthor{\bsnm{Liu}, \binits{X.}},
\oauthor{\bsnm{Gao}, \binits{J.}},
\oauthor{\bsnm{Chen}, \binits{W.}}:
{DeBERTa}: {Decoding}-enhanced {BERT} with {Disentangled} {Attention}.
arXiv:2006.03654 [cs]
(2021).
arXiv: 2006.03654.
Accessed 2022-05-16
\end{botherref}
\endbibitem

\bibitem[\protect\citeauthoryear{He et~al.}{2021b}]{he_debertav3_2021}
\begin{botherref}
\oauthor{\bsnm{He}, \binits{P.}},
\oauthor{\bsnm{Gao}, \binits{J.}},
\oauthor{\bsnm{Chen}, \binits{W.}}:
{DeBERTaV3}: {Improving} {DeBERTa} using {ELECTRA}-{Style} {Pre}-{Training} with {Gradient}-{Disentangled} {Embedding} {Sharing}.
arXiv:2111.09543 [cs]
(2021).
arXiv: 2111.09543.
Accessed 2022-05-20
\end{botherref}
\endbibitem

\bibitem[\protect\citeauthoryear{Pfeiffer et~al.}{2022}]{pfeiffer_lifting_2022}
\begin{bchapter}
\bauthor{\bsnm{Pfeiffer}, \binits{J.}},
\bauthor{\bsnm{Goyal}, \binits{N.}},
\bauthor{\bsnm{Lin}, \binits{X.}},
\bauthor{\bsnm{Li}, \binits{X.}},
\bauthor{\bsnm{Cross}, \binits{J.}},
\bauthor{\bsnm{Riedel}, \binits{S.}},
\bauthor{\bsnm{Artetxe}, \binits{M.}}:
\bctitle{Lifting the {Curse} of {Multilinguality} by {Pre}-training {Modular} {Transformers}}.
In: \bbtitle{Proceedings of the 2022 {Conference} of the {North} {American} {Chapter} of the {Association} for {Computational} {Linguistics}: {Human} {Language} {Technologies}},
pp. \bfpage{3479}--\blpage{3495}.
\bpublisher{Association for Computational Linguistics},
\blocation{Seattle, United States}
(\byear{2022}).
\doiurl{10.18653/v1/2022.naacl-main.255} .
\burl{https://aclanthology.org/2022.naacl-main.255}
\end{bchapter}
\endbibitem

\bibitem[\protect\citeauthoryear{Vamvas et~al.}{2023}]{vamvas2023swissbert}
\begin{botherref}
\oauthor{\bsnm{Vamvas}, \binits{J.}},
\oauthor{\bsnm{Gra{\"e}n}, \binits{J.}},
\oauthor{\bsnm{Sennrich}, \binits{R.}}:
Swissbert: The multilingual language model for switzerland.
arXiv e-prints,
2303
(2023)
\end{botherref}
\endbibitem

\bibitem[\protect\citeauthoryear{OpenAI}{2023}]{openai_gpt-4_2023}
\begin{botherref}
\oauthor{\bsnm{OpenAI}}:
{GPT}-4 {Technical} {Report}.
arXiv.
arXiv:2303.08774 [cs]
(2023).
\doiurl{10.48550/arXiv.2303.08774} .
\url{http://arxiv.org/abs/2303.08774}
Accessed 2023-05-25
\end{botherref}
\endbibitem

\bibitem[\protect\citeauthoryear{Niklaus et~al.}{2023}]{niklaus_multilegalpile_2023}
\begin{botherref}
\oauthor{\bsnm{Niklaus}, \binits{J.}},
\oauthor{\bsnm{Matoshi}, \binits{V.}},
\oauthor{\bsnm{Stürmer}, \binits{M.}},
\oauthor{\bsnm{Chalkidis}, \binits{I.}},
\oauthor{\bsnm{Ho}, \binits{D.E.}}:
{MultiLegalPile}: {A} {689GB} {Multilingual} {Legal} {Corpus}.
arXiv.
arXiv:2306.02069 [cs]
(2023).
\url{http://arxiv.org/abs/2306.02069}
Accessed 2023-06-14
\end{botherref}
\endbibitem

\bibitem[\protect\citeauthoryear{Conneau and Lample}{2019}]{conneau_cross-lingual_2019}
\begin{bchapter}
\bauthor{\bsnm{Conneau}, \binits{A.}},
\bauthor{\bsnm{Lample}, \binits{G.}}:
\bctitle{Cross-lingual {Language} {Model} {Pretraining}}.
In: \bbtitle{Advances in {Neural} {Information} {Processing} {Systems}},
vol. \bseriesno{32}.
\bpublisher{Curran Associates, Inc.}, \blocation{???}
(\byear{2019}).
\burl{https://proceedings.neurips.cc/paper/2019/hash/c04c19c2c2474dbf5f7ac4372c5b9af1-Abstract.html}
Accessed 2021-10-05
\end{bchapter}
\endbibitem

\bibitem[\protect\citeauthoryear{Wei et~al.}{2021}]{wei-etal-2022}
\begin{botherref}
\oauthor{\bsnm{Wei}, \binits{J.}},
\oauthor{\bsnm{Bosma}, \binits{M.}},
\oauthor{\bsnm{Zhao}, \binits{V.Y.}},
\oauthor{\bsnm{Guu}, \binits{K.}},
\oauthor{\bsnm{Yu}, \binits{A.W.}},
\oauthor{\bsnm{Lester}, \binits{B.}},
\oauthor{\bsnm{Du}, \binits{N.}},
\oauthor{\bsnm{Dai}, \binits{A.M.}},
\oauthor{\bsnm{Le}, \binits{Q.V.}}:
Finetuned language models are zero-shot learners.
CoRR
\textbf{abs/2109.01652}
(2021)
{\href{https://arxiv.org/abs/2109.01652}{{2109.01652}}}
\end{botherref}
\endbibitem

\bibitem[\protect\citeauthoryear{Ouyang et~al.}{2022}]{instructgpt}
\begin{botherref}
\oauthor{\bsnm{Ouyang}, \binits{L.}},
\oauthor{\bsnm{Wu}, \binits{J.}},
\oauthor{\bsnm{Jiang}, \binits{X.}},
\oauthor{\bsnm{Almeida}, \binits{D.}},
\oauthor{\bsnm{Wainwright}, \binits{C.L.}},
\oauthor{\bsnm{Mishkin}, \binits{P.}},
\oauthor{\bsnm{Zhang}, \binits{C.}},
\oauthor{\bsnm{Agarwal}, \binits{S.}},
\oauthor{\bsnm{Slama}, \binits{K.}},
\oauthor{\bsnm{Ray}, \binits{A.}},
\oauthor{\bsnm{Schulman}, \binits{J.}},
\oauthor{\bsnm{Hilton}, \binits{J.}},
\oauthor{\bsnm{Kelton}, \binits{F.}},
\oauthor{\bsnm{Miller}, \binits{L.}},
\oauthor{\bsnm{Simens}, \binits{M.}},
\oauthor{\bsnm{Askell}, \binits{A.}},
\oauthor{\bsnm{Welinder}, \binits{P.}},
\oauthor{\bsnm{Christiano}, \binits{P.}},
\oauthor{\bsnm{Leike}, \binits{J.}},
\oauthor{\bsnm{Lowe}, \binits{R.}}:
Training language models to follow instructions with human feedback.
arXiv
(2022).
\doiurl{10.48550/ARXIV.2203.02155} .
\url{https://arxiv.org/abs/2203.02155}
\end{botherref}
\endbibitem

\bibitem[\protect\citeauthoryear{Pfeiffer et~al.}{2021}]{pfeiffer_unks_2021}
\begin{bchapter}
\bauthor{\bsnm{Pfeiffer}, \binits{J.}},
\bauthor{\bsnm{Vulić}, \binits{I.}},
\bauthor{\bsnm{Gurevych}, \binits{I.}},
\bauthor{\bsnm{Ruder}, \binits{S.}}:
\bctitle{{UNKs} {Everywhere}: {Adapting} {Multilingual} {Language} {Models} to {New} {Scripts}}.
In: \bbtitle{Proceedings of the 2021 {Conference} on {Empirical} {Methods} in {Natural} {Language} {Processing}},
pp. \bfpage{10186}--\blpage{10203}.
\bpublisher{Association for Computational Linguistics},
\blocation{Online and Punta Cana, Dominican Republic}
(\byear{2021}).
\doiurl{10.18653/v1/2021.emnlp-main.800} .
\burl{https://aclanthology.org/2021.emnlp-main.800}
\end{bchapter}
\endbibitem

\bibitem[\protect\citeauthoryear{Wettig et~al.}{2023}]{wettig2022should}
\begin{bchapter}
\bauthor{\bsnm{Wettig}, \binits{A.}},
\bauthor{\bsnm{Gao}, \binits{T.}},
\bauthor{\bsnm{Zhong}, \binits{Z.}},
\bauthor{\bsnm{Chen}, \binits{D.}}:
\bctitle{Should you mask 15{\%} in masked language modeling?}
In: \bbtitle{Proceedings of the 17th Conference of the European Chapter of the Association for Computational Linguistics},
pp. \bfpage{2985}--\blpage{3000}.
\bpublisher{Association for Computational Linguistics},
\blocation{Dubrovnik, Croatia}
(\byear{2023}).
\burl{https://aclanthology.org/2023.eacl-main.217}
\end{bchapter}
\endbibitem

\bibitem[\protect\citeauthoryear{Raffel et~al.}{2020}]{raffel_exploring_2020}
\begin{barticle}
\bauthor{\bsnm{Raffel}, \binits{C.}},
\bauthor{\bsnm{Shazeer}, \binits{N.}},
\bauthor{\bsnm{Roberts}, \binits{A.}},
\bauthor{\bsnm{Lee}, \binits{K.}},
\bauthor{\bsnm{Narang}, \binits{S.}},
\bauthor{\bsnm{Matena}, \binits{M.}},
\bauthor{\bsnm{Zhou}, \binits{Y.}},
\bauthor{\bsnm{Li}, \binits{W.}},
\bauthor{\bsnm{Liu}, \binits{P.J.}}:
\batitle{Exploring the {Limits} of {Transfer} {Learning} with a {Unified} {Text}-to-{Text} {Transformer}}.
\bjtitle{Journal of Machine Learning Research}
\bvolume{21}(\bissue{140}),
\bfpage{1}--\blpage{67}
(\byear{2020}).
Accessed 2023-05-17
\end{barticle}
\endbibitem

\bibitem[\protect\citeauthoryear{Beltagy et~al.}{2020}]{beltagy_longformer_2020}
\begin{botherref}
\oauthor{\bsnm{Beltagy}, \binits{I.}},
\oauthor{\bsnm{Peters}, \binits{M.E.}},
\oauthor{\bsnm{Cohan}, \binits{A.}}:
Longformer: {The} {Long}-{Document} {Transformer}.
arXiv:2004.05150 [cs]
(2020).
arXiv: 2004.05150.
Accessed 2021-03-05
\end{botherref}
\endbibitem

\bibitem[\protect\citeauthoryear{Zhang et~al.}{2020}]{zhang_bertscore_2020}
\begin{botherref}
\oauthor{\bsnm{Zhang}, \binits{T.}},
\oauthor{\bsnm{Kishore}, \binits{V.}},
\oauthor{\bsnm{Wu}, \binits{F.}},
\oauthor{\bsnm{Weinberger}, \binits{K.Q.}},
\oauthor{\bsnm{Artzi}, \binits{Y.}}:
{BERTScore}: {Evaluating} {Text} {Generation} with {BERT}.
arXiv:1904.09675 [cs]
(2020).
arXiv: 1904.09675.
Accessed 2022-03-08
\end{botherref}
\endbibitem

\bibitem[\protect\citeauthoryear{Papineni et~al.}{2002}]{BLEU}
\begin{bchapter}
\bauthor{\bsnm{Papineni}, \binits{K.}},
\bauthor{\bsnm{Roukos}, \binits{S.}},
\bauthor{\bsnm{Ward}, \binits{T.}},
\bauthor{\bsnm{Zhu}, \binits{W.-J.}}:
\bctitle{Bleu: A method for automatic evaluation of machine translation}.
In: \bbtitle{Proceedings of the 40th Annual Meeting on Association for Computational Linguistics}.
\bsertitle{ACL '02},
pp. \bfpage{311}--\blpage{318}.
\bpublisher{Association for Computational Linguistics},
\blocation{USA}
(\byear{2002}).
\doiurl{10.3115/1073083.1073135} .
\burl{https://doi.org/10.3115/1073083.1073135}
\end{bchapter}
\endbibitem

\bibitem[\protect\citeauthoryear{Banerjee and Lavie}{2005}]{banerjee_meteor_2005}
\begin{bchapter}
\bauthor{\bsnm{Banerjee}, \binits{S.}},
\bauthor{\bsnm{Lavie}, \binits{A.}}:
\bctitle{{METEOR}: {An} {Automatic} {Metric} for {MT} {Evaluation} with {Improved} {Correlation} with {Human} {Judgments}}.
In: \bbtitle{Proceedings of the {ACL} {Workshop} on {Intrinsic} and {Extrinsic} {Evaluation} {Measures} for {Machine} {Translation} And/or {Summarization}},
pp. \bfpage{65}--\blpage{72}.
\bpublisher{Association for Computational Linguistics},
\blocation{Ann Arbor, Michigan}
(\byear{2005}).
\burl{https://aclanthology.org/W05-0909}
\end{bchapter}
\endbibitem

\bibitem[\protect\citeauthoryear{Lin}{2004}]{lin_rouge_2004}
\begin{bchapter}
\bauthor{\bsnm{Lin}, \binits{C.-Y.}}:
\bctitle{{ROUGE}: {A} {Package} for {Automatic} {Evaluation} of {Summaries}}.
In: \bbtitle{Text {Summarization} {Branches} {Out}},
pp. \bfpage{74}--\blpage{81}.
\bpublisher{Association for Computational Linguistics},
\blocation{Barcelona, Spain}
(\byear{2004}).
\burl{https://aclanthology.org/W04-1013}
\end{bchapter}
\endbibitem

\bibitem[\protect\citeauthoryear{Chalkidis}{2023}]{Chalkidis2023ChatGPTMP}
\begin{botherref}
\oauthor{\bsnm{Chalkidis}, \binits{I.}}:
{ChatGPT may Pass the Bar Exam soon, but has a Long Way to Go for the LexGLUE benchmark}.
ArXiv
\textbf{abs/2304.1}
(2023)
\end{botherref}
\endbibitem

\bibitem[\protect\citeauthoryear{Trautmann}{2023}]{Trautmann2023}
\begin{botherref}
\oauthor{\bsnm{Trautmann}, \binits{D.}}:
Large Language Model Prompt Chaining for Long Legal Document Classification
(2023).
\url{http://arxiv.org/abs/2308.04138}
\end{botherref}
\endbibitem

\bibitem[\protect\citeauthoryear{Robertson and Zaragoza}{2009}]{INR-019}
\begin{barticle}
\bauthor{\bsnm{Robertson}, \binits{S.}},
\bauthor{\bsnm{Zaragoza}, \binits{H.}}:
\batitle{The probabilistic relevance framework: Bm25 and beyond}.
\bjtitle{Foundations and Trends® in Information Retrieval}
\bvolume{3}(\bissue{4}),
\bfpage{333}--\blpage{389}
(\byear{2009})
\doiurl{10.1561/1500000019}
\end{barticle}
\endbibitem

\bibitem[\protect\citeauthoryear{Zhan et~al.}{2021}]{10.1145/3404835.3462880}
\begin{bchapter}
\bauthor{\bsnm{Zhan}, \binits{J.}},
\bauthor{\bsnm{Mao}, \binits{J.}},
\bauthor{\bsnm{Liu}, \binits{Y.}},
\bauthor{\bsnm{Guo}, \binits{J.}},
\bauthor{\bsnm{Zhang}, \binits{M.}},
\bauthor{\bsnm{Ma}, \binits{S.}}:
\bctitle{Optimizing dense retrieval model training with hard negatives}.
In: \bbtitle{Proceedings of the 44th International ACM SIGIR Conference on Research and Development in Information Retrieval}.
\bsertitle{SIGIR '21},
pp. \bfpage{1503}--\blpage{1512}.
\bpublisher{Association for Computing Machinery},
\blocation{New York, NY, USA}
(\byear{2021}).
\doiurl{10.1145/3404835.3462880} .
\burl{https://doi.org/10.1145/3404835.3462880}
\end{bchapter}
\endbibitem

\bibitem[\protect\citeauthoryear{Wang et~al.}{2020}]{10.5555/3495724.3496209}
\begin{bchapter}
\bauthor{\bsnm{Wang}, \binits{W.}},
\bauthor{\bsnm{Wei}, \binits{F.}},
\bauthor{\bsnm{Dong}, \binits{L.}},
\bauthor{\bsnm{Bao}, \binits{H.}},
\bauthor{\bsnm{Yang}, \binits{N.}},
\bauthor{\bsnm{Zhou}, \binits{M.}}:
\bctitle{Minilm: Deep self-attention distillation for task-agnostic compression of pre-trained transformers}.
In: \bbtitle{Proceedings of the 34th International Conference on Neural Information Processing Systems}.
\bsertitle{NIPS'20}.
\bpublisher{Curran Associates Inc.},
\blocation{Red Hook, NY, USA}
(\byear{2020})
\end{bchapter}
\endbibitem

\bibitem[\protect\citeauthoryear{Wang et~al.}{2013}]{Wang2013ATA}
\begin{bchapter}
\bauthor{\bsnm{Wang}, \binits{Y.}},
\bauthor{\bsnm{Wang}, \binits{L.}},
\bauthor{\bsnm{Li}, \binits{Y.}},
\bauthor{\bsnm{He}, \binits{D.}},
\bauthor{\bsnm{Liu}, \binits{T.-Y.}}:
\bctitle{A theoretical analysis of ndcg type ranking measures}.
In: \beditor{\bsnm{Shalev-Shwartz}, \binits{S.}},
\beditor{\bsnm{Steinwart}, \binits{I.}} (eds.)
\bbtitle{Proceedings of the 26th Annual Conference on Learning Theory}.
\bsertitle{Proceedings of Machine Learning Research},
vol. \bseriesno{30},
pp. \bfpage{25}--\blpage{54}.
\bpublisher{PMLR},
\blocation{Princeton, NJ, USA}
(\byear{2013}).
\burl{https://proceedings.mlr.press/v30/Wang13.html}
\end{bchapter}
\endbibitem

\bibitem[\protect\citeauthoryear{Chalkidis et~al.}{2021}]{chalkidis_regulatory_2021}
\begin{bchapter}
\bauthor{\bsnm{Chalkidis}, \binits{I.}},
\bauthor{\bsnm{Fergadiotis}, \binits{M.}},
\bauthor{\bsnm{Manginas}, \binits{N.}},
\bauthor{\bsnm{Katakalou}, \binits{E.}},
\bauthor{\bsnm{Malakasiotis}, \binits{P.}}:
\bctitle{Regulatory {Compliance} through {Doc2Doc} {Information} {Retrieval}: {A} case study in {EU}/{UK} legislation where text similarity has limitations}.
In: \bbtitle{Proceedings of the 16th {Conference} of the {European} {Chapter} of the {Association} for {Computational} {Linguistics}: {Main} {Volume}},
pp. \bfpage{3498}--\blpage{3511}.
\bpublisher{Association for Computational Linguistics},
\blocation{Online}
(\byear{2021}).
\doiurl{10.18653/v1/2021.eacl-main.305} .
\burl{https://aclanthology.org/2021.eacl-main.305}
Accessed 2023-04-24
\end{bchapter}
\endbibitem

\bibitem[\protect\citeauthoryear{Henderson et~al.}{2017}]{Henderson2017EfficientNL}
\begin{botherref}
\oauthor{\bsnm{Henderson}, \binits{M.}},
\oauthor{\bsnm{Al-Rfou}, \binits{R.}},
\oauthor{\bsnm{Strope}, \binits{B.}},
\oauthor{\bsnm{Sung}, \binits{Y.-H.}},
\oauthor{\bsnm{Luk{\'a}cs}, \binits{L.}},
\oauthor{\bsnm{Guo}, \binits{R.}},
\oauthor{\bsnm{Kumar}, \binits{S.}},
\oauthor{\bsnm{Miklos}, \binits{B.}},
\oauthor{\bsnm{Kurzweil}, \binits{R.}}:
Efficient natural language response suggestion for smart reply.
ArXiv
\textbf{abs/1705.00652}
(2017)
\end{botherref}
\endbibitem

\bibitem[\protect\citeauthoryear{Schluter}{2017}]{schluter-2017-limits}
\begin{bchapter}
\bauthor{\bsnm{Schluter}, \binits{N.}}:
\bctitle{The limits of automatic summarisation according to {ROUGE}}.
In: \beditor{\bsnm{Lapata}, \binits{M.}},
\beditor{\bsnm{Blunsom}, \binits{P.}},
\beditor{\bsnm{Koller}, \binits{A.}} (eds.)
\bbtitle{Proceedings of the 15th Conference of the {E}uropean Chapter of the Association for Computational Linguistics: Volume 2, Short Papers},
pp. \bfpage{41}--\blpage{45}.
\bpublisher{Association for Computational Linguistics},
\blocation{Valencia, Spain}
(\byear{2017}).
\burl{https://aclanthology.org/E17-2007}
\end{bchapter}
\endbibitem

\bibitem[\protect\citeauthoryear{Lewis et~al.}{2021}]{lewis_retrieval-augmented_2021}
\begin{botherref}
\oauthor{\bsnm{Lewis}, \binits{P.}},
\oauthor{\bsnm{Perez}, \binits{E.}},
\oauthor{\bsnm{Piktus}, \binits{A.}},
\oauthor{\bsnm{Petroni}, \binits{F.}},
\oauthor{\bsnm{Karpukhin}, \binits{V.}},
\oauthor{\bsnm{Goyal}, \binits{N.}},
\oauthor{\bsnm{Küttler}, \binits{H.}},
\oauthor{\bsnm{Lewis}, \binits{M.}},
\oauthor{\bsnm{Yih}, \binits{W.-t.}},
\oauthor{\bsnm{Rocktäschel}, \binits{T.}},
\oauthor{\bsnm{Riedel}, \binits{S.}},
\oauthor{\bsnm{Kiela}, \binits{D.}}:
Retrieval-{Augmented} {Generation} for {Knowledge}-{Intensive} {NLP} {Tasks}.
arXiv.
arXiv:2005.11401 [cs]
(2021).
\doiurl{10.48550/arXiv.2005.11401} .
\url{http://arxiv.org/abs/2005.11401}
Accessed 2023-07-31
\end{botherref}
\endbibitem

\bibitem[\protect\citeauthoryear{Schick et~al.}{2023}]{schick_toolformer_2023}
\begin{botherref}
\oauthor{\bsnm{Schick}, \binits{T.}},
\oauthor{\bsnm{Dwivedi-Yu}, \binits{J.}},
\oauthor{\bsnm{Dessì}, \binits{R.}},
\oauthor{\bsnm{Raileanu}, \binits{R.}},
\oauthor{\bsnm{Lomeli}, \binits{M.}},
\oauthor{\bsnm{Zettlemoyer}, \binits{L.}},
\oauthor{\bsnm{Cancedda}, \binits{N.}},
\oauthor{\bsnm{Scialom}, \binits{T.}}:
Toolformer: {Language} {Models} {Can} {Teach} {Themselves} to {Use} {Tools}.
arXiv.
arXiv:2302.04761 [cs]
(2023).
\doiurl{10.48550/arXiv.2302.04761} .
\url{http://arxiv.org/abs/2302.04761}
Accessed 2023-02-13
\end{botherref}
\endbibitem

\bibitem[\protect\citeauthoryear{T.Y.S.S. et~al.}{2024}]{tyss_towards_2024}
\begin{botherref}
\oauthor{\bsnm{T.Y.S.S.}, \binits{S.}},
\oauthor{\bsnm{Baumgartner}, \binits{N.}},
\oauthor{\bsnm{Stürmer}, \binits{M.}},
\oauthor{\bsnm{Grabmair}, \binits{M.}},
\oauthor{\bsnm{Niklaus}, \binits{J.}}:
Towards {Explainability} and {Fairness} in {Swiss} {Judgement} {Prediction}: {Benchmarking} on a {Multilingual} {Dataset}.
arXiv.
arXiv:2402.17013 [cs]
(2024).
\url{http://arxiv.org/abs/2402.17013}
Accessed 2024-03-01
\end{botherref}
\endbibitem

\bibitem[\protect\citeauthoryear{Chalkidis et~al.}{2022}]{chalkidis_fairlex_2022}
\begin{botherref}
\oauthor{\bsnm{Chalkidis}, \binits{I.}},
\oauthor{\bsnm{Pasini}, \binits{T.}},
\oauthor{\bsnm{Zhang}, \binits{S.}},
\oauthor{\bsnm{Tomada}, \binits{L.}},
\oauthor{\bsnm{Schwemer}, \binits{S.F.}},
\oauthor{\bsnm{Søgaard}, \binits{A.}}:
{FairLex}: {A} {Multilingual} {Benchmark} for {Evaluating} {Fairness} in {Legal} {Text} {Processing}.
arXiv:2203.07228 [cs]
(2022).
arXiv: 2203.07228.
Accessed 2022-04-14
\end{botherref}
\endbibitem

\bibitem[\protect\citeauthoryear{Chen et~al.}{2019}]{chen_charge-based_2019}
\begin{bchapter}
\bauthor{\bsnm{Chen}, \binits{H.}},
\bauthor{\bsnm{Cai}, \binits{D.}},
\bauthor{\bsnm{Dai}, \binits{W.}},
\bauthor{\bsnm{Dai}, \binits{Z.}},
\bauthor{\bsnm{Ding}, \binits{Y.}}:
\bctitle{Charge-{Based} {Prison} {Term} {Prediction} with {Deep} {Gating} {Network}}.
In: \bbtitle{Proceedings of the 2019 {Conference} on {Empirical} {Methods} in {Natural} {Language} {Processing} and the 9th {International} {Joint} {Conference} on {Natural} {Language} {Processing} ({EMNLP}-{IJCNLP})},
pp. \bfpage{6362}--\blpage{6367}.
\bpublisher{Association for Computational Linguistics},
\blocation{Hong Kong, China}
(\byear{2019}).
\doiurl{10.18653/v1/D19-1667} .
\burl{https://aclanthology.org/D19-1667}
Accessed 2022-07-14
\end{bchapter}
\endbibitem

\bibitem[\protect\citeauthoryear{Leins et~al.}{2020}]{leins_give_2020}
\begin{bchapter}
\bauthor{\bsnm{Leins}, \binits{K.}},
\bauthor{\bsnm{Lau}, \binits{J.H.}},
\bauthor{\bsnm{Baldwin}, \binits{T.}}:
\bctitle{Give {Me} {Convenience} and {Give} {Her} {Death}: {Who} {Should} {Decide} {What} {Uses} of {NLP} are {Appropriate}, and on {What} {Basis}?}
In: \bbtitle{Proceedings of the 58th {Annual} {Meeting} of the {Association} for {Computational} {Linguistics}},
pp. \bfpage{2908}--\blpage{2913}.
\bpublisher{Association for Computational Linguistics},
\blocation{Online}
(\byear{2020}).
\doiurl{10.18653/v1/2020.acl-main.261} .
\burl{https://aclanthology.org/2020.acl-main.261}
Accessed 2021-08-23
\end{bchapter}
\endbibitem

\bibitem[\protect\citeauthoryear{Bender and Koller}{2020}]{bender_climbing_2020}
\begin{bchapter}
\bauthor{\bsnm{Bender}, \binits{E.M.}},
\bauthor{\bsnm{Koller}, \binits{A.}}:
\bctitle{Climbing towards {NLU}: {On} {Meaning}, {Form}, and {Understanding} in the {Age} of {Data}}.
In: \bbtitle{Proceedings of the 58th {Annual} {Meeting} of the {Association} for {Computational} {Linguistics}},
pp. \bfpage{5185}--\blpage{5198}.
\bpublisher{Association for Computational Linguistics},
\blocation{Online}
(\byear{2020}).
\doiurl{10.18653/v1/2020.acl-main.463} .
\burl{https://aclanthology.org/2020.acl-main.463}
\end{bchapter}
\endbibitem

\bibitem[\protect\citeauthoryear{Lee et~al.}{2019}]{leeBioBERTPretrainedBiomedical2020}
\begin{barticle}
\bauthor{\bsnm{Lee}, \binits{J.}},
\bauthor{\bsnm{Yoon}, \binits{W.}},
\bauthor{\bsnm{Kim}, \binits{S.}},
\bauthor{\bsnm{Kim}, \binits{D.}},
\bauthor{\bsnm{Kim}, \binits{S.}},
\bauthor{\bsnm{So}, \binits{C.H.}},
\bauthor{\bsnm{Kang}, \binits{J.}}:
\batitle{{BioBERT: a pre-trained biomedical language representation model for biomedical text mining}}.
\bjtitle{Bioinformatics}
\bvolume{36}(\bissue{4}),
\bfpage{1234}--\blpage{1240}
(\byear{2019})
\doiurl{10.1093/bioinformatics/btz682}
{\href{https://arxiv.org/abs/https://academic.oup.com/bioinformatics/article-pdf/36/4/1234/48983216/bioinformatics\_36\_4\_1234.pdf}{{https://academic.oup.com/bioinformatics/article-pdf/36/4/1234/48983216/bioinformatics\_36\_4\_1234.pdf}}}
\end{barticle}
\endbibitem

\bibitem[\protect\citeauthoryear{Naseem et~al.}{2022}]{naseem2022benchmarking}
\begin{barticle}
\bauthor{\bsnm{Naseem}, \binits{U.}},
\bauthor{\bsnm{Dunn}, \binits{A.G.}},
\bauthor{\bsnm{Khushi}, \binits{M.}},
\bauthor{\bsnm{Kim}, \binits{J.}}:
\batitle{Benchmarking for biomedical natural language processing tasks with a domain specific albert}.
\bjtitle{BMC bioinformatics}
\bvolume{23}(\bissue{1}),
\bfpage{1}--\blpage{15}
(\byear{2022})
\end{barticle}
\endbibitem

\bibitem[\protect\citeauthoryear{Lan et~al.}{2020}]{lan_albert_2020}
\begin{botherref}
\oauthor{\bsnm{Lan}, \binits{Z.}},
\oauthor{\bsnm{Chen}, \binits{M.}},
\oauthor{\bsnm{Goodman}, \binits{S.}},
\oauthor{\bsnm{Gimpel}, \binits{K.}},
\oauthor{\bsnm{Sharma}, \binits{P.}},
\oauthor{\bsnm{Soricut}, \binits{R.}}:
{ALBERT}: {A} {Lite} {BERT} for {Self}-supervised {Learning} of {Language} {Representations}.
arXiv:1909.11942 [cs]
(2020).
arXiv: 1909.11942.
Accessed 2020-10-29
\end{botherref}
\endbibitem

\bibitem[\protect\citeauthoryear{Johnson et~al.}{2016}]{johnson_mimic-iii_2016}
\begin{barticle}
\bauthor{\bsnm{Johnson}, \binits{A.E.W.}},
\bauthor{\bsnm{Pollard}, \binits{T.J.}},
\bauthor{\bsnm{Shen}, \binits{L.}},
\bauthor{\bsnm{Lehman}, \binits{L.-W.H.}},
\bauthor{\bsnm{Feng}, \binits{M.}},
\bauthor{\bsnm{Ghassemi}, \binits{M.}},
\bauthor{\bsnm{Moody}, \binits{B.}},
\bauthor{\bsnm{Szolovits}, \binits{P.}},
\bauthor{\bsnm{Celi}, \binits{L.A.}},
\bauthor{\bsnm{Mark}, \binits{R.G.}}:
\batitle{{MIMIC}-{III}, a freely accessible critical care database}.
\bjtitle{Scientific data}
\bvolume{3},
\bfpage{160035}
(\byear{2016})
\doiurl{10.1038/sdata.2016.35}
\end{barticle}
\endbibitem

\bibitem[\protect\citeauthoryear{Gu et~al.}{2021}]{Gu2021BLURBbenchmark}
\begin{botherref}
\oauthor{\bsnm{Gu}, \binits{Y.}},
\oauthor{\bsnm{Tinn}, \binits{R.}},
\oauthor{\bsnm{Cheng}, \binits{H.}},
\oauthor{\bsnm{Lucas}, \binits{M.}},
\oauthor{\bsnm{Usuyama}, \binits{N.}},
\oauthor{\bsnm{Liu}, \binits{X.}},
\oauthor{\bsnm{Naumann}, \binits{T.}},
\oauthor{\bsnm{Gao}, \binits{J.}},
\oauthor{\bsnm{Poon}, \binits{H.}}:
Domain-specific language model pretraining for biomedical natural language processing.
ACM Trans. Comput. Healthcare
\textbf{3}(1)
(2021)
\doiurl{10.1145/3458754}
\end{botherref}
\endbibitem

\bibitem[\protect\citeauthoryear{Peng et~al.}{2019}]{pengBLUE2019}
\begin{botherref}
\oauthor{\bsnm{Peng}, \binits{Y.}},
\oauthor{\bsnm{Yan}, \binits{S.}},
\oauthor{\bsnm{Lu}, \binits{Z.}}:
Transfer Learning in Biomedical Natural Language Processing: An Evaluation of BERT and ELMo on Ten Benchmarking Datasets
(2019).
\doiurl{10.48550/arXiv.1906.05474} .
\url{http://arxiv.org/abs/1906.05474}
Accessed 2023-05-27
\end{botherref}
\endbibitem

\bibitem[\protect\citeauthoryear{Yang et~al.}{2020}]{yang-FinBERT-2020}
\begin{botherref}
\oauthor{\bsnm{Yang}, \binits{Y.}},
\oauthor{\bsnm{UY}, \binits{M.C.S.}},
\oauthor{\bsnm{Huang}, \binits{A.}}:
{FinBERT}: A Pretrained Language Model for Financial Communications
(2020).
\doiurl{10.48550/arXiv.2006.08097} .
\url{http://arxiv.org/abs/2006.08097}
Accessed 2023-05-27
\end{botherref}
\endbibitem

\bibitem[\protect\citeauthoryear{Malo et~al.}{2014}]{maloGoodDebtBad2014}
\begin{barticle}
\bauthor{\bsnm{Malo}, \binits{P.}},
\bauthor{\bsnm{Sinha}, \binits{A.}},
\bauthor{\bsnm{Korhonen}, \binits{P.}},
\bauthor{\bsnm{Wallenius}, \binits{J.}},
\bauthor{\bsnm{Takala}, \binits{P.}}:
\batitle{Good debt or bad debt: Detecting semantic orientations in economic texts}.
\bjtitle{J. Assoc. Inf. Sci. Technol.}
\bvolume{65}(\bissue{4}),
\bfpage{782}--\blpage{796}
(\byear{2014})
\doiurl{10.1002/asi.23062}
\end{barticle}
\endbibitem

\bibitem[\protect\citeauthoryear{Huang et~al.}{2014}]{huangEvidenceInformationContent2014}
\begin{botherref}
\oauthor{\bsnm{Huang}, \binits{A.H.}},
\oauthor{\bsnm{Zang}, \binits{A.Y.}},
\oauthor{\bsnm{Zheng}, \binits{R.}}:
Evidence on the information content of text in analyst reports.
ERN: Econometric Modeling in Financial Economics (Topic)
(2014)
\end{botherref}
\endbibitem

\bibitem[\protect\citeauthoryear{Shah et~al.}{2022}]{shah-FLANG-FLUE-2022}
\begin{botherref}
\oauthor{\bsnm{Shah}, \binits{R.S.}},
\oauthor{\bsnm{Chawla}, \binits{K.}},
\oauthor{\bsnm{Eidnani}, \binits{D.}},
\oauthor{\bsnm{Shah}, \binits{A.}},
\oauthor{\bsnm{Du}, \binits{W.}},
\oauthor{\bsnm{Chava}, \binits{S.}},
\oauthor{\bsnm{Raman}, \binits{N.}},
\oauthor{\bsnm{Smiley}, \binits{C.}},
\oauthor{\bsnm{Chen}, \binits{J.}},
\oauthor{\bsnm{Yang}, \binits{D.}}:
{WHEN FLUE MEETS FLANG}: Benchmarks and Large Pre-trained Language Model for Financial Domain
(2022).
\doiurl{10.48550/arXiv.2211.00083} .
\url{http://arxiv.org/abs/2211.00083}
Accessed 2023-05-27
\end{botherref}
\endbibitem

\bibitem[\protect\citeauthoryear{Wu et~al.}{2023}]{wu-BloombergGPT-2023}
\begin{botherref}
\oauthor{\bsnm{Wu}, \binits{S.}},
\oauthor{\bsnm{Irsoy}, \binits{O.}},
\oauthor{\bsnm{Lu}, \binits{S.}},
\oauthor{\bsnm{Dabravolski}, \binits{V.}},
\oauthor{\bsnm{Dredze}, \binits{M.}},
\oauthor{\bsnm{Gehrmann}, \binits{S.}},
\oauthor{\bsnm{Kambadur}, \binits{P.}},
\oauthor{\bsnm{Rosenberg}, \binits{D.}},
\oauthor{\bsnm{Mann}, \binits{G.}}:
{BloombergGPT}: A Large Language Model for Finance
(2023).
\doiurl{10.48550/arXiv.2303.17564} .
\url{http://arxiv.org/abs/2303.17564}
Accessed 2023-05-27
\end{botherref}
\endbibitem

\bibitem[\protect\citeauthoryear{Beltagy et~al.}{2019}]{beltagy_scibert_2019}
\begin{botherref}
\oauthor{\bsnm{Beltagy}, \binits{I.}},
\oauthor{\bsnm{Lo}, \binits{K.}},
\oauthor{\bsnm{Cohan}, \binits{A.}}:
{SciBERT}: {A} {Pretrained} {Language} {Model} for {Scientific} {Text}.
arXiv:1903.10676 [cs]
(2019).
arXiv: 1903.10676.
Accessed 2021-11-18
\end{botherref}
\endbibitem

\bibitem[\protect\citeauthoryear{Hu et~al.}{2022}]{huConfliBERTPretrainedLanguage2022}
\begin{bchapter}
\bauthor{\bsnm{Hu}, \binits{Y.}},
\bauthor{\bsnm{Hosseini}, \binits{M.}},
\bauthor{\bsnm{Skorupa~Parolin}, \binits{E.}},
\bauthor{\bsnm{Osorio}, \binits{J.}},
\bauthor{\bsnm{Khan}, \binits{L.}},
\bauthor{\bsnm{Brandt}, \binits{P.}},
\bauthor{\bsnm{D'Orazio}, \binits{V.}}:
\bctitle{{{ConfliBERT}}: {{A Pre-trained Language Model}} for {{Political Conflict}} and {{Violence}}}.
In: \bbtitle{Proceedings of the 2022 {{Conference}} of the {{North American Chapter}} of the {{Association}} for {{Computational Linguistics}}: {{Human Language Technologies}}},
pp. \bfpage{5469}--\blpage{5482}.
\bpublisher{{Association for Computational Linguistics}}, \blocation{???}
(\byear{2022}).
\doiurl{10.18653/v1/2022.naacl-main.400} .
\burl{https://aclanthology.org/2022.naacl-main.400}
Accessed 2023-05-31
\end{bchapter}
\endbibitem

\bibitem[\protect\citeauthoryear{Kawintiranon and Singh}{2022}]{kawintiranonPoliBERTweetPretrainedLanguage2022}
\begin{bchapter}
\bauthor{\bsnm{Kawintiranon}, \binits{K.}},
\bauthor{\bsnm{Singh}, \binits{L.}}:
\bctitle{{{PoliBERTweet}}: {{A Pre-trained Language Model}} for {{Analyzing Political Content}} on {{Twitter}}}.
In: \bbtitle{Proceedings of the {{Thirteenth Language Resources}} and {{Evaluation Conference}}},
pp. \bfpage{7360}--\blpage{7367}.
\bpublisher{{European Language Resources Association}}, \blocation{???}
(\byear{2022}).
\burl{https://aclanthology.org/2022.lrec-1.801}
Accessed 2023-05-31
\end{bchapter}
\endbibitem

\bibitem[\protect\citeauthoryear{Aghaei et~al.}{2023}]{aghaeiSecureBERTDomainSpecificLanguage2023}
\begin{bchapter}
\bauthor{\bsnm{Aghaei}, \binits{E.}},
\bauthor{\bsnm{Niu}, \binits{X.}},
\bauthor{\bsnm{Shadid}, \binits{W.}},
\bauthor{\bsnm{Al-Shaer}, \binits{E.}}:
\bctitle{{{SecureBERT}}: {{A Domain-Specific Language Model}} for~{{Cybersecurity}}}.
In: \beditor{\bsnm{Li}, \binits{F.}},
\beditor{\bsnm{Liang}, \binits{K.}},
\beditor{\bsnm{Lin}, \binits{Z.}},
\beditor{\bsnm{Katsikas}, \binits{S.K.}} (eds.)
\bbtitle{Security and {{Privacy}} in {{Communication Networks}}}.
\bsertitle{Lecture {{Notes}} of the {{Institute}} for {{Computer Sciences}}, {{Social Informatics}} and {{Telecommunications Engineering}}},
pp. \bfpage{39}--\blpage{56}.
\bpublisher{{Springer Nature Switzerland}}, \blocation{???}
(\byear{2023}).
\doiurl{10.1007/978-3-031-25538-0_3}
\end{bchapter}
\endbibitem

\bibitem[\protect\citeauthoryear{Zheng et~al.}{2022}]{zhengPretrainedDomainspecificLanguage2022}
\begin{botherref}
\oauthor{\bsnm{Zheng}, \binits{Z.}},
\oauthor{\bsnm{Lu}, \binits{X.-Z.}},
\oauthor{\bsnm{Chen}, \binits{K.-Y.}},
\oauthor{\bsnm{Zhou}, \binits{Y.-C.}},
\oauthor{\bsnm{Lin}, \binits{J.-R.}}:
Pretrained domain-specific language model for natural language processing tasks in the aec domain.
Comput. Ind.
\textbf{142}(C)
(2022)
\doiurl{10.1016/j.compind.2022.103733}
\end{botherref}
\endbibitem

\bibitem[\protect\citeauthoryear{Zhou et~al.}{2022}]{zhouCancerBERTCancerDomainspecific2022}
\begin{barticle}
\bauthor{\bsnm{Zhou}, \binits{S.}},
\bauthor{\bsnm{Wang}, \binits{N.}},
\bauthor{\bsnm{Wang}, \binits{L.}},
\bauthor{\bsnm{Liu}, \binits{H.}},
\bauthor{\bsnm{Zhang}, \binits{R.}}:
\batitle{{CancerBERT: a cancer domain-specific language model for extracting breast cancer phenotypes from electronic health records}}.
\bjtitle{Journal of the American Medical Informatics Association}
\bvolume{29}(\bissue{7}),
\bfpage{1208}--\blpage{1216}
(\byear{2022})
\doiurl{10.1093/jamia/ocac040}
{\href{https://arxiv.org/abs/https://academic.oup.com/jamia/article-pdf/29/7/1208/44062229/ocac040.pdf}{{https://academic.oup.com/jamia/article-pdf/29/7/1208/44062229/ocac040.pdf}}}
\end{barticle}
\endbibitem

\bibitem[\protect\citeauthoryear{Chalkidis et~al.}{2020}]{chalkidis_legal-bert_2020}
\begin{bchapter}
\bauthor{\bsnm{Chalkidis}, \binits{I.}},
\bauthor{\bsnm{Fergadiotis}, \binits{M.}},
\bauthor{\bsnm{Malakasiotis}, \binits{P.}},
\bauthor{\bsnm{Aletras}, \binits{N.}},
\bauthor{\bsnm{Androutsopoulos}, \binits{I.}}:
\bctitle{{LEGAL}-{BERT}: {The} {Muppets} straight out of {Law} {School}}.
In: \bbtitle{Findings of the {Association} for {Computational} {Linguistics}: {EMNLP} 2020},
pp. \bfpage{2898}--\blpage{2904}
(\byear{2020})
\end{bchapter}
\endbibitem

\bibitem[\protect\citeauthoryear{Zheng et~al.}{2021}]{zheng_when_2021}
\begin{botherref}
\oauthor{\bsnm{Zheng}, \binits{L.}},
\oauthor{\bsnm{Guha}, \binits{N.}},
\oauthor{\bsnm{Anderson}, \binits{B.R.}},
\oauthor{\bsnm{Henderson}, \binits{P.}},
\oauthor{\bsnm{Ho}, \binits{D.E.}}:
When {Does} {Pretraining} {Help}? {Assessing} {Self}-{Supervised} {Learning} for {Law} and the {CaseHOLD} {Dataset}.
arXiv:2104.08671 [cs]
(2021).
arXiv: 2104.08671 version: 3.
Accessed 2021-08-27
\end{botherref}
\endbibitem

\bibitem[\protect\citeauthoryear{Henderson et~al.}{2022}]{henderson_pile_2022}
\begin{botherref}
\oauthor{\bsnm{Henderson}, \binits{P.}},
\oauthor{\bsnm{Krass}, \binits{M.S.}},
\oauthor{\bsnm{Zheng}, \binits{L.}},
\oauthor{\bsnm{Guha}, \binits{N.}},
\oauthor{\bsnm{Manning}, \binits{C.D.}},
\oauthor{\bsnm{Jurafsky}, \binits{D.}},
\oauthor{\bsnm{Ho}, \binits{D.E.}}:
Pile of {Law}: {Learning} {Responsible} {Data} {Filtering} from the {Law} and a {256GB} {Open}-{Source} {Legal} {Dataset}.
arXiv.
arXiv:2207.00220 [cs]
(2022).
\url{http://arxiv.org/abs/2207.00220}
Accessed 2022-07-19
\end{botherref}
\endbibitem

\bibitem[\protect\citeauthoryear{Niklaus and Giofré}{2022}]{niklaus_budgetlongformer_2022}
\begin{botherref}
\oauthor{\bsnm{Niklaus}, \binits{J.}},
\oauthor{\bsnm{Giofré}, \binits{D.}}:
{BudgetLongformer}: {Can} we {Cheaply} {Pretrain} a {SotA} {Legal} {Language} {Model} {From} {Scratch}?
arXiv.
arXiv:2211.17135 [cs]
(2022).
\doiurl{10.48550/arXiv.2211.17135} .
\url{http://arxiv.org/abs/2211.17135}
Accessed 2023-01-23
\end{botherref}
\endbibitem

\bibitem[\protect\citeauthoryear{Clark et~al.}{2020}]{clark_electra_2020}
\begin{botherref}
\oauthor{\bsnm{Clark}, \binits{K.}},
\oauthor{\bsnm{Luong}, \binits{M.-T.}},
\oauthor{\bsnm{Le}, \binits{Q.V.}},
\oauthor{\bsnm{Manning}, \binits{C.D.}}:
{ELECTRA}: {Pre}-training {Text} {Encoders} as {Discriminators} {Rather} {Than} {Generators}.
arXiv:2003.10555 [cs]
(2020).
arXiv: 2003.10555.
Accessed 2022-03-08
\end{botherref}
\endbibitem

\bibitem[\protect\citeauthoryear{Hua et~al.}{2022}]{hua_legalrelectra_2022}
\begin{botherref}
\oauthor{\bsnm{Hua}, \binits{W.}},
\oauthor{\bsnm{Zhang}, \binits{Y.}},
\oauthor{\bsnm{Chen}, \binits{Z.}},
\oauthor{\bsnm{Li}, \binits{J.}},
\oauthor{\bsnm{Weber}, \binits{M.}}:
{LegalRelectra}: {Mixed}-domain {Language} {Modeling} for {Long}-range {Legal} {Text} {Comprehension}.
arXiv.
arXiv:2212.08204 [cs]
(2022).
\doiurl{10.48550/arXiv.2212.08204} .
\url{http://arxiv.org/abs/2212.08204}
Accessed 2023-03-29
\end{botherref}
\endbibitem

\bibitem[\protect\citeauthoryear{Feng et~al.}{2022}]{ijcai2022p765}
\begin{bchapter}
\bauthor{\bsnm{Feng}, \binits{Y.}},
\bauthor{\bsnm{Li}, \binits{C.}},
\bauthor{\bsnm{Ng}, \binits{V.}}:
\bctitle{Legal judgment prediction: A survey of the state of the art}.
In: \beditor{\bsnm{Raedt}, \binits{L.D.}} (ed.)
\bbtitle{Proceedings of the Thirty-First International Joint Conference on Artificial Intelligence, {IJCAI-22}},
pp. \bfpage{5461}--\blpage{5469}.
\bpublisher{International Joint Conferences on Artificial Intelligence Organization}, \blocation{???}
(\byear{2022}).
\doiurl{10.24963/ijcai.2022/765} .
\bcomment{Survey Track}.
\burl{https://doi.org/10.24963/ijcai.2022/765}
\end{bchapter}
\endbibitem

\bibitem[\protect\citeauthoryear{Aletras et~al.}{2016}]{aletras_predicting_2016}
\begin{barticle}
\bauthor{\bsnm{Aletras}, \binits{N.}},
\bauthor{\bsnm{Tsarapatsanis}, \binits{D.}},
\bauthor{\bsnm{Preoţiuc-Pietro}, \binits{D.}},
\bauthor{\bsnm{Lampos}, \binits{V.}}:
\batitle{Predicting judicial decisions of the {European} {Court} of {Human} {Rights}: a {Natural} {Language} {Processing} perspective}.
\bjtitle{PeerJ Computer Science}
\bvolume{2},
\bfpage{93}
(\byear{2016})
\doiurl{10.7717/peerj-cs.93} .
\bcomment{Publisher: PeerJ Inc.}
Accessed 2021-05-26
\end{barticle}
\endbibitem

\bibitem[\protect\citeauthoryear{Şulea et~al.}{2017}]{sulea_predicting_2017}
\begin{bchapter}
\bauthor{\bsnm{Şulea}, \binits{O.-M.}},
\bauthor{\bsnm{Zampieri}, \binits{M.}},
\bauthor{\bsnm{Vela}, \binits{M.}},
\bauthor{\bsnm{Genabith}, \binits{J.}}:
\bctitle{Predicting the {Law} {Area} and {Decisions} of {French} {Supreme} {Court} {Cases}}.
In: \bbtitle{Proceedings of the {International} {Conference} {Recent} {Advances} in {Natural} {Language} {Processing}, {RANLP} 2017},
pp. \bfpage{716}--\blpage{722}.
\bpublisher{INCOMA Ltd.},
\blocation{Varna, Bulgaria}
(\byear{2017})
\end{bchapter}
\endbibitem

\bibitem[\protect\citeauthoryear{Medvedeva et~al.}{2018}]{medvedeva_judicial_2018}
\begin{botherref}
\oauthor{\bsnm{Medvedeva}, \binits{M.}},
\oauthor{\bsnm{Vols}, \binits{M.}},
\oauthor{\bsnm{Wieling}, \binits{M.}}:
Judicial decisions of the {European} {Court} of {Human} {Rights}: looking into the crystall ball.
Proceedings of the Conference on Empirical Legal Studies in Europe 2018
(2018).
Accessed 2021-05-26
\end{botherref}
\endbibitem

\bibitem[\protect\citeauthoryear{Chalkidis et~al.}{2019}]{chalkidis_neural_2019}
\begin{bchapter}
\bauthor{\bsnm{Chalkidis}, \binits{I.}},
\bauthor{\bsnm{Androutsopoulos}, \binits{I.}},
\bauthor{\bsnm{Aletras}, \binits{N.}}:
\bctitle{Neural {Legal} {Judgment} {Prediction} in {English}}.
In: \bbtitle{Proceedings of the 57th {Annual} {Meeting} of the {Association} for {Computational} {Linguistics}},
pp. \bfpage{4317}--\blpage{4323}.
\bpublisher{Association for Computational Linguistics},
\blocation{Florence, Italy}
(\byear{2019}).
\doiurl{10.18653/v1/P19-1424} .
\burl{https://www.aclweb.org/anthology/P19-1424}
Accessed 2021-03-02
\end{bchapter}
\endbibitem

\bibitem[\protect\citeauthoryear{Xiao et~al.}{2018}]{xiao_cail2018_2018}
\begin{botherref}
\oauthor{\bsnm{Xiao}, \binits{C.}},
\oauthor{\bsnm{Zhong}, \binits{H.}},
\oauthor{\bsnm{Guo}, \binits{Z.}},
\oauthor{\bsnm{Tu}, \binits{C.}},
\oauthor{\bsnm{Liu}, \binits{Z.}},
\oauthor{\bsnm{Sun}, \binits{M.}},
\oauthor{\bsnm{Feng}, \binits{Y.}},
\oauthor{\bsnm{Han}, \binits{X.}},
\oauthor{\bsnm{Hu}, \binits{Z.}},
\oauthor{\bsnm{Wang}, \binits{H.}},
\oauthor{\bsnm{Xu}, \binits{J.}}:
{CAIL2018}: {A} {Large}-{Scale} {Legal} {Dataset} for {Judgment} {Prediction}.
arXiv:1807.02478 [cs]
(2018).
arXiv: 1807.02478.
Accessed 2021-06-25
\end{botherref}
\endbibitem

\bibitem[\protect\citeauthoryear{Xiao et~al.}{2021}]{xiao_lawformer_2021}
\begin{barticle}
\bauthor{\bsnm{Xiao}, \binits{C.}},
\bauthor{\bsnm{Hu}, \binits{X.}},
\bauthor{\bsnm{Liu}, \binits{Z.}},
\bauthor{\bsnm{Tu}, \binits{C.}},
\bauthor{\bsnm{Sun}, \binits{M.}}:
\batitle{Lawformer: {A} pre-trained language model for {Chinese} legal long documents}.
\bjtitle{AI Open}
\bvolume{2},
\bfpage{79}--\blpage{84}
(\byear{2021})
\doiurl{10.1016/j.aiopen.2021.06.003} .
Accessed 2022-07-19
\end{barticle}
\endbibitem

\bibitem[\protect\citeauthoryear{Malik et~al.}{2021}]{malik_ildc_2021}
\begin{bchapter}
\bauthor{\bsnm{Malik}, \binits{V.}},
\bauthor{\bsnm{Sanjay}, \binits{R.}},
\bauthor{\bsnm{Nigam}, \binits{S.K.}},
\bauthor{\bsnm{Ghosh}, \binits{K.}},
\bauthor{\bsnm{Guha}, \binits{S.K.}},
\bauthor{\bsnm{Bhattacharya}, \binits{A.}},
\bauthor{\bsnm{Modi}, \binits{A.}}:
\bctitle{{ILDC} for {CJPE}: {Indian} {Legal} {Documents} {Corpus} for {Court} {Judgment} {Prediction} and {Explanation}}.
In: \bbtitle{Proceedings of the 59th {Annual} {Meeting} of the {Association} for {Computational} {Linguistics} and the 11th {International} {Joint} {Conference} on {Natural} {Language} {Processing} ({Volume} 1: {Long} {Papers})},
pp. \bfpage{4046}--\blpage{4062}.
\bpublisher{Association for Computational Linguistics},
\blocation{Online}
(\byear{2021}).
\doiurl{10.18653/v1/2021.acl-long.313} .
\burl{https://aclanthology.org/2021.acl-long.313}
Accessed 2021-12-15
\end{bchapter}
\endbibitem

\bibitem[\protect\citeauthoryear{Chalkidis et~al.}{2021}]{chalkidis2021paragraph}
\begin{bchapter}
\bauthor{\bsnm{Chalkidis}, \binits{I.}},
\bauthor{\bsnm{Fergadiotis}, \binits{M.}},
\bauthor{\bsnm{Tsarapatsanis}, \binits{D.}},
\bauthor{\bsnm{Aletras}, \binits{N.}},
\bauthor{\bsnm{Androutsopoulos}, \binits{I.}},
\bauthor{\bsnm{Malakasiotis}, \binits{P.}}:
\bctitle{Paragraph-level rationale extraction through regularization: A case study on european court of human rights cases}.
In: \bbtitle{Proceedings of the 2021 Conference of the North American Chapter of the Association for Computational Linguistics: Human Language Technologies},
pp. \bfpage{226}--\blpage{241}
(\byear{2021})
\end{bchapter}
\endbibitem

\bibitem[\protect\citeauthoryear{Prasad et~al.}{2023}]{Prasad2023}
\begin{botherref}
\oauthor{\bsnm{Prasad}, \binits{N.}},
\oauthor{\bsnm{Boughanem}, \binits{M.}},
\oauthor{\bsnm{Dkaki}, \binits{T.}}:
A Hierarchical Neural Framework for Classification and its Explanation in Large Unstructured Legal Documents
(2023).
\url{http://arxiv.org/abs/2309.10563}
\end{botherref}
\endbibitem

\bibitem[\protect\citeauthoryear{Soh et~al.}{2019}]{soh_legal_2019}
\begin{bchapter}
\bauthor{\bsnm{Soh}, \binits{J.}},
\bauthor{\bsnm{Lim}, \binits{H.K.}},
\bauthor{\bsnm{Chai}, \binits{I.E.}}:
\bctitle{Legal {Area} {Classification}: {A} {Comparative} {Study} of {Text} {Classifiers} on {Singapore} {Supreme} {Court} {Judgments}}.
In: \bbtitle{Proceedings of the {Natural} {Legal} {Language} {Processing} {Workshop} 2019},
pp. \bfpage{67}--\blpage{77}.
\bpublisher{Association for Computational Linguistics},
\blocation{Minneapolis, Minnesota}
(\byear{2019}).
\doiurl{10.18653/v1/W19-2208} .
\burl{http://aclweb.org/anthology/W19-2208}
Accessed 2021-03-02
\end{bchapter}
\endbibitem

\bibitem[\protect\citeauthoryear{Katz et~al.}{2023}]{katz_natural_2023}
\begin{botherref}
\oauthor{\bsnm{Katz}, \binits{D.M.}},
\oauthor{\bsnm{Hartung}, \binits{D.}},
\oauthor{\bsnm{Gerlach}, \binits{L.}},
\oauthor{\bsnm{Jana}, \binits{A.}},
\oauthor{\bsnm{Bommarito}, \binits{M.J.}}:
Natural {Language} {Processing} in the {Legal} {Domain},
Rochester, NY
(2023).
\url{https://papers.ssrn.com/abstract=4336224}
Accessed 2023-01-24
\end{botherref}
\endbibitem

\bibitem[\protect\citeauthoryear{Ye et~al.}{2018}]{ye_interpretable_2018}
\begin{bchapter}
\bauthor{\bsnm{Ye}, \binits{H.}},
\bauthor{\bsnm{Jiang}, \binits{X.}},
\bauthor{\bsnm{Luo}, \binits{Z.}},
\bauthor{\bsnm{Chao}, \binits{W.}}:
\bctitle{Interpretable {Charge} {Predictions} for {Criminal} {Cases}: {Learning} to {Generate} {Court} {Views} from {Fact} {Descriptions}}.
In: \bbtitle{Proceedings of the 2018 {Conference} of the {North} {American} {Chapter} of the {Association} for {Computational} {Linguistics}: {Human} {Language} {Technologies}, {Volume} 1 ({Long} {Papers})},
pp. \bfpage{1854}--\blpage{1864}.
\bpublisher{Association for Computational Linguistics},
\blocation{New Orleans, Louisiana}
(\byear{2018}).
\doiurl{10.18653/v1/N18-1168} .
\burl{https://aclanthology.org/N18-1168}
Accessed 2022-07-14
\end{bchapter}
\endbibitem

\bibitem[\protect\citeauthoryear{Wu et~al.}{2020}]{wu_-biased_2020}
\begin{bchapter}
\bauthor{\bsnm{Wu}, \binits{Y.}},
\bauthor{\bsnm{Kuang}, \binits{K.}},
\bauthor{\bsnm{Zhang}, \binits{Y.}},
\bauthor{\bsnm{Liu}, \binits{X.}},
\bauthor{\bsnm{Sun}, \binits{C.}},
\bauthor{\bsnm{Xiao}, \binits{J.}},
\bauthor{\bsnm{Zhuang}, \binits{Y.}},
\bauthor{\bsnm{Si}, \binits{L.}},
\bauthor{\bsnm{Wu}, \binits{F.}}:
\bctitle{De-{Biased} {Court}'s {View} {Generation} with {Causality}}.
In: \bbtitle{Proceedings of the 2020 {Conference} on {Empirical} {Methods} in {Natural} {Language} {Processing} ({EMNLP})},
pp. \bfpage{763}--\blpage{780}.
\bpublisher{Association for Computational Linguistics},
\blocation{Online}
(\byear{2020}).
\doiurl{10.18653/v1/2020.emnlp-main.56} .
\burl{https://aclanthology.org/2020.emnlp-main.56}
Accessed 2022-07-14
\end{bchapter}
\endbibitem

\bibitem[\protect\citeauthoryear{Li and Zhang}{2021}]{Li_Zhang_2021}
\begin{barticle}
\bauthor{\bsnm{Li}, \binits{Q.}},
\bauthor{\bsnm{Zhang}, \binits{Q.}}:
\batitle{Court opinion generation from case fact description with legal basis}.
\bjtitle{Proceedings of the AAAI Conference on Artificial Intelligence}
\bvolume{35}(\bissue{17}),
\bfpage{14840}--\blpage{14848}
(\byear{2021})
\doiurl{10.1609/aaai.v35i17.17742}
\end{barticle}
\endbibitem

\bibitem[\protect\citeauthoryear{Yue et~al.}{2021}]{10.1145/3404835.3462984}
\begin{bchapter}
\bauthor{\bsnm{Yue}, \binits{L.}},
\bauthor{\bsnm{Liu}, \binits{Q.}},
\bauthor{\bsnm{Wu}, \binits{H.}},
\bauthor{\bsnm{An}, \binits{Y.}},
\bauthor{\bsnm{Wang}, \binits{L.}},
\bauthor{\bsnm{Yuan}, \binits{S.}},
\bauthor{\bsnm{Wu}, \binits{D.}}:
\bctitle{Circumstances enhanced criminal court view generation}.
In: \bbtitle{Proceedings of the 44th International ACM SIGIR Conference on Research and Development in Information Retrieval}.
\bsertitle{SIGIR '21},
pp. \bfpage{1855}--\blpage{1859}.
\bpublisher{Association for Computing Machinery},
\blocation{New York, NY, USA}
(\byear{2021}).
\doiurl{10.1145/3404835.3462984} .
\burl{https://doi.org/10.1145/3404835.3462984}
\end{bchapter}
\endbibitem

\bibitem[\protect\citeauthoryear{Grover et~al.}{2004}]{grover_holj_2004}
\begin{bchapter}
\bauthor{\bsnm{Grover}, \binits{C.}},
\bauthor{\bsnm{Hachey}, \binits{B.}},
\bauthor{\bsnm{Hughson}, \binits{I.}}:
\bctitle{The {HOLJ} {Corpus}. {Supporting} {Summarisation} of {Legal} {Texts}}.
In: \bbtitle{Proceedings of the 5th {International} {Workshop} on {Linguistically} {Interpreted} {Corpora}},
pp. \bfpage{47}--\blpage{54}.
\bpublisher{COLING},
\blocation{Geneva, Switzerland}
(\byear{2004}).
\burl{https://aclanthology.org/W04-1907}
\end{bchapter}
\endbibitem

\bibitem[\protect\citeauthoryear{Hachey and Grover}{2006}]{hachey-grover-2006-extractive}
\begin{barticle}
\bauthor{\bsnm{Hachey}, \binits{B.}},
\bauthor{\bsnm{Grover}, \binits{C.}}:
\batitle{Extractive summarisation of legal texts}.
\bjtitle{Artificial Intelligence and Law}
\bvolume{14}(\bissue{4}),
\bfpage{305}--\blpage{345}
(\byear{2006})
\doiurl{10.1007/s10506-007-9039-z}
\end{barticle}
\endbibitem

\bibitem[\protect\citeauthoryear{Kim et~al.}{2013}]{10.1007/978-3-642-39931-2_14}
\begin{bchapter}
\bauthor{\bsnm{Kim}, \binits{M.-Y.}},
\bauthor{\bsnm{Xu}, \binits{Y.}},
\bauthor{\bsnm{Goebel}, \binits{R.}}:
\bctitle{Summarization of legal texts with high cohesion and automatic compression rate}.
In: \beditor{\bsnm{Motomura}, \binits{Y.}},
\beditor{\bsnm{Butler}, \binits{A.}},
\beditor{\bsnm{Bekki}, \binits{D.}} (eds.)
\bbtitle{New Frontiers in Artificial Intelligence},
pp. \bfpage{190}--\blpage{204}.
\bpublisher{Springer},
\blocation{Berlin, Heidelberg}
(\byear{2013})
\end{bchapter}
\endbibitem

\bibitem[\protect\citeauthoryear{Jain et~al.}{2021}]{JAIN2021100388}
\begin{barticle}
\bauthor{\bsnm{Jain}, \binits{D.}},
\bauthor{\bsnm{Borah}, \binits{M.D.}},
\bauthor{\bsnm{Biswas}, \binits{A.}}:
\batitle{Summarization of legal documents: Where are we now and the way forward}.
\bjtitle{Computer Science Review}
\bvolume{40},
\bfpage{100388}
(\byear{2021})
\doiurl{10.1016/j.cosrev.2021.100388}
\end{barticle}
\endbibitem

\bibitem[\protect\citeauthoryear{Shukla et~al.}{2022}]{shukla2022legal}
\begin{bchapter}
\bauthor{\bsnm{Shukla}, \binits{A.}},
\bauthor{\bsnm{Bhattacharya}, \binits{P.}},
\bauthor{\bsnm{Poddar}, \binits{S.}},
\bauthor{\bsnm{Mukherjee}, \binits{R.}},
\bauthor{\bsnm{Ghosh}, \binits{K.}},
\bauthor{\bsnm{Goyal}, \binits{P.}},
\bauthor{\bsnm{Ghosh}, \binits{S.}}:
\bctitle{Legal case document summarization: Extractive and abstractive methods and their evaluation}.
In: \bbtitle{Proceedings of the 2nd Conference of the Asia-Pacific Chapter of the Association for Computational Linguistics and the 12th International Joint Conference on Natural Language Processing},
pp. \bfpage{1048}--\blpage{1064}
(\byear{2022})
\end{bchapter}
\endbibitem

\bibitem[\protect\citeauthoryear{Kornilova and Eidelman}{2019}]{kornilova2019billsum}
\begin{bchapter}
\bauthor{\bsnm{Kornilova}, \binits{A.}},
\bauthor{\bsnm{Eidelman}, \binits{V.}}:
\bctitle{Billsum: A corpus for automatic summarization of us legislation}.
In: \bbtitle{Proceedings of the 2nd Workshop on New Frontiers in Summarization},
pp. \bfpage{48}--\blpage{56}
(\byear{2019})
\end{bchapter}
\endbibitem

\bibitem[\protect\citeauthoryear{Shen et~al.}{2022}]{shen_multi-lexsum_2022}
\begin{botherref}
\oauthor{\bsnm{Shen}, \binits{Z.}},
\oauthor{\bsnm{Lo}, \binits{K.}},
\oauthor{\bsnm{Yu}, \binits{L.}},
\oauthor{\bsnm{Dahlberg}, \binits{N.}},
\oauthor{\bsnm{Schlanger}, \binits{M.}},
\oauthor{\bsnm{Downey}, \binits{D.}}:
Multi-{LexSum}: {Real}-{World} {Summaries} of {Civil} {Rights} {Lawsuits} at {Multiple} {Granularities}.
arXiv.
arXiv:2206.10883 [cs]
(2022).
\doiurl{10.48550/arXiv.2206.10883} .
\url{http://arxiv.org/abs/2206.10883}
Accessed 2022-07-25
\end{botherref}
\endbibitem

\bibitem[\protect\citeauthoryear{Lewis et~al.}{2020}]{lewis2020bart}
\begin{bchapter}
\bauthor{\bsnm{Lewis}, \binits{M.}},
\bauthor{\bsnm{Liu}, \binits{Y.}},
\bauthor{\bsnm{Goyal}, \binits{N.}},
\bauthor{\bsnm{Ghazvininejad}, \binits{M.}},
\bauthor{\bsnm{Mohamed}, \binits{A.}},
\bauthor{\bsnm{Levy}, \binits{O.}},
\bauthor{\bsnm{Stoyanov}, \binits{V.}},
\bauthor{\bsnm{Zettlemoyer}, \binits{L.}}:
\bctitle{Bart: Denoising sequence-to-sequence pre-training for natural language generation, translation, and comprehension}.
In: \bbtitle{Proceedings of the 58th Annual Meeting of the Association for Computational Linguistics},
pp. \bfpage{7871}--\blpage{7880}
(\byear{2020})
\end{bchapter}
\endbibitem

\bibitem[\protect\citeauthoryear{Zhang et~al.}{2020}]{zhang_pegasus_2020}
\begin{botherref}
\oauthor{\bsnm{Zhang}, \binits{J.}},
\oauthor{\bsnm{Zhao}, \binits{Y.}},
\oauthor{\bsnm{Saleh}, \binits{M.}},
\oauthor{\bsnm{Liu}, \binits{P.J.}}:
{PEGASUS}: {Pre}-training with {Extracted} {Gap}-sentences for {Abstractive} {Summarization}.
arXiv:1912.08777 [cs]
(2020).
arXiv: 1912.08777.
Accessed 2022-03-08
\end{botherref}
\endbibitem

\bibitem[\protect\citeauthoryear{Aumiller et~al.}{2022}]{aumiller_eur-lex-sum_2022}
\begin{botherref}
\oauthor{\bsnm{Aumiller}, \binits{D.}},
\oauthor{\bsnm{Chouhan}, \binits{A.}},
\oauthor{\bsnm{Gertz}, \binits{M.}}:
{EUR}-{Lex}-{Sum}: {A} {Multi}- and {Cross}-lingual {Dataset} for {Long}-form {Summarization} in the {Legal} {Domain}.
arXiv.
arXiv:2210.13448 [cs]
(2022).
\url{http://arxiv.org/abs/2210.13448}
Accessed 2022-10-28
\end{botherref}
\endbibitem

\bibitem[\protect\citeauthoryear{Datta et~al.}{2023}]{datta-etal-2023-mildsum}
\begin{bchapter}
\bauthor{\bsnm{Datta}, \binits{D.}},
\bauthor{\bsnm{Soni}, \binits{S.}},
\bauthor{\bsnm{Mukherjee}, \binits{R.}},
\bauthor{\bsnm{Ghosh}, \binits{S.}}:
\bctitle{{MILDS}um: A novel benchmark dataset for multilingual summarization of {I}ndian legal case judgments}.
In: \beditor{\bsnm{Bouamor}, \binits{H.}},
\beditor{\bsnm{Pino}, \binits{J.}},
\beditor{\bsnm{Bali}, \binits{K.}} (eds.)
\bbtitle{Proceedings of the 2023 Conference on Empirical Methods in Natural Language Processing},
pp. \bfpage{5291}--\blpage{5302}.
\bpublisher{Association for Computational Linguistics},
\blocation{Singapore}
(\byear{2023}).
\burl{https://aclanthology.org/2023.emnlp-main.321}
\end{bchapter}
\endbibitem

\bibitem[\protect\citeauthoryear{Bauer et~al.}{2023}]{bauer2023legal}
\begin{botherref}
\oauthor{\bsnm{Bauer}, \binits{E.}},
\oauthor{\bsnm{Stammbach}, \binits{D.}},
\oauthor{\bsnm{Gu}, \binits{N.}},
\oauthor{\bsnm{Ash}, \binits{E.}}:
Legal Extractive Summarization of U.S. Court Opinions
(2023)
\end{botherref}
\endbibitem

\bibitem[\protect\citeauthoryear{Martínez-González et~al.}{2005}]{martinez-gonzalez_reference_2005}
\begin{bchapter}
\bauthor{\bsnm{Martínez-González}, \binits{M.}},
\bauthor{\bsnm{Fuente}, \binits{P.}},
\bauthor{\bsnm{Vicente}, \binits{D.-J.}}:
\bctitle{Reference {Extraction} and {Resolution} for {Legal} {Texts}}.
In: \beditor{\bsnm{Pal}, \binits{S.K.}},
\beditor{\bsnm{Bandyopadhyay}, \binits{S.}},
\beditor{\bsnm{Biswas}, \binits{S.}} (eds.)
\bbtitle{Pattern {Recognition} and {Machine} {Intelligence}}.
\bsertitle{Lecture {Notes} in {Computer} {Science}},
pp. \bfpage{218}--\blpage{221}.
\bpublisher{Springer},
\blocation{Berlin, Heidelberg}
(\byear{2005}).
\doiurl{10.1007/11590316_29}
\end{bchapter}
\endbibitem

\bibitem[\protect\citeauthoryear{Nambanoor~Kunnath et~al.}{2022}]{nambanoor_kunnath_dynamic_2022}
\begin{bchapter}
\bauthor{\bsnm{Nambanoor~Kunnath}, \binits{S.}},
\bauthor{\bsnm{Pride}, \binits{D.}},
\bauthor{\bsnm{Knoth}, \binits{P.}}:
\bctitle{Dynamic {Context} {Extraction} for {Citation} {Classification}}.
In: \bbtitle{Proceedings of the 2nd {Conference} of the {Asia}-{Pacific} {Chapter} of the {Association} for {Computational} {Linguistics} and the 12th {International} {Joint} {Conference} on {Natural} {Language} {Processing} ({Volume} 1: {Long} {Papers})},
pp. \bfpage{539}--\blpage{549}.
\bpublisher{Association for Computational Linguistics},
\blocation{Online only}
(\byear{2022}).
\burl{https://aclanthology.org/2022.aacl-main.41}
Accessed 2023-06-15
\end{bchapter}
\endbibitem

\bibitem[\protect\citeauthoryear{Lawrie et~al.}{2023}]{lawrie2023neural}
\begin{botherref}
\oauthor{\bsnm{Lawrie}, \binits{D.}},
\oauthor{\bsnm{Yang}, \binits{E.}},
\oauthor{\bsnm{Oard}, \binits{D.W.}},
\oauthor{\bsnm{Mayfield}, \binits{J.}}:
Neural Approaches to Multilingual Information Retrieval
(2023)
\end{botherref}
\endbibitem

\bibitem[\protect\citeauthoryear{Bajaj et~al.}{2018}]{bajaj2018ms}
\begin{botherref}
\oauthor{\bsnm{Bajaj}, \binits{P.}},
\oauthor{\bsnm{Campos}, \binits{D.}},
\oauthor{\bsnm{Craswell}, \binits{N.}},
\oauthor{\bsnm{Deng}, \binits{L.}},
\oauthor{\bsnm{Gao}, \binits{J.}},
\oauthor{\bsnm{Liu}, \binits{X.}},
\oauthor{\bsnm{Majumder}, \binits{R.}},
\oauthor{\bsnm{McNamara}, \binits{A.}},
\oauthor{\bsnm{Mitra}, \binits{B.}},
\oauthor{\bsnm{Nguyen}, \binits{T.}},
\oauthor{\bsnm{Rosenberg}, \binits{M.}},
\oauthor{\bsnm{Song}, \binits{X.}},
\oauthor{\bsnm{Stoica}, \binits{A.}},
\oauthor{\bsnm{Tiwary}, \binits{S.}},
\oauthor{\bsnm{Wang}, \binits{T.}}:
MS MARCO: A Human Generated MAchine Reading COmprehension Dataset
(2018)
\end{botherref}
\endbibitem

\bibitem[\protect\citeauthoryear{Conneau et~al.}{2020}]{conneau2020unsupervised}
\begin{botherref}
\oauthor{\bsnm{Conneau}, \binits{A.}},
\oauthor{\bsnm{Khandelwal}, \binits{K.}},
\oauthor{\bsnm{Goyal}, \binits{N.}},
\oauthor{\bsnm{Chaudhary}, \binits{V.}},
\oauthor{\bsnm{Wenzek}, \binits{G.}},
\oauthor{\bsnm{Guzmán}, \binits{F.}},
\oauthor{\bsnm{Grave}, \binits{E.}},
\oauthor{\bsnm{Ott}, \binits{M.}},
\oauthor{\bsnm{Zettlemoyer}, \binits{L.}},
\oauthor{\bsnm{Stoyanov}, \binits{V.}}:
Unsupervised Cross-lingual Representation Learning at Scale
(2020)
\end{botherref}
\endbibitem

\bibitem[\protect\citeauthoryear{Leveling}{2012}]{Leveling2012OnTE}
\begin{bchapter}
\bauthor{\bsnm{Leveling}, \binits{J.}}:
\bctitle{On the effect of stopword removal for sms-based faq retrieval}.
In: \bbtitle{International Conference on Applications of Natural Language to Data Bases}
(\byear{2012})
\end{bchapter}
\endbibitem

\bibitem[\protect\citeauthoryear{Khattab and Zaharia}{2020}]{ColBERT2020}
\begin{bchapter}
\bauthor{\bsnm{Khattab}, \binits{O.}},
\bauthor{\bsnm{Zaharia}, \binits{M.}}:
\bctitle{Colbert: Efficient and effective passage search via contextualized late interaction over bert}.
In: \bbtitle{Proceedings of the 43rd International ACM SIGIR Conference on Research and Development in Information Retrieval},
pp. \bfpage{39}--\blpage{48}.
\bpublisher{Association for Computing Machinery}, \blocation{???}
(\byear{2020}).
\doiurl{10.1145/3397271.3401075} .
\burl{https://doi.org/10.1145/3397271.3401075}
\end{bchapter}
\endbibitem

\bibitem[\protect\citeauthoryear{Gao et~al.}{2021}]{Gao2021}
\begin{bchapter}
\bauthor{\bsnm{Gao}, \binits{L.}},
\bauthor{\bsnm{Dai}, \binits{Z.}},
\bauthor{\bsnm{Callan}, \binits{J.}}:
\bctitle{Coil: Revisit exact lexical match in information retrieval with contextualized inverted list}.
In: \beditor{\bsnm{Toutanova}, \binits{K.}},
\beditor{\bsnm{Rumshisky}, \binits{A.}},
\beditor{\bsnm{Zettlemoyer}, \binits{L.}},
\beditor{\bsnm{Hakkani-Tur}, \binits{D.}},
\beditor{\bsnm{Beltagy}, \binits{I.}},
\beditor{\bsnm{Bethard}, \binits{S.}},
\beditor{\bsnm{Cotterell}, \binits{R.}},
\beditor{\bsnm{Chakraborty}, \binits{T.}},
\beditor{\bsnm{Zhou}, \binits{Y.}} (eds.)
\bbtitle{Proceedings of the 2021 Conference of the North American Chapter of the Association for Computational Linguistics: Human Language Technologies},
pp. \bfpage{3030}--\blpage{3042}.
\bpublisher{Association for Computational Linguistics}, \blocation{???}
(\byear{2021}).
\doiurl{10.18653/v1/2021.naacl-main.241} .
\burl{https://aclanthology.org/2021.naacl-main.241}
\end{bchapter}
\endbibitem

\bibitem[\protect\citeauthoryear{Lin and Lin}{2023}]{Lin2023}
\begin{botherref}
\oauthor{\bsnm{Lin}, \binits{S.C.}},
\oauthor{\bsnm{Lin}, \binits{J.}}:
A dense representation framework for lexical and semantic matching.
ACM Transactions on Information Systems
\textbf{41}
(2023)
\doiurl{10.1145/3582426}
\end{botherref}
\endbibitem

\bibitem[\protect\citeauthoryear{Wang et~al.}{2020}]{MiniLM2020}
\begin{bchapter}
\bauthor{\bsnm{Wang}, \binits{W.}},
\bauthor{\bsnm{Wei}, \binits{F.}},
\bauthor{\bsnm{Dong}, \binits{L.}},
\bauthor{\bsnm{Bao}, \binits{H.}},
\bauthor{\bsnm{Yang}, \binits{N.}},
\bauthor{\bsnm{Zhou}, \binits{M.}}:
\bctitle{Minilm: Deep self-attention distillation for task-agnostic compression of pre-trained transformers}.
In: \bbtitle{Proceedings of the 34th International Conference on Neural Information Processing Systems}.
\bpublisher{Curran Associates Inc.}, \blocation{???}
(\byear{2020})
\end{bchapter}
\endbibitem

\end{thebibliography}

\end{document}